\documentclass[10pt,twocolumn,letterpaper]{article}

\usepackage[pagenumbers]{wacv} 

\usepackage[accsupp]{axessibility}
\usepackage{tcolorbox}
\usepackage{wrapfig}
\usepackage{graphicx}
\usepackage[percent]{overpic}
\usepackage{pict2e}
\usepackage{multirow}
\usepackage{bm}
\usepackage{booktabs}
\usepackage{multicol}
\usepackage{makecell}
\usepackage[normalem]{ulem}

\newcommand{\I}{\textbf{I}}
\newcommand{\T}{\textbf{T}}
\newcommand{\B}{\textbf{B}}
\newcommand{\R}{\textbf{R}}
\newcommand{\MR}{\textbf{MR}}

\newcommand{\PT}{\mathcal{T}}
\newcommand{\PR}{\mathcal{R}}

\newcommand{\flux}{\texttt{FLUX.1 Kontext}}

\newcommand{\n}{\bm{n}}
\newcommand{\w}{\bm\omega}

\newcommand{\wo}{\bm\omega_\mathrm{o}}

\newcommand{\dotp}[2]{|#1 \cdot #2|}
\newcommand{\f}{f}

\newcommand{\Sph}{{{\mathcal S}^2}}

\newcommand{\intd}{\,\mathrm{d}}

\newlength{\resLen}

\definecolor{wacvblue}{rgb}{0.21,0.49,0.74}
\usepackage[pagebackref,breaklinks,colorlinks,allcolors=wacvblue]{hyperref}

\def\wacvPaperID{0005} 
\def\confName{WACVW}
\def\confYear{2026}

\title{SIRR-LMM: Single-image Reflection Removal via Large Multimodal Model}

\author{Yu Guo, Zhiqiang Lao, Xiyun Song, Yubin Zhou and Heather Yu\\
Futurewei Technologies, US\\
{\tt\small \{yguo1,zlao,xsong,yzhou2,hyu\}@futurewei.com}}

\begin{document}
\maketitle
\begin{abstract}
Glass surfaces create complex interactions of reflected and transmitted light, making single-image reflection removal (SIRR) challenging. Existing datasets suffer from limited physical realism in synthetic data or insufficient scale in real captures. We introduce a synthetic dataset generation framework that path-traces 3D glass models over real background imagery to create physically accurate reflection scenarios with varied glass properties, camera settings, and post-processing effects. To leverage the capabilities of Large Multimodal Model (LMM), we concatenate the image layers into a single composite input, apply joint captioning, and fine-tune the model using task-specific LoRA rather than full-parameter training. This enables our approach to achieve improved reflection removal and separation performance compared to state-of-the-art methods.
\end{abstract}

\section{Introduction}
\label{sec:intro}

Glass surfaces are common in everyday environments, appearing as windows, doors, picture frames, and display cases, \etc. They introduce complex interactions between reflected and transmitted light, making accurate reflection removal or separation essential for applications such as photo retouching, autonomous driving, augmented reality, and digital heritage preservation.

Single-image reflection removal (SIRR) is inherently ill-posed because reflections and transmissions are tightly coupled, especially in over exposed regions. Real-world conditions further complicate the task through diverse lighting, scene geometry, and glass properties such as thickness or color, which limiting the effectiveness of classical decomposition methods.

Deep learning methods have made significant progress in SIRR through the use of sophisticated neural architectures and large-scale training. Recent advances in Large Multimodal Models (LMMs) further demonstrate strong capabilities in understanding and manipulating complex visual phenomena through text-based interactions. Since these foundation models are trained on billions of real-world images, they naturally acquire prior knowledge about transparent materials, including glass. This makes them promising candidates for reflection removal, especially when a pre-trained general purpose model is further fine-tuned on a smaller dataset tailored to this task.

Despite these advantages, a major bottleneck in advancing reflection removal is the limited availability of high-quality training data. Real-world data collection is hindered by the difficulty of capturing accurate ground truth to separate reflection and transmission, while traditional synthetic datasets that simply blend two images fail to capture the physical realism needed for effective learning. These challenges restrict the development of more robust and reliable reflection removal systems.


We address this gap with a path-traced synthetic dataset that uses physically accurate 3D glass models combined with real background imagery. Our rendering pipeline simulates diverse glass and camera conditions using High Dynamic Range (HDR) environment maps for lighting and RGB images for near-field content. Instead of relying on large datasets, we fine-tune an LMM with task-specific Low-Rank Adaptation (LoRA), enabling strong performance with relatively small but high-fidelity synthetic data.
Our main contributions include:
\begin{itemize}
    \item A physically accurate synthetic dataset generation framework using ray tracing and real-world imagery, with an efficient approach that combines HDR environment maps with RGB images to avoid complex 3D scene modeling.
    \item To the best of our knowledge, this is the first work to explore the fine-tuned Large Multimodal Models for single-image reflection removal task. 
    \item Empirical validation showing that appropriately adapted LMM with our synthetic data achieves superior reflection removal performance compared to previous approaches.
\end{itemize}
\section{Related Work}




\subsection{Reflection Removal and Separation}

Most SIRR methods focus on reflection removal to obtain clean images that contain only transmission content \cite{fan2017generic, yang2018seeing, wan2018crrn, wen2019single, wei2019single, wan2019corrn, zou2020deep, kim2020single, zheng2021single, prasad2021v, wang2022background, zhu2023hue, song2023robust, li2023two, rosh2024r2sfd, zhu2024revisiting, wei2024dereflectformer, zhao2025reversible, cai2025f2t2}. 

In certain applications, such as environment estimation \cite{guo2025epbr}, reflection separation becomes equally important. Several works have demonstrated the ability to generate both reflection and transmission images simultaneously \cite{wan2017benchmarking, zhang2018single, wan2020reflection, li2020single, hu2021trash, dong2021location, chang2021single, wan2022benchmarking, lei2022categorized, he2023reflection, hu2023single, hu2024single, chen2025high}. Compared to reflection removal, reflection recovery presents greater challenges due to the inherent blurring and ghosting effects caused by double reflections and out-of-focus phenomena, making accurate recovery significantly more difficult.

To overcome the limitations of single-image approaches, researchers have explored various multi-modal strategies. Some works utilize polarized images \cite{wieschollek2018separating, lyu2019reflection, lei2020polarized, yao2025polarfree}, while others rely on additional flash images \cite{chang2020siamese, lei2021robust, lei2023robust, wang2025flash}. Alternative approaches include processing image sequences \cite{liu2020learning}, incorporating semantic information \cite{liu2022semantic}, using text descriptions \cite{zhong2024language, hong2024differ}, utilizing selfie images \cite{kee2025removing}, or employing interactive methods \cite{chen2025firm}. For a complete coverage of these approaches, see more details in recent surveys \cite{amanlou2022single, yang2025survey, wang2025review}.

\subsection{Existing Datasets}


Real captured datasets are built by capturing images with reflections, then blocking transmission to obtain reflection-only images, or removing reflective surface to produce clean transmission images. Popular datasets include SIR$^2$ \cite{wan2017benchmarking}, Real \cite{zhang2018single}, Nature \cite{li2020single}, CDR \cite{lei2022categorized}, and RRW \cite{zhu2024revisiting}. Although video-based capture with camera or glass movement enables rapid dataset collection, it sacrifices diversity and often fails to achieve complete separation of reflection and transmission, leaving residual reflections inconsistencies. It also fails with changes in motion or lighting, making the resulting data unreliable for training.

Real data are often insufficient for comprehensive training, making synthetic data a necessary complement. A common strategy synthesizes reflections by mixing two tone-mapped images, typically expressed as $I = f(T) + g(R)$ or $I = f(T, R)$, using datasets such as PASCAL VOC \cite{everingham2010pascal}. However, these approaches rely on simplified reflection models that do not capture real glass behavior. To address this, Kim \etal \cite{kim2020single} estimate scene geometry from RGBD inputs and use path tracing to simulate light transport between scene meshes and glass. Since tone-mapped images do not preserve the true relationship between pixel values and scene luminance, Kee \etal \cite{kee2025removing} instead mixes linear RAW photos to construct more faithful data. 

With these insights, our work integrates HDR environment maps with path tracing to generate synthetic data that better represent the lighting and reflection in the real-world. In addition, our method can easily extend to the non-planar surface, such as prescription eyeglasses \cite{zhang2017virtual}.

\subsection{Image synthesis LMM conditioned on text}

Diffusion Transformers (DiT) \cite{peebles2023scalable} represent a major advancement in generative modeling by combining transformer architectures with diffusion processes. DiT uses self-attention to capture long-range dependencies and complex spatial relationships, enabling more sophisticated visual understanding and higher-quality image generation.

\flux~\cite{batifol2025flux} builds on the DiT to create a state-of-the-art generative model for complex visual manipulation tasks. Through Flow Matching training and context-aware processing, it can handle intricate phenomena such as glass reflections with high fidelity while providing precise and flexible control over the generation process.

LoRA \cite{huang2024context} offers an efficient method for adapting large foundation models to specific tasks by adding low-rank matrices that encode task-specific behavior. This allows models like \flux~to be customized for reflection removal using limited data while preserving their broad pre-trained knowledge.

We build our approach on \flux~and train a LoRA specifically for our tasks.
\section{Data Generation}

\begin{figure}[t]
    \centering
    \setlength{\resLen}{0.195\linewidth}
    \addtolength{\tabcolsep}{-5.5pt}
    \begin{tabular}{ccccc}
        \includegraphics[width=\resLen]{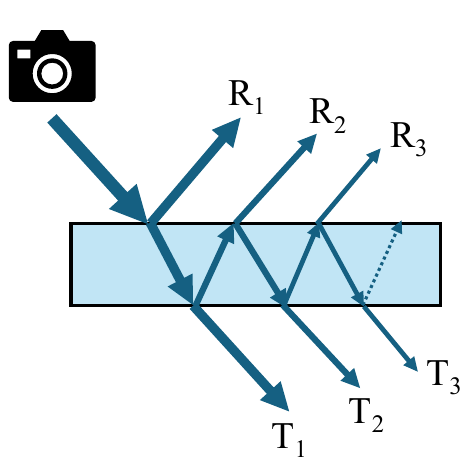} &
        \includegraphics[width=\resLen]{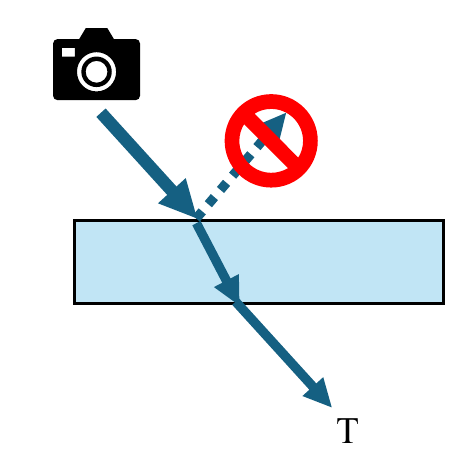} &
        \includegraphics[width=\resLen]{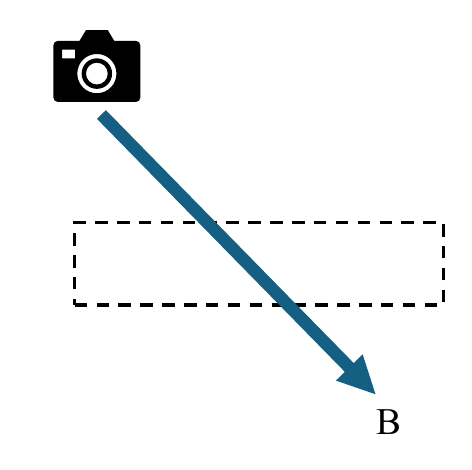} &
        \includegraphics[width=\resLen]{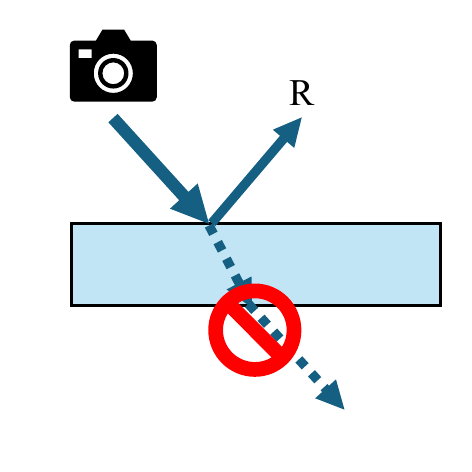} &
        \includegraphics[width=\resLen]{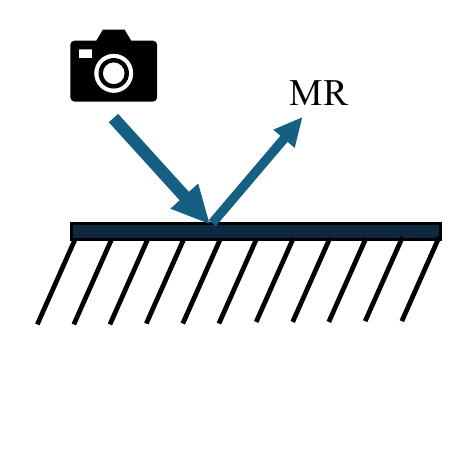}
        \\
        \includegraphics[width=\resLen, trim=150 50 20 120, clip]{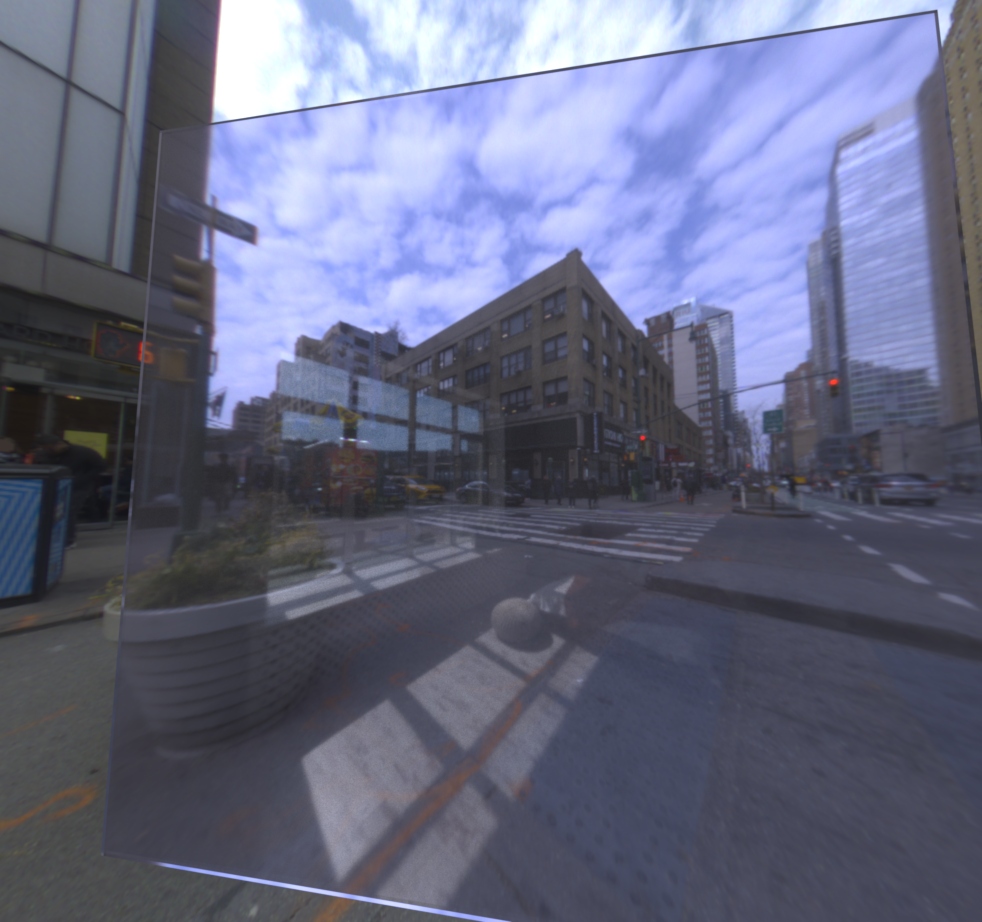} &
        \includegraphics[width=\resLen, trim=150 50 20 120, clip]{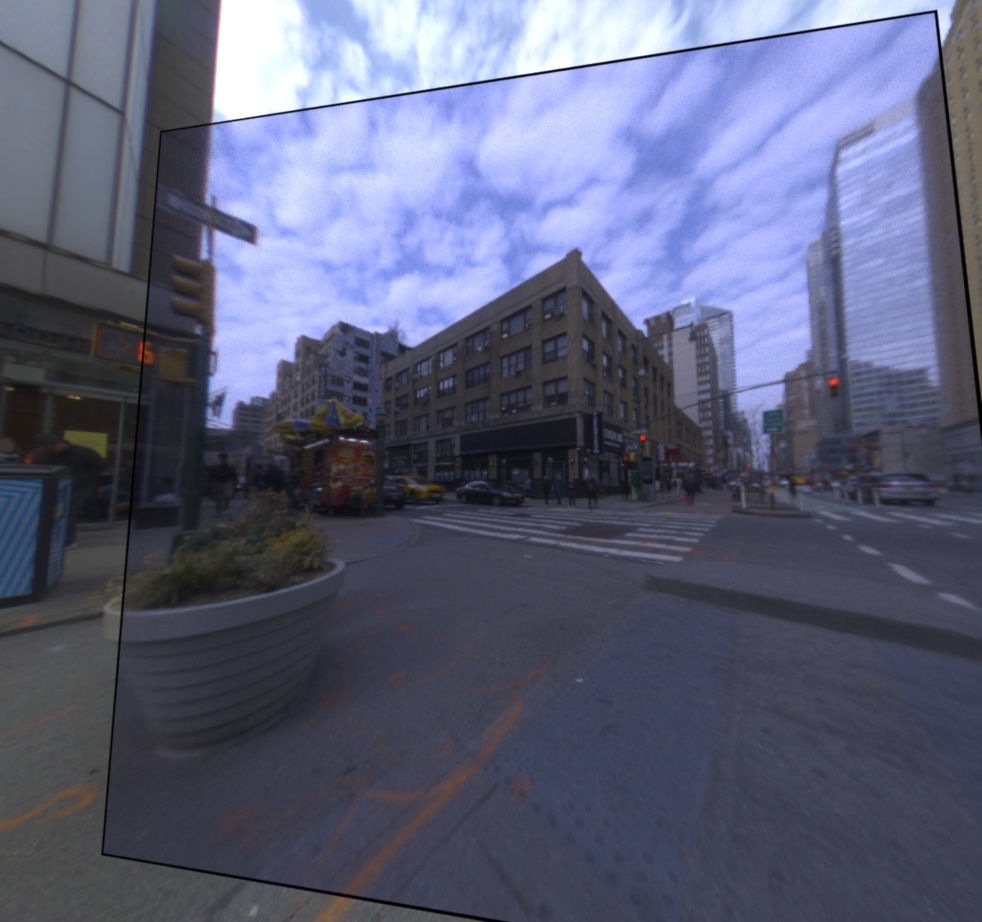} &
        \includegraphics[width=\resLen, trim=150 50 20 120, clip]{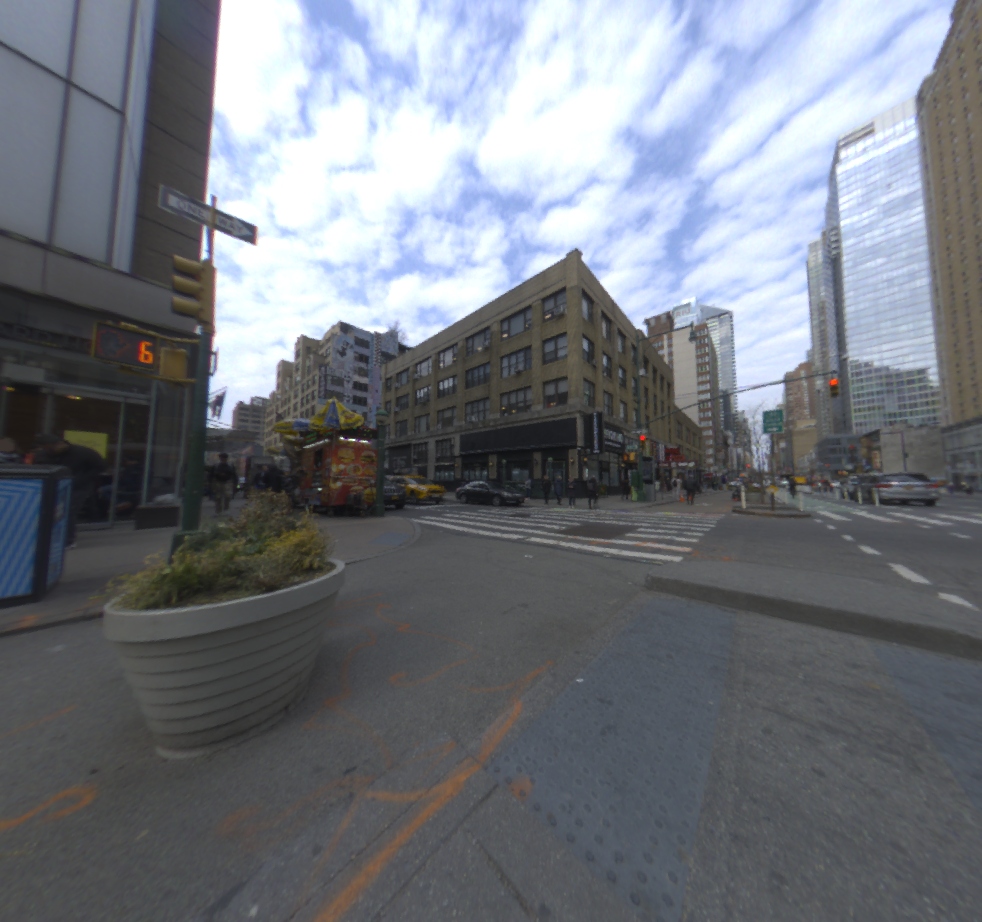} &
        \includegraphics[width=\resLen, trim=150 50 20 120, clip]{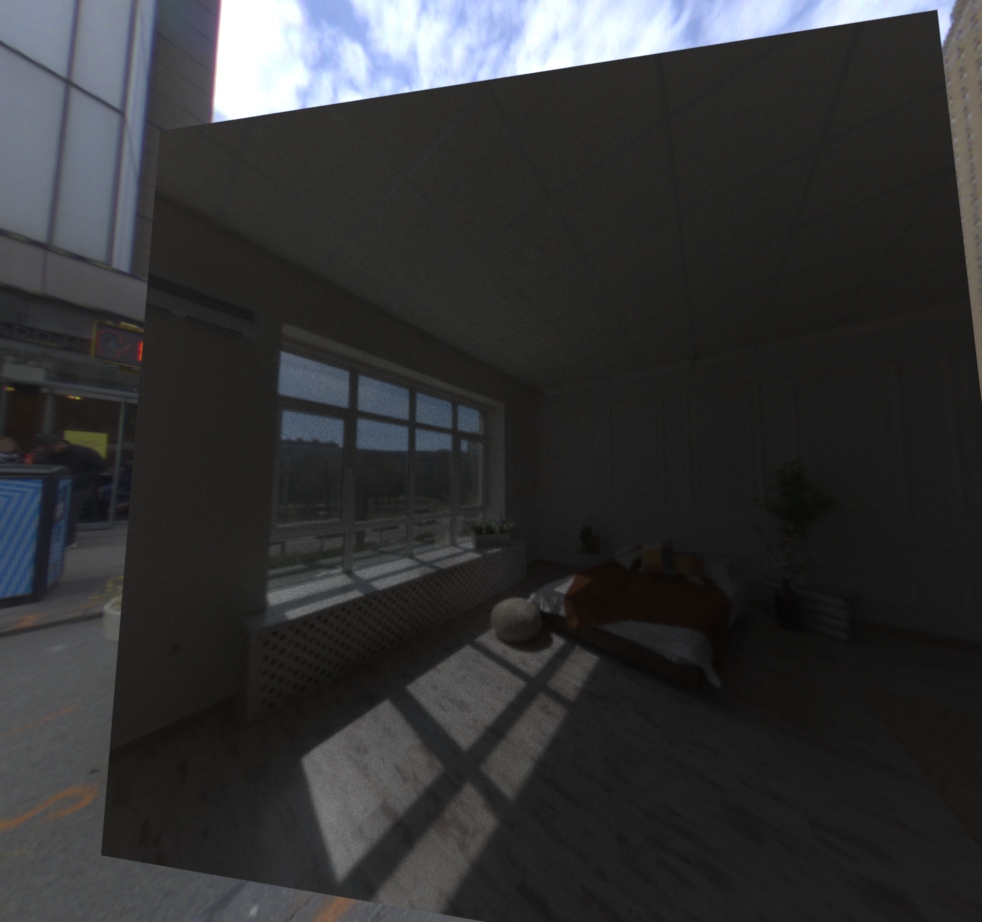} &
        \includegraphics[width=\resLen, trim=150 50 20 120, clip]{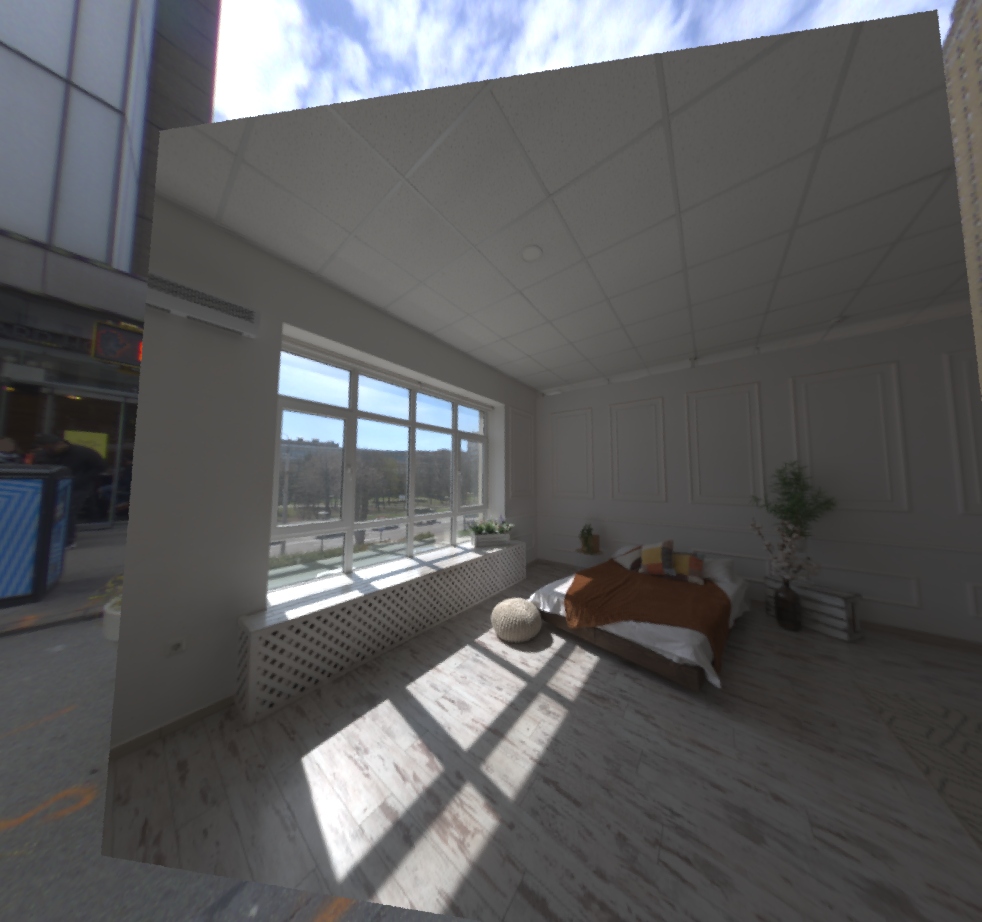}   
        \\
        (a) \I & (b) \T & (c) \B & (d) \R & (e) \MR
    \end{tabular}
    \vspace{-8pt}
    \caption{
        \textbf{Reflection and transmission.} 
        (a) The \I nput image with glass reflection; 
        (b) The \T ransmission component of \I; 
        (c) The \B ackground image without glass;
        (d) The \R eflection component of \I;
        (e) The \textbf{M}irror \textbf{R}eflection, which is considered as back scene image without any glass effects.
        We use blue tint glass to better illustrate the difference between \T~and \B.
    }
    \vspace{-10pt}
    \label{fig:data_example}
\end{figure}

Our dataset generation framework combines path-traced 3D glass models with real-world background imagery to create physically accurate reflection scenarios. 

\subsection{Path-traced scene setup}
When light passes through glass, it generates effects such as blurring, ghosting, and attenuation. Image space synthesis methods must design appropriate degradation for reflection and transmission, which could involve using a weight, Gaussian, or even a trained network. Unlike previous synthetic methods, we simulate glass effects by physically modeling the glass and rendering it with path tracing \cite{guo2018position}. The rendering equation for a non-emissive transparent surface point is,
\vspace{-10pt}
\begin{equation}
    \label{eqn:rendering_bsdf}
    L(\wo) = \int_{\Sph} \f(\wo, \w) \, L(\w) \, \dotp{\w}{\n} \intd \w 
    \vspace{-5pt}
\end{equation}
where $L$ is the radiance in the viewing direction $\wo$. The final rendering or pixel color is the result of integrating the energy from all light paths, weighted by the Bidirectional Reflectance Distribution Function (BRDF, $\f$) and the cosine of the angle between the normal surface ($\n$) and the direction of the light path ($\w$).

For a perfect glass slab, we can write down all the light paths (Figure \ref{fig:data_example}-(a), top) by using $\PT$ (light hit glass and transmit/refract) and $\PR$ (light hit glass and reflect).
All the transmissions, 
\vspace{-7pt}
\begin{equation}
  T_1 + T_2 + T_3 + \cdots
    = \PT\PT + \PT\PR^2\PT + \PT\PR^4\PT + \cdots
    \vspace{-5pt}
\end{equation}
and all the reflections,
\vspace{-7pt}
\begin{equation}
  R_1 + R_2 + R_3 + \cdots
    = \PR + \PT\PR\PT + \PT\PR^3\PT + \cdots
    \vspace{-5pt}
\end{equation}
For each $\PT$ and $\PR$, the glass effects, such as blur or attenuation, contribute to the brightness of the image. 

\begin{figure}[t]
    \centering
    \setlength{\resLen}{0.26\linewidth}
    \addtolength{\tabcolsep}{0pt}
    \begin{tabular}{ccc}
        \includegraphics[width=\resLen]{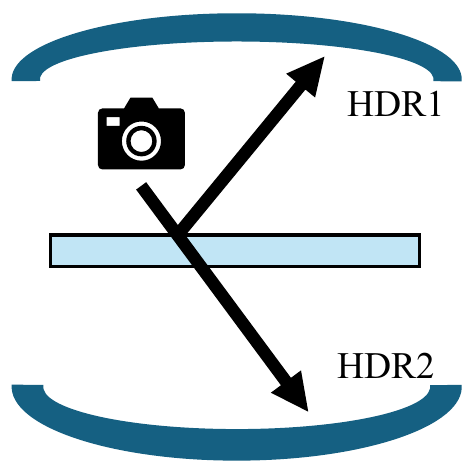} &
        \includegraphics[width=\resLen]{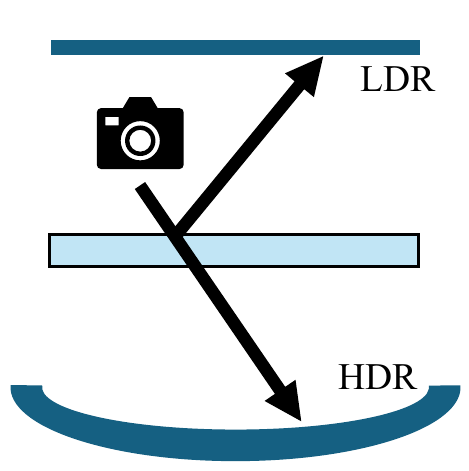} &
        \includegraphics[width=\resLen]{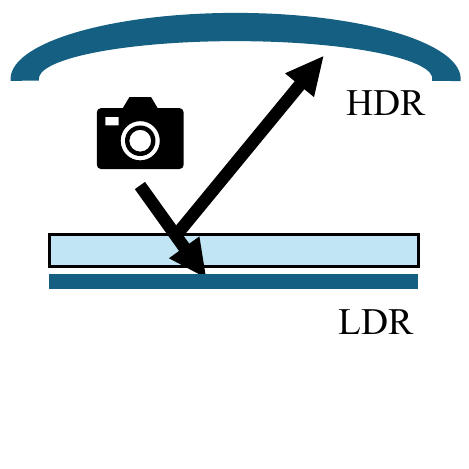}
        \\
        \includegraphics[width=\resLen]{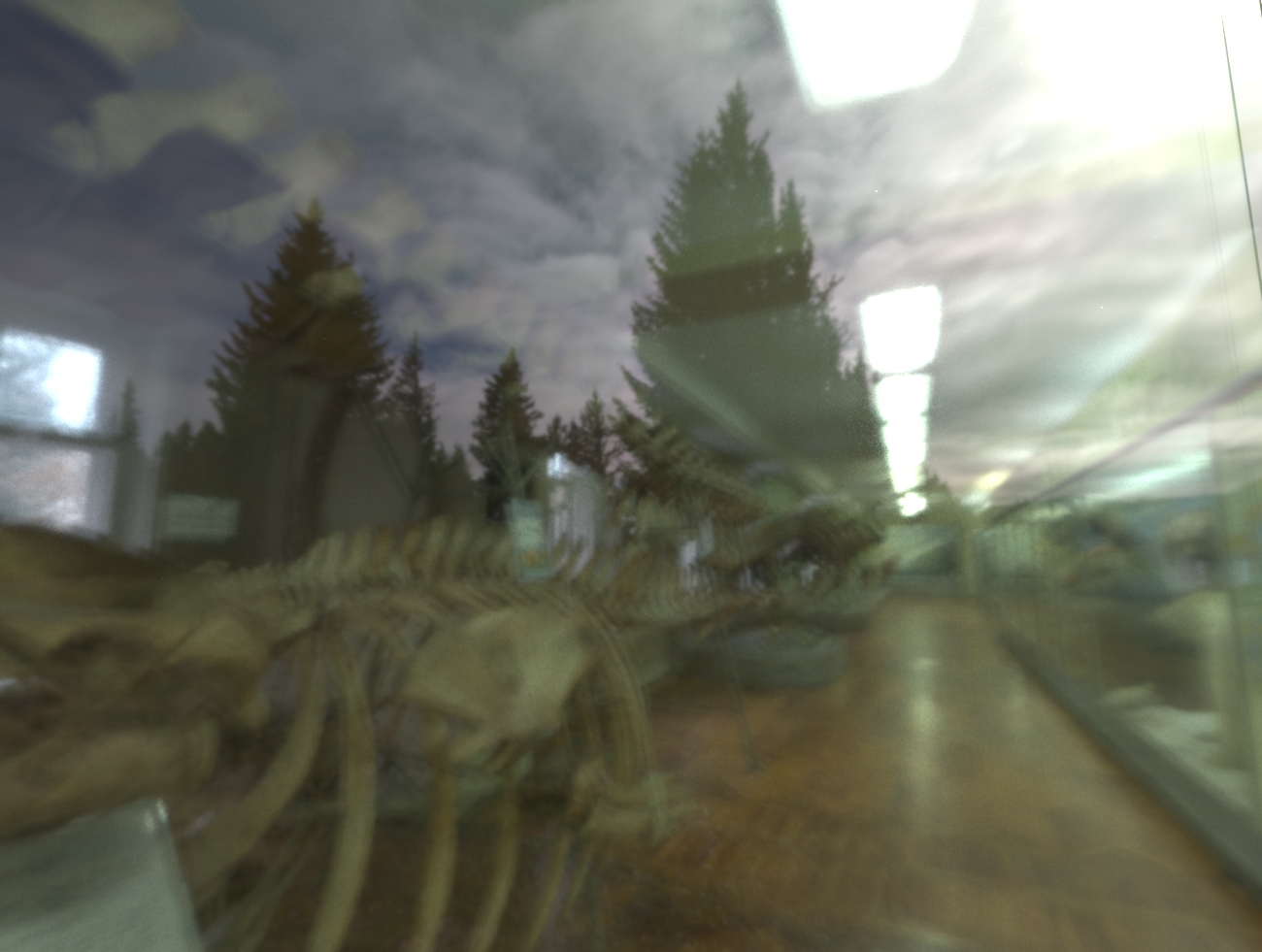} &
        \includegraphics[width=\resLen]{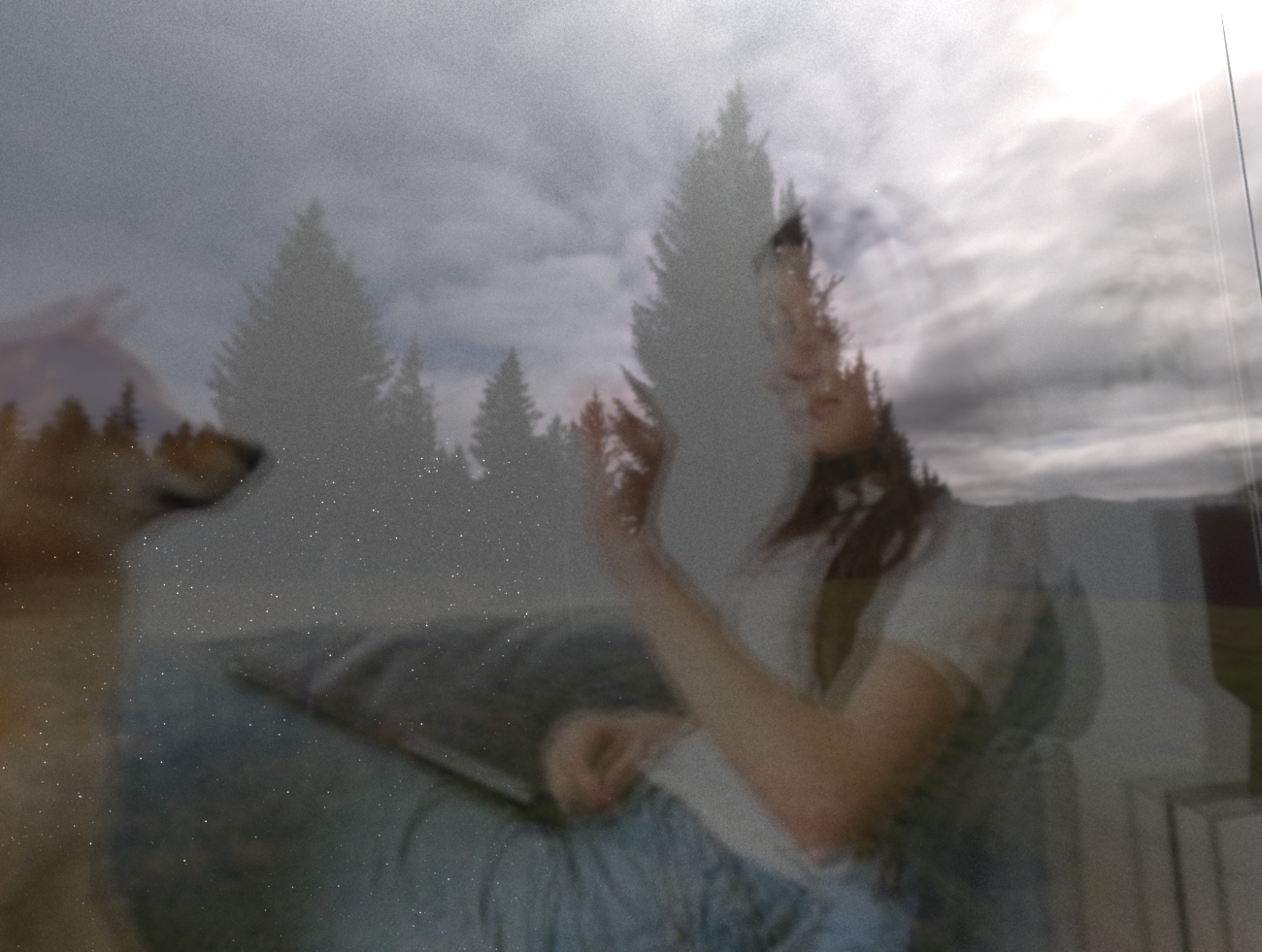} &
        \includegraphics[width=\resLen]{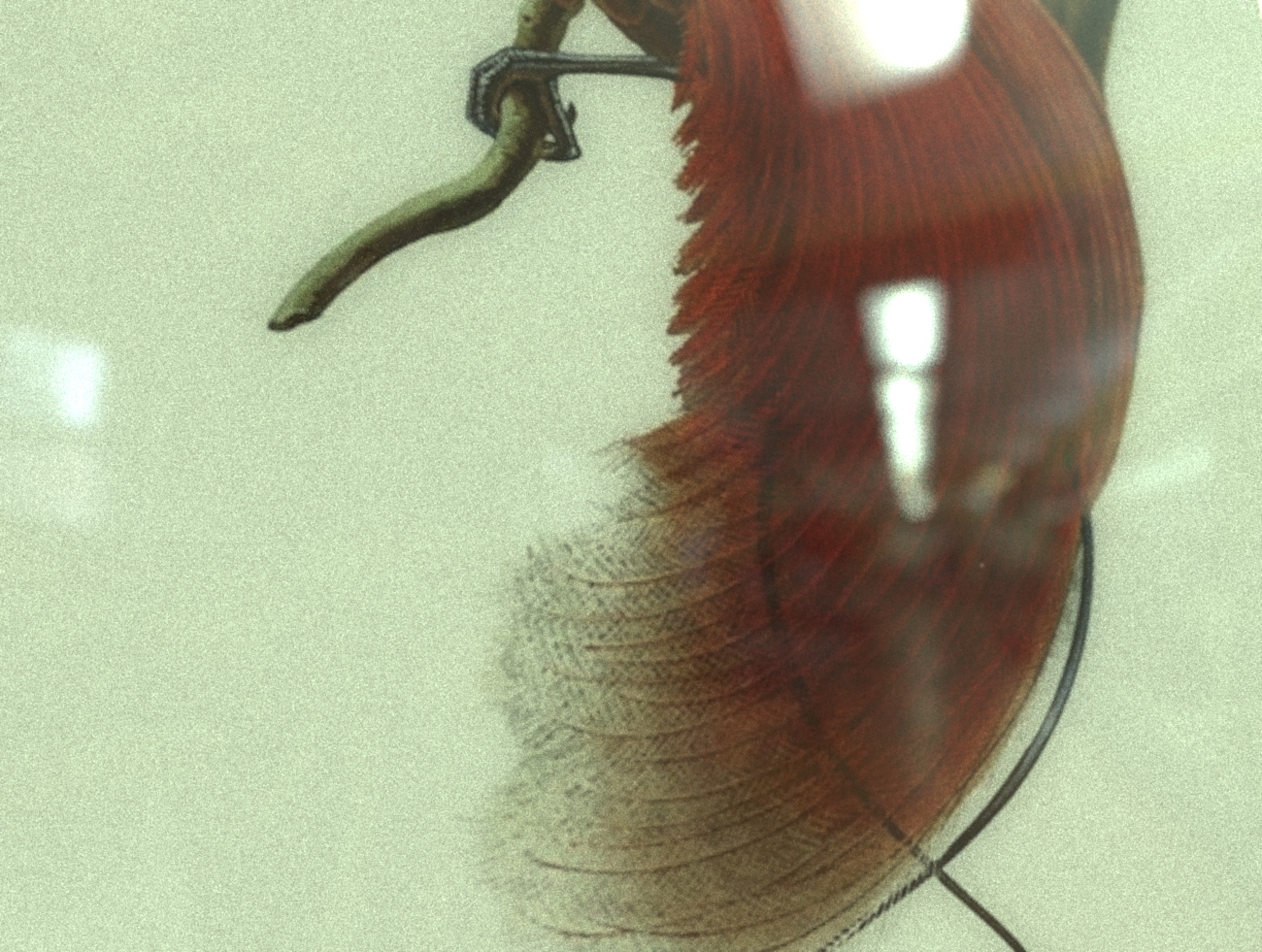}
        \\
        (a) HDR|HDR & (b) HDR|LDR & (c) LDR|HDR
    \end{tabular}
    \vspace{-8pt}
    \caption{
        \textbf{Three different scenes setup.} 
        (a) Use HDR environment maps for both transmission and reflection. 
        (b) Use HDR map for transmission and put a LDR image behind camera for reflection.
        (c) Use HDR map for reflection and put a LDR image behind the glass for transmission. 
    }
    \vspace{-10pt}
    \label{fig:scene_setup}
\end{figure}

Figure \ref{fig:data_example} illustrates five rendered images and the illustration of their corresponding light paths.
\begin{itemize}
\item \I: Input image with full glass effects. 
\item \T: Transmission of the rendered glass with reflection disabled, showing the background. 
\item \B: Background with the glass removed. 
\item \R: Reflection part of the rendered glass with transmission disabled, displaying a dim scene reflection. 
\item \MR: Mirror reflection, where changing the glass to a perfect mirror results in a sharp reflected scene. 
\end{itemize}
The \I~contains both \T~($T_1=\PT\PT$) and \R~($R_1=\PR$), as well as the higher-order ($\PT\PR^k\PT$) light interaction within the glass slab. 
\T~has typically been regarded as an estimate of \B, until \cite{lei2020polarized, zheng2021single} provided a detailed discourse. Compared to \B, \T~is adding refraction shift and energy loss, which makes it look darker. Moreover, \cite{kim2020single, wan2020reflection} explicitly discusses the differences between \R~and \MR. In short, \MR~has the same light path ($\PR$) as \R, but with no energy loss. Our data generation pipeline is capable of producing all of these effects. \textbf{For the single-image reflection removal task, only \I, \T, and \R~are used.} However, the generated images retain broader potential, for instance, the pairs of \R~and \MR~can be applied to image enhancement tasks.

In rendering images with glass effects, simply having the glass is insufficient; a complete 3D scene is necessary both in front of the glass (for reflections) and behind it (for transmissions). Creating such a 3D scene from scratch is complex and difficult to modify. As an alternative, \cite{kim2020single} proposes estimating the depth from an RGB image to input into the renderer. However, this approach has a significant limitation: it cannot simulate light sources, which are crucial for tasks involving reflection removal. Consequently, highlights become overexposed and cannot be effectively simulated using LDR images. To address this, we used HDR environment maps to model the scene on both sides of the glass, as shown in Figure \ref{fig:scene_setup}-(a).

To further enrich the dataset with scenarios that involve close proximity to glass, we introduce two additional scene configurations: I) since environment maps typically exclude humans or animals, we provide an alternative setup (Figure \ref{fig:scene_setup}-(b)) that captures greater reflection variability, such as the frequent appearance of the photographer in the glass; II) Considering cases of photographing artwork behind protective glass, where reflections are concentrated within the frame while the artwork itself remains largely unaffected. As illustrated in Figure \ref{fig:scene_setup}-(c), a scenario common in real-world settings such as museums and galleries, where reflective regions are ambiguous and can easily be mistaken for artwork content.

\begin{figure}[t]
    \centering
    \setlength{\resLen}{0.22\linewidth}
    \addtolength{\tabcolsep}{-5pt}
    \begin{tabular}{cc@{\hspace{5pt}}cc}
        \includegraphics[height=\resLen]{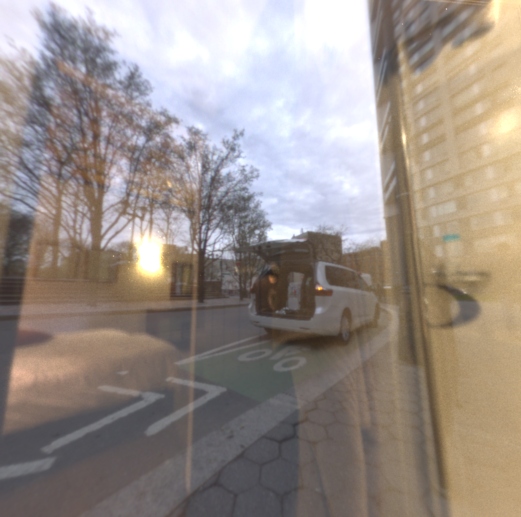} &
        \includegraphics[height=\resLen]{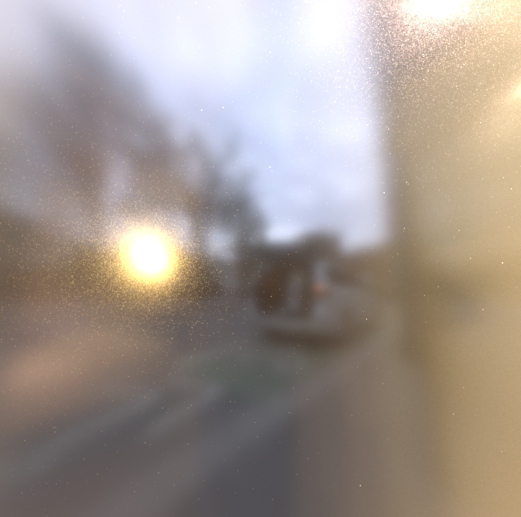} &
        \includegraphics[height=\resLen]{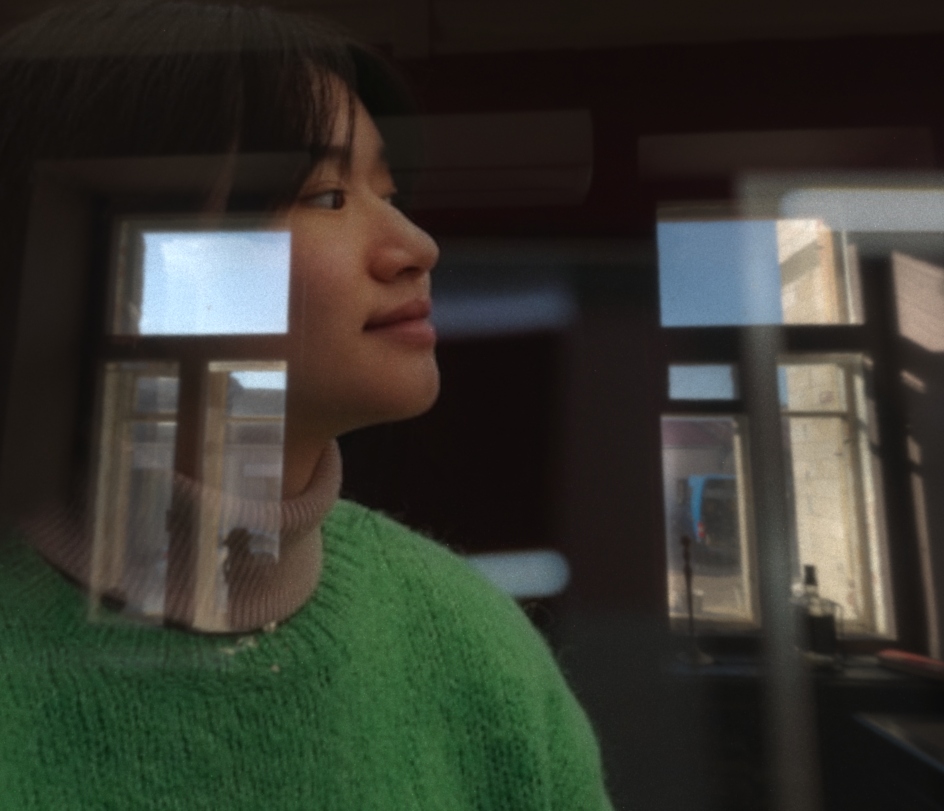} &
        \includegraphics[height=\resLen]{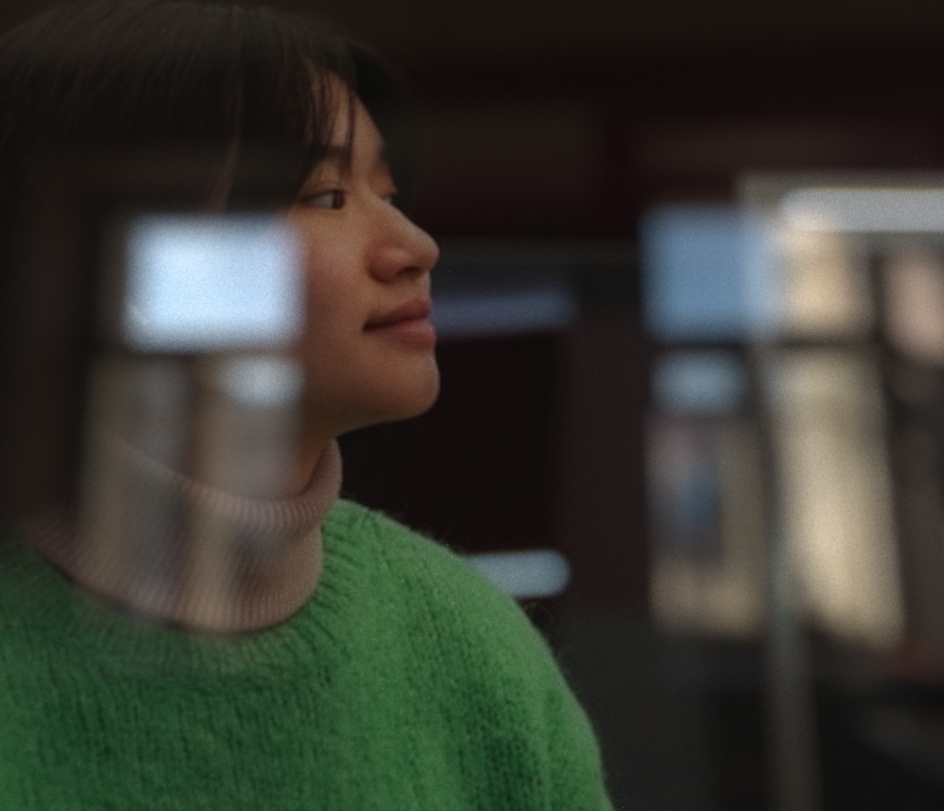}
        \\
        (a) & (b) & (c) & (d)
    \end{tabular}
    \vspace{-8pt}
    \caption{
        \textbf{Blurry effects.}
        (a) and (b): different roughness of the glass surface.
        (c) and (d): different aperture sizes of the camera.
    }
    \vspace{-10pt}
    \label{fig:other_effects}
\end{figure}

\subsection{Scene parameter variation}
\begin{figure}[t]
    \centering
    \setlength{\resLen}{0.24\linewidth}
    \addtolength{\tabcolsep}{-5pt}
    \hskip -10pt
    \begin{tabular}{cc@{\hspace{10pt}}cc}
        \includegraphics[height=\resLen]{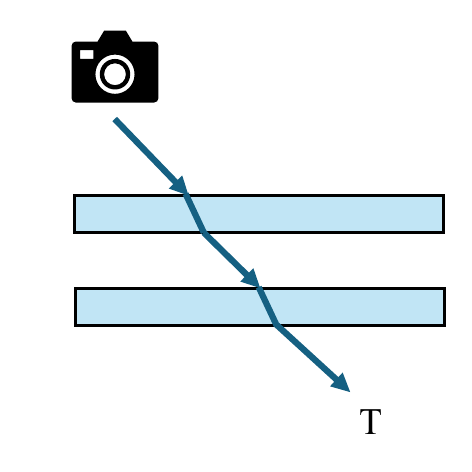} &
        \includegraphics[height=\resLen]{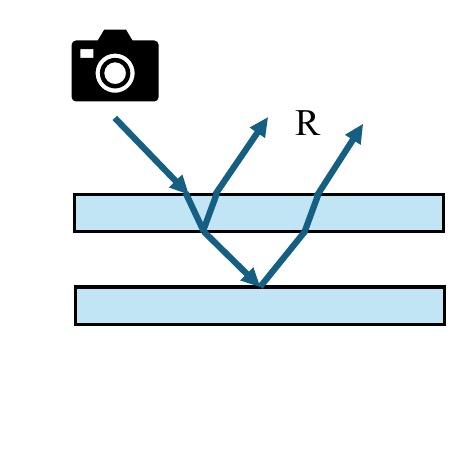} &
        \includegraphics[height=\resLen]{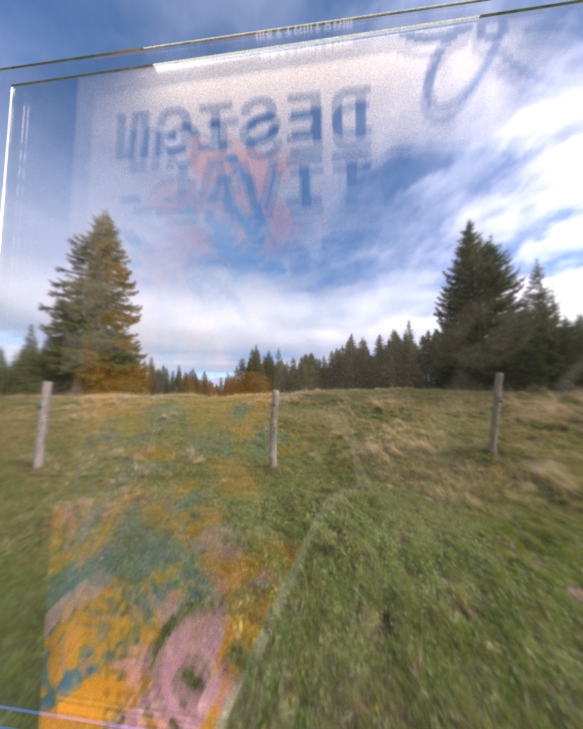} &
        \includegraphics[height=\resLen]{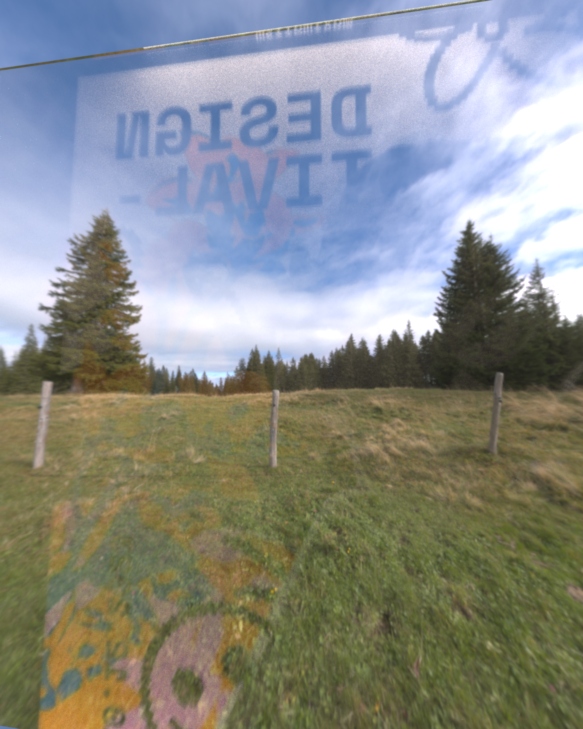}
        \\
        (a) & (b) & (c) & (d)
    \end{tabular}
    \vspace{-8pt}
    \caption{
        \textbf{Double reflection illustration.} 
        (a) Transmission: turn off the reflection for both glass layers. 
        (b) Reflection: only turn off the transmission for the bottom glass. 
        (c) An example of double reflection.
        (d) A compared single reflection.
    }
    \vspace{-10pt}
    \label{fig:double_reflect}
\end{figure}

To accurately represent the variety scenes in real-world applications, we parameterize several properties for glass. The thickness (residential windows could be 3 to 6 mm and structural/safety glass could be as thick as 10 to 40 mm), the color of the glass (from clear to diverse tints and color temperatures), the index of refraction (IOR) (from 1.45 to 1.65 to reflect different glass types), and the roughness of the surface (for the purpose of privacy, decoration, or light diffusion), see Figure \ref{fig:other_effects}-(a)(b). We also introduce double-layer glass (for thermal/sound insulation) by adding an additional glass layer and altering the interlayer distance. Double layers could generate the double reflection effect and also induce reflection, see Figure \ref{fig:double_reflect}. 

To ensure the dataset captures realistic imaging scenarios, we systematically vary camera parameters. Camera position is adjusted to provide multiple viewpoints relative to the glass surfaces, while the field of view is changed from wide-angle to telephoto perspectives. Focal distance and aperture size are also varied to produce different depth of field effects and defocus characteristics, further enhancing the diversity of the dataset, see Figure \ref{fig:other_effects}-(c)(d). Path-traced rendering is employed to guarantee physically accurate light transport. We use a sufficient number of ray bounces to capture complex light interactions, and adaptive sampling strategies are implemented for efficient rendering. The material models for glass surfaces are defined using physically-based BSDF to ensure that the rendered images closely match real-world observations.

To simulate real-world imaging conditions, we apply several post-processing effects to the rendered images. Gamma correction is performed using the standard sRGB curve, and auto white balance is applied to adjust the color temperature automatically. We also introduce JPEG compression artifacts to mimic the imperfections commonly found in digital photographs.

\section{Text-to-image Model Fine-tuning}

Most of current text-to-image models, such as \flux~model, could generate images conditioned jointly on a text prompt and a reference image. The conditional distribution is calculated as 
\vspace{-5pt}
\begin{equation}
    p (x \,|\, y, c)
    \vspace{-5pt}
\end{equation}
where $x$ is target image, $y$ is a condition image or guiding image, and $c$ is a natural-language prompt.

\begin{figure}[t]
    \centering
    \setlength{\resLen}{\linewidth}
    \includegraphics[width=\resLen]{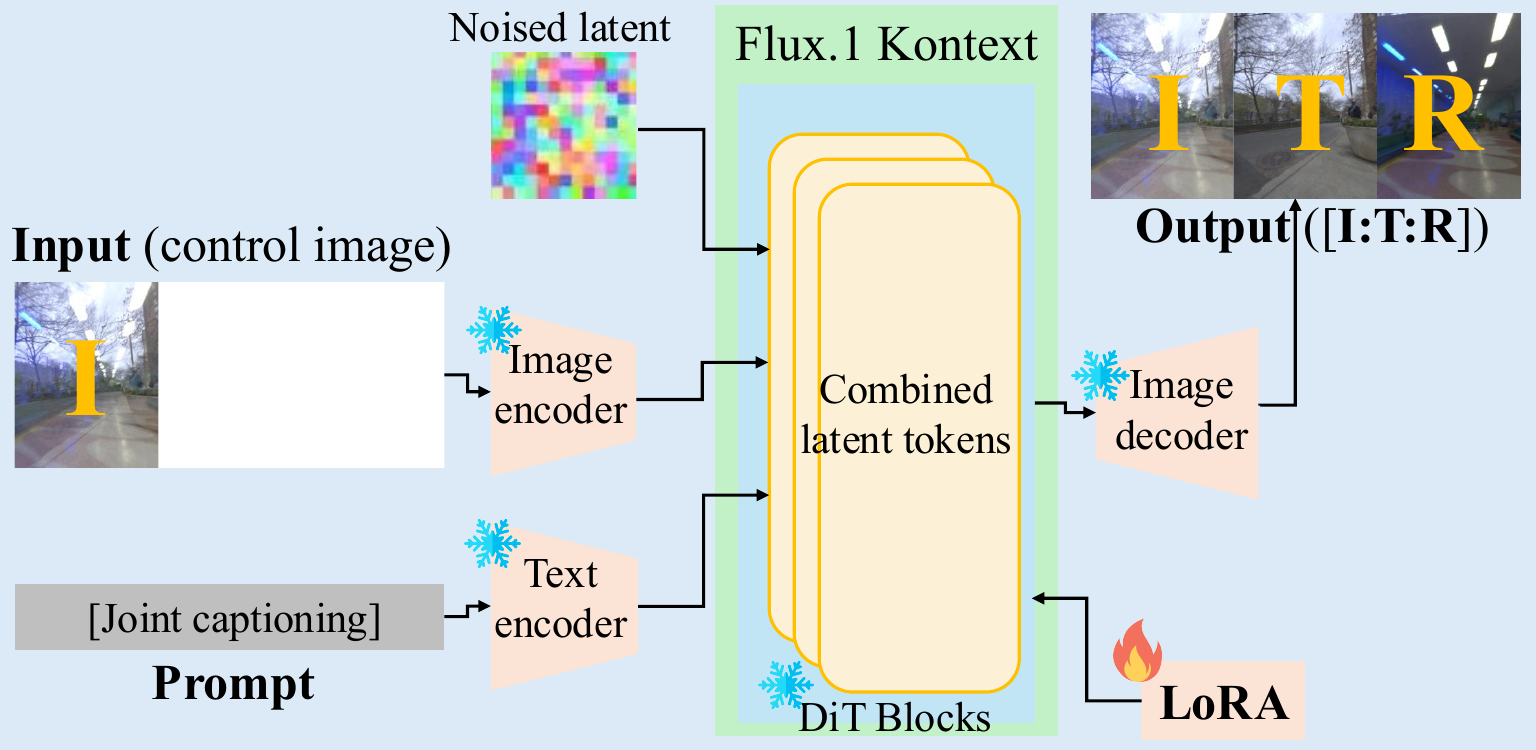}     
    \vspace{-20pt}
    \caption{
        \textbf{Fine-tuning pipeline.} A overview of LoRA fine-tuning \flux~with consolidated image pairs ([\I:\T:\R]) and consolidated prompts. 
    }
    \vspace{-10pt}
    \label{fig:pipeline}
\end{figure}

Following the idea of In-Context LoRA \cite{huang2024context}, we generate the transmission and reflection components of a glare image simultaneously by directly concatenating them into a single large image [\I:\T:\R] during training (see \textbf{Output} in Figure \ref{fig:pipeline}), while consolidating their captions into a merged prompt with a high-level description for each panel. See PROMPT we used in Section \ref{sec:exp}. 

The single consolidated prompt for the entire image set begins with an overall description, followed by individual prompts for each image. 
The correlations between the input image (\I) and the decomposed images (\T~and \R) are implicitly maintained through the consolidated prompt. This unified prompt design is more compatible with existing text-to-image models and allows the overall description to naturally convey the task.

Instead of providing different image sets with different prompts, we embed the prompt into the model as a feature specific to the task. In this way, the prompt does not need to be considered during inference, allowing our method to still be considered as a single-image reflection removal/separation. At the same time, we preserve the capability for in-context text-to-image generation to better handle the complex intrinsic relationships between each layer.

We keep the image \I~both in both generated image and in the conditioning image to further enhance the correlation of the generated and conditioned images. We also reformulate this task to an image inpainting task by adding white space to the conditioning image (see \textbf{Input} in Figure \ref{fig:pipeline}), which is not necessary, but it turns out easier to converge.

\begin{figure*}[t]
    \centering
    \setlength{\resLen}{0.12\linewidth}
    \addtolength{\tabcolsep}{-5pt}
    \renewcommand{\arraystretch}{0.8}
    \hskip 0pt
    \begin{tabular}{cccccccc}
        \begin{overpic}[width=\resLen]{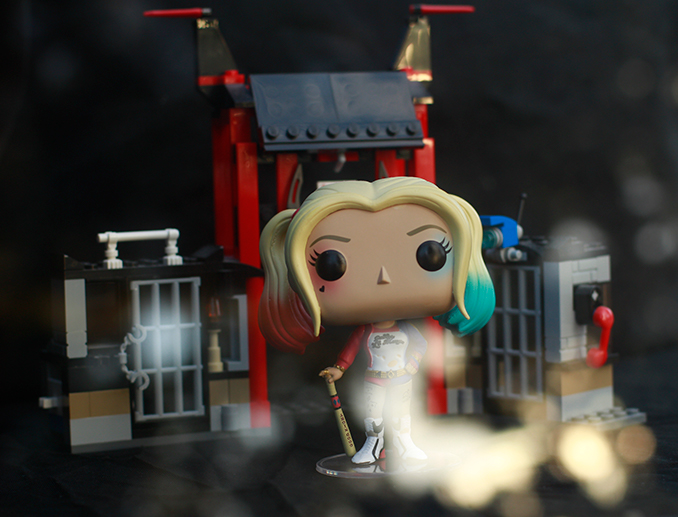}
            \put(0,3){\scriptsize \colorbox{white}{\color{black} (a)}}
        \end{overpic} &
        \includegraphics[width=\resLen]{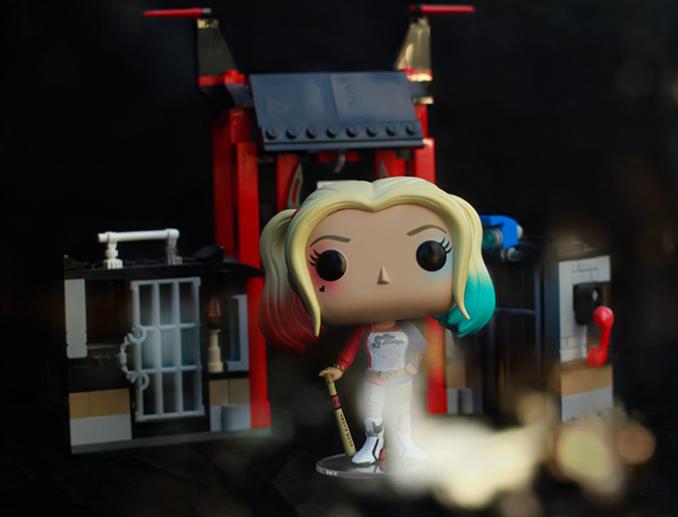} &
        \includegraphics[width=\resLen]{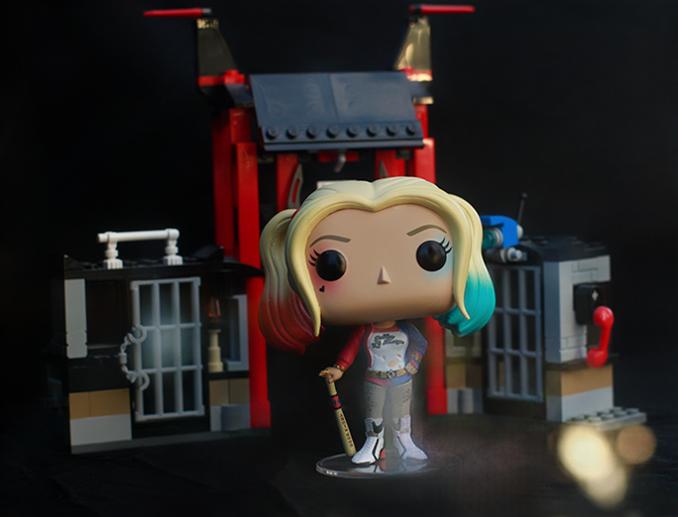} &
        \includegraphics[width=\resLen]{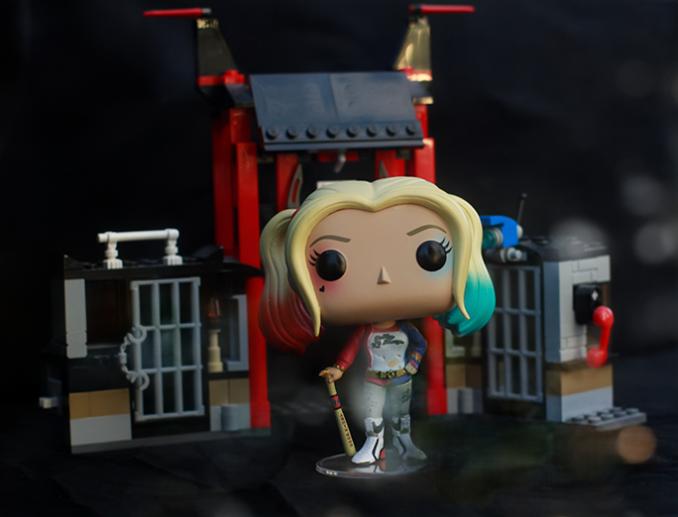} &
        \includegraphics[width=\resLen]{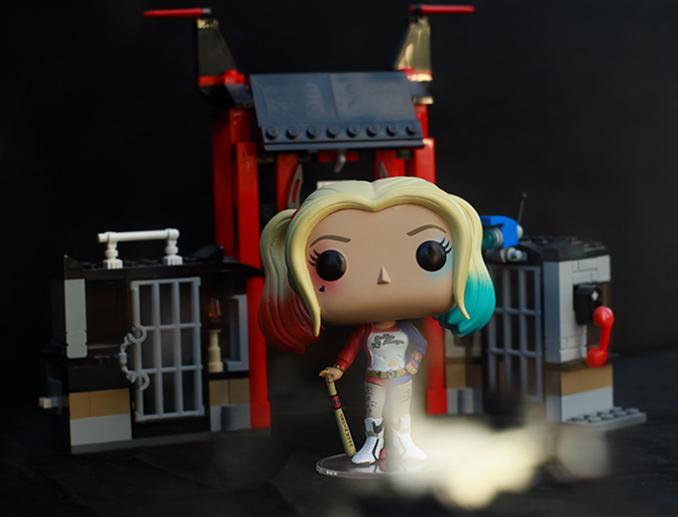} &
        \includegraphics[width=\resLen]{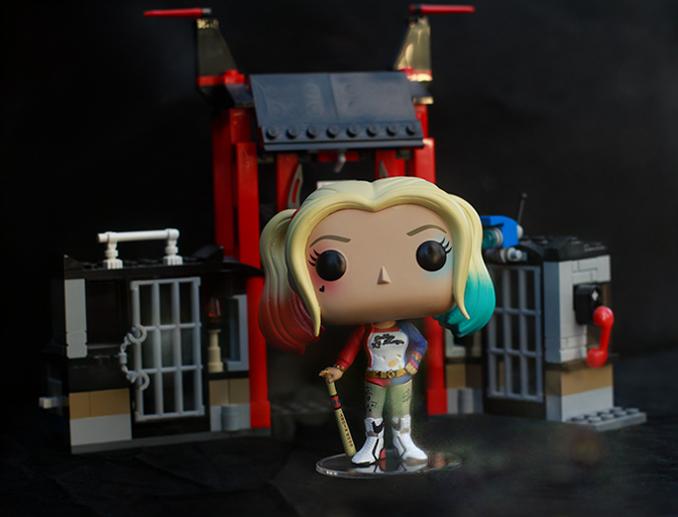} &
        \includegraphics[width=\resLen]{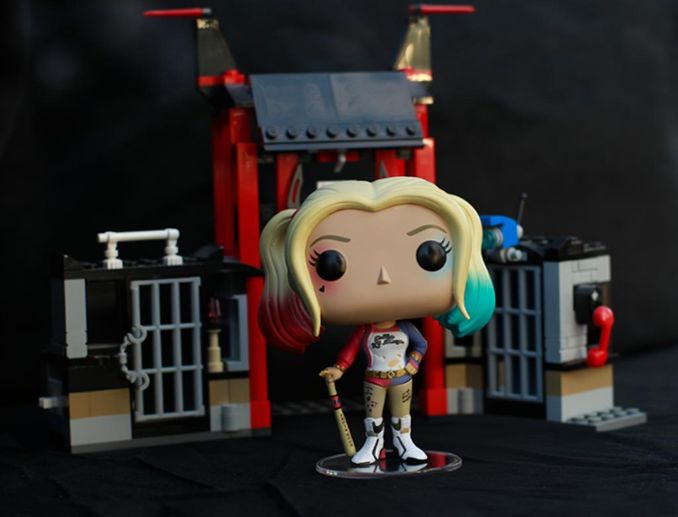} &
        \includegraphics[width=\resLen]{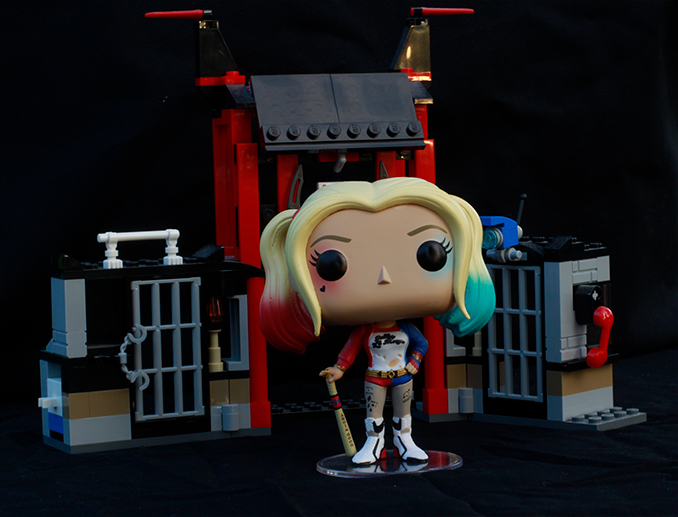} \\
        
        \begin{overpic}[width=\resLen]{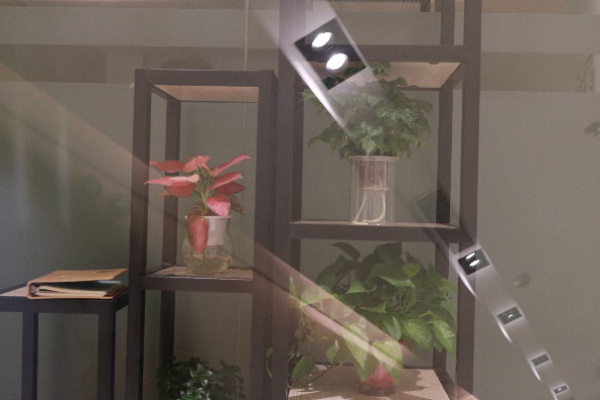}
            \put(0,3){\scriptsize \colorbox{white}{\color{black} (b)}}    
        \end{overpic} &
        \includegraphics[width=\resLen]{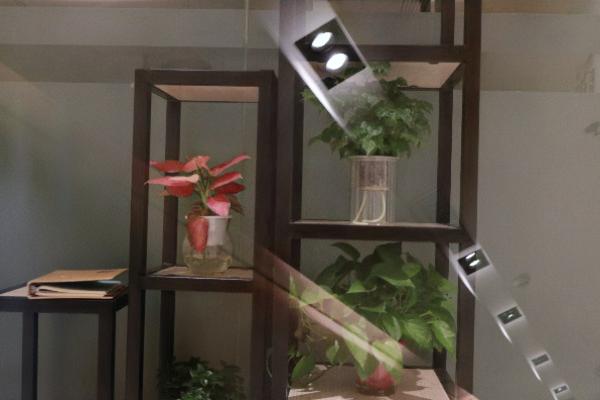} &
        \includegraphics[width=\resLen]{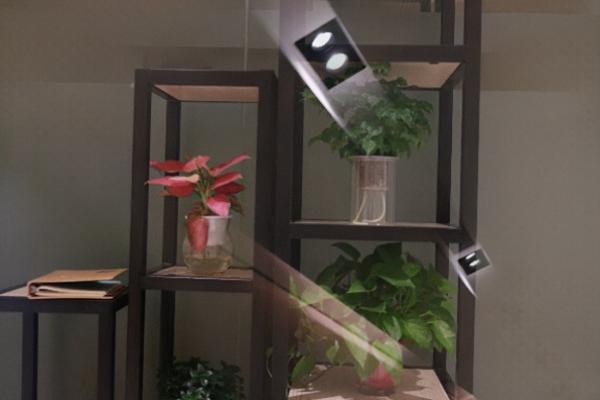} &
        \includegraphics[width=\resLen]{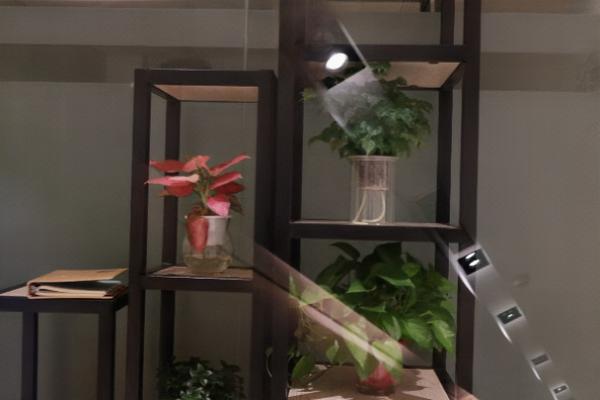} &
        \includegraphics[width=\resLen]{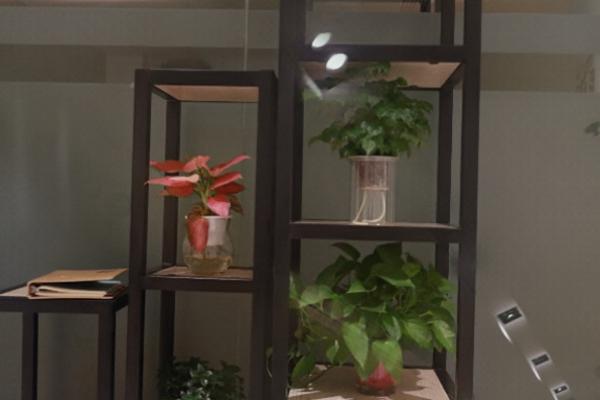} &
        \includegraphics[width=\resLen]{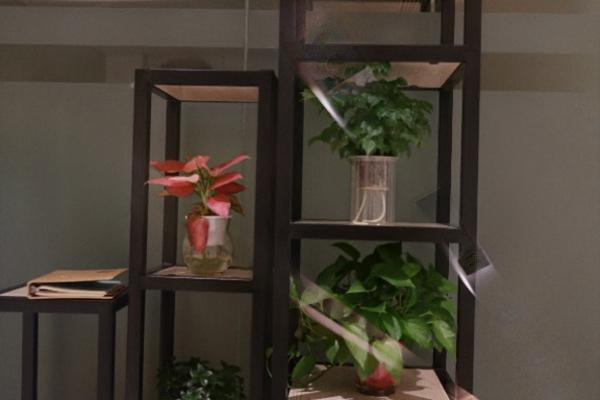} &
        \includegraphics[width=\resLen]{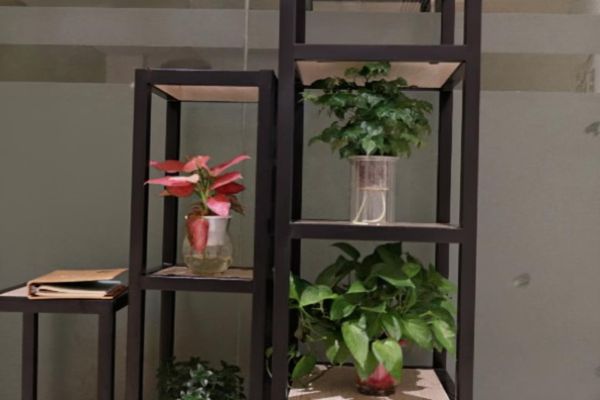} &
        \includegraphics[width=\resLen]{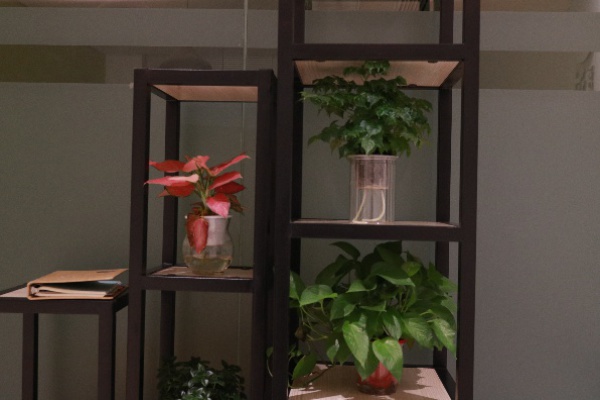} \\
        
        \begin{overpic}[width=\resLen]{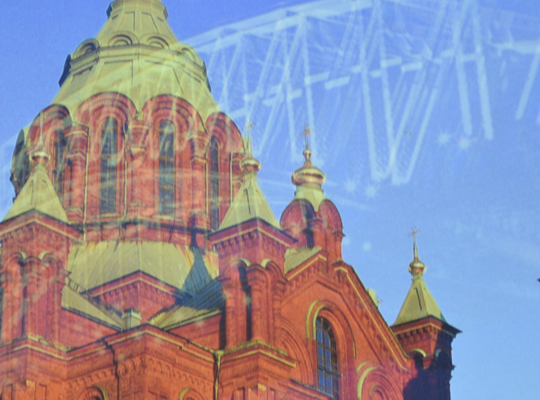}
            \put(0,3){\scriptsize \colorbox{white}{\color{black} (c)}}
        \end{overpic} &
        \includegraphics[width=\resLen]{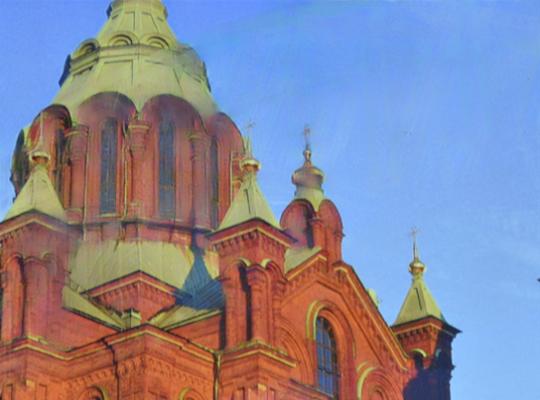} &
        \includegraphics[width=\resLen]{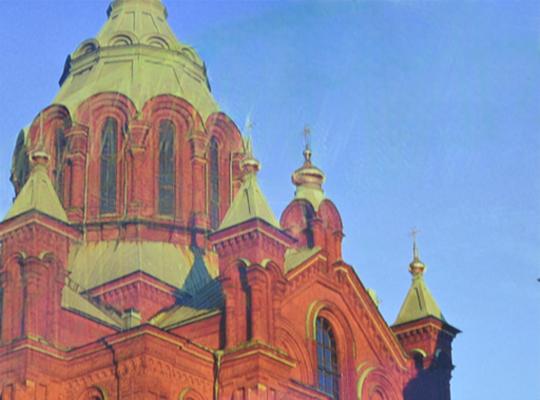} &
        \includegraphics[width=\resLen]{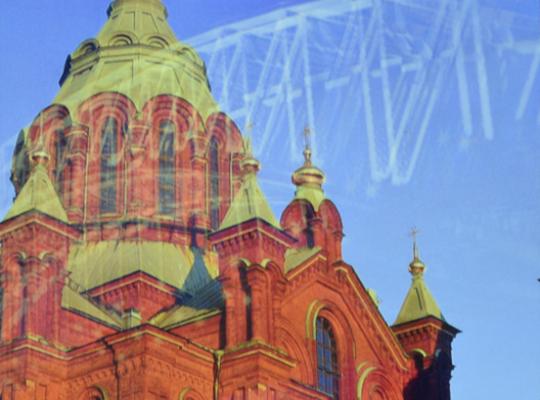} &
        \includegraphics[width=\resLen]{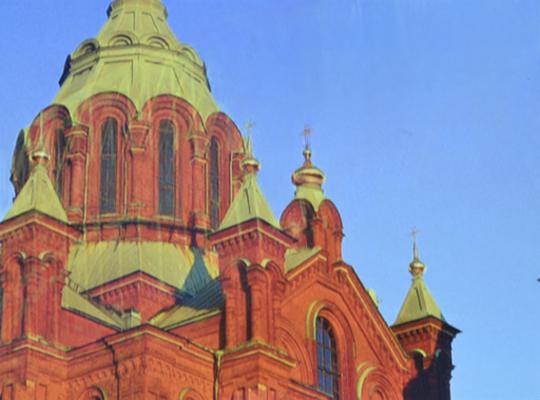} &
        \includegraphics[width=\resLen]{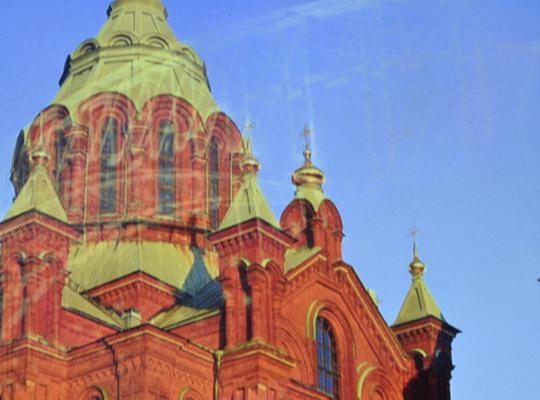} &
        \includegraphics[width=\resLen]{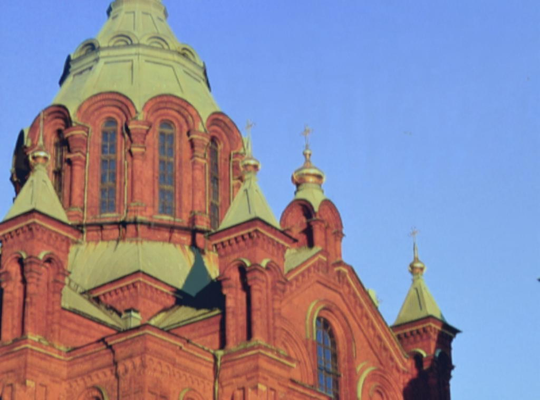} &
        \includegraphics[width=\resLen]{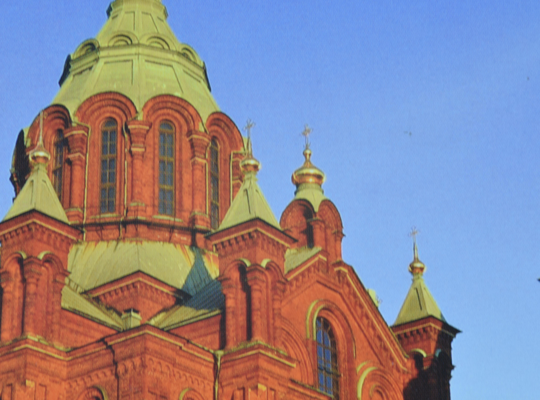} \\
        
        \begin{overpic}[width=\resLen]{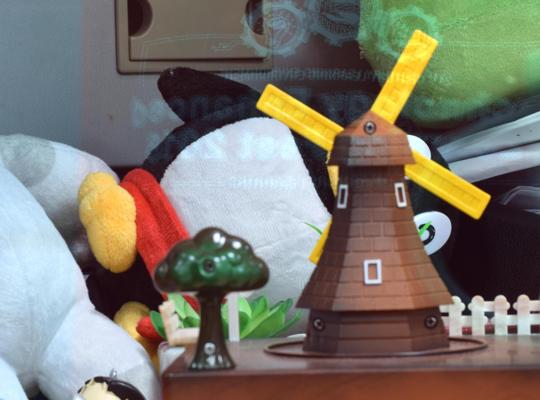}
            \put(0,3){\scriptsize \colorbox{white}{\color{black} (d)}}
        \end{overpic} &
        \includegraphics[width=\resLen]{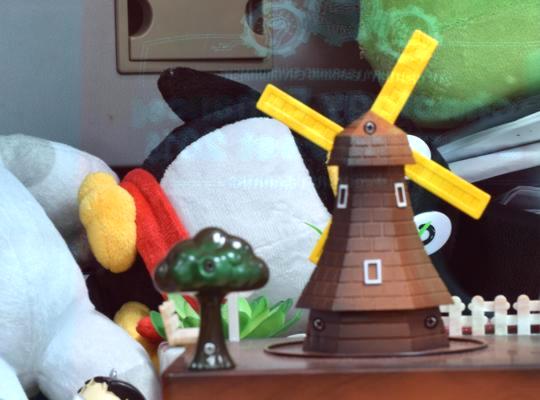} &
        \includegraphics[width=\resLen]{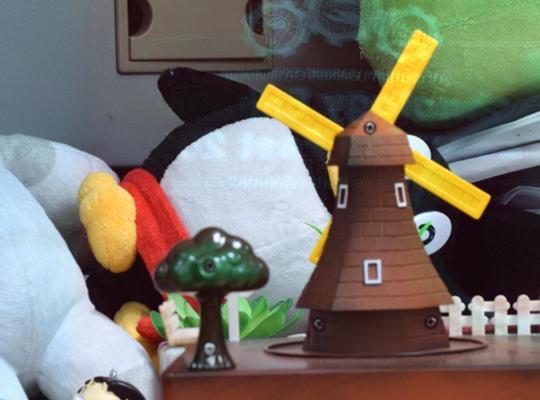} &
        \includegraphics[width=\resLen]{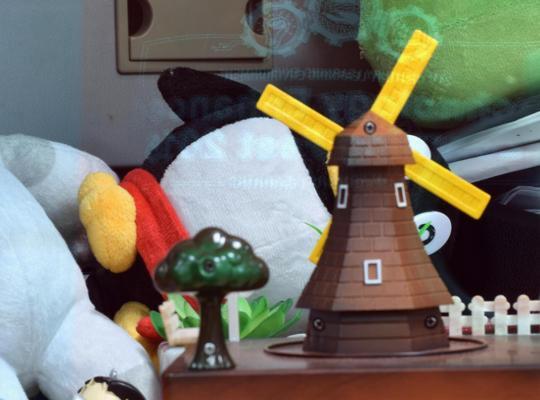} &
        \includegraphics[width=\resLen]{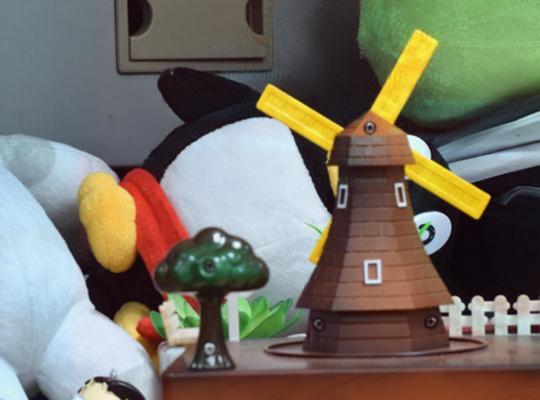} &
        \includegraphics[width=\resLen]{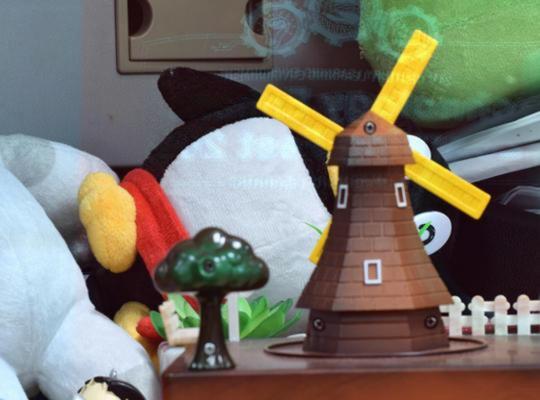} &
        \includegraphics[width=\resLen]{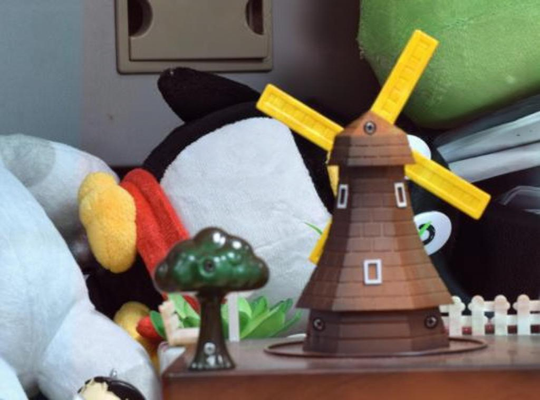} &
        \includegraphics[width=\resLen]{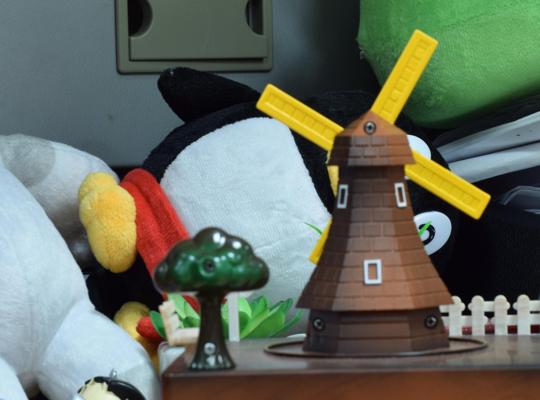} \\
        
        \begin{overpic}[width=\resLen]{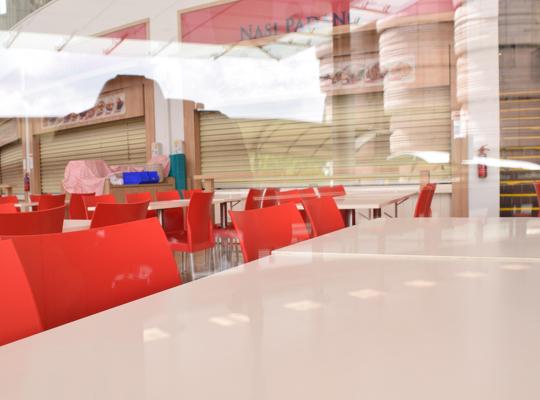} 
            \put(0,3){\scriptsize \colorbox{white}{\color{black} (e)}}
        \end{overpic} &
        \includegraphics[width=\resLen]{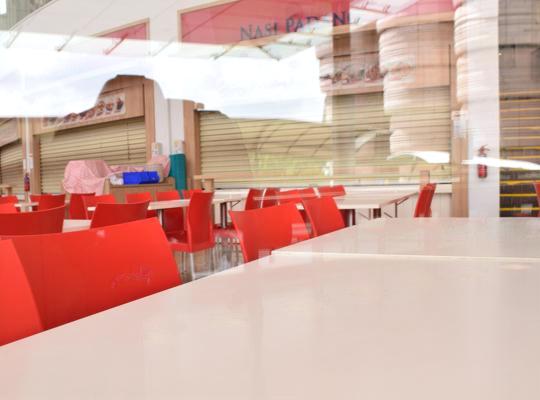} &
        \includegraphics[width=\resLen]{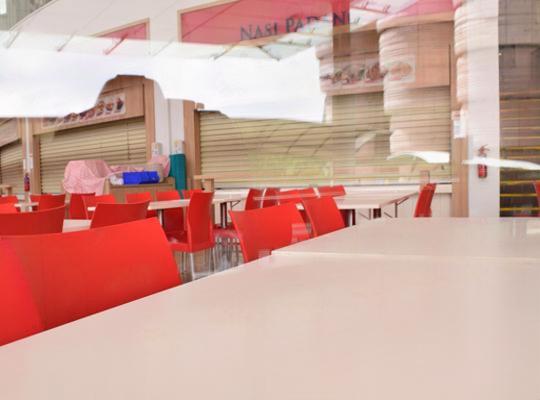} &
        \includegraphics[width=\resLen]{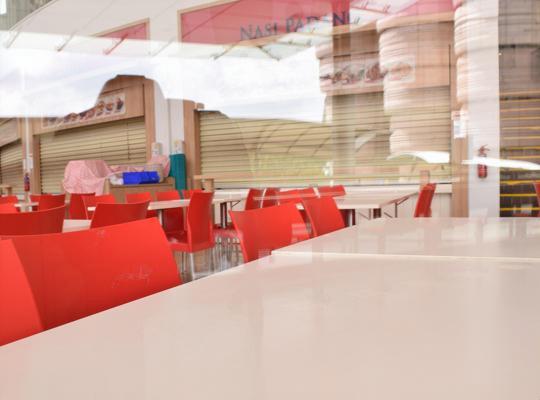} &
        \includegraphics[width=\resLen]{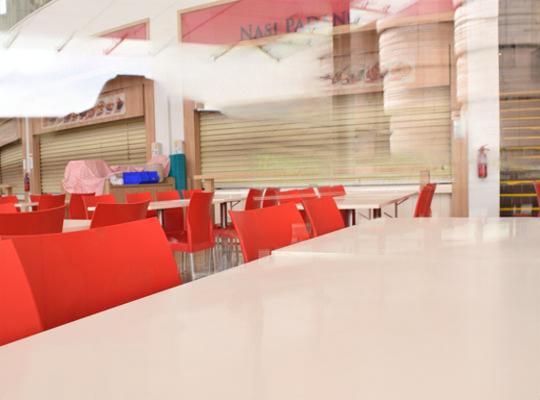} &
        \includegraphics[width=\resLen]{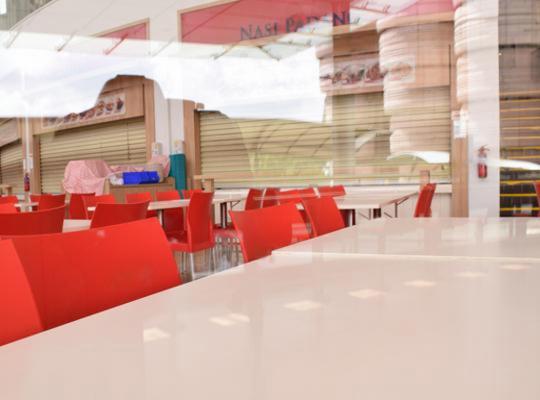} &
        \includegraphics[width=\resLen]{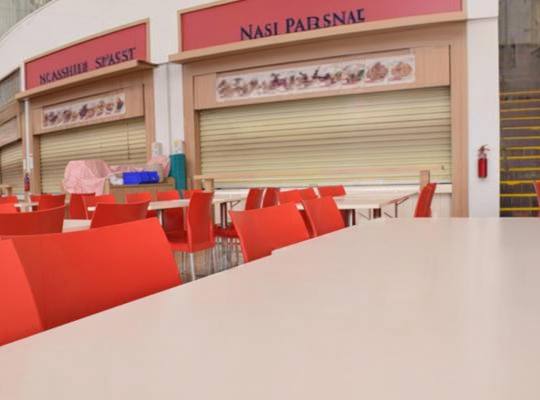} &
        \includegraphics[width=\resLen]{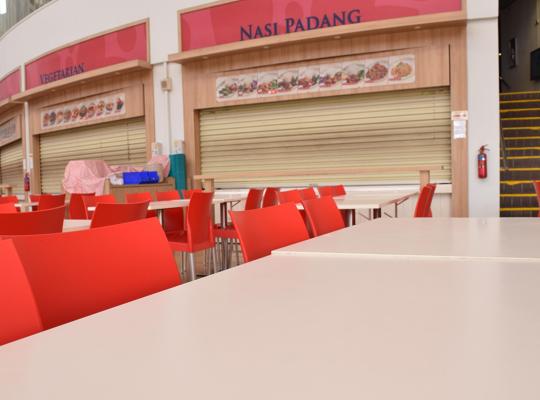} \\
        
        Input 
        & Dong \cite{dong2021location} 
        & DSRNet \cite{hu2023single} 
        & Zhu \cite{zhu2024revisiting} 
        & DSIT \cite{hu2024single}
        & RDNet \cite{zhao2025reversible}
        & Ours & GT
    \end{tabular}
    \vspace{-8pt}
    \caption{
        \textbf{Visual comparison of single-image reflection removal results.} The 5 examples are picking from \textit{Real}, \textit{Nature}, \textit{SIR}$^2$-\textit{Postcard}, \textit{SIR}$^2$-\textit{SolidObject} and \textit{SIR}$^2$-\textit{Wild}. In general, our method could generate clean image and has a better prediction in the overexposed region.
    }
    \label{fig:main_t}
\end{figure*}
\begin{figure*}[t]
    \centering
    \setlength{\resLen}{0.137\linewidth}
    \addtolength{\tabcolsep}{-5pt}
    \renewcommand{\arraystretch}{0.8}
    \begin{tabular}{ccccccc}
        \includegraphics[width=\resLen]{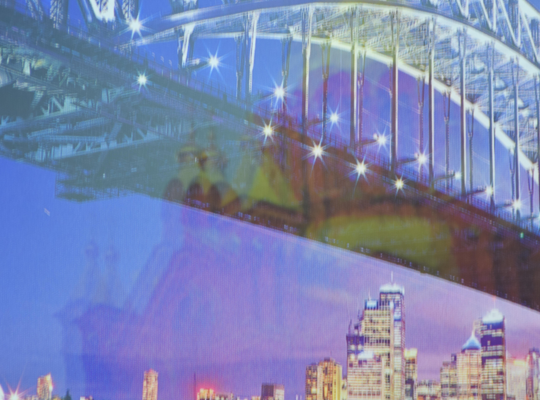} &
        \includegraphics[width=\resLen]{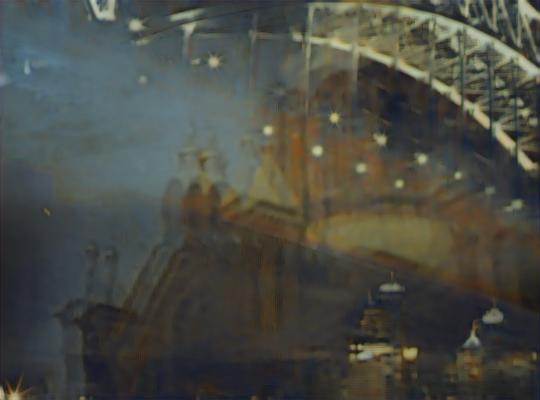} &
        \includegraphics[width=\resLen]{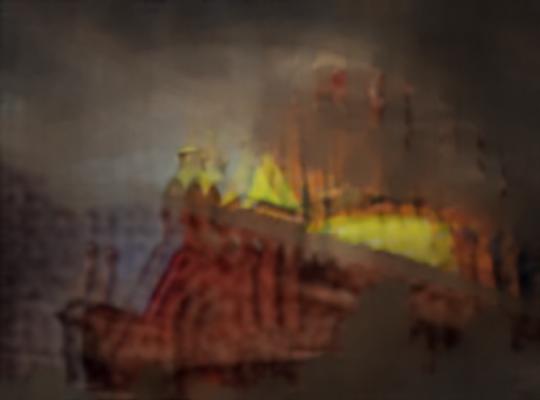} &
        \includegraphics[width=\resLen]{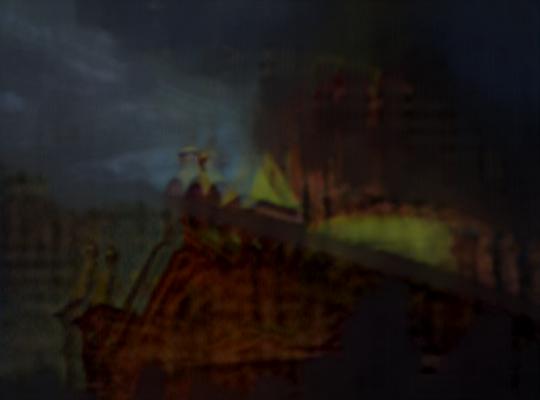} &
        \includegraphics[width=\resLen]{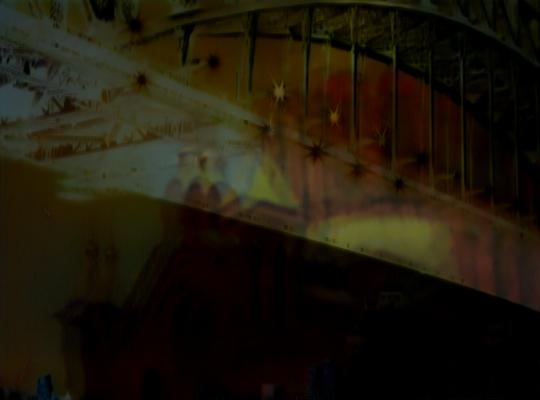} &
        \includegraphics[width=\resLen]{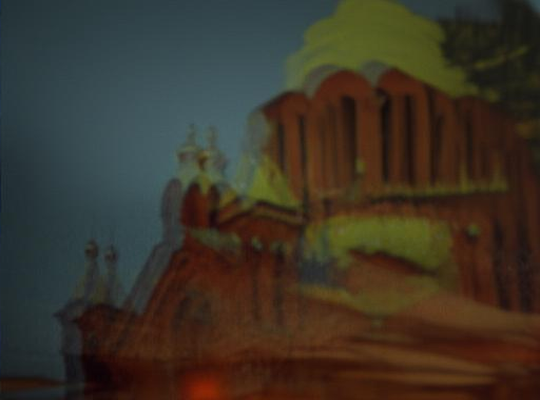} &
        \includegraphics[width=\resLen]{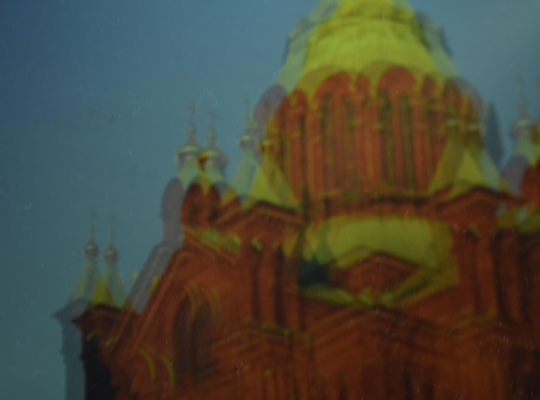}
        \\
        \includegraphics[width=\resLen]{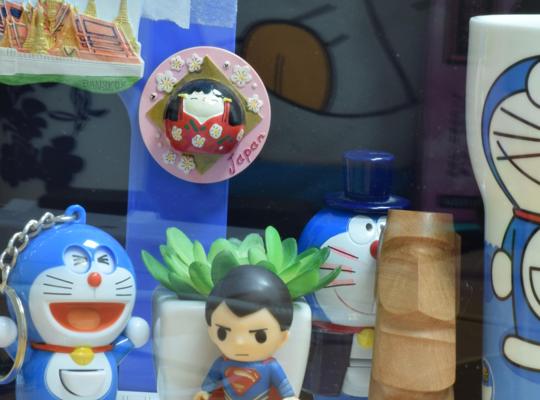} &
        \includegraphics[width=\resLen]{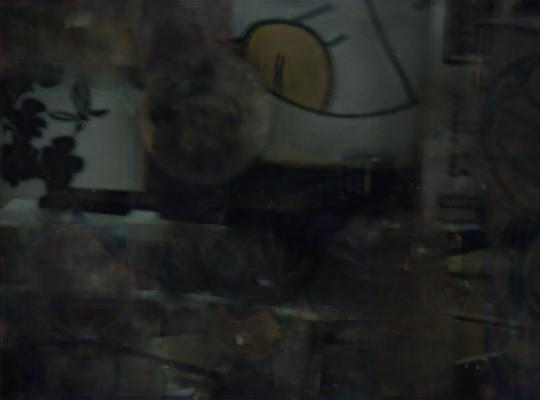} &
        \includegraphics[width=\resLen]{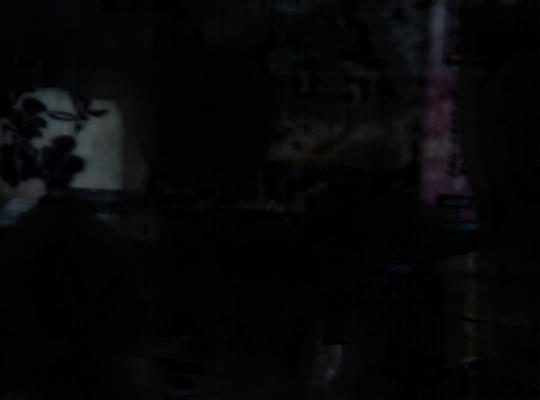} &
        \includegraphics[width=\resLen]{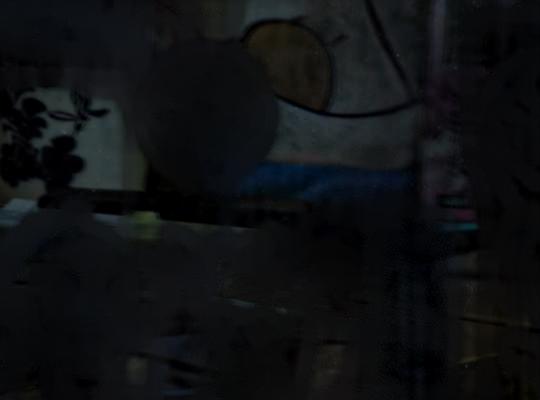} &
        \includegraphics[width=\resLen]{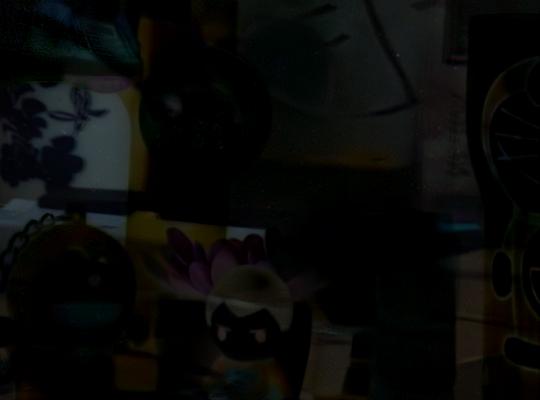} &
        \includegraphics[width=\resLen]{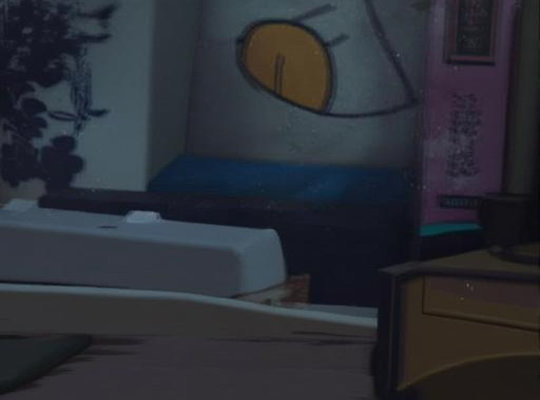} &
        \includegraphics[width=\resLen]{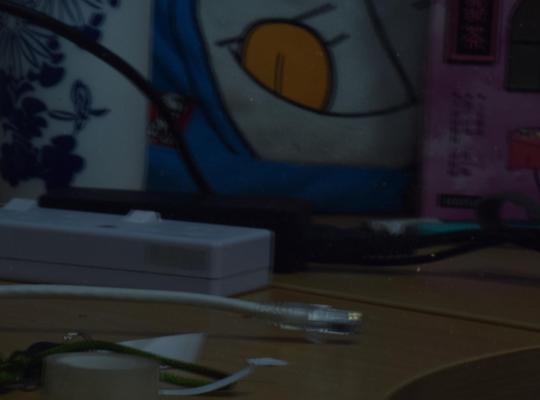}
        \\
        \includegraphics[width=\resLen]{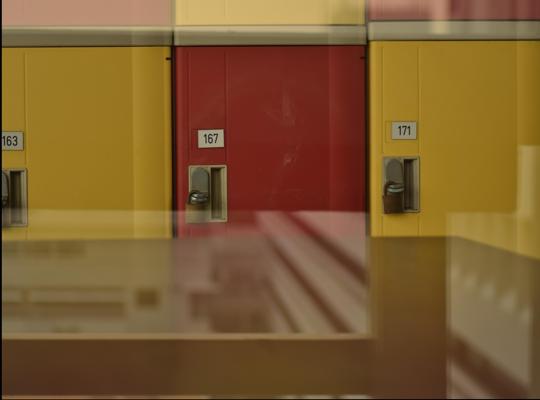} &
        \includegraphics[width=\resLen]{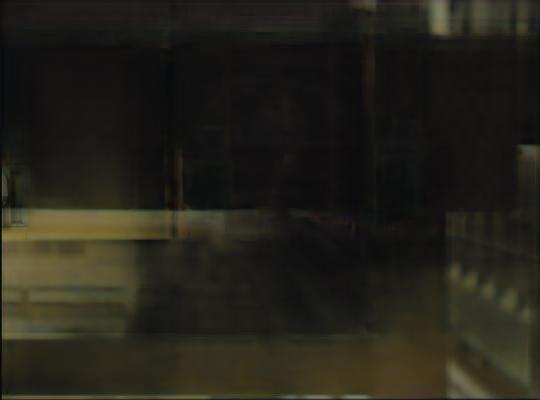} &
        \includegraphics[width=\resLen]{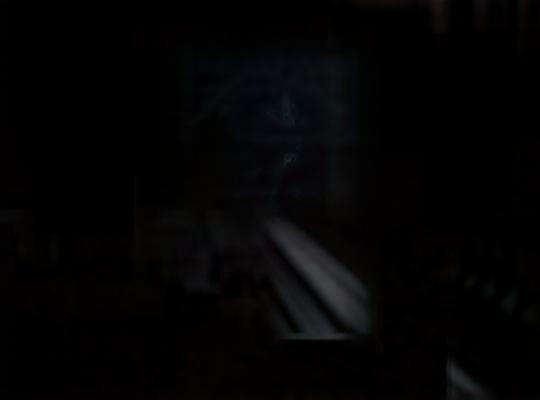} &
        \includegraphics[width=\resLen]{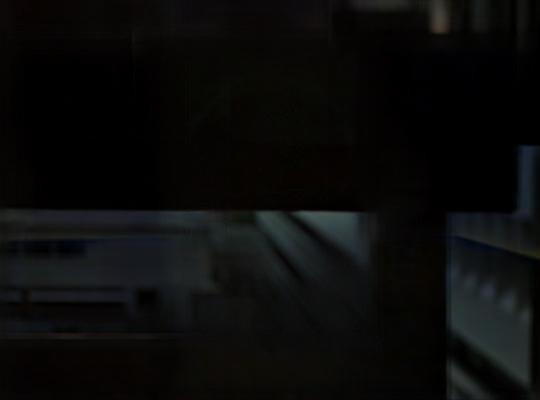} &
        \includegraphics[width=\resLen]{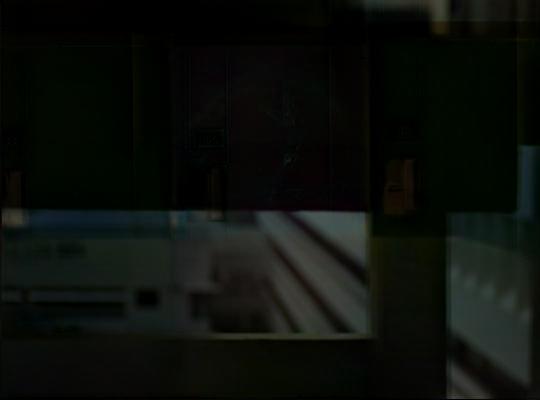} &
        \includegraphics[width=\resLen]{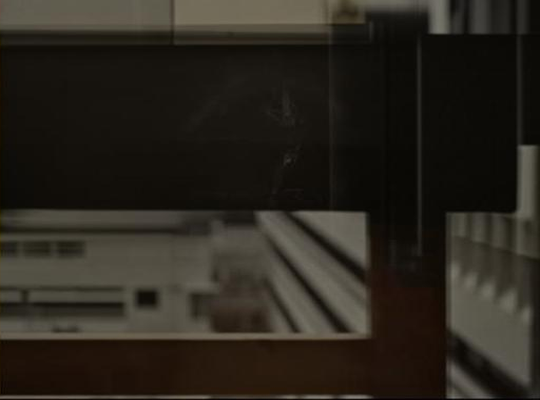} &
        \includegraphics[width=\resLen]{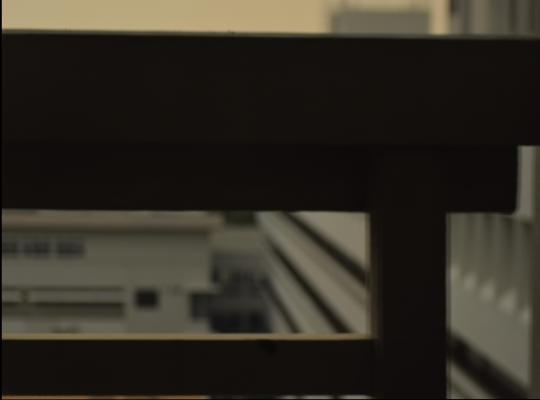}
        \\
        Input 
        & Dong \cite{dong2021location} 
        & DSRNet \cite{hu2023single}
        & DSIT \cite{hu2024single}
        & RDNet \cite{zhao2025reversible}
        & Ours & GT
    \end{tabular}
    \vspace{-8pt}
    \caption{
        \textbf{Visual comparison of single-image reflection separation results.} The 3 examples are picking from \textit{SIR}$^2$-\textit{Postcard}, \textit{SIR}$^2$-\textit{SolidObject} and \textit{SIR}$^2$-\textit{Wild}. Ours is the only method which could generate clean and meaningful reflection image.
    }
    \label{fig:main_r}
\end{figure*}

\section{Experimental Setup}
\label{sec:exp}

We provide a Python script to render the synthetic images. Each image set includes 5 JPEG images (see examples in Figure \ref{fig:data_preview} from Appendix): the original rendered image with reflections, a ground truth transmission layer, a ground truth background image, a ground truth reflection layer, a ground truth mirror reflection image, and metadata describing the glass properties and camera parameters used in the rendering process. Original high dynamic linear images (.exr) are also provided. For HDR environment maps, we chose 140 indoor scenes and 30 outdoor nature scenes from Polyhaven\footnote{https://polyhaven.com/hdris}, and 110 outdoor city scenes from UrbanSky\footnote{https://cave.cs.columbia.edu/repository/UrbanSky}. 200 high resolution LDR images are from the Internet
with keyword searches, such as "human/pets" or "artwork".

With 1,000 image pairs rendered with our script, we build our approach on the \flux~model and train an In-Context LoRA specifically for the task. \textbf{T} and \textbf{R} are concatenated into a single composite image, \textbf{I} is used as control image, as we illustrated in \ref{fig:pipeline}. And captions for these images are fixed (see below PROMPT) during both training and inference, which makes it an internal parameter and does not offset the \textit{single} image reflection removal rule.  

\begin{tcolorbox}[left=1pt,right=0pt,top=0pt,bottom=0pt, colback=blue!10!white, colframe=blue!20!white, fontupper=\ttfamily\scriptsize, title=PROMPT]
This set of three images showcases an image decomposition task;

[IMAGE1] captures an image looking through a transparent glass, both the scene behind the glass and the reflection of the glass could be seen; 

[IMAGE2] displays the transmission of glass with reflection removed; 

[IMAGE3] shows only the reflection of glass without transmission; 

[IMAGE1] could be decomposed to [IMAGE2] and [IMAGE3].
\end{tcolorbox}
\vspace{-5pt}
It is trained on a single 4090 GPU (24 G) for 4,000 steps with a batch size of 1 and a LoRA rank of 16. For inference, we employ 20 sampling steps with a guide scale of 4, which matches the distillation guide scale of \flux.

\begin{figure*}
    \centering
    \setlength{\resLen}{0.16\textwidth}
    \addtolength{\tabcolsep}{-5pt}
    \renewcommand{\arraystretch}{0.6}
    \begin{tabular}{cccccc}
        \begin{overpic}[width=\resLen, height=0.75\resLen, keepaspectratio=False]{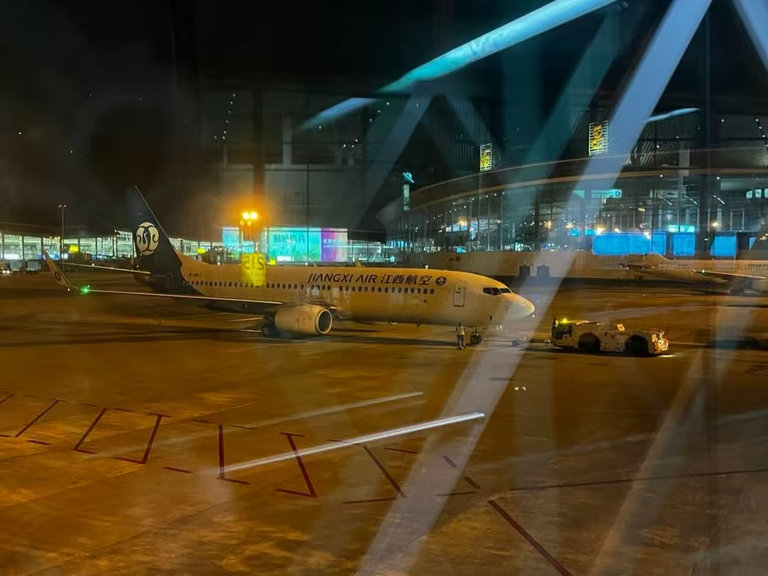}
            \put(0,3){\scriptsize \colorbox{white}{\color{black} (a)}}
        \end{overpic} &
        \includegraphics[width=\resLen, height=0.75\resLen, keepaspectratio=False]{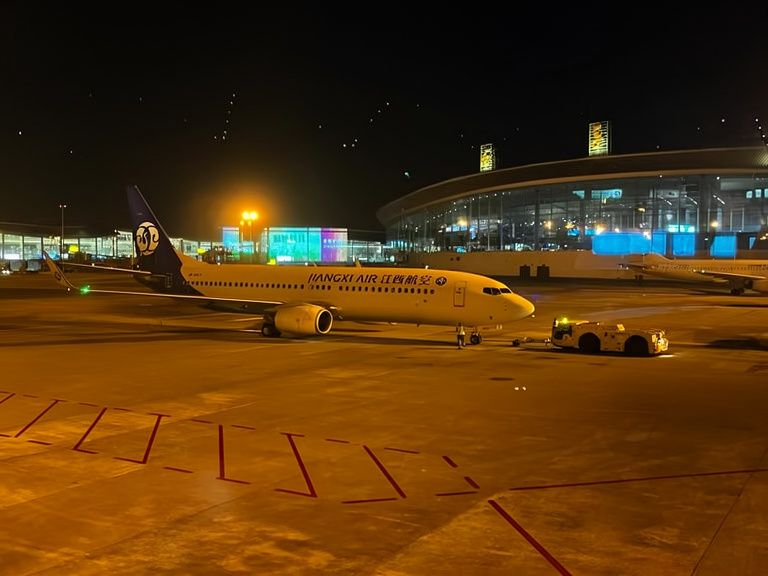} &
        \includegraphics[width=\resLen, height=0.75\resLen, keepaspectratio=False]{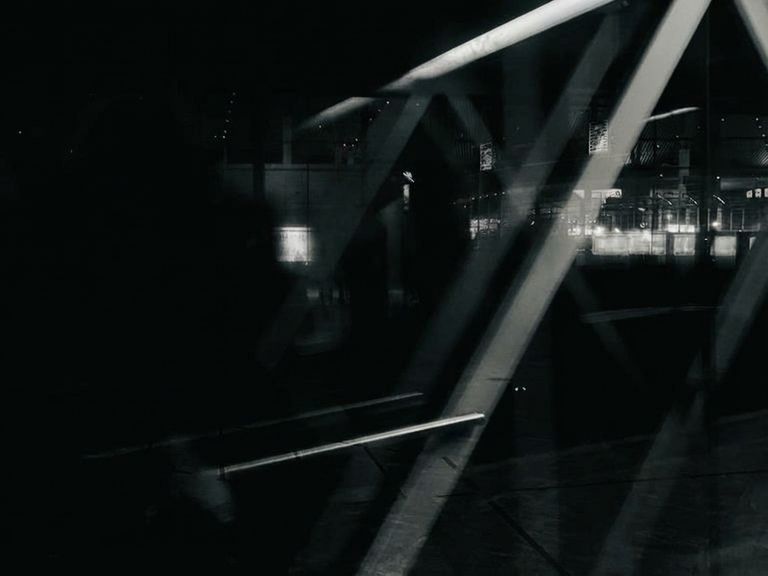} 
        &
        \begin{overpic}[width=\resLen, height=0.75\resLen, keepaspectratio=False]{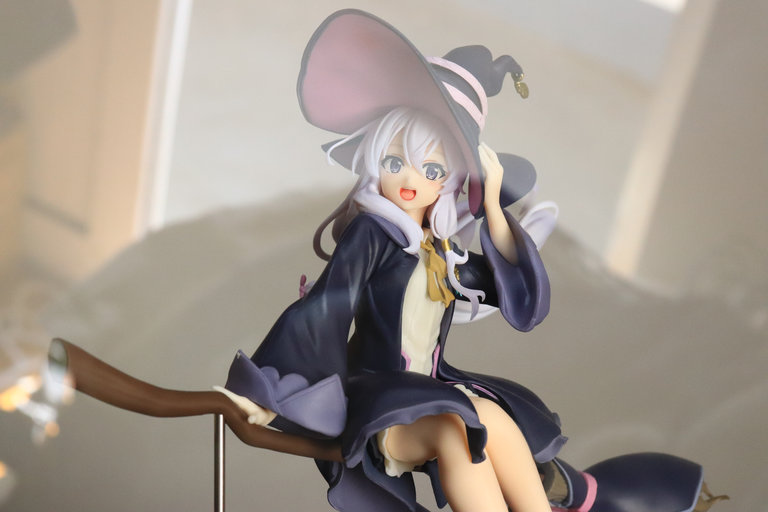}
            \put(0,3){\scriptsize \colorbox{white}{\color{black} (b)}}
        \end{overpic} &
        \includegraphics[width=\resLen, height=0.75\resLen, keepaspectratio=False]{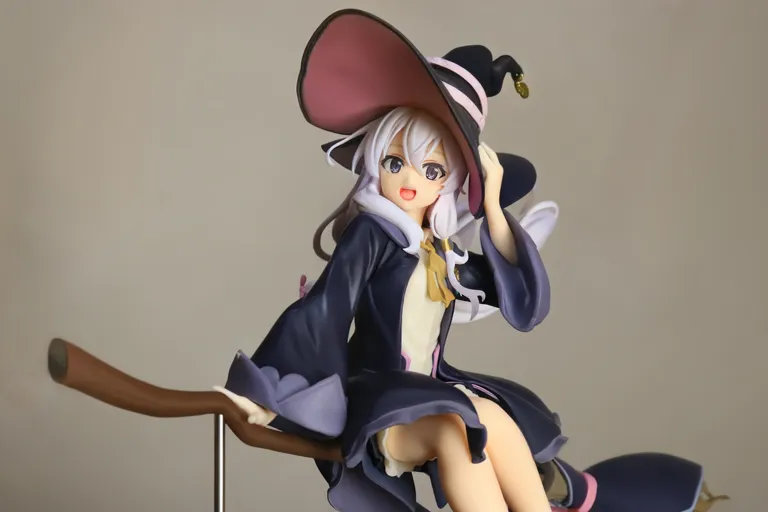} &
        \includegraphics[width=\resLen, height=0.75\resLen, keepaspectratio=False]{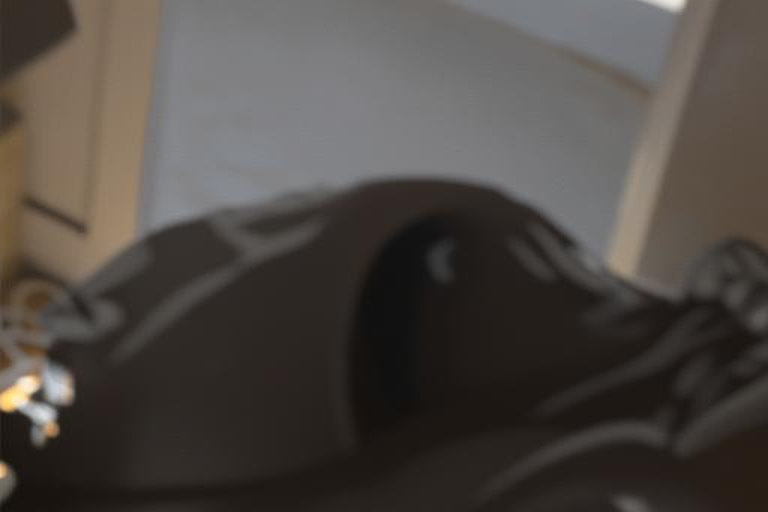} 
        \\        
        \begin{overpic}[width=\resLen, height=0.75\resLen, keepaspectratio=False, trim=0 0 0 10,clip]{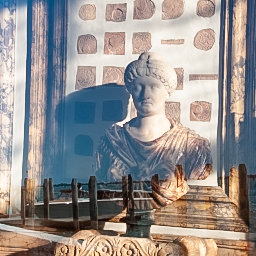}
            \put(0,3){\scriptsize \colorbox{white}{\color{black} (c)}}
        \end{overpic} &
        \includegraphics[width=\resLen, height=0.75\resLen, keepaspectratio=False, trim=0 0 0 35,clip]{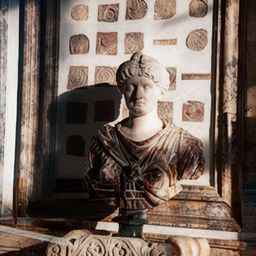} &
        \includegraphics[width=\resLen, height=0.75\resLen, keepaspectratio=False, trim=0 0 0 35,clip]{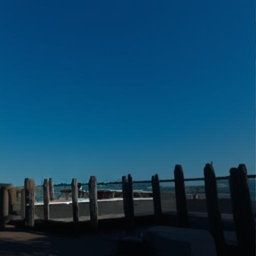}
        &
        \begin{overpic}[width=\resLen, height=0.75\resLen, keepaspectratio=False, trim=0 0 0 10,clip]{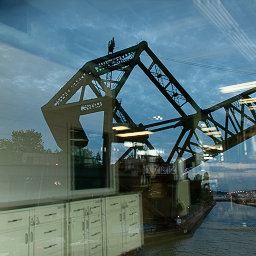} 
            \put(0,3){\scriptsize \colorbox{white}{\color{black} (d)}}
        \end{overpic} &
        \includegraphics[width=\resLen, height=0.75\resLen, keepaspectratio=False, trim=0 0 0 35,clip]{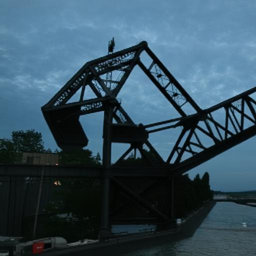} &
        \includegraphics[width=\resLen, height=0.75\resLen, keepaspectratio=False, trim=0 0 0 35,clip]{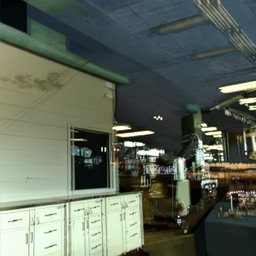}
        \\
        \begin{overpic}[width=\resLen]{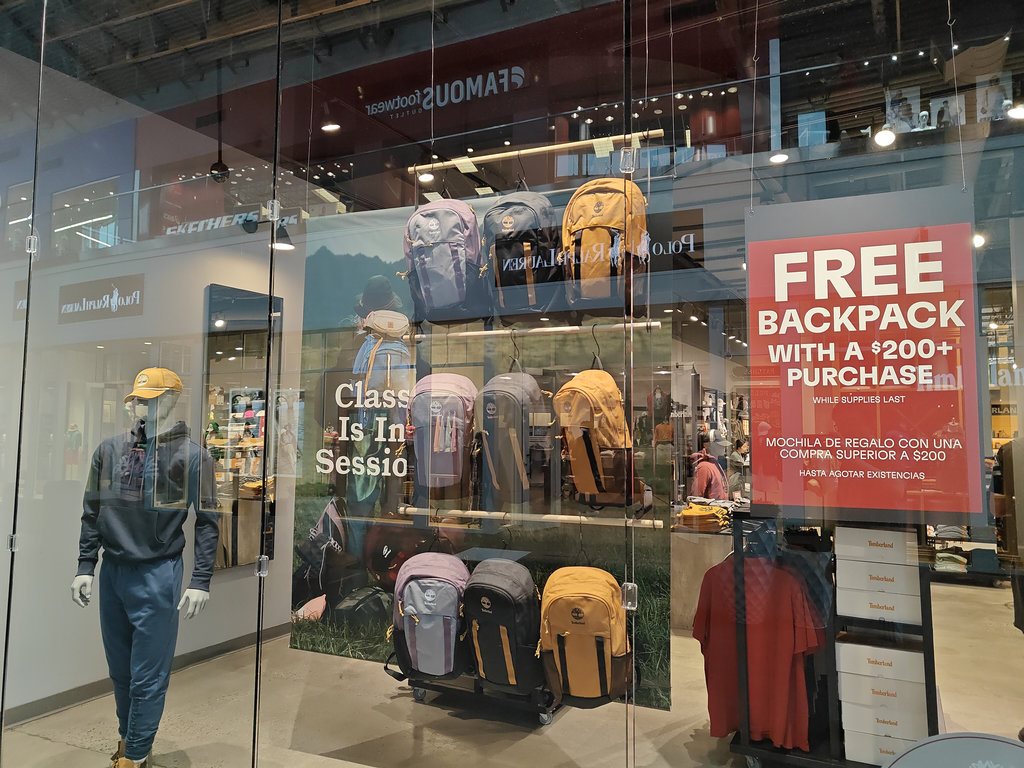}
            \put(0,3){\scriptsize \colorbox{white}{\color{black} (e)}}
        \end{overpic} &
        \includegraphics[width=\resLen]{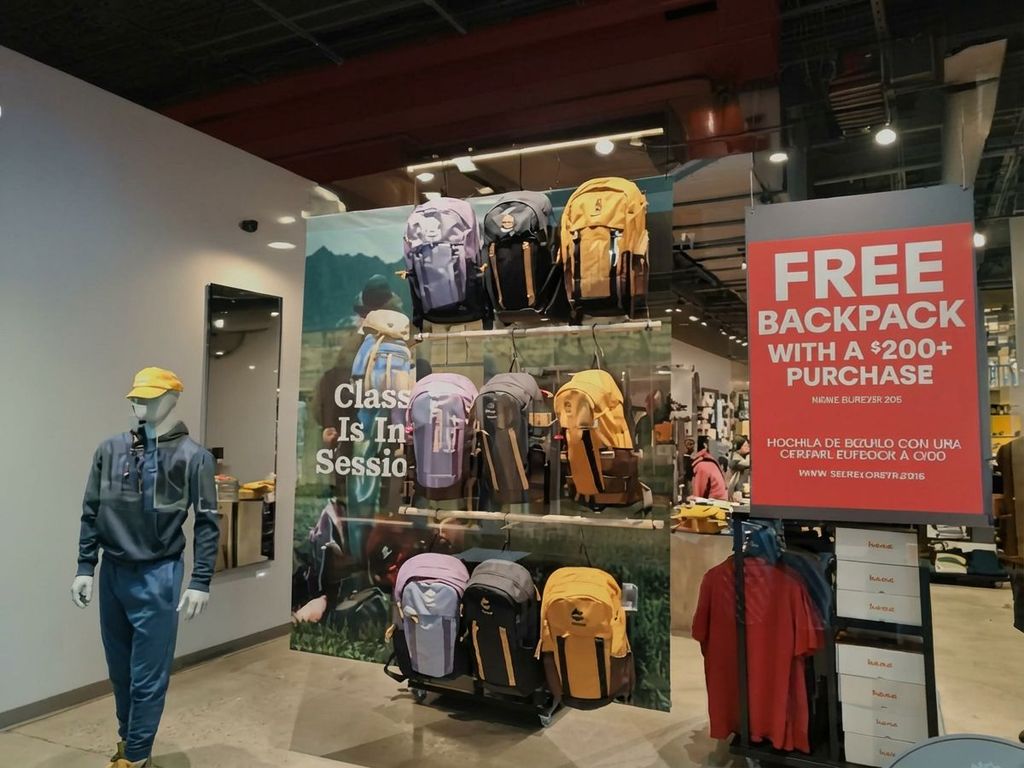} &
        \includegraphics[width=\resLen]{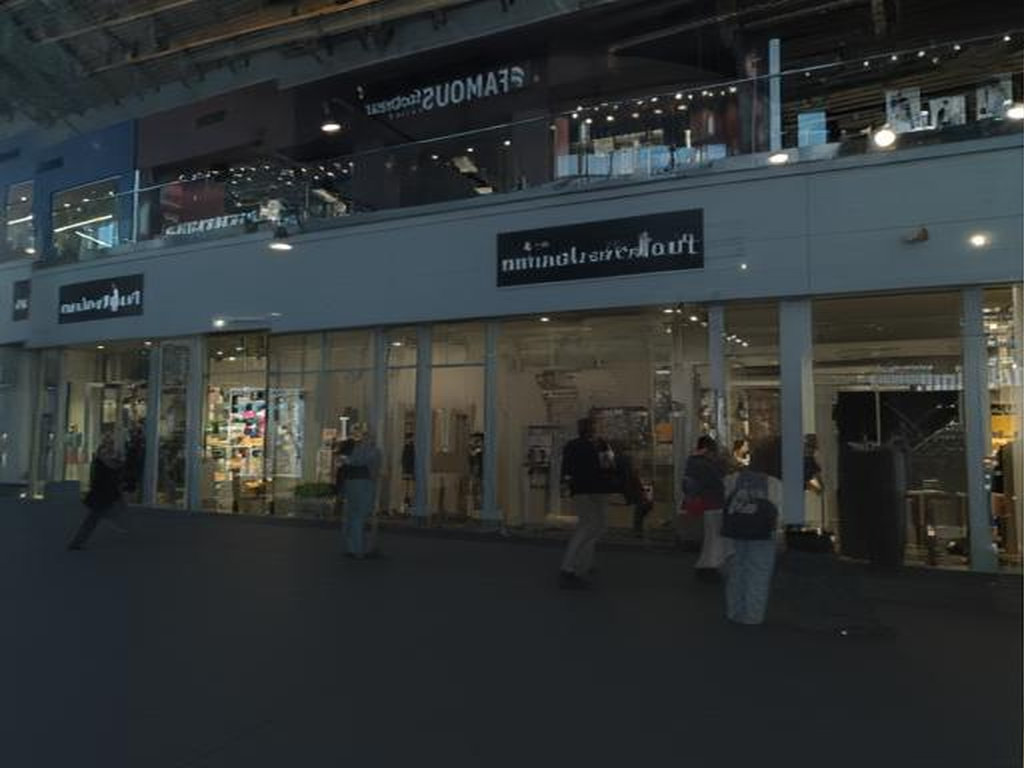}
        &
        \begin{overpic}[width=\resLen]{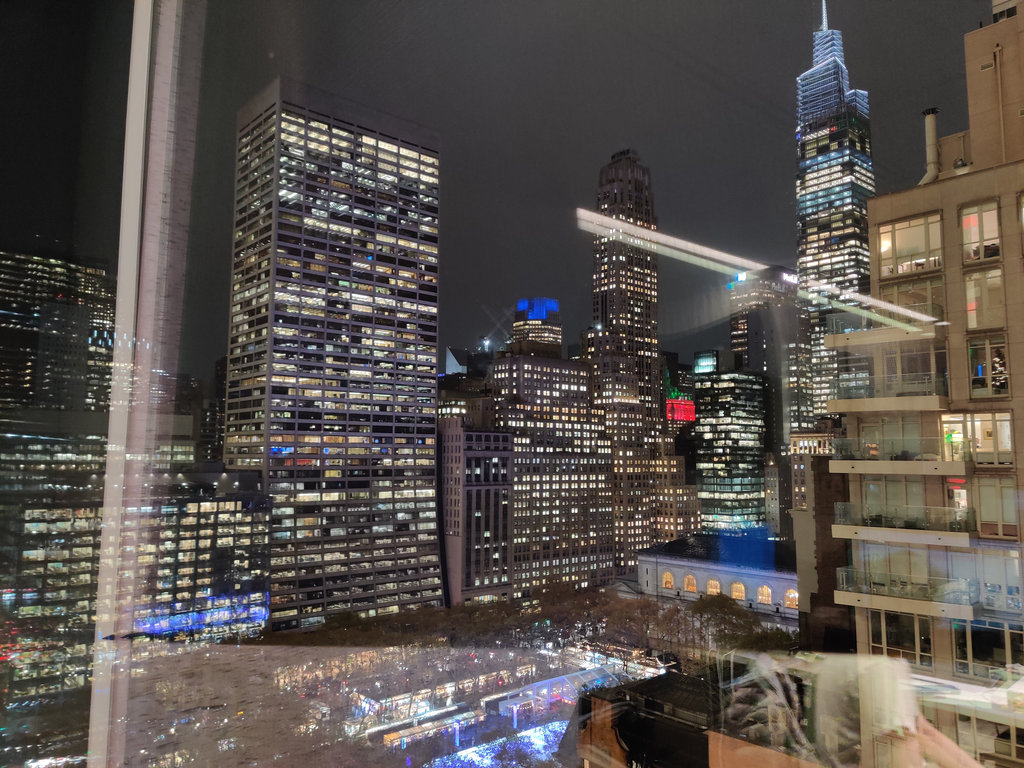}
            \put(0,3){\scriptsize \colorbox{white}{\color{black} (f)}}
        \end{overpic} &
        \includegraphics[width=\resLen]{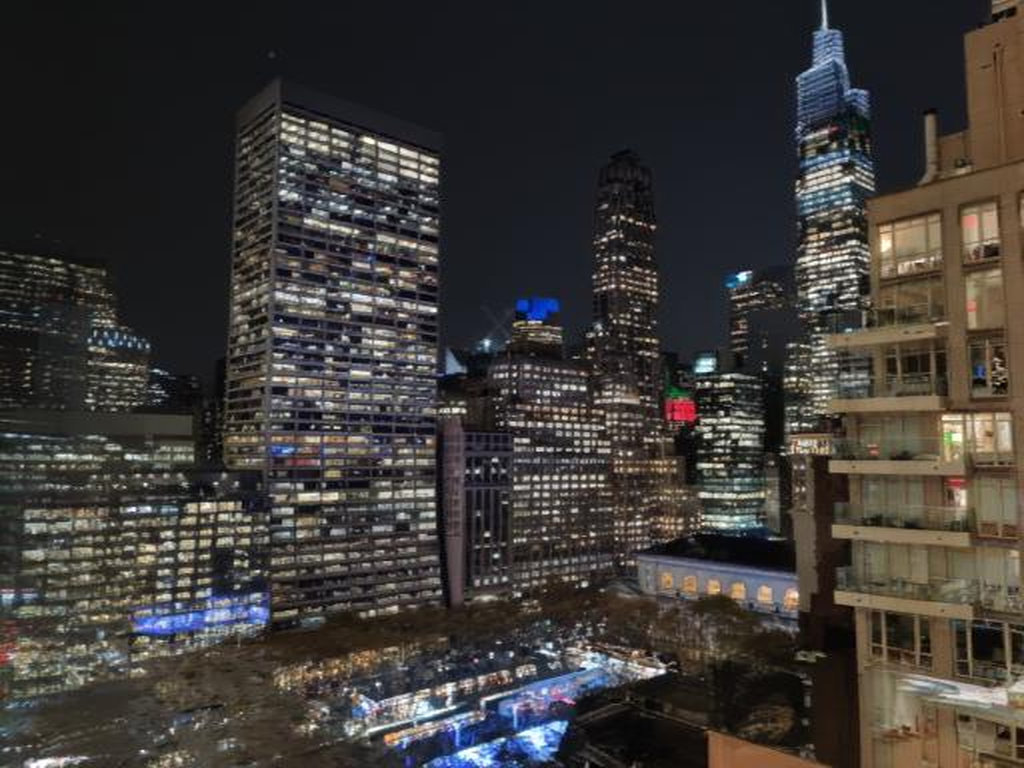} &
        \includegraphics[width=\resLen]{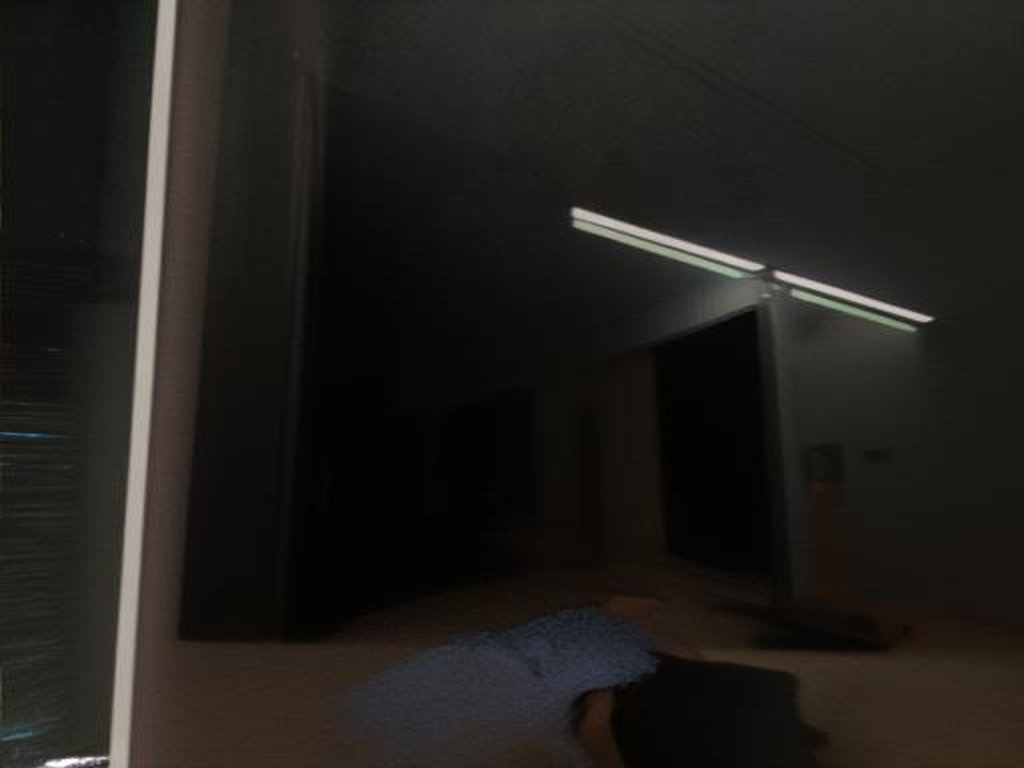}
        \\
        Input
        & Transmission
        & Reflection
        & Input
        & Transmission
        & Reflection
    \end{tabular}
    \vspace{-8pt}
    \caption{
        \textbf{More SIRR results.}(a) (b) from \cite{hu2025dereflection}; (c) (d) from \cite{kee2025removing}; (e) (f) from our smartphone captures. 
    }
    \label{fig:main_other_showcase}
\end{figure*}
\newcommand{\PSNR}{P$\uparrow$}
\newcommand{\SSIM}{S$\uparrow$}
\newcommand{\LPIPS}{L$\downarrow$}
\newcommand{\first}[1]{\textbf{#1}}
\newcommand{\second}[1]{\underline{#1}}
\newcommand{\third}[1]{\uuline{#1}}

\begin{table*}[t]
\small
\caption{\textbf{Quantitative comparisons on the real reflection benchmarks.}
The top-3 performing results are highlighted in \first{first}, \second{second} and \third{third}. \PSNR~is PSNR, \SSIM~is SSIM and \LPIPS~is LPIPS. Ours (syn) is not ranked since it is the only one that does not include real training data. All the visual results in this paper are generated by Ours (syn).}
\vspace{-8pt}
\label{tab:compare}
\setlength{\tabcolsep}{1.8pt}
\def\arraystretch{1.2}
\begin{tabular}{l|ccc|ccc|ccccccccc|ccc}
\toprule
\hline
\multirow{2}{*}{\small Method} & \multicolumn{3}{c|}{\multirow{2}{*}{\small\textit{Real} (20) \cite{zhang2018single}}} 
                        & \multicolumn{3}{c|}{\multirow{2}{*}{\small\textit{Nature} (20) \cite{li2020single}}} 
                        & \multicolumn{9}{c|}{\small\textit{SIR}$^2$ (141) \cite{wan2017benchmarking}} 
                        & \multicolumn{3}{c}{\multirow{2}{*}{\small Average (181)}}\\
\cline{8-16} 
                        & \multicolumn{3}{c|}{}                      
                        & \multicolumn{3}{c|}{}                        
                        & \multicolumn{3}{c|}{\footnotesize Postcard (20)}   
                        & \multicolumn{3}{c|}{\footnotesize SolidObject (20)}  
                        & \multicolumn{3}{c|}{\footnotesize Wildscene (101)} 
                        & \multicolumn{3}{c}{} \\ 
\hline
& \PSNR & \SSIM & \LPIPS & \PSNR & \SSIM & \LPIPS & \PSNR & \SSIM & \multicolumn{1}{c|}{\LPIPS} & \PSNR & \SSIM & \multicolumn{1}{c|}{\LPIPS} & \PSNR & \SSIM & \LPIPS & \PSNR & \SSIM & \LPIPS\\ 
\hline
\footnotesize Dong \cite{dong2021location}    
& 20.36 
& 0.809 
& 0.284 
& 23.41 
& 0.822 
& 0.156 
& 23.24 
& 0.882 
& \multicolumn{1}{c|}{0.165} 
& 23.87 
& 0.909 
& \multicolumn{1}{c|}{0.093} 
& 23.97 
& 0.914 
& 0.109 
& 23.418
& 0.8881
& 0.1380 \\ 
\footnotesize DSRNet \cite{hu2023single}      
& 20.94 
& 0.813 
& 0.286 
& 24.75 
& 0.825 
& 0.180 
& 23.65 
& 0.898 
& \multicolumn{1}{c|}{0.154} 
& \third{26.55} 
& 0.931 
& \multicolumn{1}{c|}{\third{0.074}} 
& 26.56 
& 0.932 
& 0.088 
& 25.416
& 0.9032
& 0.1258 \\ 
\footnotesize Zhu \cite{zhu2024revisiting}    
& 22.85 
& 0.829 
& 0.255 
& \third{25.64} 
& \third{0.835} 
& \second{0.157} 
& 23.69 
& 0.865 
& \multicolumn{1}{c|}{0.200} 
& \first{26.79} 
& \first{0.936} 
& \multicolumn{1}{c|}{0.075} 
& \first{26.88} 
& 0.929 
& 0.089 
& 25.935
& 0.9013
& 0.1256 \\ 

\footnotesize DSIT \cite{hu2024single}        
& \third{23.82} 
& \third{0.852} 
& \third{0.231} 
& \first{26.07} 
& 0.833 
& 0.183 
& \second{25.08} 
& \second{0.909} 
& \multicolumn{1}{c|}{\third{0.147}} 
& 26.53 
& 0.932 
& \multicolumn{1}{c|}{0.076} 
& \third{26.84} 
& \first{0.937} 
& \first{0.075} 
& \second{26.192}
& \second{0.9125}
& \second{0.1122} \\ 

\footnotesize RDNet \cite{zhao2025reversible} 
& \second{24.22} 
& \first{0.859}
& \second{0.228}  
& \second{25.78} 
& \second{0.838} 
& \third{0.171} 
& \third{25.06} 
& \third{0.906}
& \multicolumn{1}{c|}{\second{0.140}} 
& \second{26.60}
& \second{0.934} 
& \multicolumn{1}{c|}{\first{0.072}} 
& 26.80 
& \second{0.934} 
& \third{0.081} 
& \third{26.188}
& \third{0.9120}
& \third{0.1127} \\ 
       
\footnotesize Ours \scriptsize{(syn+real)}                           
& \first{24.89} 
& \second{0.858} 
& \first{0.227} 
& 25.14 
& \first{0.845} 
& \first{0.124} 
& \first{26.18} 
& \first{0.925} 
& \multicolumn{1}{c|}{\first{0.100}} 
& 26.41 
& \third{0.933} 
& \multicolumn{1}{c|}{\second{0.073}} 
& \second{26.86} 
& \third{0.933} 
& \second{0.076} 
& \first{26.327}
& \first{0.9141}
& \first{0.1003} \\     

\hline
\footnotesize Ours \scriptsize{(syn)}                            
& 24.20 
& 0.847 
& 0.245 
& 24.81 
& 0.844 
& 0.135 
& 26.20 
& 0.933 
& \multicolumn{1}{c|}{0.099} 
& 26.34 
& 0.931 
& \multicolumn{1}{c|}{0.072} 
& 26.85 
& 0.932 
& 0.078 
& 26.204
& 0.9129
& 0.1044 \\    

\hline 
\bottomrule
\end{tabular}
\vspace{-8pt}
\end{table*}

\section{Results and Analysis}
\vspace{0pt}
\noindent\textbf{The testing dataset.} Following prior work, we evaluate our fine-tuned LMM on three real-world reflection benchmarks: \textit{Real} \cite{zhang2018single}, \textit{Nature} \cite{li2020single}, and \textit{SIR}$^{2}$ \cite{wan2017benchmarking}, which collectively cover diverse real-world reflection scenarios and are standard for assessing reflection removal performance.
Note that there are 20 testing images in \textit{Real} and 20 in \textit{Nature}, while for \textit{SIR}$^{2}$, we tested 3 subcategories: for each scene setup in \textit{Postcard}, there are 10 different glass/lens variances and we random choose one from each scene, and in total there are 20 images; we did the same to \textit{SolidObject}; and we use all the 101 images in \textit{Wildscene} as testing data.

\noindent\textbf{Comparison methods}. 
Five most recent representative single-image reflection removal methods, Dong \cite{dong2021location}, DSRNet \cite{hu2023single}, Zhu \cite{zhu2024revisiting}, DSIT \cite{hu2024single} and RDNet \cite{zhao2025reversible}, are compared with our method. All methods could predict both \T~and \R~except Zhu. DSIT is based on the Vision Transformers structure, and RDNet wins the NTIRE 2025 single-image reflection removal in the wild challenge \cite{yang2025ntire}. Both of them are considered state-of-the-art methods.

\noindent\textbf{Evaluation indicators}. 
We use PSNR \cite{huynh2008scope} and SSIM \cite{wang2003multiscale} as error indicators following previous methods, and further adopt LPIPS \cite{zhang2018unreasonable} to measure the perceptual quality of the predicted results. Evaluations are performed using the same test scripts and the same testing data to ensure fairness.

\subsection{Quantitative Performance}
For all quantitative comparisons, we use the authors' official code and publicly released pre-trained weights without modification.
To align with previous work, 289 image pairs from \textit{Real} (89) and \textit{Nature} (200) are added to the training data for quantitative comparisons. LoRA training takes 4 to 5 hours, and different training and testing image ratios and resolutions are supported. 

Table \ref{tab:compare} shows that our approach outperforms most competing methods across all test datasets. Given the diverse scenes, lightings in these real-world datasets, achieving consistently top results on all metrics is challenging. However, with only a small amount of high-quality synthetic data and reduced training time, our method still achieves performance comparable to leading techniques.

\begin{table}[t]
\small
\caption{\textbf{Regional LPIPS ($\downarrow$)} on the reflection area.
}
\vspace{-8pt}
\label{tab:locallpips}
\setlength{\tabcolsep}{2.5pt}
\def\arraystretch{1}
\begin{tabular}{c|ccccc|c}
\toprule
\hline
Method & Real & Nature & \footnotesize{Postcard} & \footnotesize{SolidObject} & \footnotesize{Wildscene} & Average \\ 
\hline
DSIT    & 0.310 & 0.195 & 0.206 & 0.121 & 0.124 & 0.161 \\
RDNet   & \textbf{0.286} & 0.191 & 0.199 & 0.122 & 0.134 & 0.163 \\
Ours    & 0.299 & \textbf{0.186} & \textbf{0.110} & \textbf{0.102} & \textbf{0.107} & \textbf{0.137} \\
\hline
\bottomrule
\end{tabular}
\vspace{-10pt}
\end{table}


In many images, reflections occupy only a small portion, making the LPIPS score over the entire image less indicative of perceptual quality. To better capture our improvement, we compute LPIPS only within the reflection regions, estimated from the difference between the input and reflection-free image. As shown in Table \ref{tab:locallpips}, our method yields an 18.2\% improvement on reflection regions, compared to 7.7\% when computed over the full image.


\begin{table}[t]
\small
\caption{\textbf{User study comparison.} 
    The win rate of our method, the tie rate and RDNet/DSIT win rate.
}
\vspace{-8pt}
\label{tab:userstudy}
\setlength{\tabcolsep}{2pt}
\def\arraystretch{1}
\begin{tabular}{c|ccc|ccc}
\toprule
\hline
Dataset & Ours & Tie & RDNet & Ours & Tie & DSIT \\ 
\hline
Real        & 49.38\% & 21.25\% & 29.37\% & 57.32\% & 21.50\% & 21.18\% \\
Nature      & 35.00\% & 42.08\% & 22.92\% & 35.00\% & 53.75\% & 11.25\% \\
Postcard    & 55.37\% & 18.66\% & 25.97\% & 67.03\% & 12.26\% & 20.71\% \\
SolidOject  & 41.43\% & 26.96\% & 31.61\% & 40.37\% & 26.06\% & 33.57\% \\
Wildscene   & 43.68\% & 29.57\% & 26.75\% & 48.57\% & 15.71\% & 35.71\% \\
\hline
Average     & 45.15\% & 27.78\% & 27.07\% & 49.17\% & 21.32\% & 29.51\% \\
\hline
\bottomrule
\end{tabular}
\vspace{-10pt}
\end{table}

\subsection{Qualitative Comparison}

\subsubsection{Single-image reflection removal}

To further illustrate our advantages, Figure \ref{fig:main_t} presents visual comparisons of transmission layers against other methods. In example (a), while recent methods remove the out-of-focus highlights, only ours accurately inpaints the occluded area and correctly learns that the stand’s reflection should be red. Example (b) contains discontinuous reflections, where others achieve only partial removal, but our approach captures the full pattern. In examples (c) and (d), the watermark-like reflections make it difficult to distinguish reflection from transmission, yet our method produces clearer and more consistent results. For strong reflections, as in example (e), our method uniquely identifies the white region as reflection and generates plausible context for the shop’s name. Overall, our outputs are visually cleaner with fewer residual artifacts.

\subsubsection{Single-image reflection separation}

In Figure \ref{fig:main_r}, we compare reflection layers and show that our method reconstructs reflection scenes more accurately than existing approaches. Unlike trivial cases dominated by highlight reflections, our goal is to recover informative reflection content embedded in the image. According to Snell’s law, most optical energy is transmitted through glass, and reflections become prominent only at grazing incidence. Even under limited information and near-dark reflections, our method uniquely estimates the reflection across the entire image while plausibly synthesizing content in regions with extremely low visibility.

\subsubsection{More results for real captured images}

We observe that existing methods perform well on current benchmarks, suggesting possible overfitting to older datasets. To further assess practicality, we evaluate our approach on additional real-world reflection images. As shown in Figure \ref{fig:main_other_showcase}, our method achieves an accurate separation of the reflection and transmission layers: the recovered transmission preserves fine details behind the glass, while the estimated reflection retains reflective elements within the frame. These results highlight the robustness of our method in complex real-world scenarios. See Figure \ref{fig:main_other_compare} in Appendix for the full comparison results.

\begin{figure}[tb]
    \centering
    \setlength{\resLen}{0.24\linewidth}
    \addtolength{\tabcolsep}{-4.5pt}
    \def\arraystretch{0.6}
    \begin{tabular}{cccc}
        \includegraphics[width=\resLen]{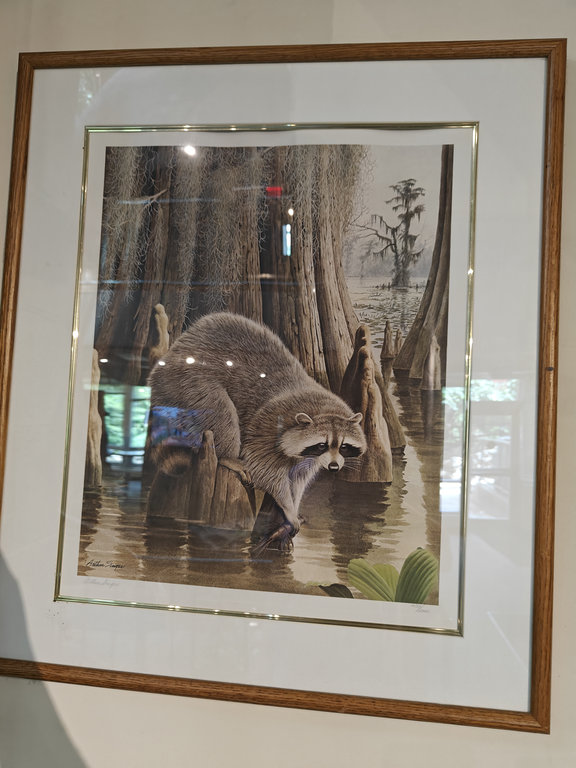} &
        \includegraphics[width=\resLen]{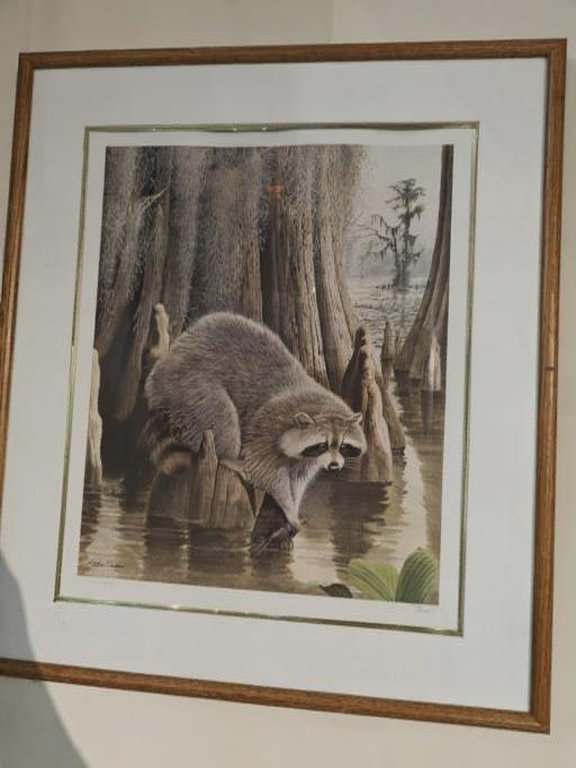} &
        \includegraphics[width=\resLen]{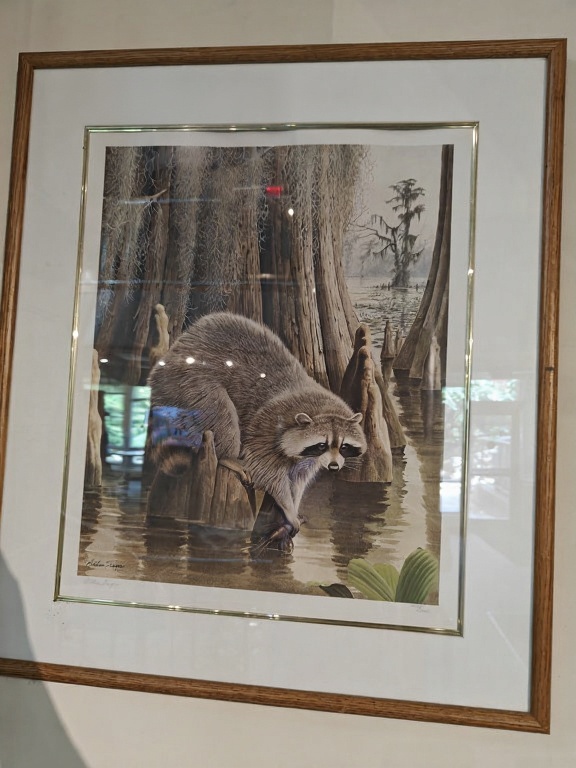} &
        \includegraphics[width=\resLen]{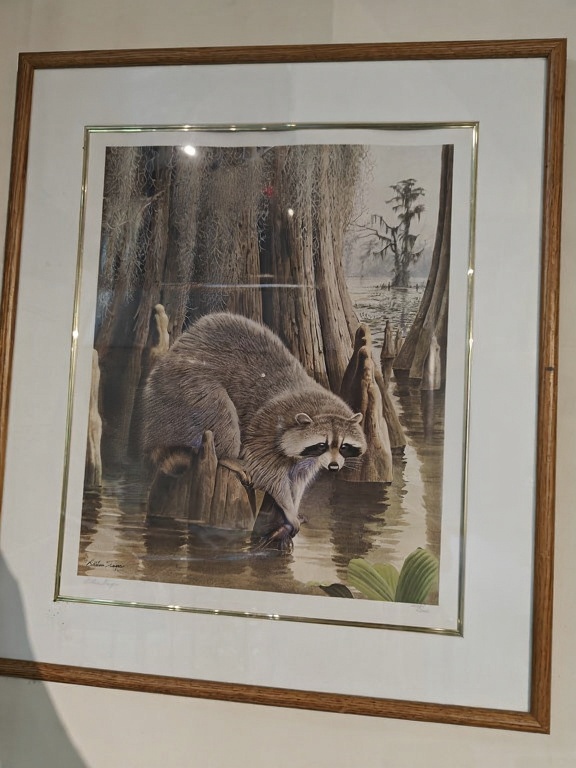}
        \\
        Input & Ours & Flux.1 & Flux.1 \\
              & \footnotesize{(generic prompt)} & \footnotesize{(generic prompt)} & \footnotesize{(detailed prompt)} 
    \end{tabular}
    \vspace{-8pt}
    \caption{
        \textbf{The Comparison of our result with basic Flux model.} LMMs have the capability in reflection removal but often require carefully designed prompts. In contrast, our method leverages limited training data to enable non-specific image prompts.
    }
    \vspace{-12pt}
    \label{fig:flux_compare}
\end{figure}

\subsection{User Study}

Although the similarity scores in Table \ref{tab:compare} are close among our method, RDNet and DSIT, our results are often visually superior in reflection removal cleanliness, fewer artifacts and overall image quality. To validate, we conducted a user study with 100 participants. Each subject was randomly assigned 40 image pairs. For each pair, they were shown an input image with reflections and two outputs, ours and a comparison method, then asked to choose the better result or indicate "similar". Our method achieved a 45.15\% win rate (72.93\% win+tie) against RDNet and 49.17\% (70.49\% win+tie) against DSIT, showing a clear subjective preference for our results. See Table \ref{tab:userstudy} for more details.


\subsection{Ablation Studies}


\subsubsection{Comparing to original Flux model}

Compared to other text-to-image LMMs, \flux~stands out as an open-source option with strong iterative editing capabilities and image quality. Same as other LMMs, it often struggles when using a fixed and generic prompt, requiring carefully tailored user input to achieve good results. In contrast, our fine-tuned model addresses this limitation by removing the dependency on user-specific prompts, consistently delivering effective edits from a constant input. See example in Figure \ref{fig:flux_compare}.

\subsubsection{Comparing to 8-bit mixture synthetic data}

\begin{figure}[t]
    \centering
    \small
    \setlength{\resLen}{0.24\linewidth}
    \addtolength{\tabcolsep}{-5pt}
    \begin{tabular}{cccc}
        \includegraphics[width=\resLen]{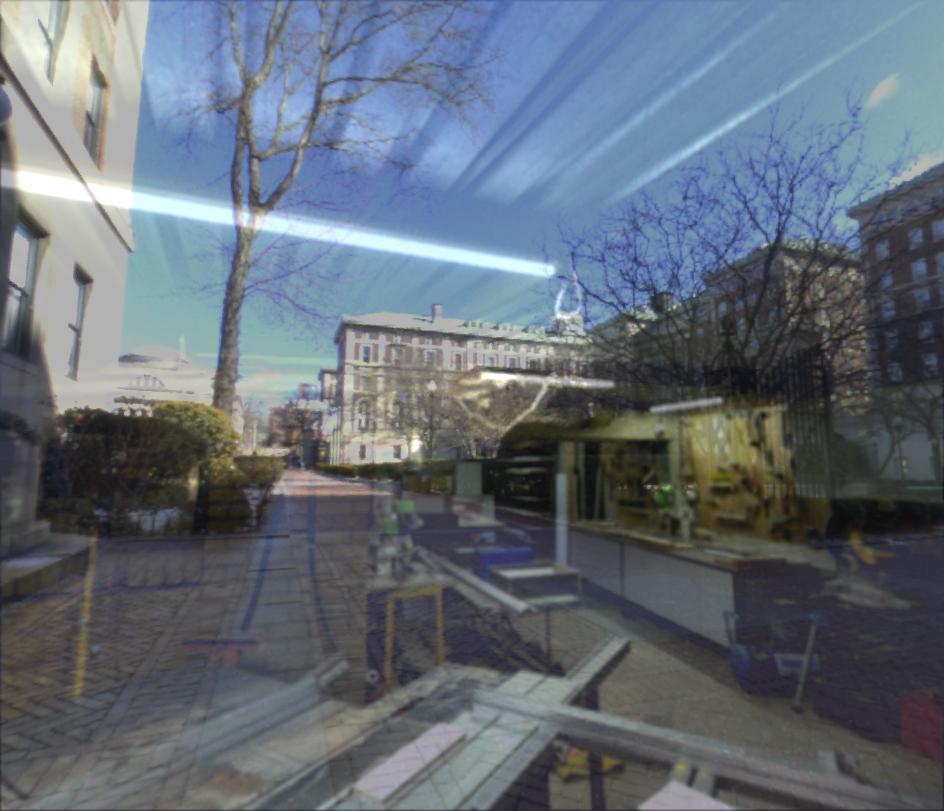} &
        \includegraphics[width=\resLen]{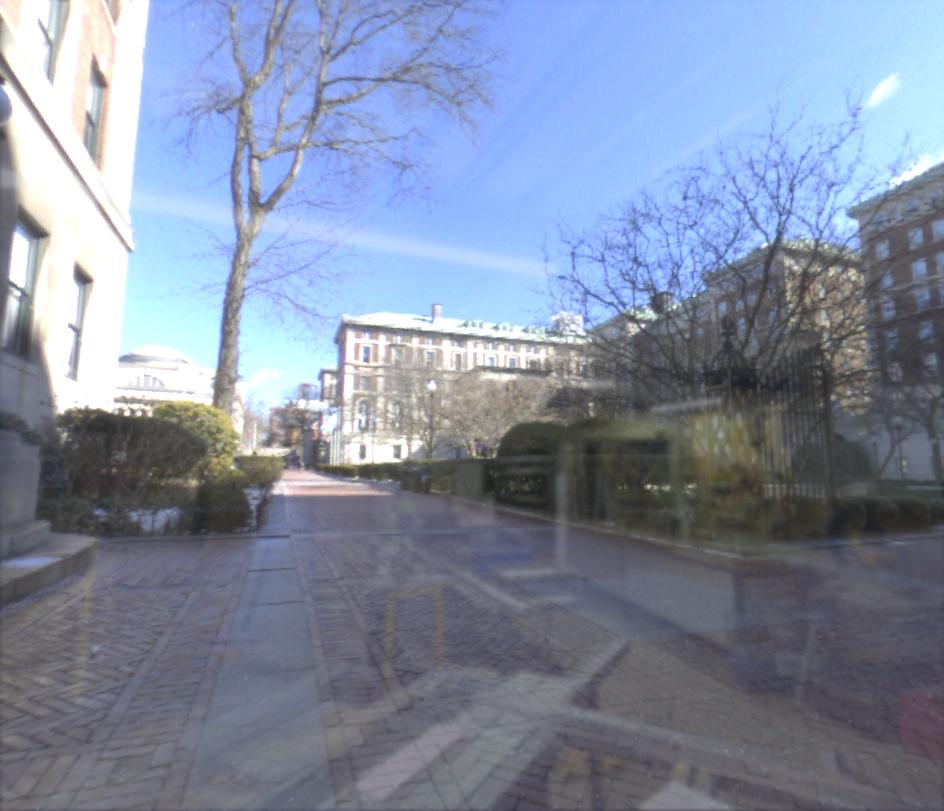} &
        \includegraphics[width=\resLen]{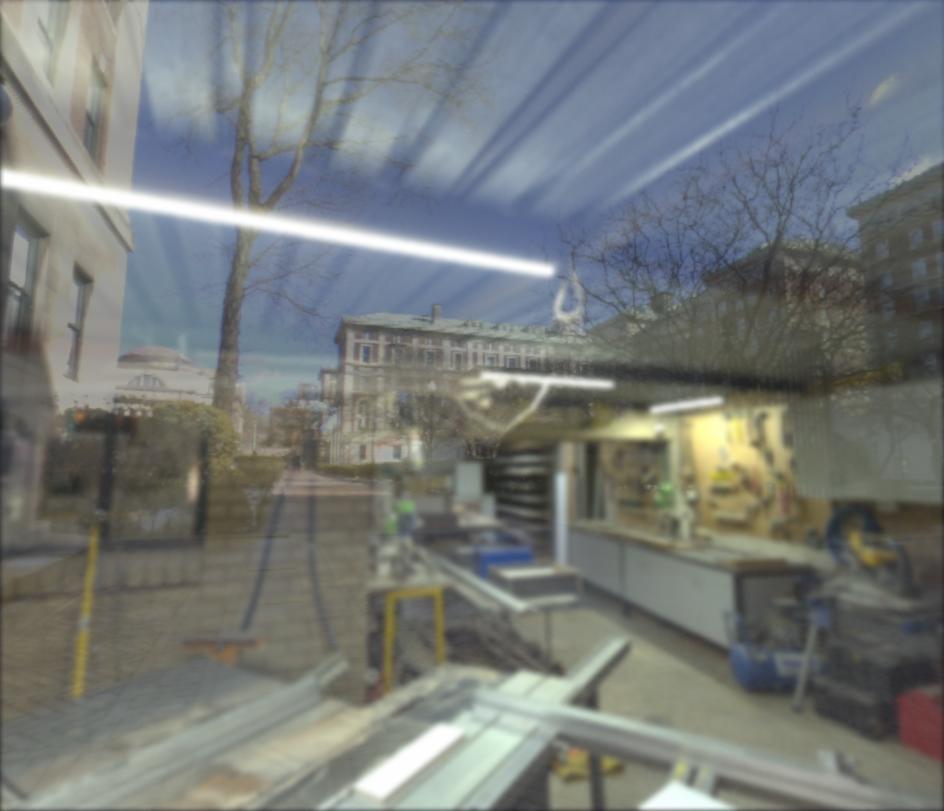} &
        \includegraphics[width=\resLen]{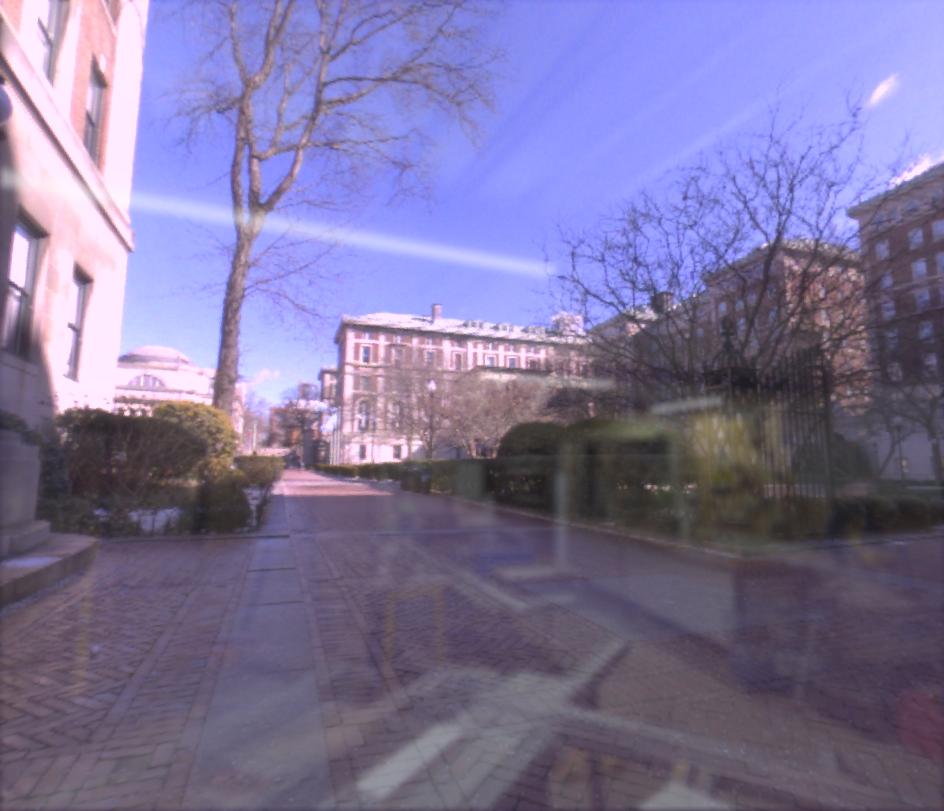} \\
        \includegraphics[width=\resLen]{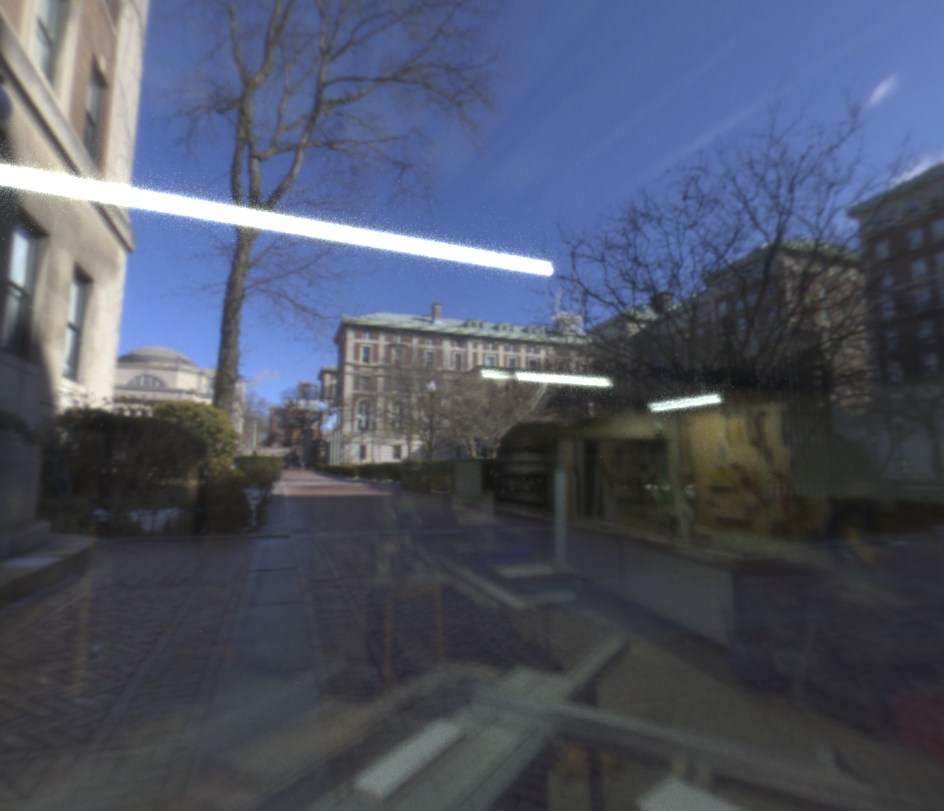} &
        \includegraphics[width=\resLen]{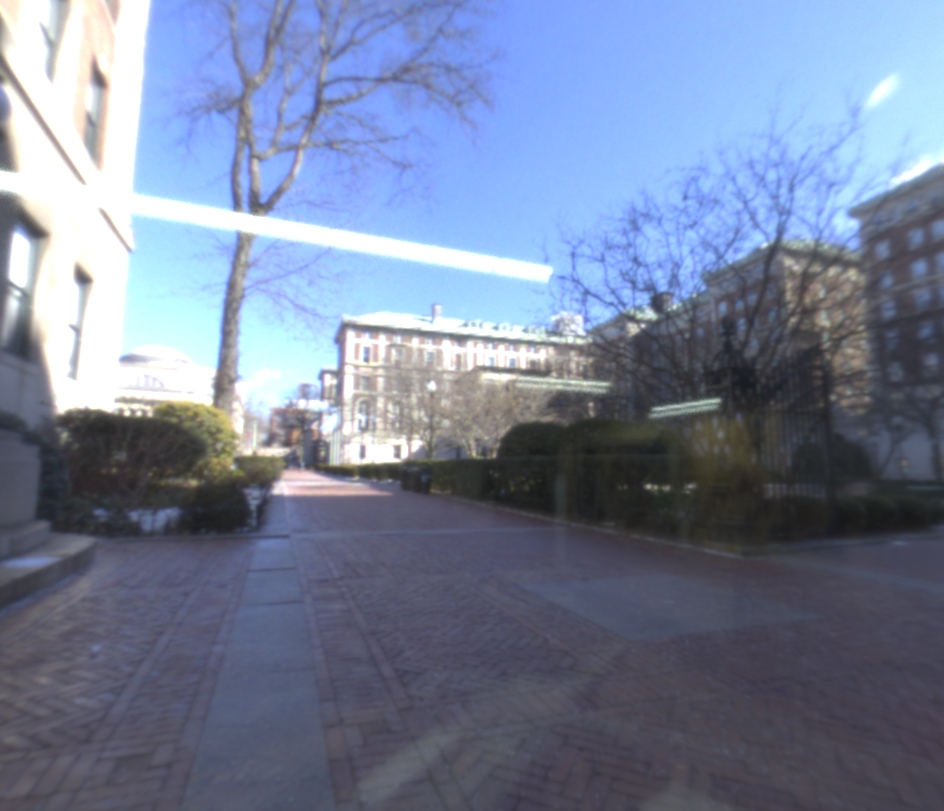} &
        \includegraphics[width=\resLen]{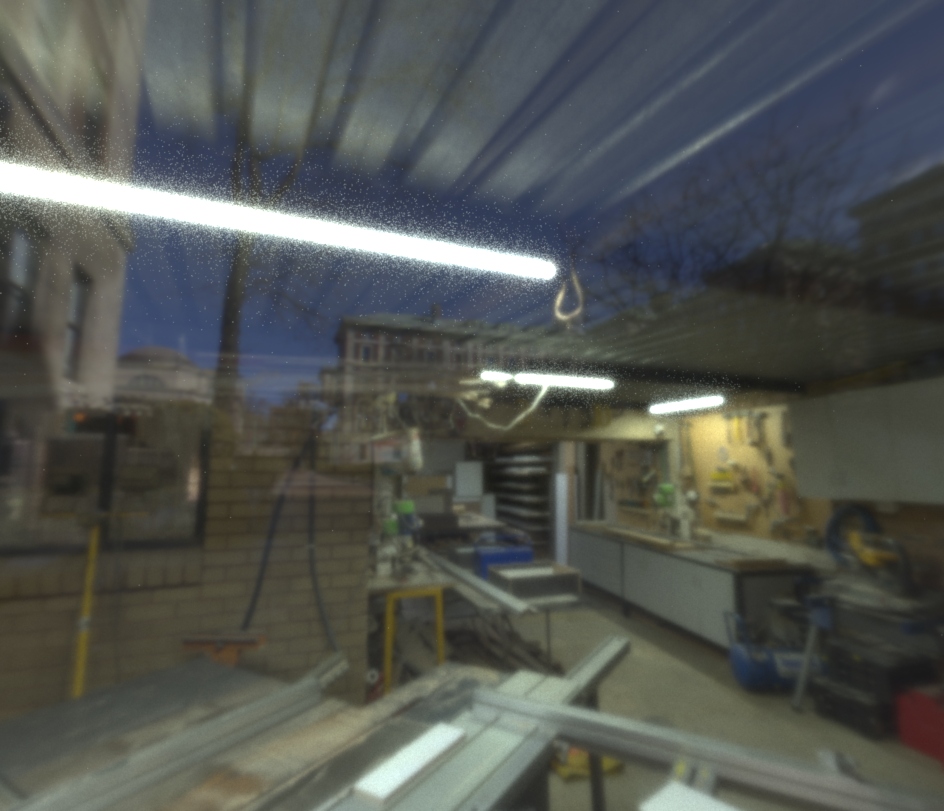} &
        \includegraphics[width=\resLen]{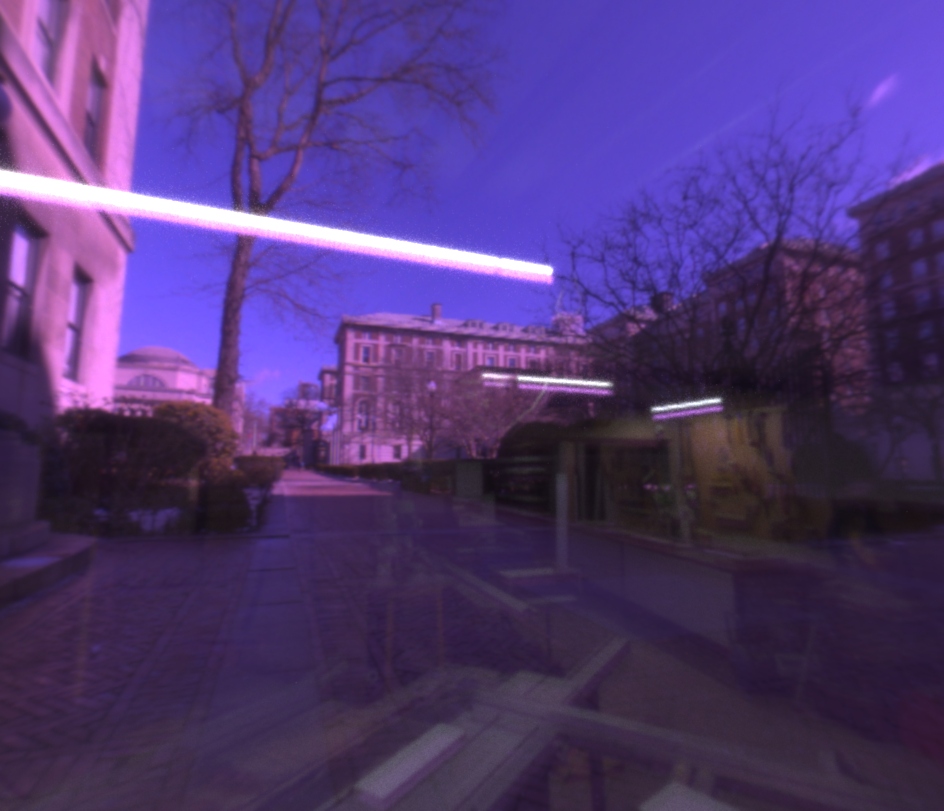} \\
        (a) & (b) & (c) & (d)
    \end{tabular}
    \vspace{-8pt}
    \caption{
        \textbf{The comparison of our synthetic data with the previous work.} The top row is the linear combination of two 8-bit LDR images that follow RDNet and the bottom row is generated using our synthesis method. The same transmission (outdoor) and reflection (indoor) scenes are used here for all the images. The top looks washed out and bottom has higher contrast which is more like a real photo.
    }
    \label{fig:syn_compare}
\end{figure}

To further assess the quality of our synthetic data, we follow RDNet and generate comparison images by linearly blending 8-bit tone-mapped images using 
$I = \alpha T + \beta R - T \circ R$. As shown in Figure~\ref{fig:syn_compare}, we use identical transmission and reflection content in both methods to ensure a fair comparison. Prior approaches synthesize reflections by mixing 8-bit tone-mapped images, often producing brighter and “flatter” results: simulated reflections lack overexposed highlights and appear disproportionately strong in shadowed regions, lifting the black levels and creating a washed-out look. In real scenes, when the outdoor environment is bright, indoor illumination is weak and reflections appear only near lights or windows (Figure~\ref{fig:syn_compare}-(b)), whereas at dusk larger portions of the indoor scene may reflect onto the outdoor view (Figure~\ref{fig:syn_compare}-(c)). Zooming into the white fluorescent light strip in Figure~\ref{fig:syn_compare}-(d), the reflection is not a single sharp rectangle but shows a “shaky” double edge and a faint halo-artifacts of refraction and ghosting caused by real glass thickness.


\begin{table}[t]
\footnotesize
\caption{\textbf{LPIPS ($\downarrow$) on different synthetic training data.} Ours is trained with our synthetic data; and Ours-RDNet is trained with the data generated using synthetic strategy from RDNet.
}
\vspace{-8pt}
\label{tab:syn_compare}
\setlength{\tabcolsep}{2.7pt}
\def\arraystretch{1.3}
\begin{tabular}{c|ccccc|c}
\toprule
\hline
Method & Real & Nature & \scriptsize{Postcard} & \scriptsize{SolidObject} & \scriptsize{Wildscene} & Average \\ 
\hline
Ours  & 0.302 & 0.218 & 0.111 & 0.108 & 0.109 & 0.142 \\
Ours-RDNet & 0.316 & 0.227 & 0.112 & 0.140 & 0.131 & 0.161 \\
\hline
\bottomrule
\end{tabular}
\vspace{-10pt}
\end{table}


To assess the impact of synthetic data quality on training, we generated 1000 image pairs using blending-based methods, kept the underlying scene fixed, and fine-tuned the Flux model. The resulting model consistently underperformed compared to training on our synthetic dataset. A detailed regional LPIPS comparison is shown in Table~\ref{tab:syn_compare}.

\subsubsection{Training size}

\begin{figure}[t]
    \centering
    \includegraphics[width=0.45\textwidth]{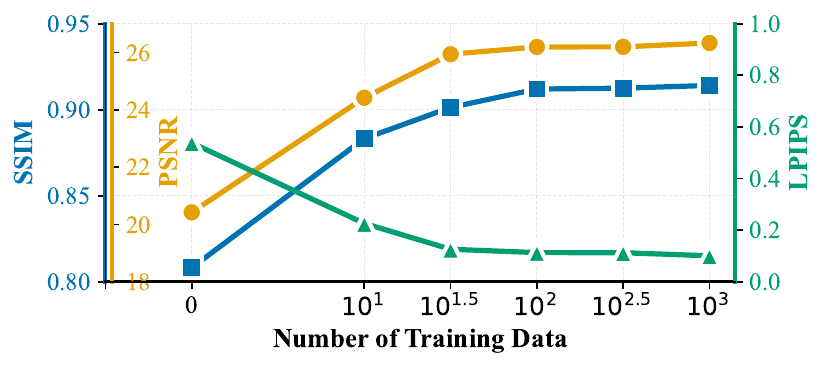}
    \vspace{-8pt}
    \caption{
        \textbf{Ablation on training data size for fine-tuning.} Performance improves rapidly with more samples and stabilizes after a few hundred.
    }
    \label{fig:training_number}
\end{figure}

To understand how training data size affects LoRA fine-tuning, we conduct an ablation study where the number of synthetic training pairs varies from 0 to 1000 (0 means no fine-tuning). As shown in Figure~\ref{fig:training_number}, the results reveal a clear and consistent trend across all metrics. Without training, performance is limited: SSIM stays below 0.81, PSNR is around 20 dB, and LPIPS remains high ($>$0.5), indicating insufficient generalization, also evident in Figure~\ref{fig:flux_compare}. With 10–50 samples, all metrics improve sharply, PSNR increases by more than 4 dB and SSIM rises above 0.88, highlighting the strong data efficiency of LoRA when adapting a pretrained LMM. Beyond roughly 300 samples, the gains taper off and the curves saturate, suggesting diminishing returns once the model has captured the task’s diversity. Overall, these findings show that while LoRA benefits from more data, a few hundred samples are sufficient for stable and competitive performance in SIRR.

\section{Limitations}
\begin{figure}[t]
    \centering
    \setlength{\resLen}{0.4\linewidth}
    \addtolength{\tabcolsep}{-5pt}
    \begin{tabular}{cc}
        \begin{overpic}[width=\resLen]{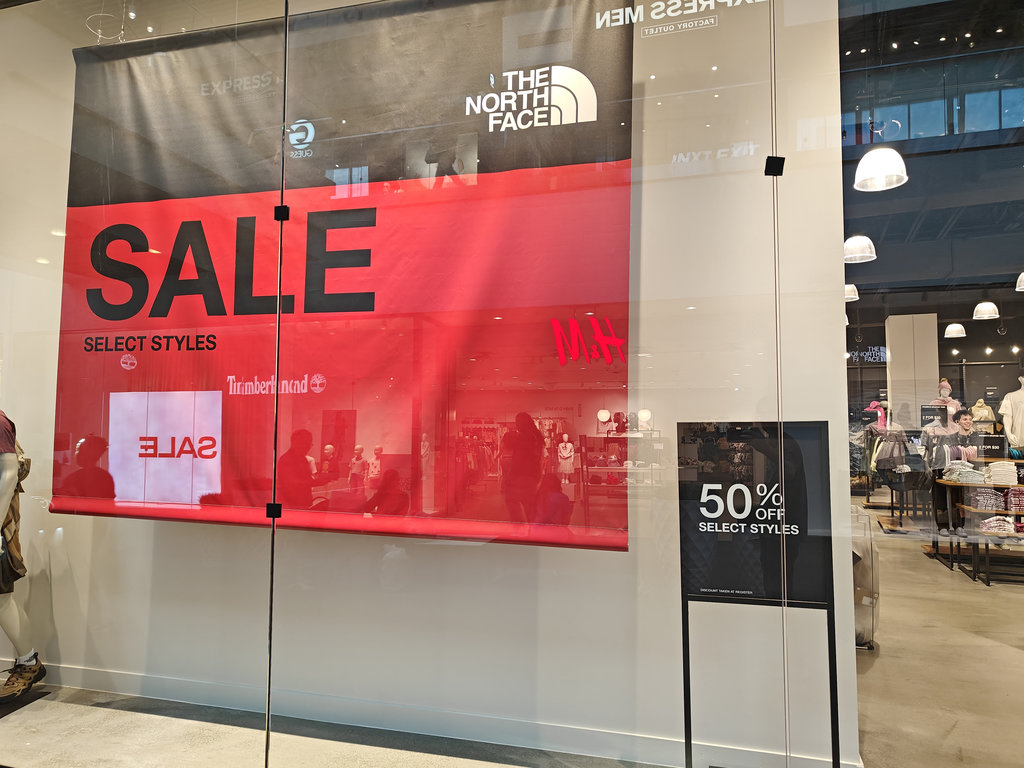}  
            \put(9,24){\color{green}\linethickness{1pt}\framebox(15,15)}
            \put(44,58){\color{yellow}\linethickness{1pt}\framebox(15,15)}
        \end{overpic}&
        \includegraphics[width=\resLen]{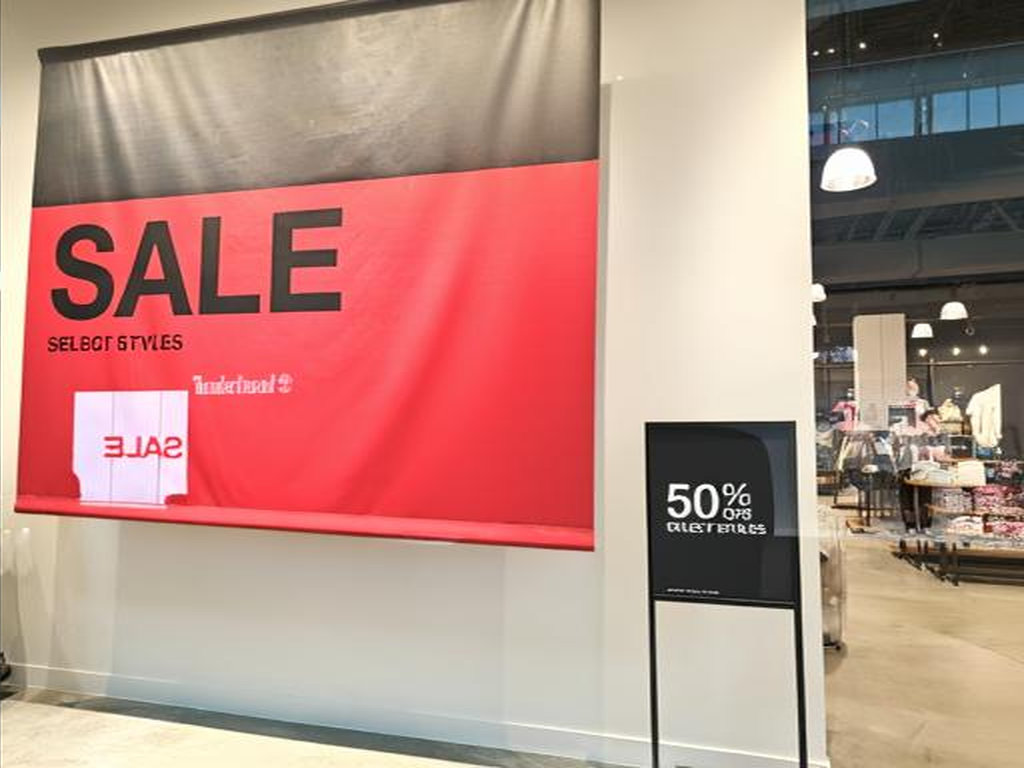}
        \\
        With reflection & Reflection removed
    \end{tabular}
    \vspace{-8pt}
    \caption{\textbf{Failure cases.}
         The white logo (in the yellow box) is on the background but was incorrectly removed as a reflection; the white reflection (in the green box) is mistakenly kept because it is treated as part of the background.
    }
    \vspace{-10pt}
    \label{fig:fail}
\end{figure}

Although our approach provides substantial improvements, several limitations remain. Path-traced rendering, despite its high physical fidelity, requires considerable computational resources, which could only generate dataset offline. In addition, white background areas are occasionally misclassified as highlight reflections to be removed and vice versa. See Figure \ref{fig:fail}.

\section{Conclusion}

We have presented a novel framework for generating physically accurate synthetic datasets for glass reflection removal. Our approach combines path-traced 3D glass models with real-world background imagery, providing a scalable solution for creating large-scale datasets with controlled parameterization. The evaluation using LoRA fine-tuning with Flux demonstrates the effectiveness of our synthetic data for real-world reflection removal tasks.
Our work addresses the critical need for high-quality training data in glass reflection removal and provides a foundation for future research in this important area of computer vision.

{
    \small
    \bibliographystyle{ieeenat_fullname}
    \bibliography{main}

@String(ECCV= {Eur. Conf. Comput. Vis.})

@String(TOG= {ACM Trans. Graph.})

@String(AAAI = {AAAI})

@String(ECCV  = {ECCV})

@String(TOG   = {ACM TOG})

@article{amanlou2022single,
  title={Single-image reflection removal using deep learning: a systematic review},
  author={Amanlou, Ali and Suratgar, Amir Abolfazl and Tavoosi, Jafar and Mohammadzadeh, Ardashir and Mosavi, Amir},
  journal={IEEE Access},
  volume={10},
  pages={29937--29953},
  year={2022},
  publisher={IEEE}
}

@article{batifol2025flux,
  title={FLUX. 1 Kontext: Flow Matching for In-Context Image Generation and Editing in Latent Space},
  author={Batifol, Stephen and Blattmann, Andreas and Boesel, Frederic and Consul, Saksham and Diagne, Cyril and Dockhorn, Tim and English, Jack and English, Zion and Esser, Patrick and Kulal, Sumith and others},
  journal={arXiv e-prints},
  pages={arXiv--2506},
  year={2025}
}

@article{cai2025f2t2,
  title={F2t2-hit: A u-shaped fft transformer and hierarchical transformer for reflection removal},
  author={Cai, Jie and Yang, Kangning and Ouyang, Ling and Fu, Lan and Ding, Jiaming and Sun, Huiming and Ho, Chiu Man and Meng, Zibo},
  journal={arXiv preprint arXiv:2506.05489},
  year={2025}
}

@article{chang2020siamese,
  title={Siamese dense network for reflection removal with flash and no-flash image pairs},
  author={Chang, Yakun and Jung, Cheolkon and Sun, Jun and Wang, Fengqiao},
  journal={International Journal of Computer Vision},
  volume={128},
  number={6},
  pages={1673--1698},
  year={2020},
  publisher={Springer}
}

@inproceedings{chang2021single,
  title={Single image reflection removal with edge guidance, reflection classifier, and recurrent decomposition},
  author={Chang, Ya-Chu and Lu, Chia-Ni and Cheng, Chia-Chi and Chiu, Wei-Chen},
  booktitle={Proceedings of the IEEE/CVF winter conference on applications of computer vision},
  pages={2033--2042},
  year={2021}
}

@inproceedings{chen2025firm,
  title={FIRM: Flexible Interactive Reflection ReMoval},
  author={Chen, Xiao and Jiang, Xudong and Tao, Yunkang and Lei, Zhen and Li, Qing and Lei, Chenyang and Zhang, Zhaoxiang},
  booktitle={Proceedings of the AAAI Conference on Artificial Intelligence},
  volume={39},
  number={2},
  pages={2230--2238},
  year={2025}
}

@article{chen2025high,
  title={High-resolution image reflection removal by Laplacian-based component-aware transformer},
  author={Chen, Songnan and Feng, Zhaoxu},
  journal={Scientific Reports},
  volume={15},
  number={1},
  pages={9972},
  year={2025},
  publisher={Nature Publishing Group UK London}
}

@inproceedings{dong2021location,
  title={Location-aware single image reflection removal},
  author={Dong, Zheng and Xu, Ke and Yang, Yin and Bao, Hujun and Xu, Weiwei and Lau, Rynson WH},
  booktitle={Proceedings of the IEEE/CVF international conference on computer vision},
  pages={5017--5026},
  year={2021}
}

@article{everingham2010pascal,
  title={The pascal visual object classes (voc) challenge},
  author={Everingham, Mark and Van Gool, Luc and Williams, Christopher KI and Winn, John and Zisserman, Andrew},
  journal={International journal of computer vision},
  volume={88},
  number={2},
  pages={303--338},
  year={2010},
  publisher={Springer}
}

@inproceedings{fan2017generic,
  title={A generic deep architecture for single image reflection removal and image smoothing},
  author={Fan, Qingnan and Yang, Jiaolong and Hua, Gang and Chen, Baoquan and Wipf, David},
  booktitle={Proceedings of the IEEE International Conference on Computer Vision},
  pages={3238--3247},
  year={2017}
}

@article{guo2018position,
  title={{Position-free Monte Carlo simulation for arbitrary layered BSDFs}},
  author={Guo, Yu and Ha{\v{s}}an, Milo{\v{s}} and Zhao, Shuang},
  journal={ACM Transactions on Graphics (ToG)},
  volume={37},
  number={6},
  pages={1--14},
  year={2018},
  publisher={ACM New York, NY, USA}
}

@inproceedings{guo2025epbr,
  title={{ePBR: Extended PBR materials in image synthesis}},
  author={Guo, Yu and Lao, Zhiqiang and Song, Xiyun and Zhou, Yubin and Lin, Zongfang and Yu, Heather},
  booktitle={Proceedings of the Computer Vision and Pattern Recognition Conference},
  pages={327--336},
  year={2025}
}

@article{he2023reflection,
  title={Reflection intensity guided single image reflection removal and transmission recovery},
  author={He, Lingzhi and Li, Feng and Cong, Runmin and Zhao, Yao},
  journal={IEEE Transactions on Multimedia},
  volume={26},
  pages={5026--5039},
  year={2023},
  publisher={IEEE}
}

@inproceedings{hong2024differ,
  title={L-differ: Single image reflection removal with language-based diffusion model},
  author={Hong, Yuchen and Zhong, Haofeng and Weng, Shuchen and Liang, Jinxiu and Shi, Boxin},
  booktitle={European Conference on Computer Vision},
  pages={58--76},
  year={2024},
  organization={Springer}
}

@article{hu2021trash,
  title={Trash or treasure? an interactive dual-stream strategy for single image reflection separation},
  author={Hu, Qiming and Guo, Xiaojie},
  journal={Advances in Neural Information Processing Systems},
  volume={34},
  pages={24683--24694},
  year={2021}
}

@inproceedings{hu2023single,
  title={Single image reflection separation via component synergy},
  author={Hu, Qiming and Guo, Xiaojie},
  booktitle={Proceedings of the IEEE/CVF international conference on computer vision},
  pages={13138--13147},
  year={2023}
}

@article{hu2024single,
  title={Single image reflection separation via dual-stream interactive transformers},
  author={Hu, Qiming and Wang, Hainuo and Guo, Xiaojie},
  journal={Advances in Neural Information Processing Systems},
  volume={37},
  pages={55228--55248},
  year={2024}
}

@article{hu2025dereflection,
  title={Dereflection Any Image with Diffusion Priors and Diversified Data},
  author={Hu, Jichen and Yang, Chen and Zhou, Zanwei and Fang, Jiemin and Yang, Xiaokang and Tian, Qi and Shen, Wei},
  journal={arXiv preprint arXiv:2503.17347},
  year={2025}
}

@article{huang2024context,
  title={In-context lora for diffusion transformers},
  author={Huang, Lianghua and Wang, Wei and Wu, Zhi-Fan and Shi, Yupeng and Dou, Huanzhang and Liang, Chen and Feng, Yutong and Liu, Yu and Zhou, Jingren},
  journal={arXiv preprint arXiv:2410.23775},
  year={2024}
}

@article{huynh2008scope,
  title={Scope of validity of PSNR in image/video quality assessment},
  author={Huynh-Thu, Quan and Ghanbari, Mohammed},
  journal={Electronics letters},
  volume={44},
  number={13},
  pages={800--801},
  year={2008},
  publisher={IET}
}

@inproceedings{kee2025removing,
  title={Removing reflections from raw photos},
  author={Kee, Eric and Pikielny, Adam and Blackburn-Matzen, Kevin and Levoy, Marc},
  booktitle={Proceedings of the Computer Vision and Pattern Recognition Conference},
  pages={161--171},
  year={2025}
}

@inproceedings{kim2020single,
  title={Single image reflection removal with physically-based training images},
  author={Kim, Soomin and Huo, Yuchi and Yoon, Sung-Eui},
  booktitle={Proceedings of the IEEE/CVF conference on computer vision and pattern recognition},
  pages={5164--5173},
  year={2020}
}

@inproceedings{lei2020polarized,
  title={Polarized reflection removal with perfect alignment in the wild},
  author={Lei, Chenyang and Huang, Xuhua and Zhang, Mengdi and Yan, Qiong and Sun, Wenxiu and Chen, Qifeng},
  booktitle={Proceedings of the IEEE/CVF conference on computer vision and pattern recognition},
  pages={1750--1758},
  year={2020}
}

@inproceedings{lei2021robust,
  title={Robust reflection removal with reflection-free flash-only cues},
  author={Lei, Chenyang and Chen, Qifeng},
  booktitle={Proceedings of the IEEE/CVF Conference on Computer Vision and Pattern Recognition},
  pages={14811--14820},
  year={2021}
}

@inproceedings{lei2022categorized,
  title={A categorized reflection removal dataset with diverse real-world scenes},
  author={Lei, Chenyang and Huang, Xuhua and Qi, Chenyang and Zhao, Yankun and Sun, Wenxiu and Yan, Qiong and Chen, Qifeng},
  booktitle={Proceedings of the IEEE/CVF Conference on Computer Vision and Pattern Recognition},
  pages={3040--3048},
  year={2022}
}

@article{lei2023robust,
  title={Robust reflection removal with flash-only cues in the wild},
  author={Lei, Chenyang and Jiang, Xudong and Chen, Qifeng},
  journal={IEEE Transactions on Pattern Analysis and Machine Intelligence},
  volume={45},
  number={12},
  pages={15530--15545},
  year={2023},
  publisher={IEEE}
}

@inproceedings{li2020single,
  title={Single image reflection removal through cascaded refinement},
  author={Li, Chao and Yang, Yixiao and He, Kun and Lin, Stephen and Hopcroft, John E},
  booktitle={Proceedings of the IEEE/CVF conference on computer vision and pattern recognition},
  pages={3565--3574},
  year={2020}
}

@article{li2023two,
  title={Two-stage single image reflection removal with reflection-aware guidance},
  author={Li, Yu and Liu, Ming and Yi, Yaling and Li, Qince and Ren, Dongwei and Zuo, Wangmeng},
  journal={Applied Intelligence},
  volume={53},
  number={16},
  pages={19433--19448},
  year={2023},
  publisher={Springer}
}

@inproceedings{liu2020learning,
  title={Learning to see through obstructions},
  author={Liu, Yu-Lun and Lai, Wei-Sheng and Yang, Ming-Hsuan and Chuang, Yung-Yu and Huang, Jia-Bin},
  booktitle={Proceedings of the IEEE/CVF Conference on Computer Vision and Pattern Recognition},
  pages={14215--14224},
  year={2020}
}

@article{liu2022semantic,
  title={Semantic guided single image reflection removal},
  author={Liu, Yunfei and Li, Yu and You, Shaodi and Lu, Feng},
  journal={ACM Transactions on Multimedia Computing, Communications and Applications},
  volume={18},
  number={3s},
  pages={1--23},
  year={2022},
  publisher={ACM New York, NY}
}

@article{lyu2019reflection,
  title={Reflection separation using a pair of unpolarized and polarized images},
  author={Lyu, Youwei and Cui, Zhaopeng and Li, Si and Pollefeys, Marc and Shi, Boxin},
  journal={Advances in neural information processing systems},
  volume={32},
  year={2019}
}

@inproceedings{peebles2023scalable,
  title={Scalable diffusion models with transformers},
  author={Peebles, William and Xie, Saining},
  booktitle={Proceedings of the IEEE/CVF international conference on computer vision},
  pages={4195--4205},
  year={2023}
}

@inproceedings{prasad2021v,
  title={V-desirr: Very fast deep embedded single image reflection removal},
  author={Prasad, BH and Boregowda, Lokesh R and Mitra, Kaushik and Chowdhury, Sanjoy and others},
  booktitle={Proceedings of the IEEE/CVF International Conference on Computer Vision},
  pages={2390--2399},
  year={2021}
}

@inproceedings{rosh2024r2sfd,
  author = {Rosh, Green and B H, Pawan Prasad and Boregowda, Lokesh R. and Mitra, Kaushik},
  title = {R2SFD: Improving Single Image Reflection Removal using Semantic Feature Dictionary},
  year = {2024},
  publisher = {Association for Computing Machinery},
  booktitle = {Proceedings of the 32nd ACM International Conference on Multimedia},
  pages = {10277–10286},
  numpages = {10},
}

@inproceedings{song2023robust,
  title={Robust single image reflection removal against adversarial attacks},
  author={Song, Zhenbo and Zhang, Zhenyuan and Zhang, Kaihao and Luo, Wenhan and Fan, Zhaoxin and Ren, Wenqi and Lu, Jianfeng},
  booktitle={Proceedings of the IEEE/CVF conference on computer vision and pattern recognition},
  pages={24688--24698},
  year={2023}
}

@inproceedings{wan2017benchmarking,
  title={Benchmarking single-image reflection removal algorithms},
  author={Wan, Renjie and Shi, Boxin and Duan, Ling-Yu and Tan, Ah-Hwee and Kot, Alex C},
  booktitle={Proceedings of the IEEE international conference on computer vision},
  pages={3922--3930},
  year={2017}
}

@inproceedings{wan2018crrn,
  title={CRRN: Multi-scale guided concurrent reflection removal network},
  author={Wan, Renjie and Shi, Boxin and Duan, Ling-Yu and Tan, Ah-Hwee and Kot, Alex C},
  booktitle={Proceedings of the IEEE Conference on Computer Vision and Pattern Recognition},
  pages={4777--4785},
  year={2018}
}

@article{wan2019corrn,
  title={CoRRN: Cooperative reflection removal network},
  author={Wan, Renjie and Shi, Boxin and Li, Haoliang and Duan, Ling-Yu and Tan, Ah-Hwee and Kot, Alex C},
  journal={IEEE transactions on pattern analysis and machine intelligence},
  volume={42},
  number={12},
  pages={2969--2982},
  year={2019},
  publisher={IEEE}
}

@inproceedings{wan2020reflection,
  title={Reflection scene separation from a single image},
  author={Wan, Renjie and Shi, Boxin and Li, Haoliang and Duan, Ling-Yu and Kot, Alex C},
  booktitle={Proceedings of the IEEE/CVF Conference on Computer Vision and Pattern Recognition},
  pages={2398--2406},
  year={2020}
}

@article{wan2022benchmarking,
  title={Benchmarking single-image reflection removal algorithms},
  author={Wan, Renjie and Shi, Boxin and Li, Haoliang and Hong, Yuchen and Duan, Ling-Yu and Kot, Alex C},
  journal={IEEE Transactions on Pattern Analysis and Machine Intelligence},
  volume={45},
  number={2},
  pages={1424--1441},
  year={2022},
  publisher={IEEE}
}

@inproceedings{wang2003multiscale,
  title={Multiscale structural similarity for image quality assessment},
  author={Wang, Zhou and Simoncelli, Eero P and Bovik, Alan C},
  booktitle={The thrity-seventh asilomar conference on signals, systems \& computers, 2003},
  volume={2},
  pages={1398--1402},
  year={2003},
  organization={Ieee}
}

@article{wang2022background,
  title={Background scene recovery from an image looking through colored glass},
  author={Wang, Ce and Xu, Dejia and Wan, Renjie and He, Bin and Shi, Boxin and Duan, Ling-Yu},
  journal={IEEE Transactions on Multimedia},
  volume={25},
  pages={2876--2887},
  year={2022},
  publisher={IEEE}
}

@inproceedings{wang2025flash,
  title={Flash-split: 2d reflection removal with flash cues and latent diffusion separation},
  author={Wang, Tianfu and Xie, Mingyang and Cai, Haoming and Shah, Sachin and Metzler, Christopher A},
  booktitle={Proceedings of the Computer Vision and Pattern Recognition Conference},
  pages={5688--5698},
  year={2025}
}

@article{wang2025review,
  title={A review on learning based image reflection removal algorithms},
  author={Wang, Xin and Zhang, Yong and Xu, Junfeng and Gao, Jun},
  journal={Intelligent Data Analysis},
  volume={29},
  number={1},
  pages={5--27},
  year={2025},
  publisher={SAGE Publications Sage UK: London, England}
}

@inproceedings{wei2019single,
  title={Single image reflection removal exploiting misaligned training data and network enhancements},
  author={Wei, Kaixuan and Yang, Jiaolong and Fu, Ying and Wipf, David and Huang, Hua},
  booktitle={Proceedings of the IEEE/CVF Conference on Computer Vision and Pattern Recognition},
  pages={8178--8187},
  year={2019}
}

@inproceedings{wei2024dereflectformer,
  title={Dereflectformer: vision transformers for single image reflection removal},
  author={Wei, Ao and Zhang, Hanbin and Zhao, Erhu},
  booktitle={International conference on pattern recognition},
  pages={257--274},
  year={2024},
  organization={Springer}
}

@inproceedings{wen2019single,
  title={Single image reflection removal beyond linearity},
  author={Wen, Qiang and Tan, Yinjie and Qin, Jing and Liu, Wenxi and Han, Guoqiang and He, Shengfeng},
  booktitle={Proceedings of the IEEE/CVF Conference on Computer Vision and Pattern Recognition},
  pages={3771--3779},
  year={2019}
}

@inproceedings{wieschollek2018separating,
  title={Separating reflection and transmission images in the wild},
  author={Wieschollek, Patrick and Gallo, Orazio and Gu, Jinwei and Kautz, Jan},
  booktitle={Proceedings of the European Conference on Computer Vision (ECCV)},
  pages={89--104},
  year={2018}
}

@inproceedings{yang2018seeing,
  title={Seeing deeply and bidirectionally: A deep learning approach for single image reflection removal},
  author={Yang, Jie and Gong, Dong and Liu, Lingqiao and Shi, Qinfeng},
  booktitle={Proceedings of the european conference on computer vision (ECCV)},
  pages={654--669},
  year={2018}
}

@inproceedings{yang2025ntire,
  title={NTIRE 2025 challenge on single image reflection removal in the wild: Datasets, methods and results},
  author={Yang, Kangning and Cai, Jie and Ouyang, Ling and Vasluianu, Florin-Alexandru and Timofte, Radu and Ding, Jiaming and Sun, Huiming and Fu, Lan and Li, Jinlong and Ho, Chiu Man and others},
  booktitle={Proceedings of the Computer Vision and Pattern Recognition Conference},
  pages={1301--1311},
  year={2025}
}

@article{yang2025survey,
  title={Survey on single-image reflection removal using deep learning techniques},
  author={Yang, Kangning and Sun, Huiming and Cai, Jie and Fu, Lan and Ding, Jiaming and Li, Jinlong and Ho, Chiu Man and Meng, Zibo},
  journal={arXiv preprint arXiv:2502.08836},
  year={2025}
}

@inproceedings{yao2025polarfree,
  title={PolarFree: Polarization-based Reflection-Free Imaging},
  author={Yao, Mingde and Wang, Menglu and Tam, King-Man and Li, Lingen and Xue, Tianfan and Gu, Jinwei},
  booktitle={Proceedings of the Computer Vision and Pattern Recognition Conference},
  pages={10890--10899},
  year={2025}
}

@article{zhang2017virtual,
  title={A virtual try-on system for prescription eyeglasses},
  author={Zhang, Qian and Guo, Yu and Laffont, Pierre-Yves and Martin, Tobias and Gross, Markus},
  journal={IEEE computer graphics and applications},
  volume={37},
  number={4},
  pages={84--93},
  year={2017},
  publisher={IEEE}
}

@inproceedings{zhang2018single,
  title={Single image reflection separation with perceptual losses},
  author={Zhang, Xuaner and Ng, Ren and Chen, Qifeng},
  booktitle={Proceedings of the IEEE conference on computer vision and pattern recognition},
  pages={4786--4794},
  year={2018}
}

@inproceedings{zhang2018unreasonable,
  title={The unreasonable effectiveness of deep features as a perceptual metric},
  author={Zhang, Richard and Isola, Phillip and Efros, Alexei A and Shechtman, Eli and Wang, Oliver},
  booktitle={Proceedings of the IEEE conference on computer vision and pattern recognition},
  pages={586--595},
  year={2018}
}

@inproceedings{zhao2025reversible,
  title={Reversible decoupling network for single image reflection removal},
  author={Zhao, Hao and Li, Mingjia and Hu, Qiming and Guo, Xiaojie},
  booktitle={Proceedings of the Computer Vision and Pattern Recognition Conference},
  pages={26430--26439},
  year={2025}
}

@inproceedings{zheng2021single,
  title={Single image reflection removal with absorption effect},
  author={Zheng, Qian and Shi, Boxin and Chen, Jinnan and Jiang, Xudong and Duan, Ling-Yu and Kot, Alex C},
  booktitle={Proceedings of the IEEE/CVF conference on computer vision and pattern recognition},
  pages={13395--13404},
  year={2021}
}

@inproceedings{zhong2024language,
  title={Language-guided image reflection separation},
  author={Zhong, Haofeng and Hong, Yuchen and Weng, Shuchen and Liang, Jinxiu and Shi, Boxin},
  booktitle={Proceedings of the IEEE/CVF Conference on Computer Vision and Pattern Recognition},
  pages={24913--24922},
  year={2024}
}

@article{zhu2023hue,
  title={Hue guidance network for single image reflection removal},
  author={Zhu, Yurui and Fu, Xueyang and Zhang, Zheyu and Liu, Aiping and Xiong, Zhiwei and Zha, Zheng-Jun},
  journal={IEEE transactions on neural networks and learning systems},
  volume={35},
  number={10},
  pages={13701--13712},
  year={2023},
  publisher={IEEE}
}

@inproceedings{zhu2024revisiting,
  title={Revisiting single image reflection removal in the wild},
  author={Zhu, Yurui and Fu, Xueyang and Jiang, Peng-Tao and Zhang, Hao and Sun, Qibin and Chen, Jinwei and Zha, Zheng-Jun and Li, Bo},
  booktitle={Proceedings of the IEEE/CVF Conference on Computer Vision and Pattern Recognition},
  pages={25468--25478},
  year={2024}
}

@inproceedings{zou2020deep,
  title={Deep adversarial decomposition: A unified framework for separating superimposed images},
  author={Zou, Zhengxia and Lei, Sen and Shi, Tianyang and Shi, Zhenwei and Ye, Jieping},
  booktitle={Proceedings of the IEEE/CVF conference on computer vision and pattern recognition},
  pages={12806--12816},
  year={2020}
}
}
\vspace{110pt}
\appendix
\section*{Appendix}

\begin{figure}[h]
    \vspace{-10pt}
    \centering
    \setlength{\resLen}{0.19\linewidth}
    \addtolength{\tabcolsep}{-5pt}
    \begin{tabular}{ccccc}
        \includegraphics[width=\resLen]{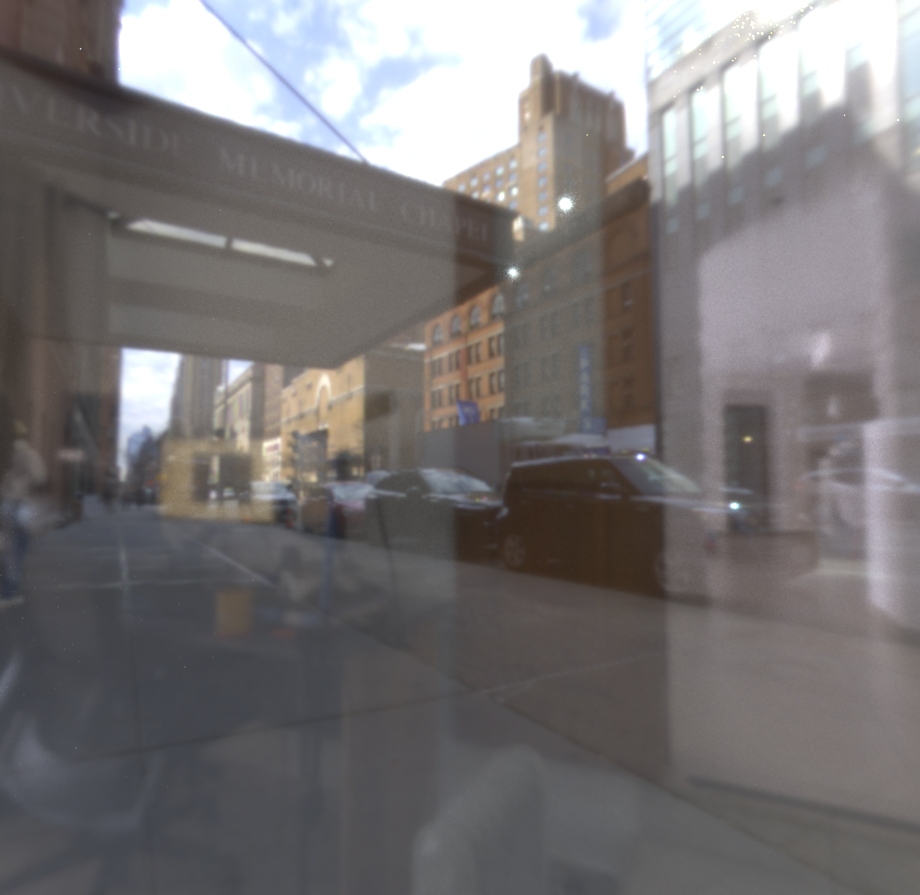} &
        \includegraphics[width=\resLen]{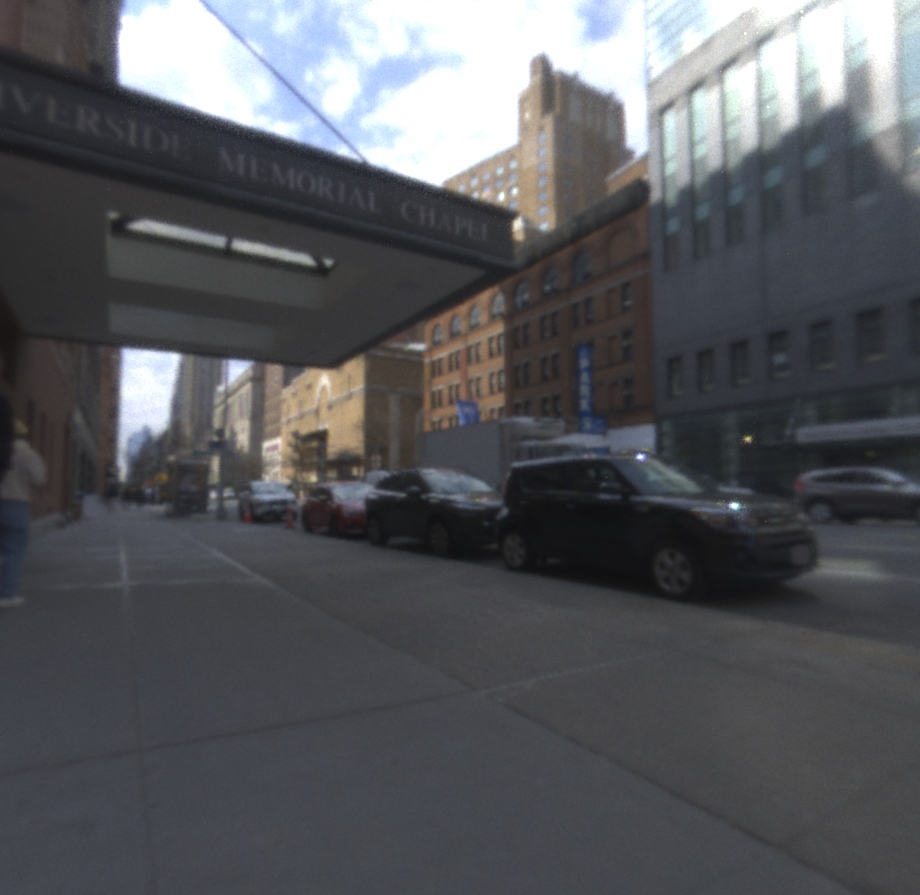} &
        \includegraphics[width=\resLen]{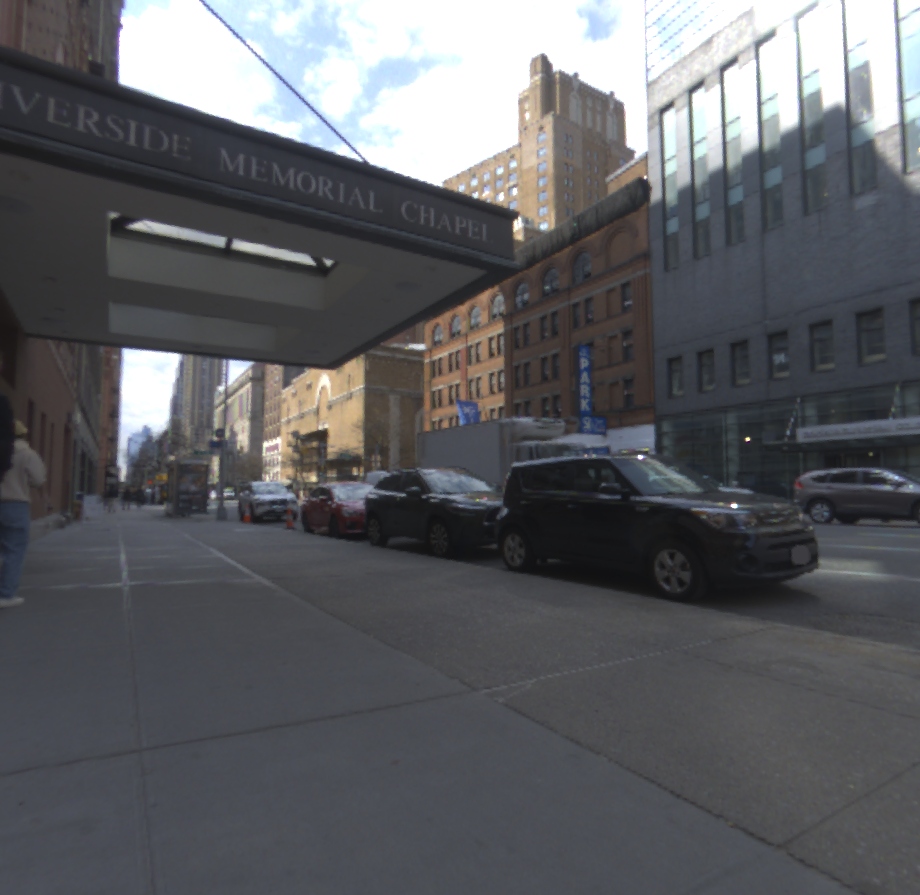} &
        \includegraphics[width=\resLen]{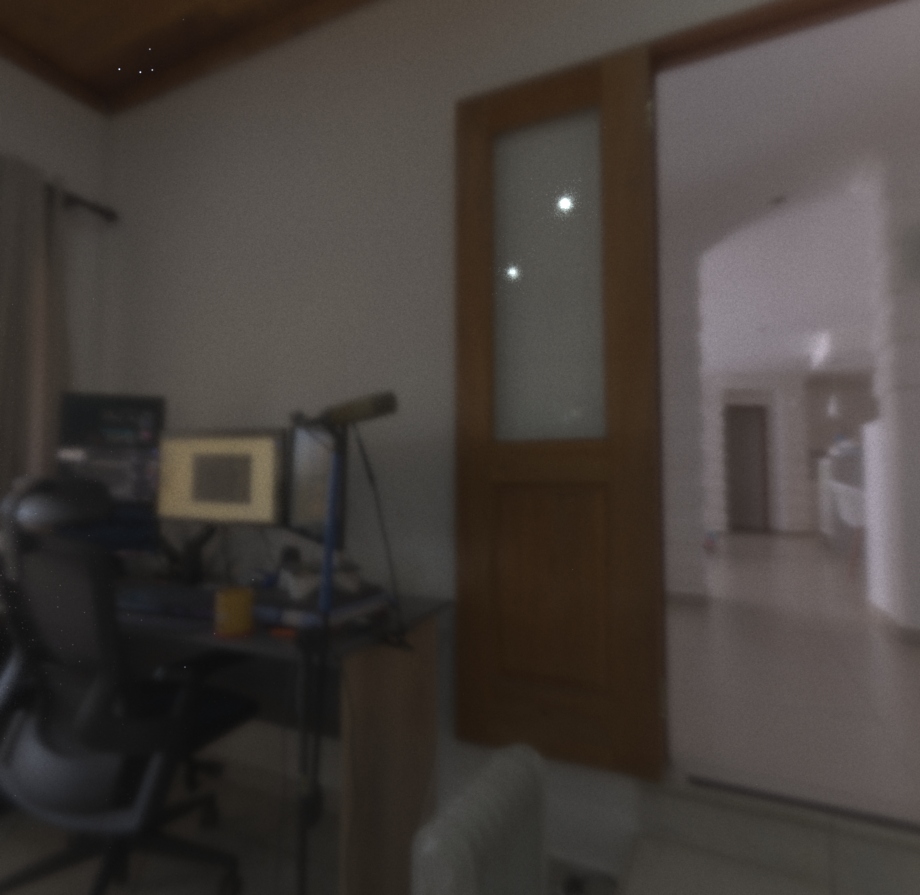} &
        \includegraphics[width=\resLen]{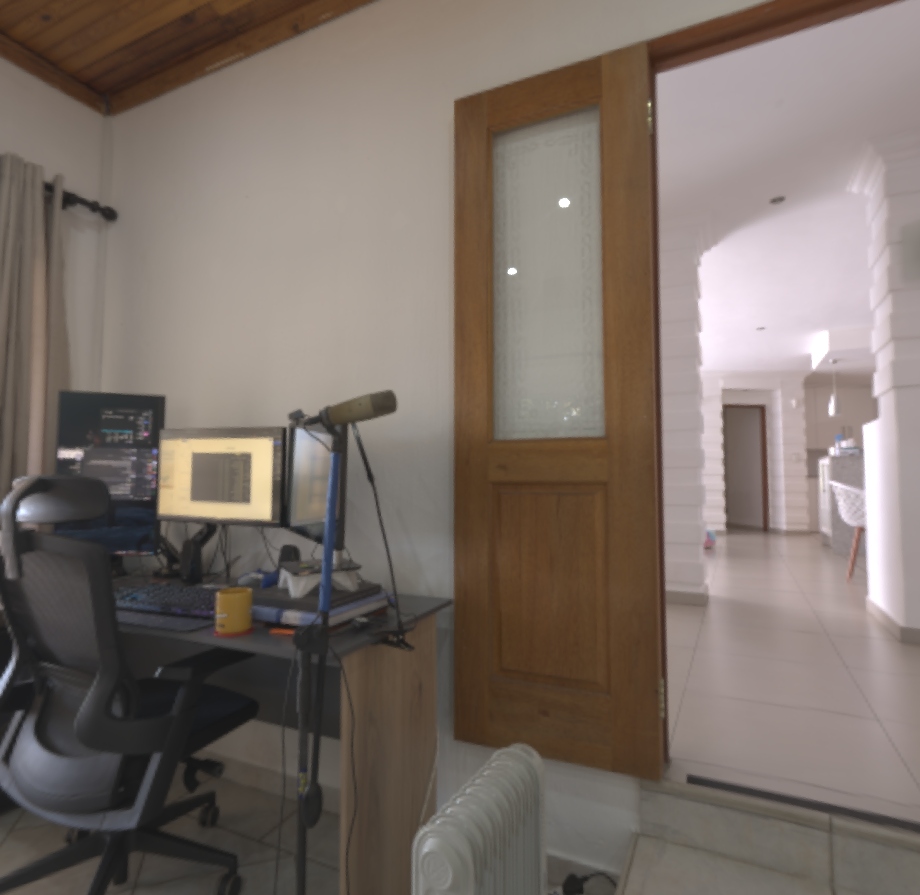} 
        \\
        \includegraphics[width=\resLen]{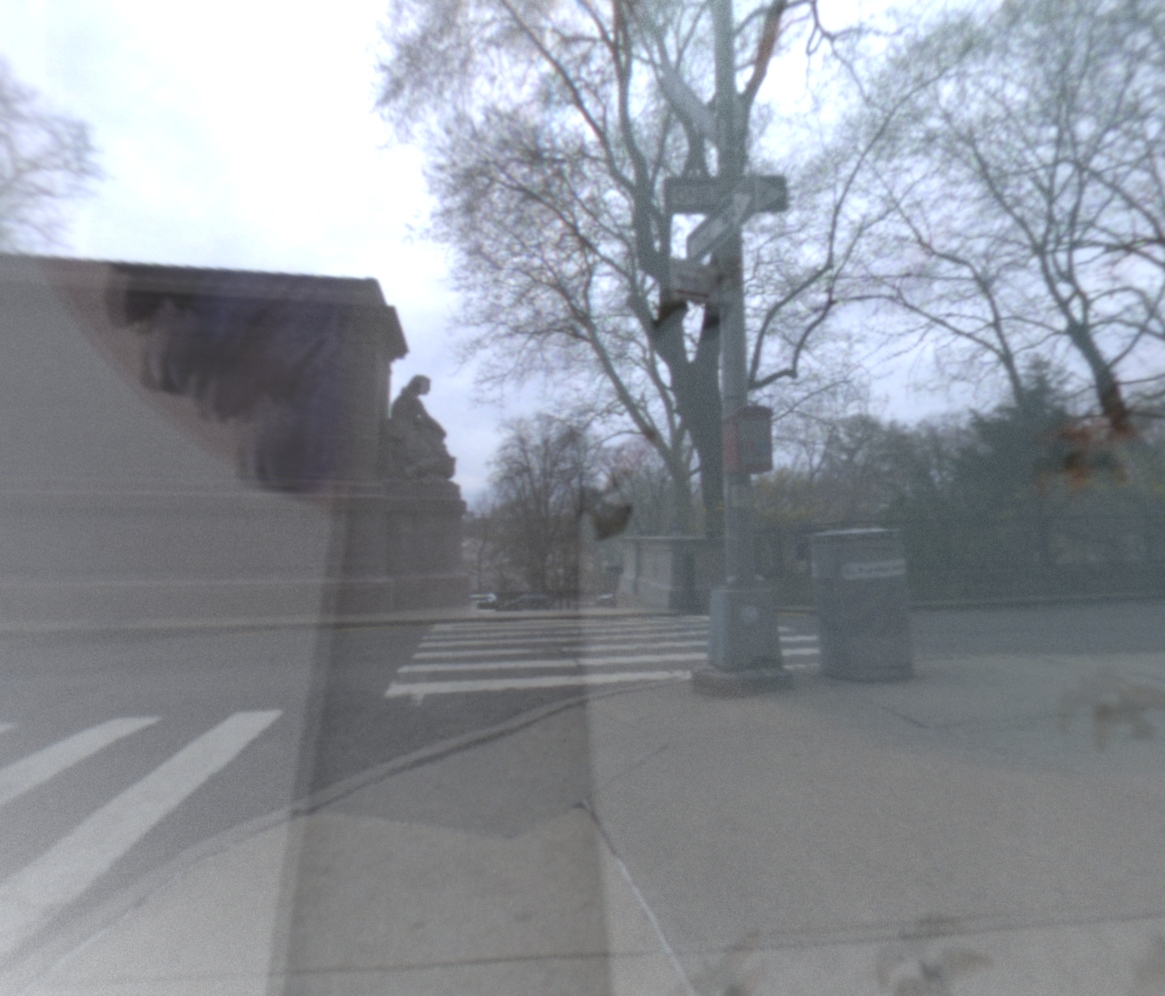} &
        \includegraphics[width=\resLen]{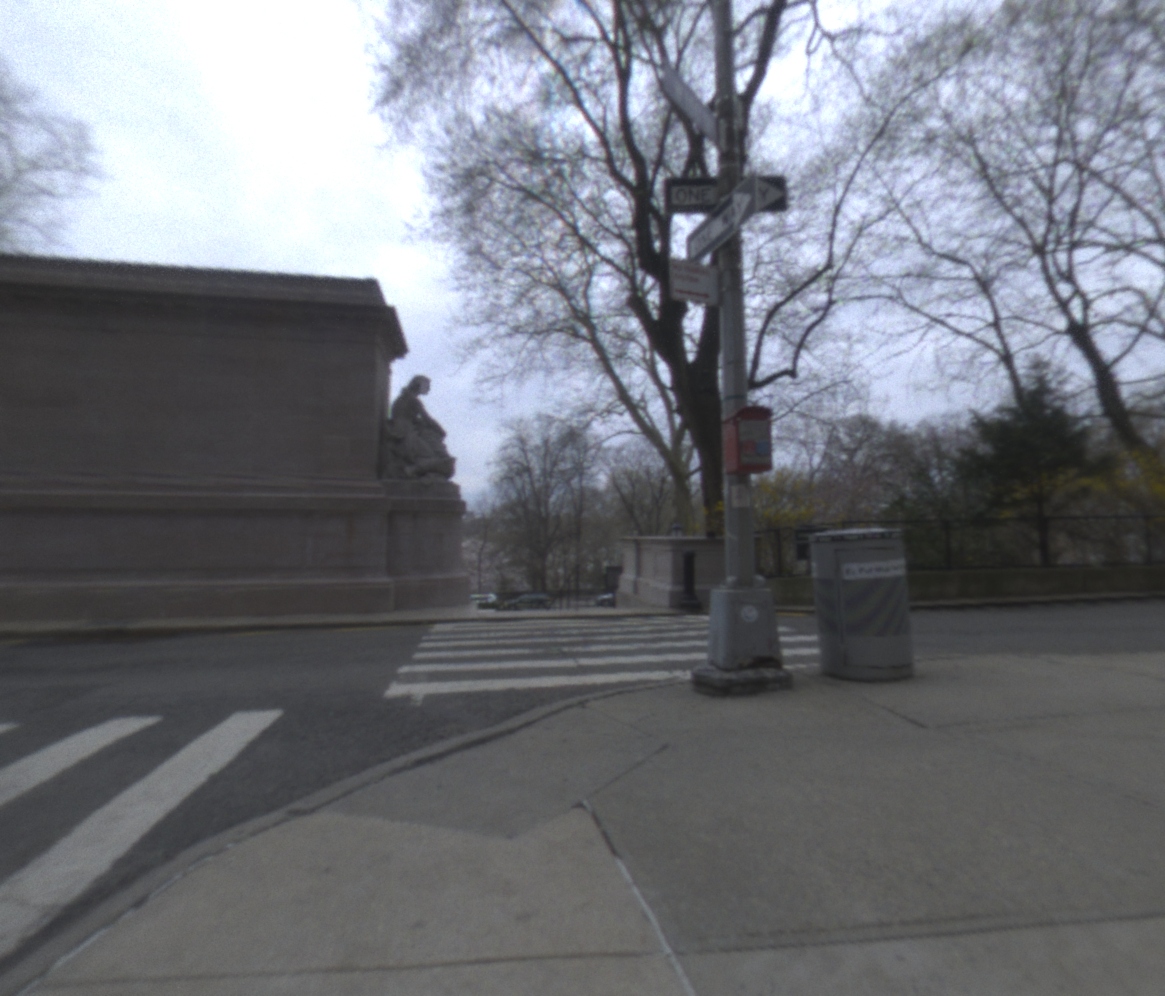} &
        \includegraphics[width=\resLen]{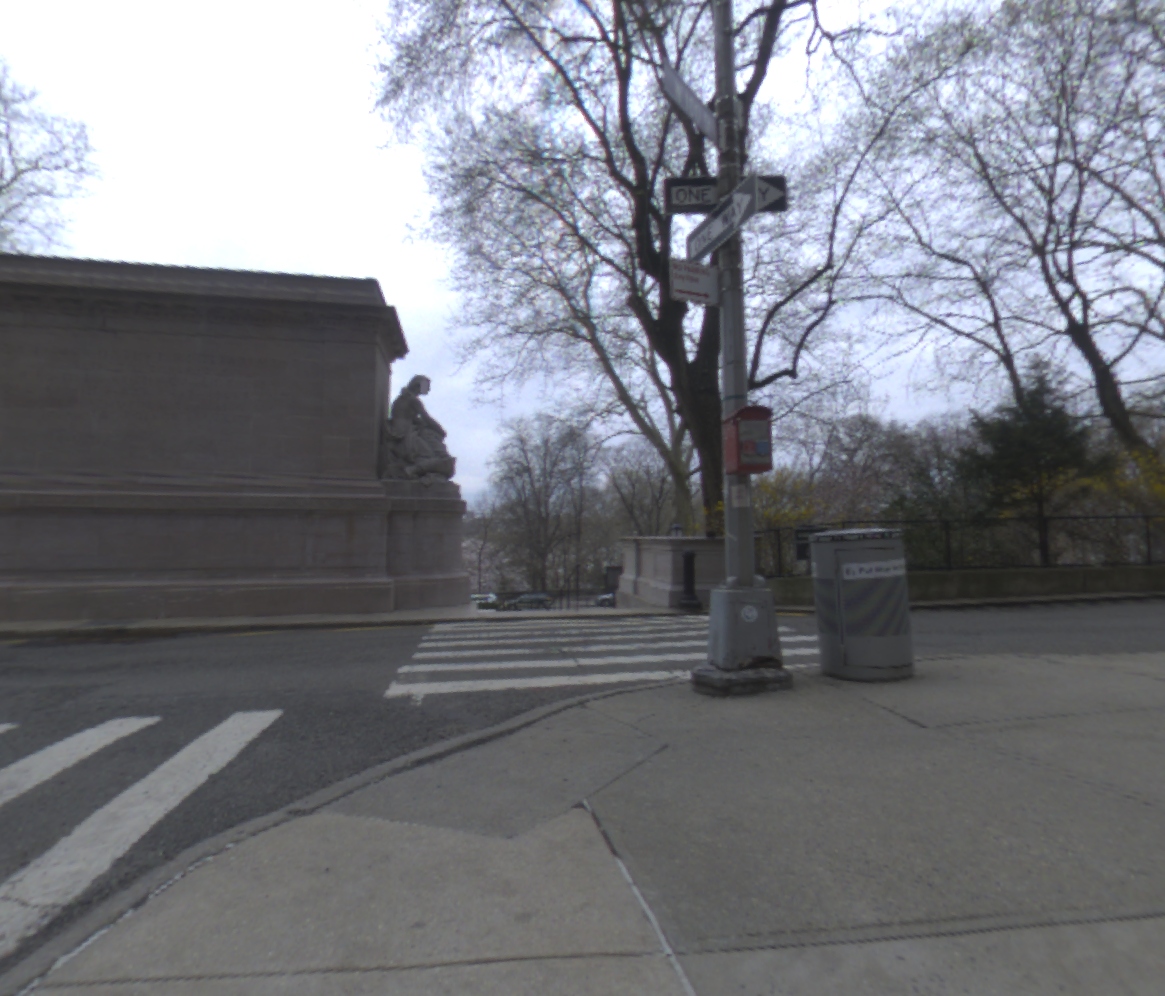} &
        \includegraphics[width=\resLen]{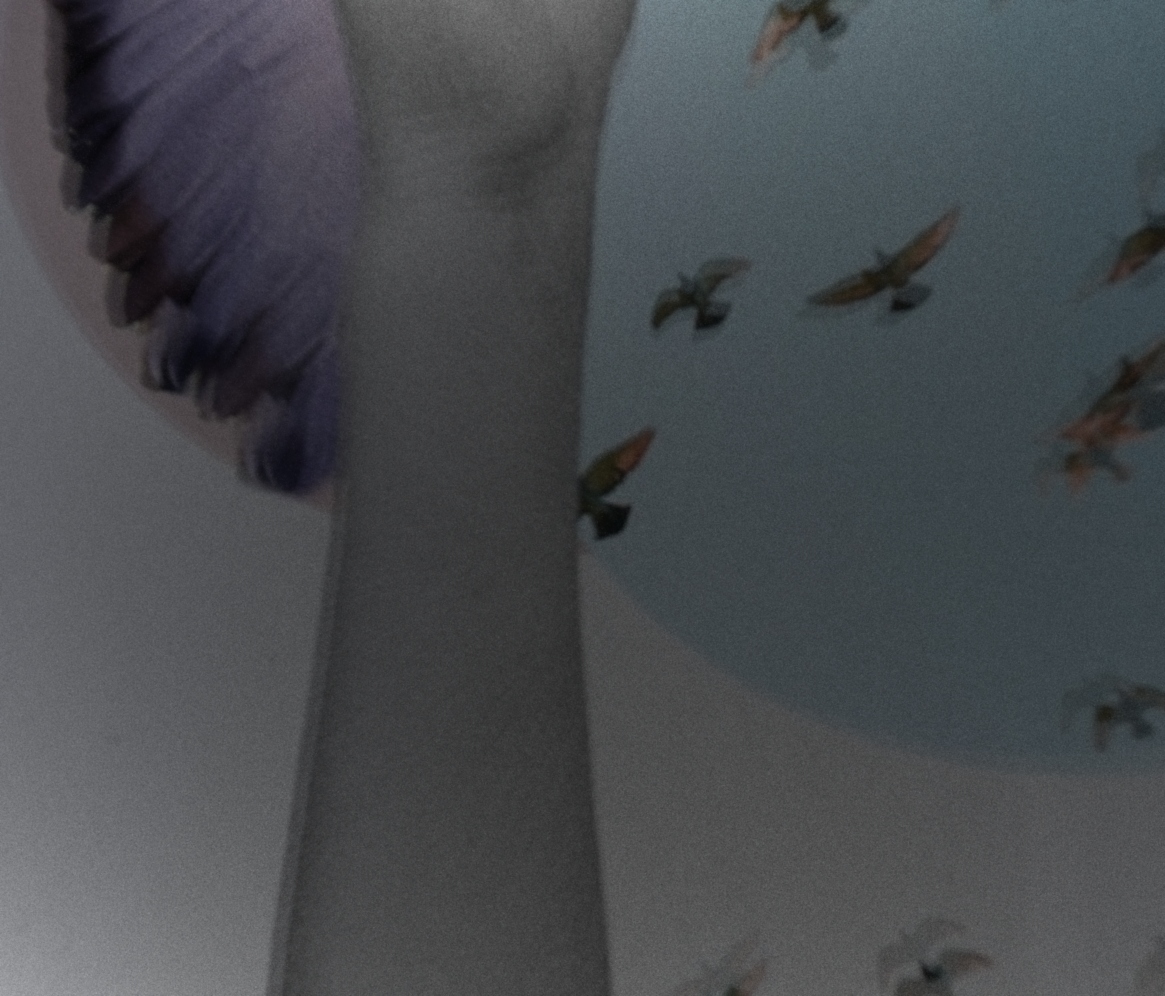} &
        \includegraphics[width=\resLen]{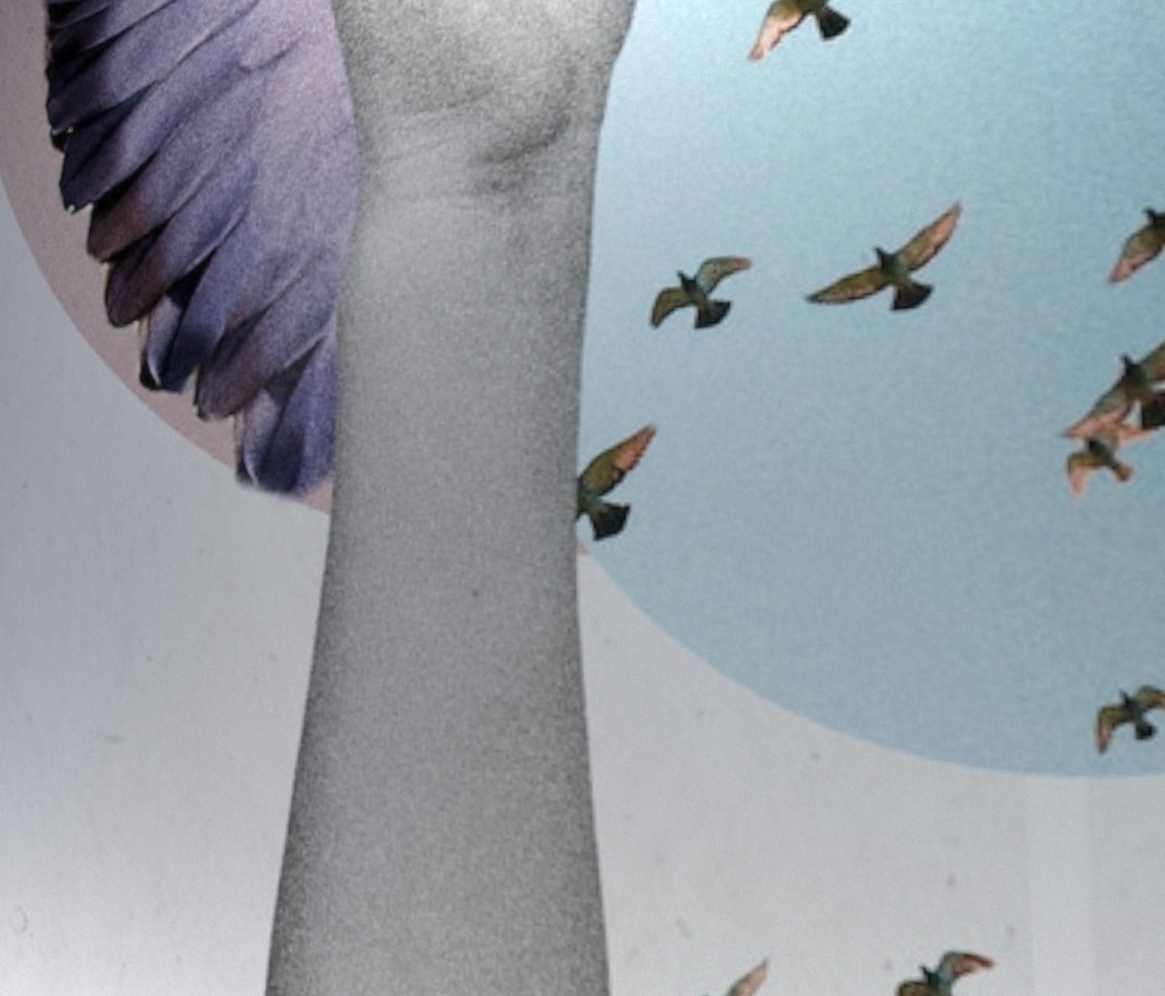} 
        \\
        \includegraphics[width=\resLen]{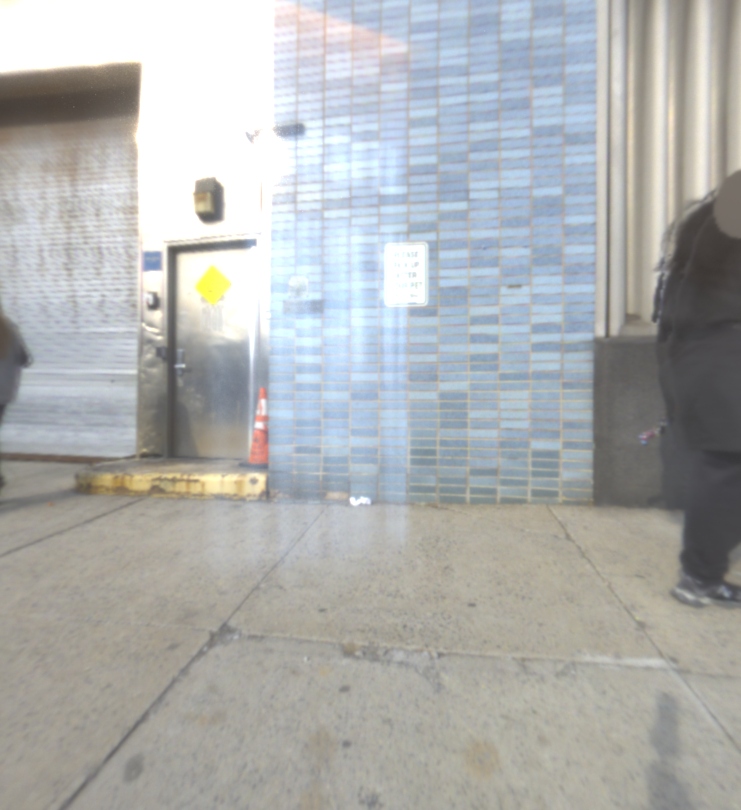} &
        \includegraphics[width=\resLen]{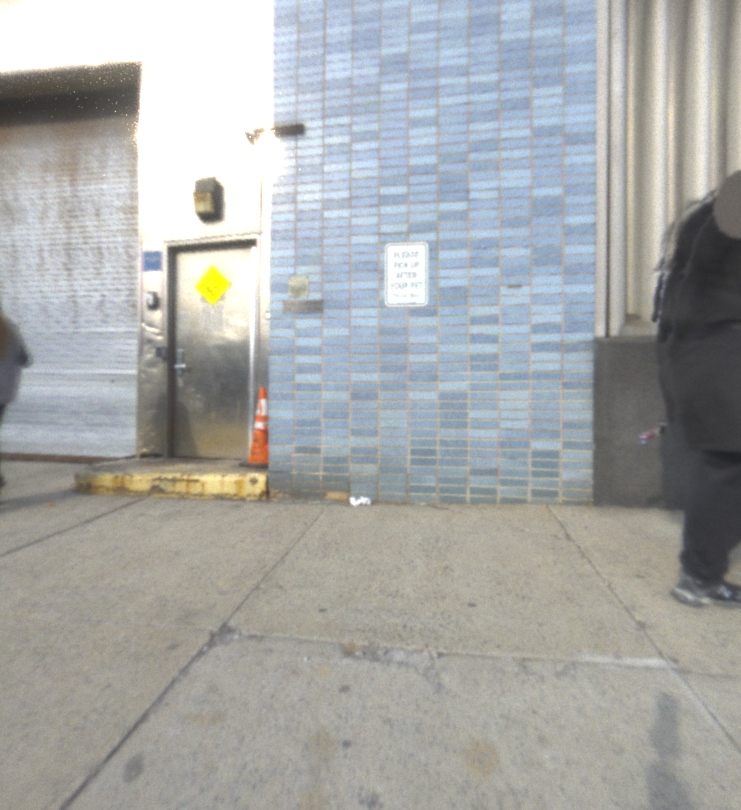} &
        \includegraphics[width=\resLen]{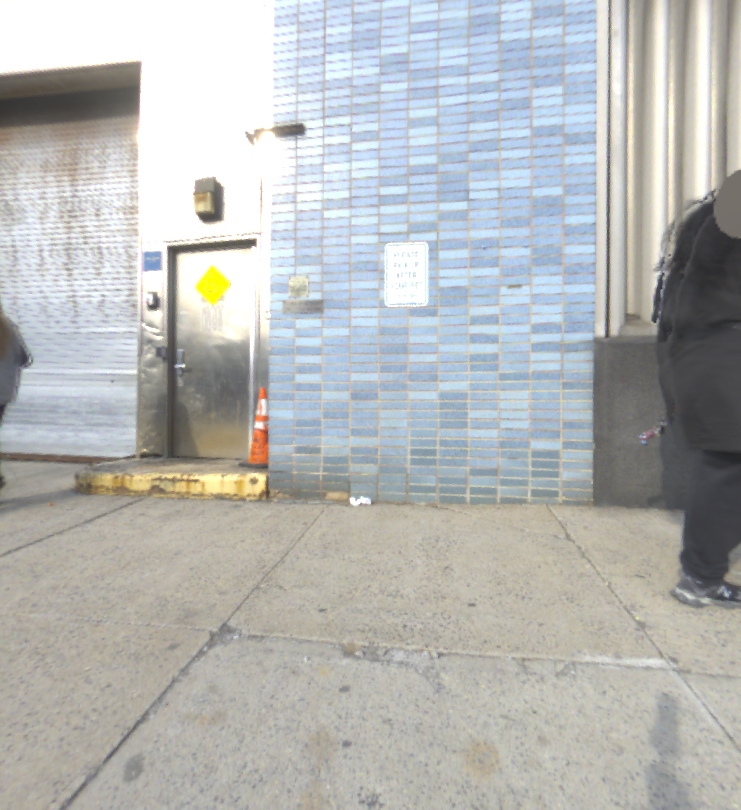} &
        \includegraphics[width=\resLen]{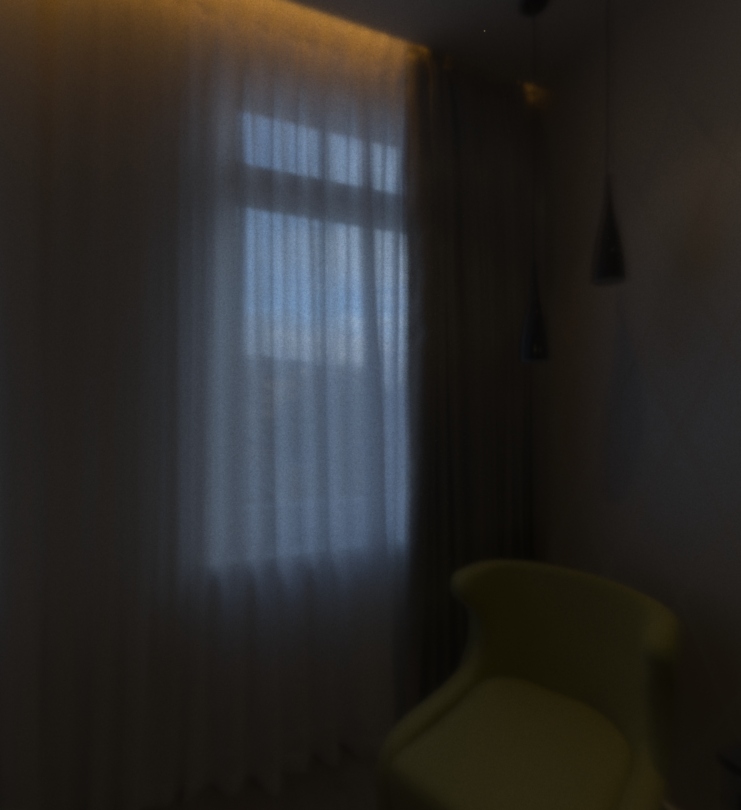} &
        \includegraphics[width=\resLen]{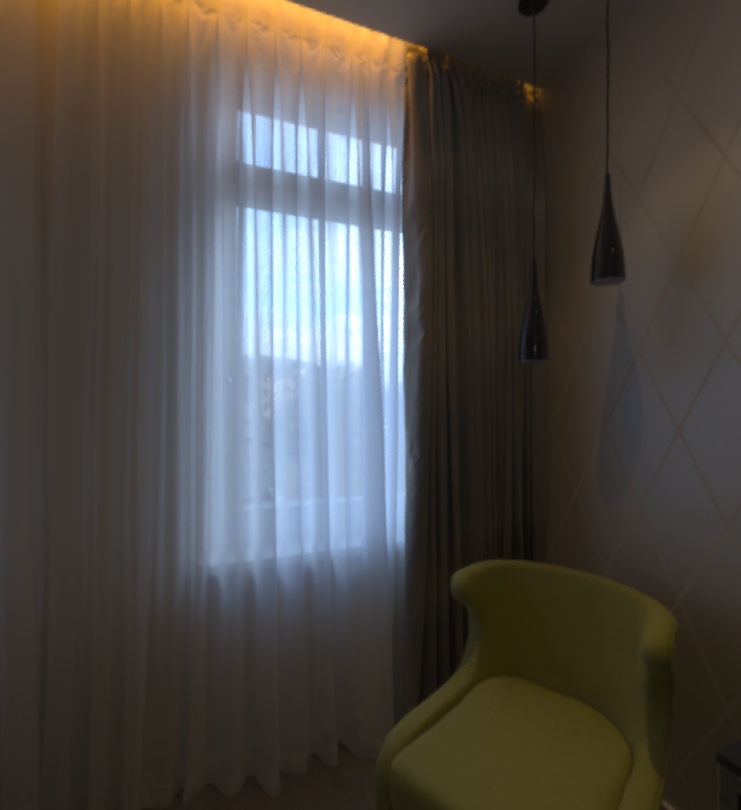} 
        \\
        \includegraphics[width=\resLen]{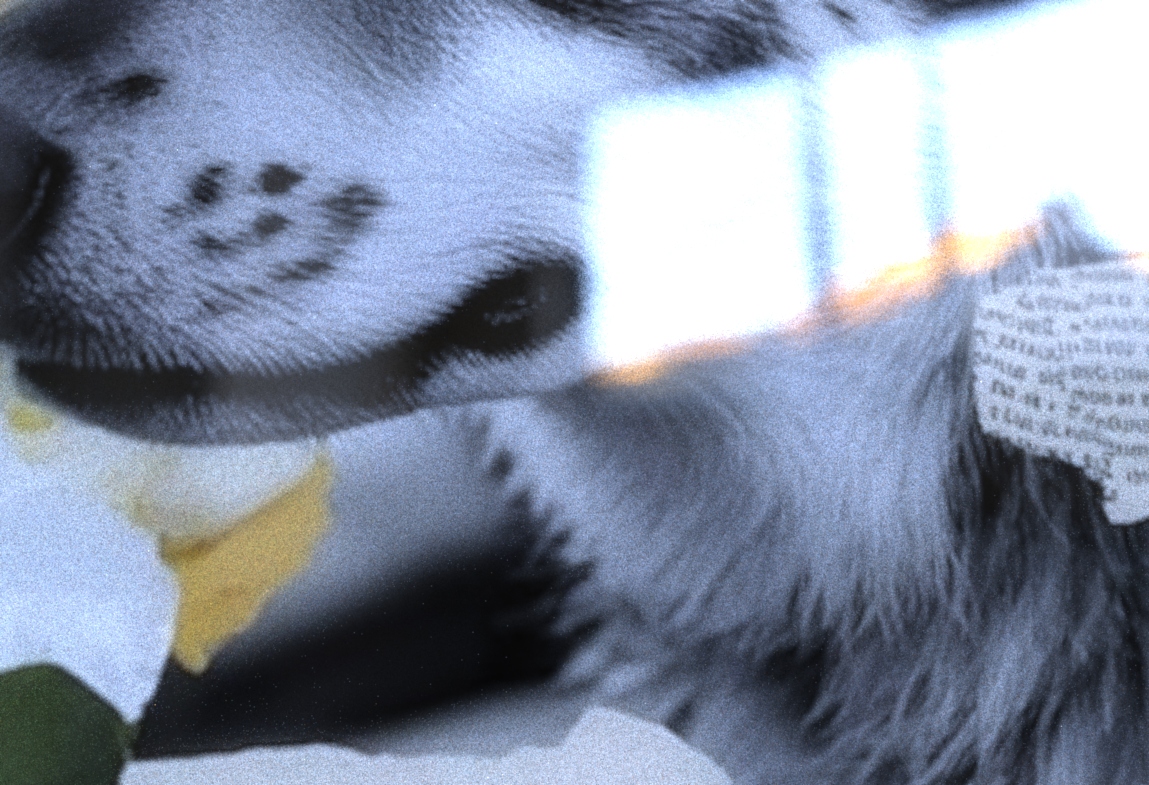} &
        \includegraphics[width=\resLen]{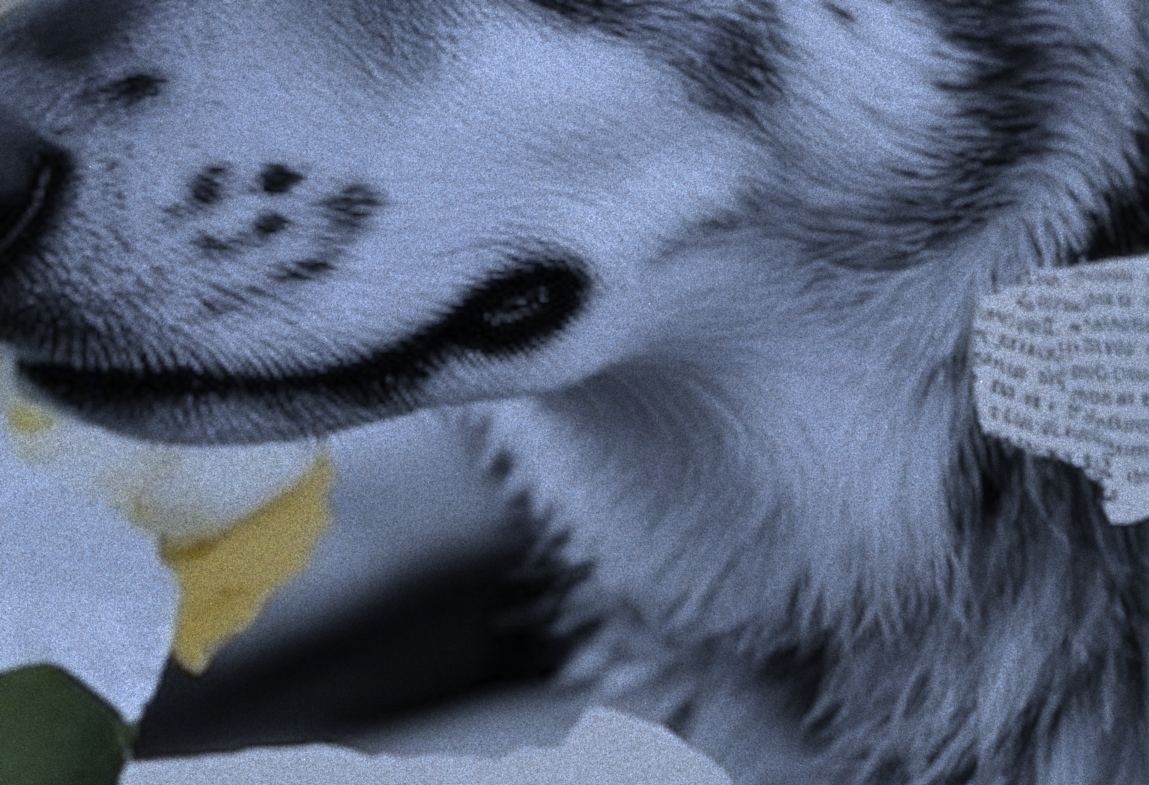} &
        \includegraphics[width=\resLen]{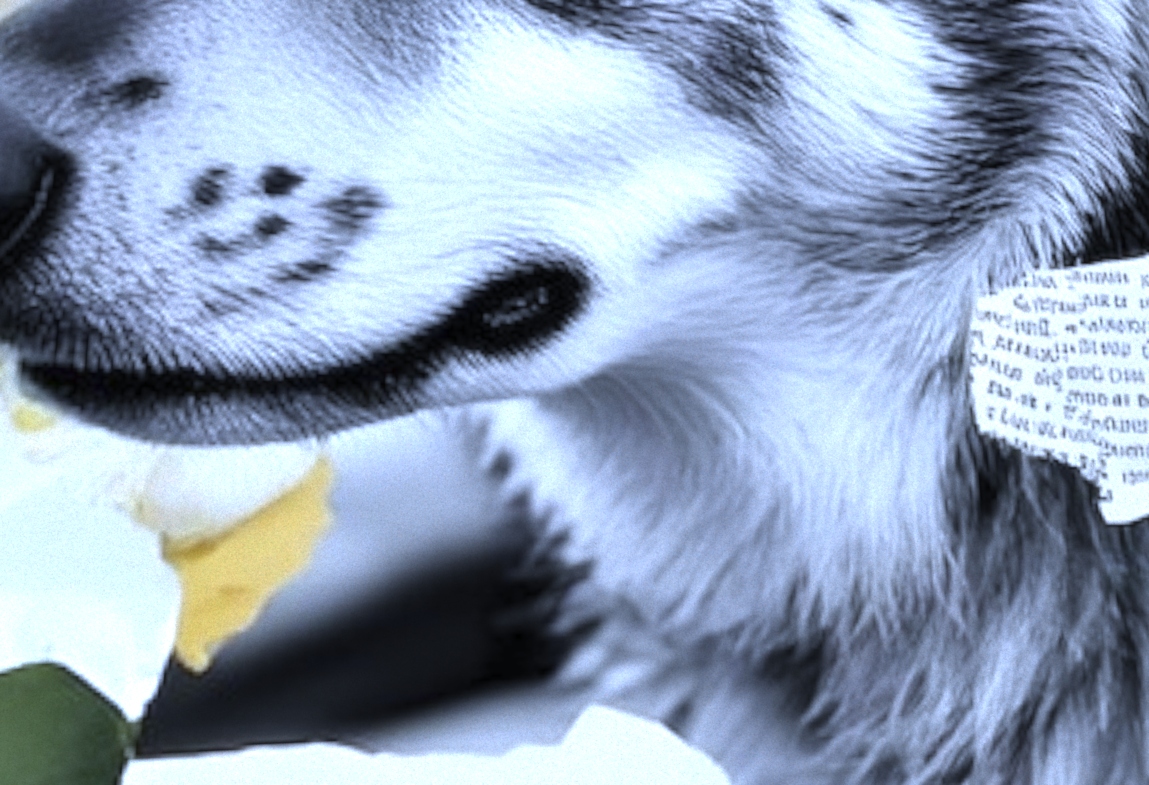} &
        \includegraphics[width=\resLen]{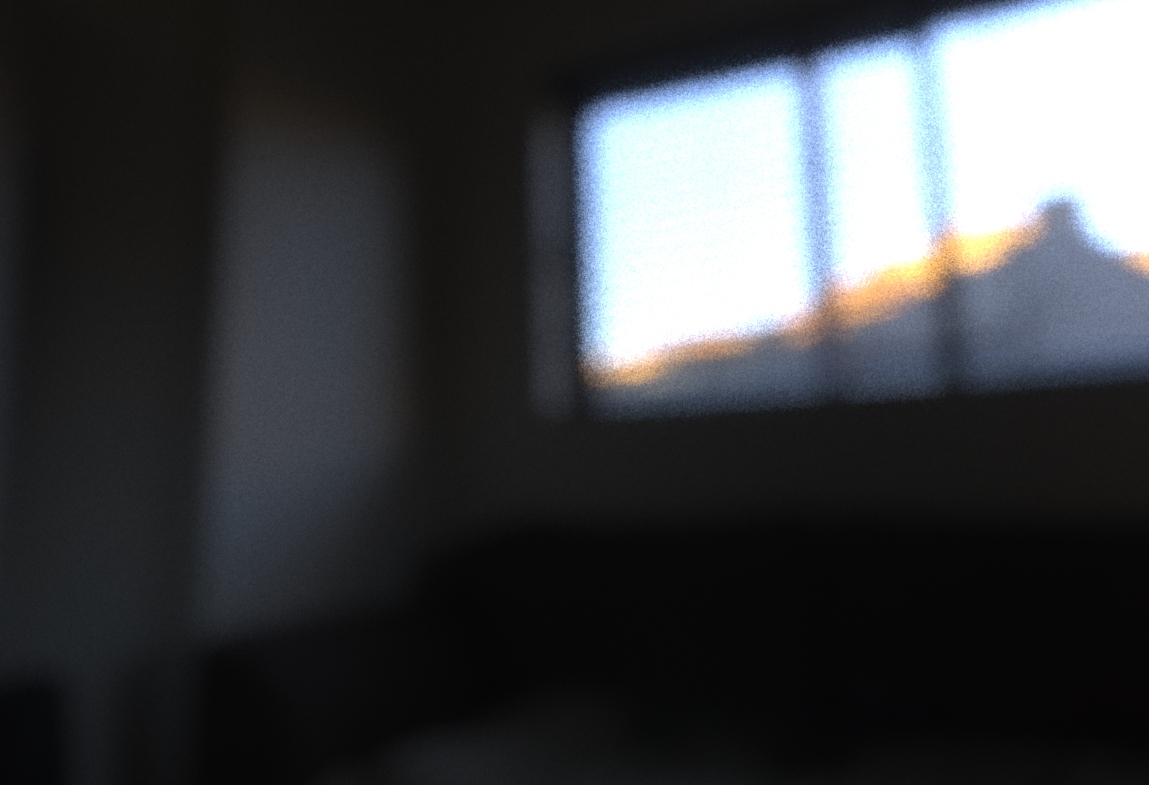} &
        \includegraphics[width=\resLen]{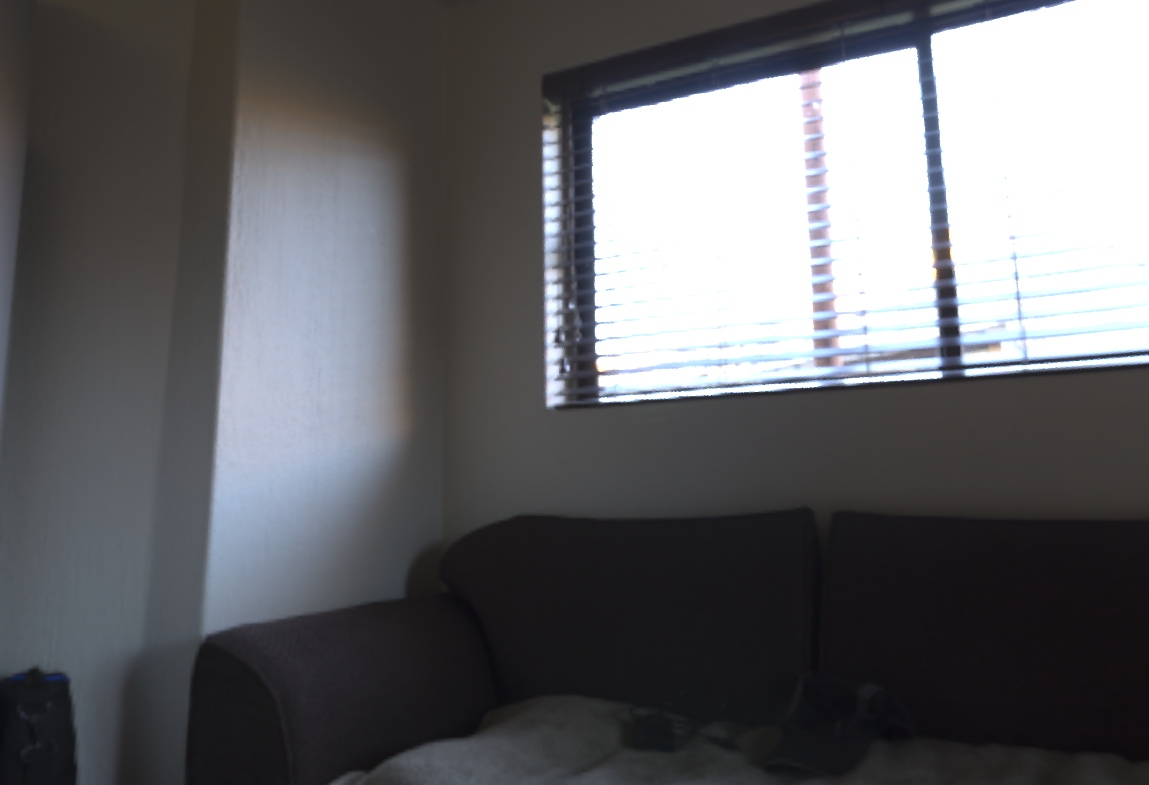} 
        \\
        \includegraphics[width=\resLen]{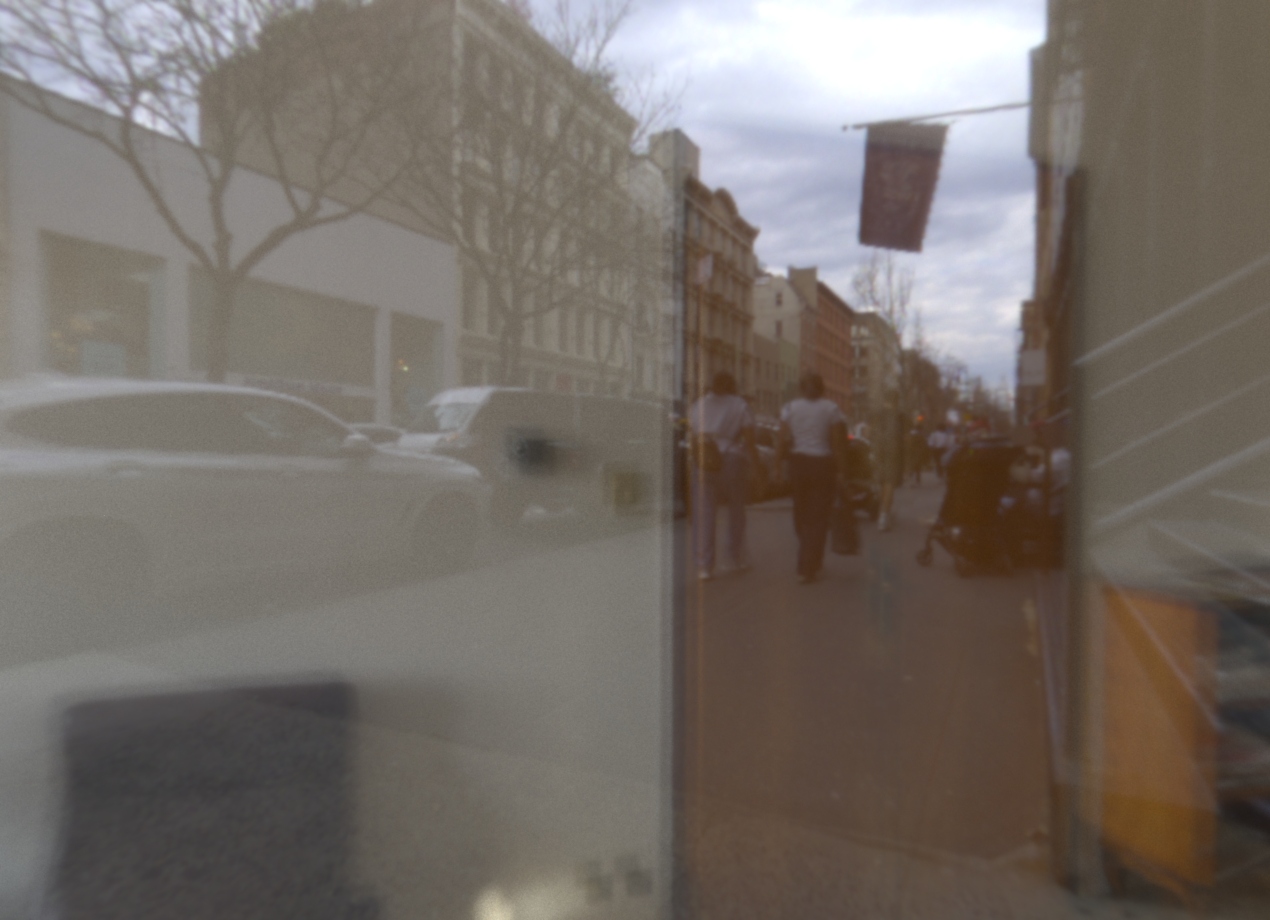} &
        \includegraphics[width=\resLen]{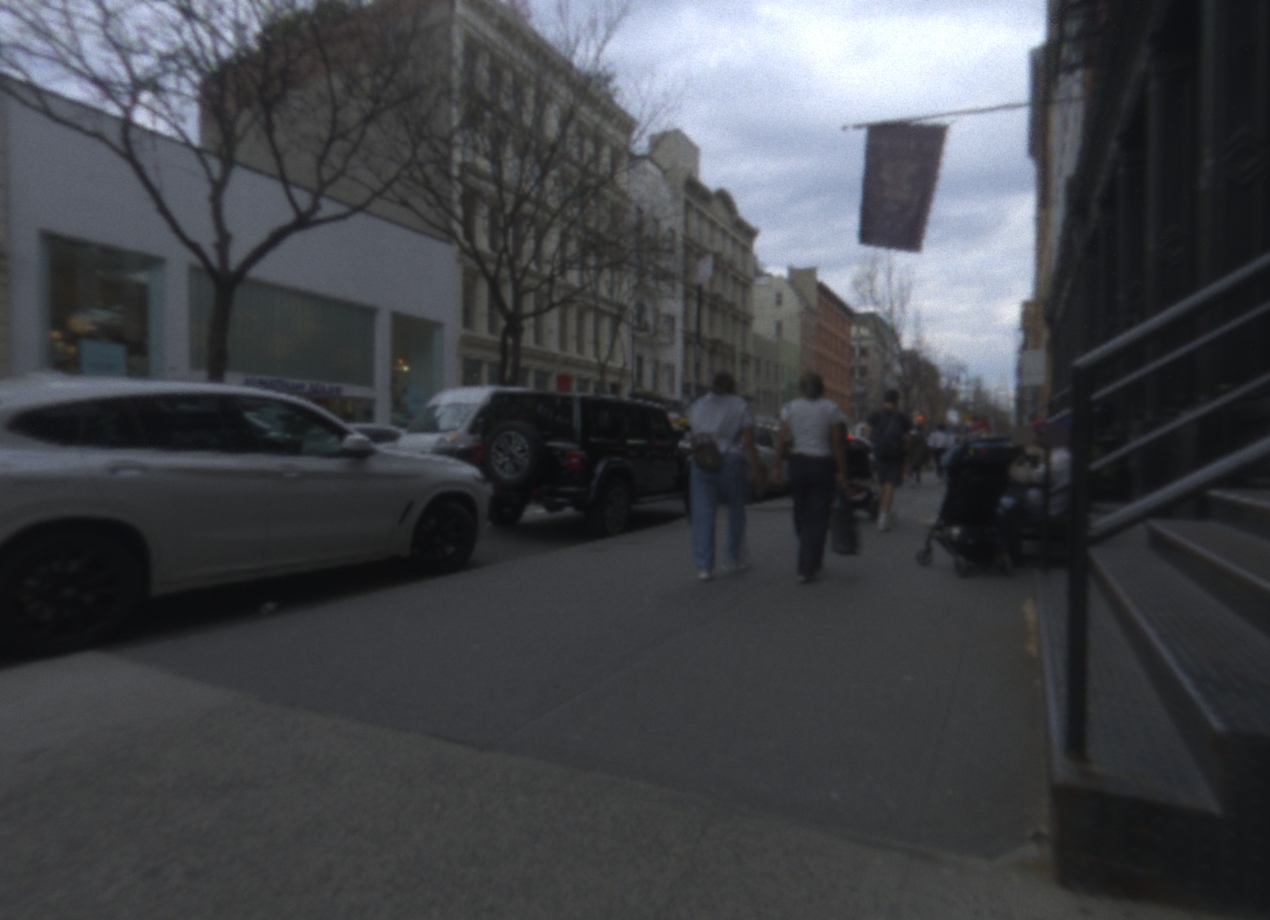} &
        \includegraphics[width=\resLen]{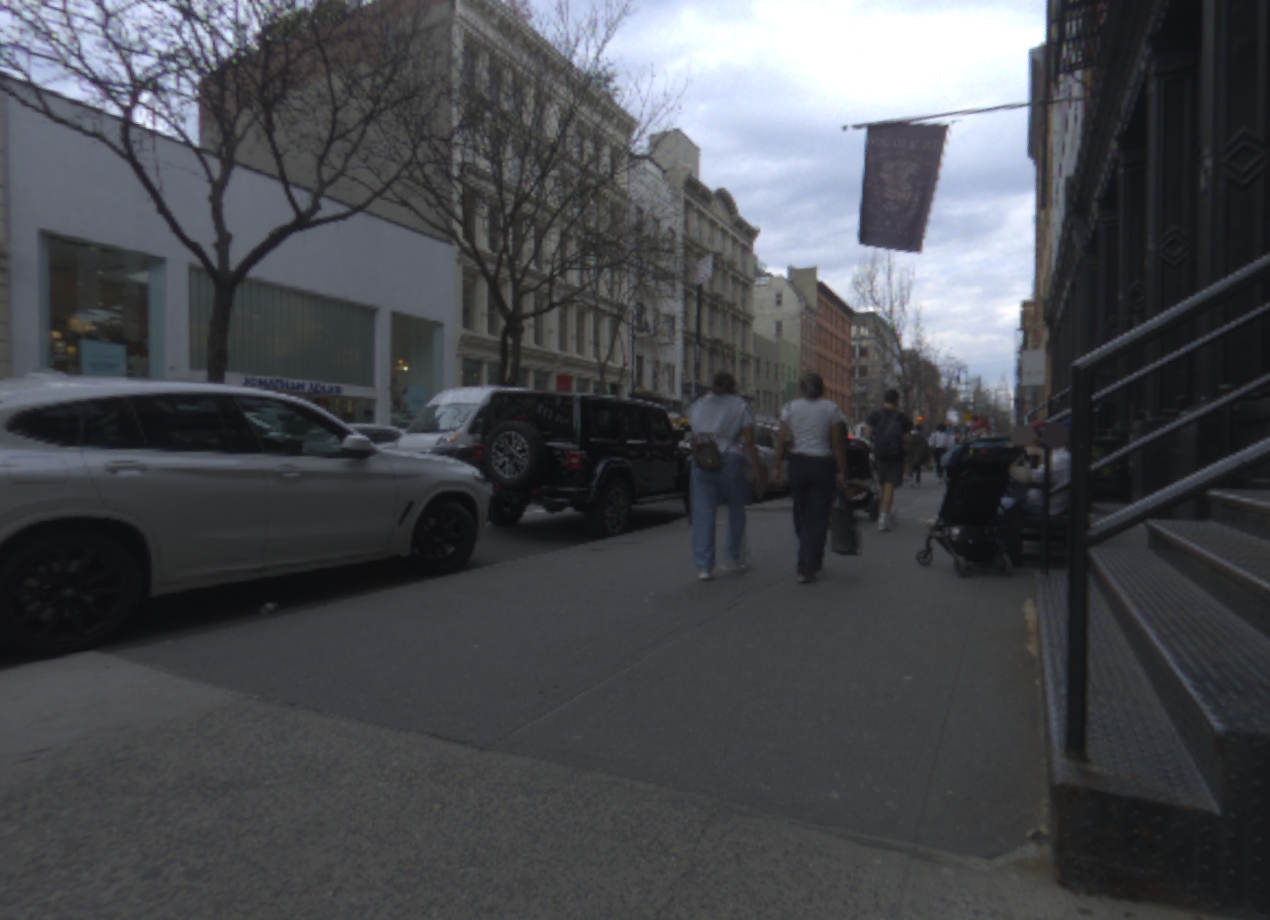} &
        \includegraphics[width=\resLen]{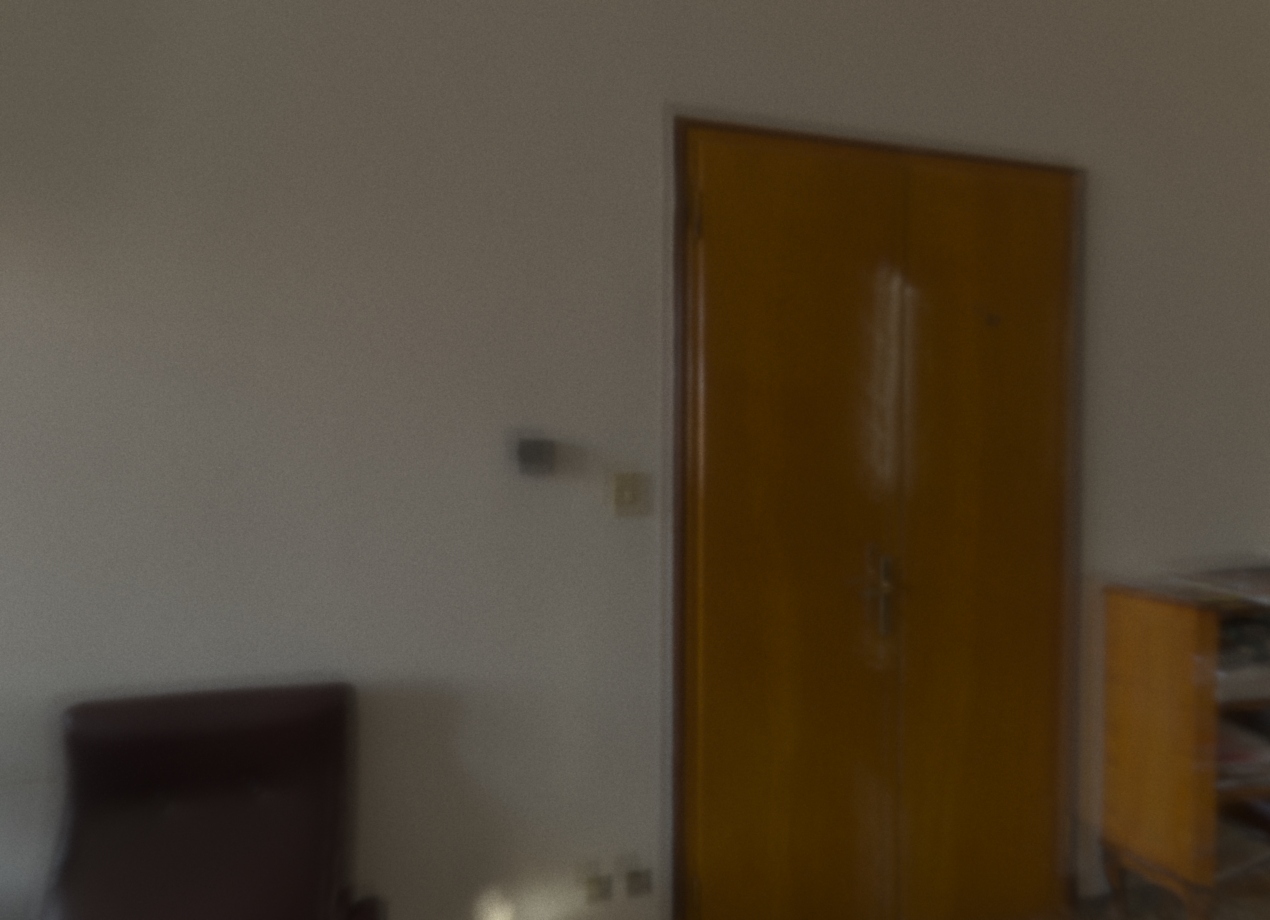} &
        \includegraphics[width=\resLen]{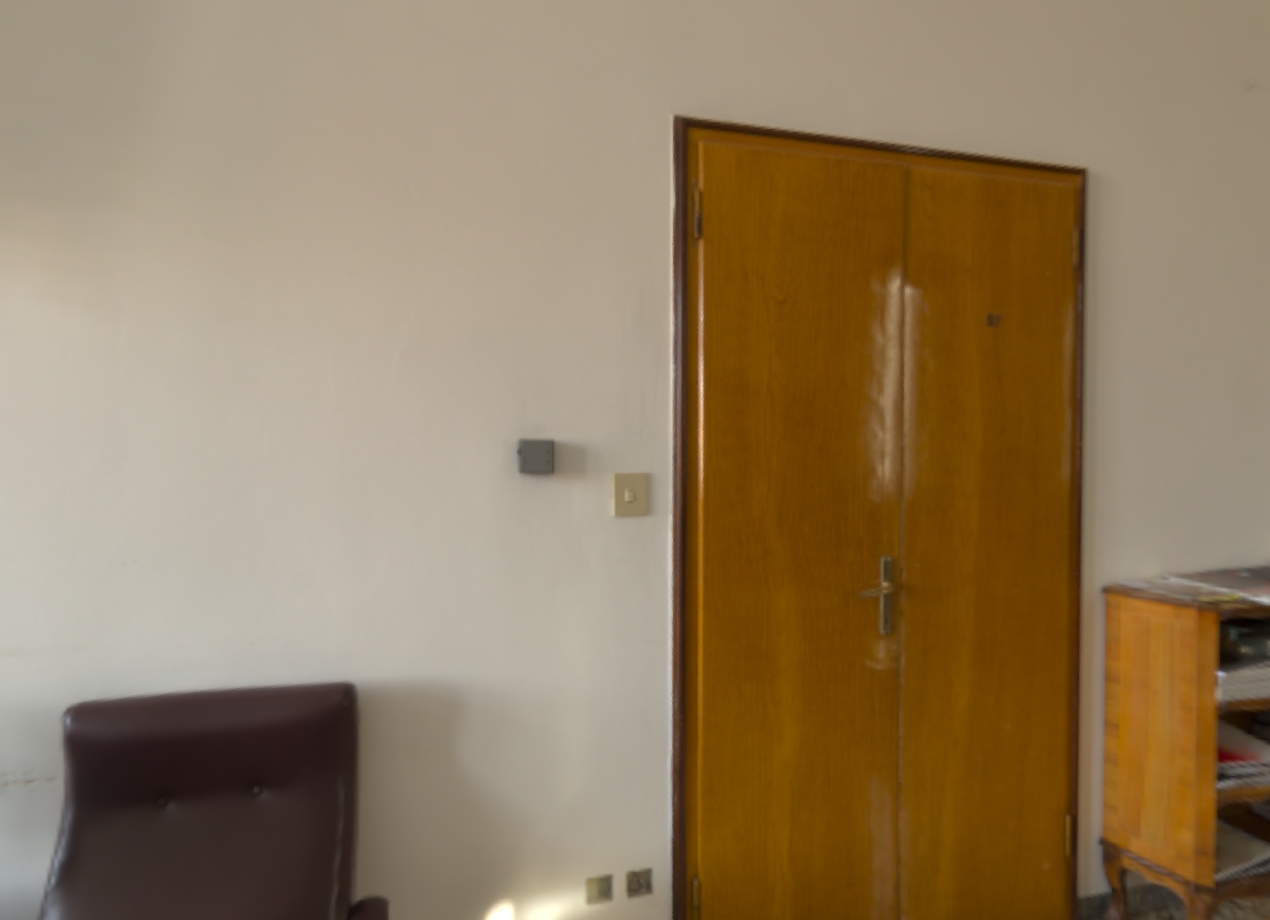} 
        \\
        \includegraphics[width=\resLen]{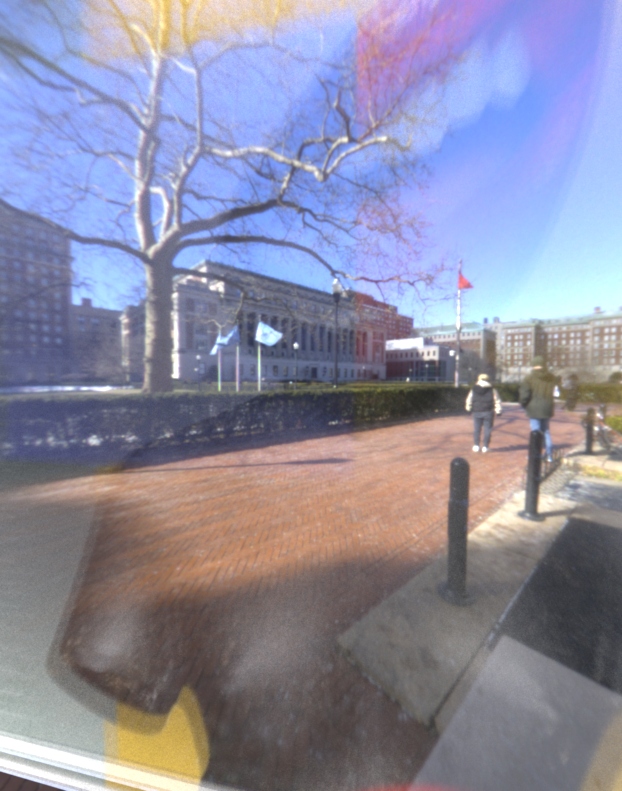} &
        \includegraphics[width=\resLen]{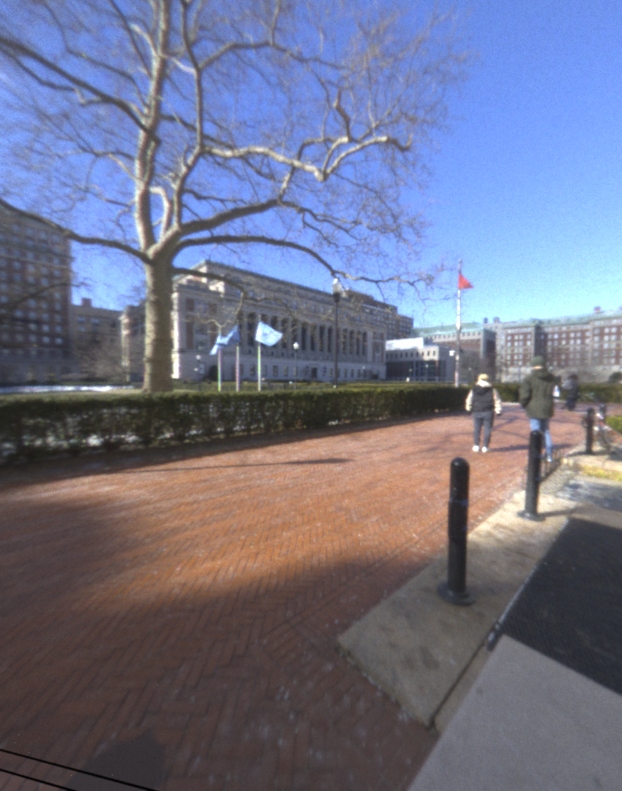} &
        \includegraphics[width=\resLen]{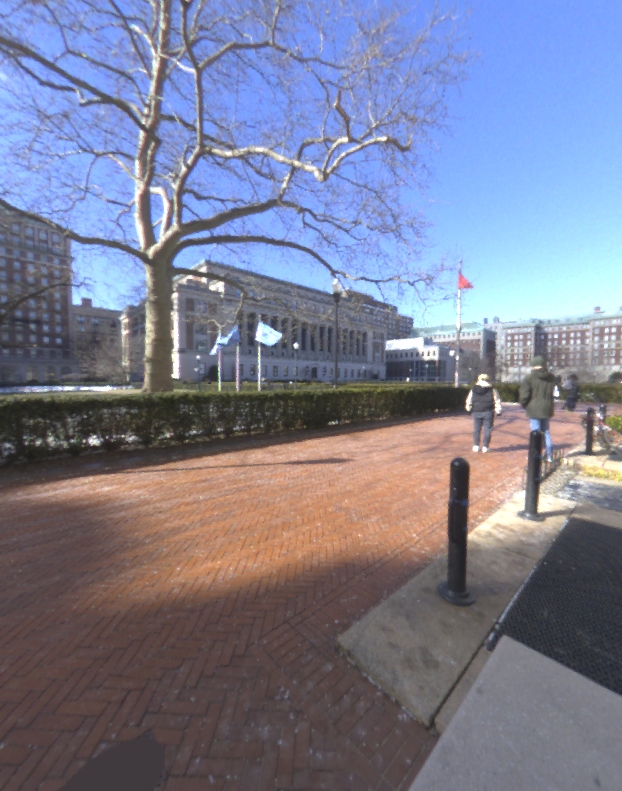} &
        \includegraphics[width=\resLen]{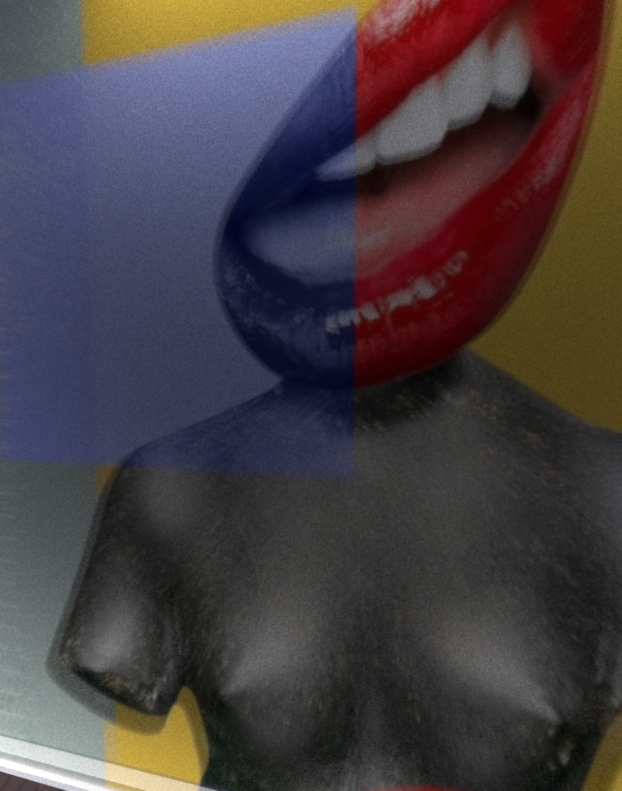} &
        \includegraphics[width=\resLen]{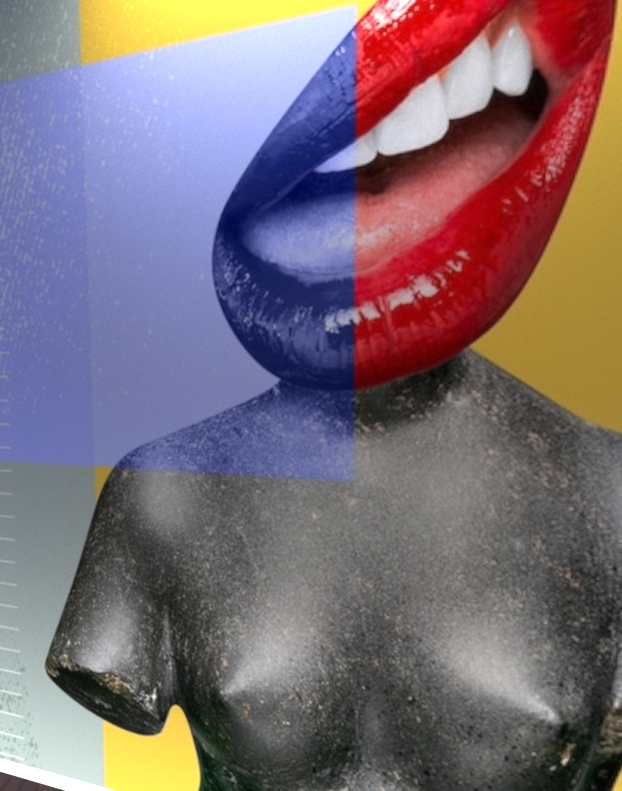} 
        \\
        \includegraphics[width=\resLen]{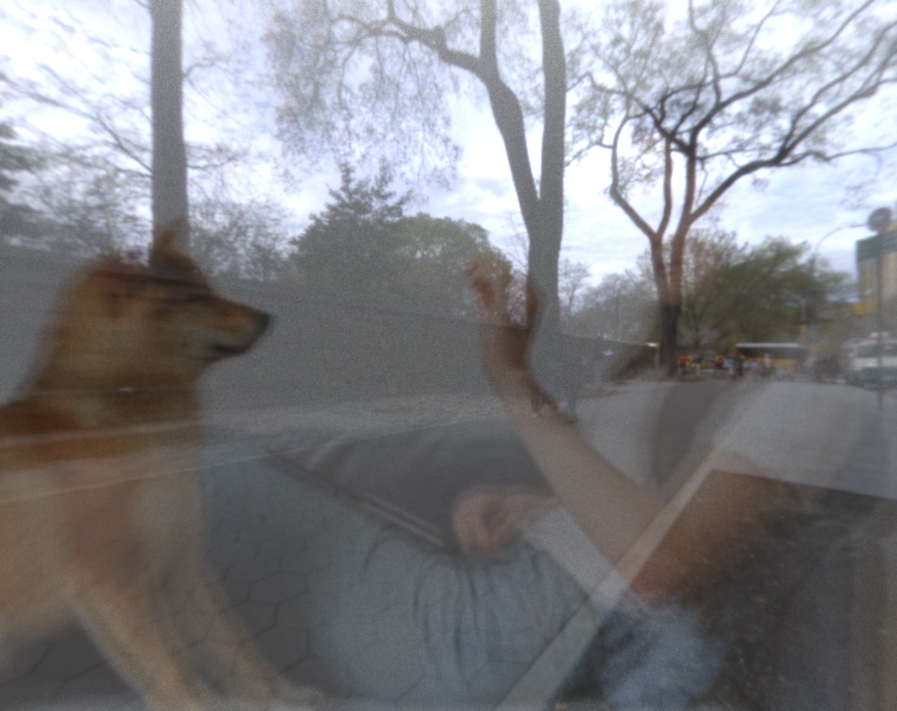} &
        \includegraphics[width=\resLen]{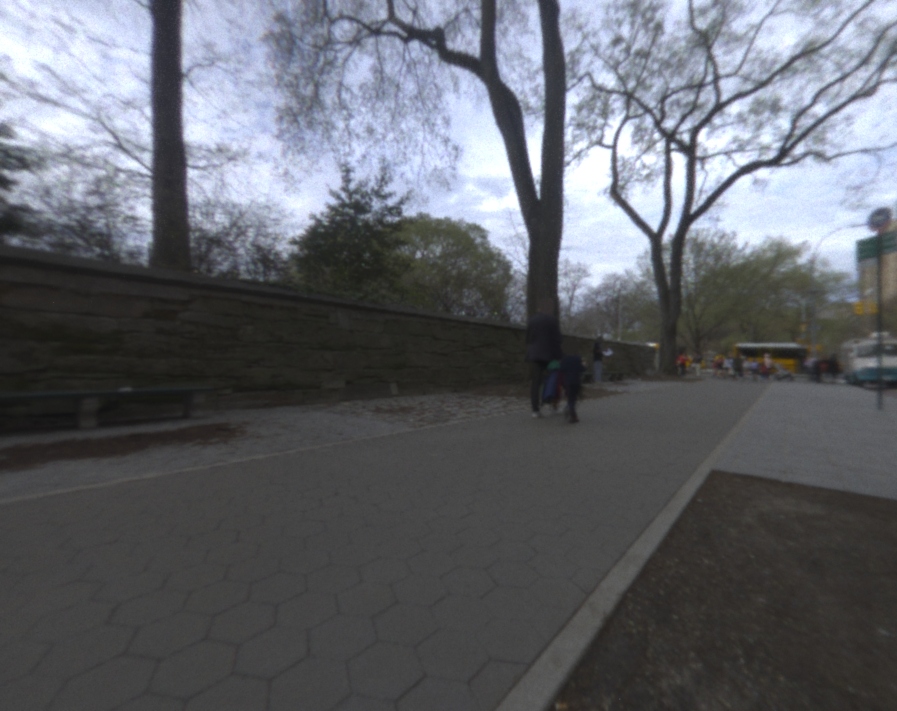} &
        \includegraphics[width=\resLen]{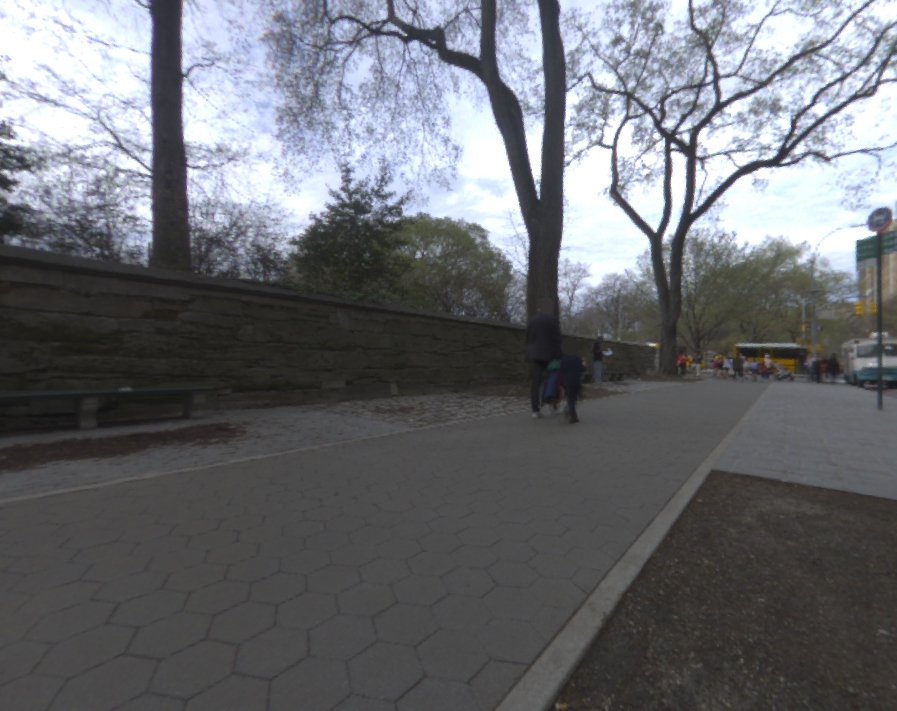} &
        \includegraphics[width=\resLen]{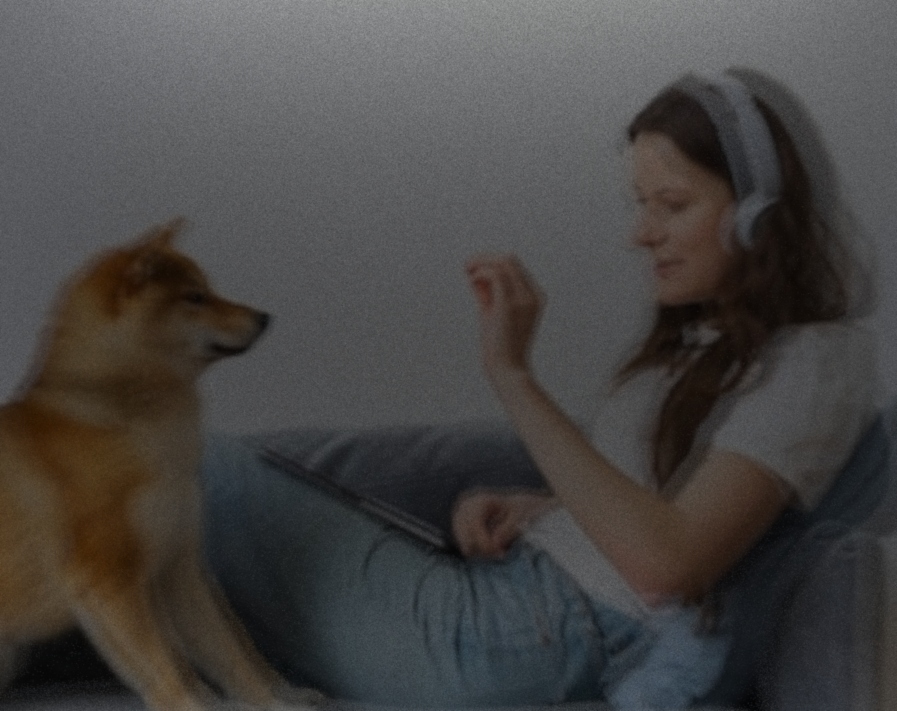} &
        \includegraphics[width=\resLen]{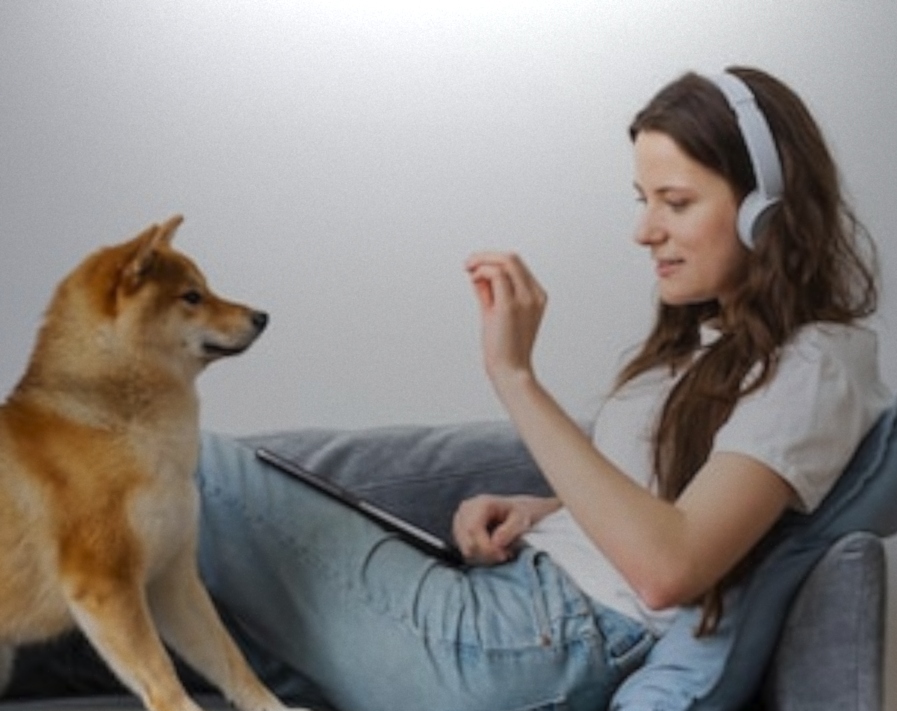} 
        \\
        \includegraphics[width=\resLen]{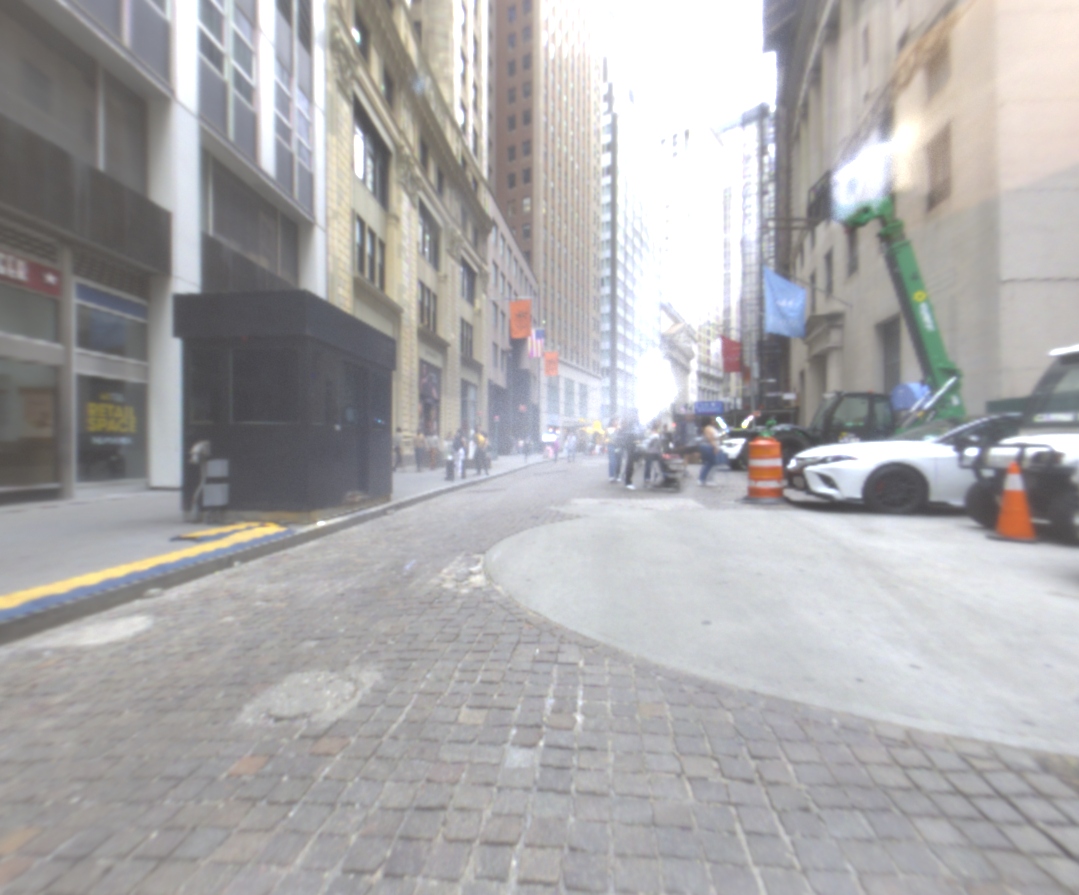} &
        \includegraphics[width=\resLen]{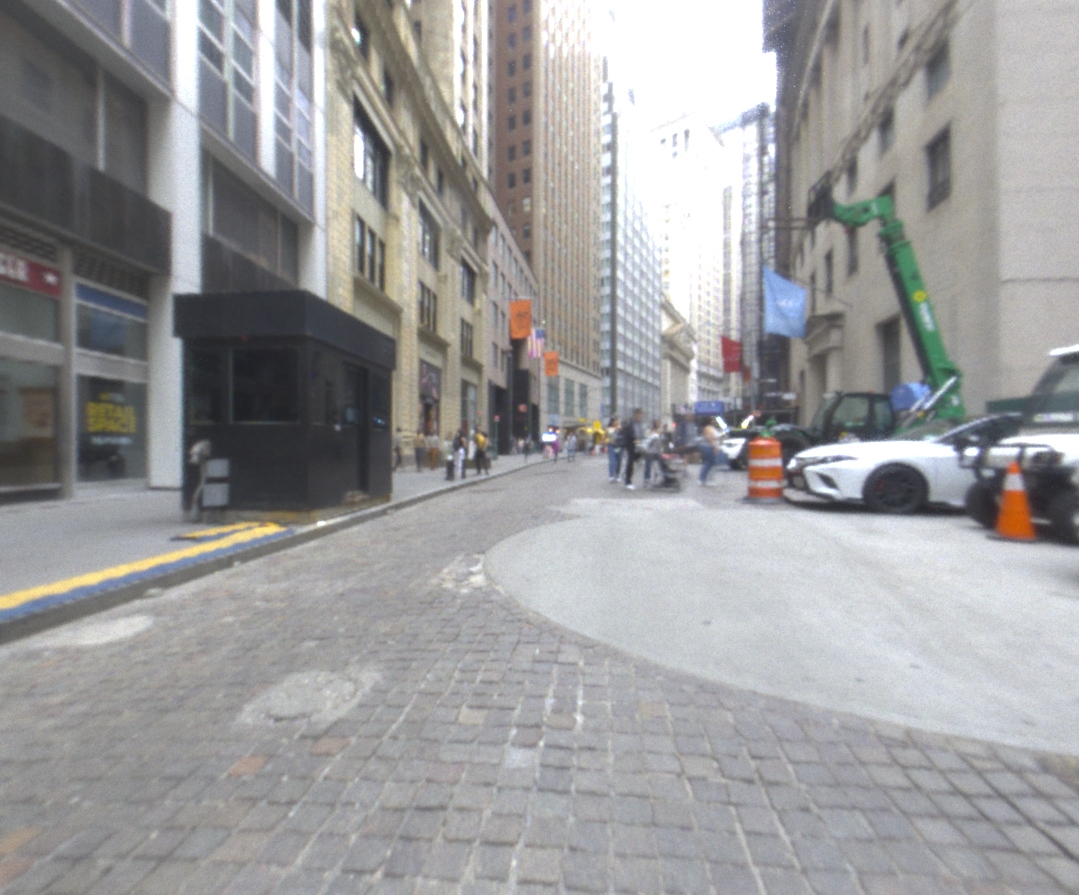} &
        \includegraphics[width=\resLen]{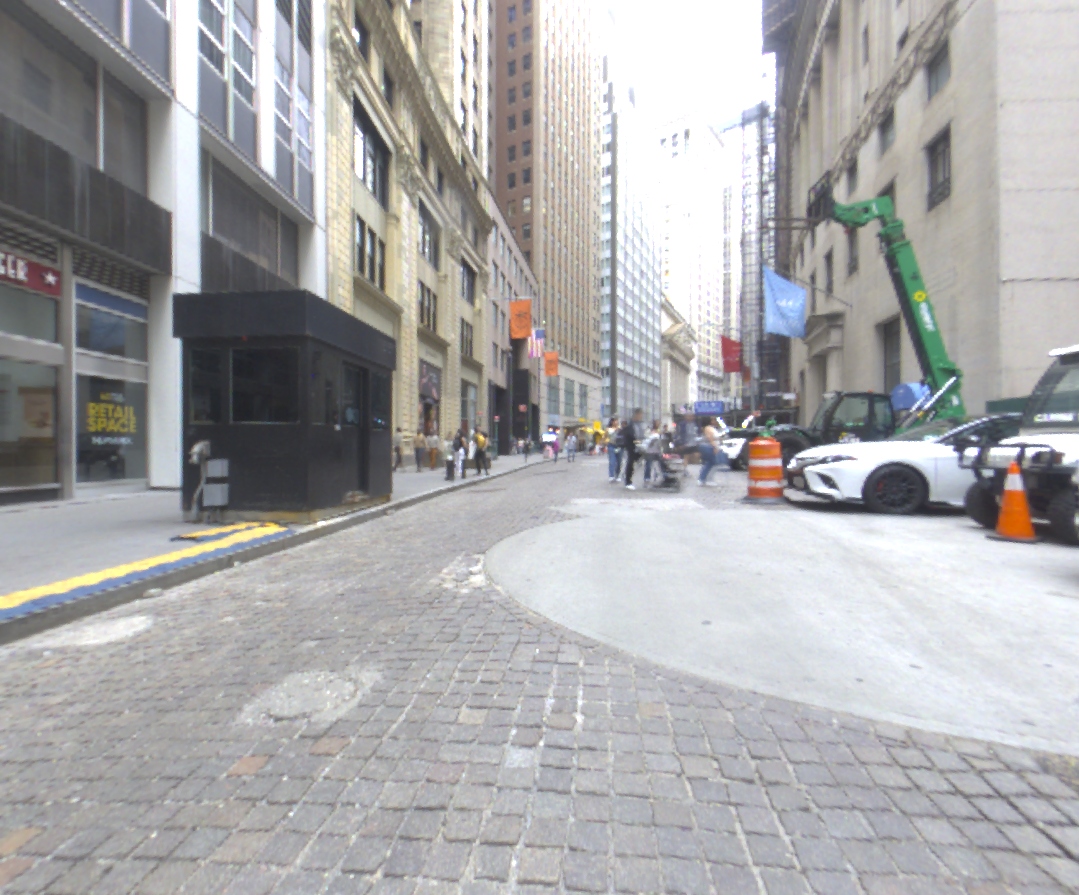} &
        \includegraphics[width=\resLen]{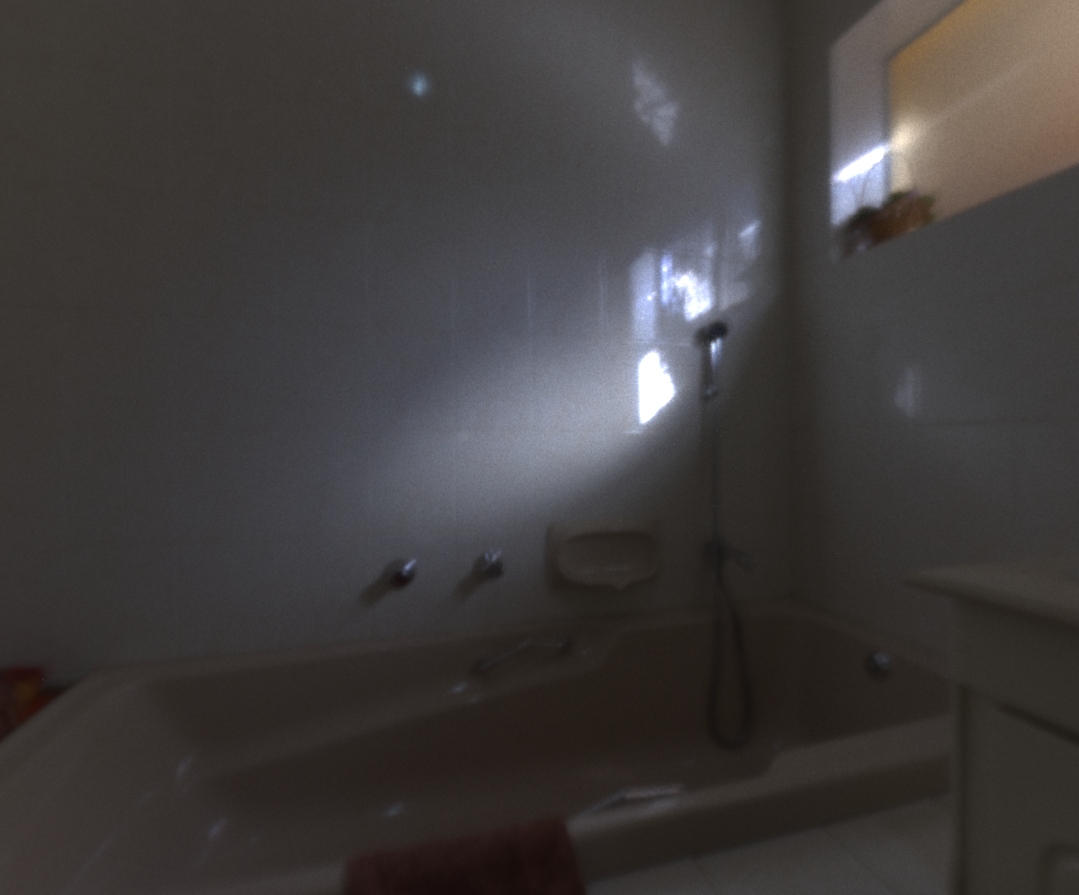} &
        \includegraphics[width=\resLen]{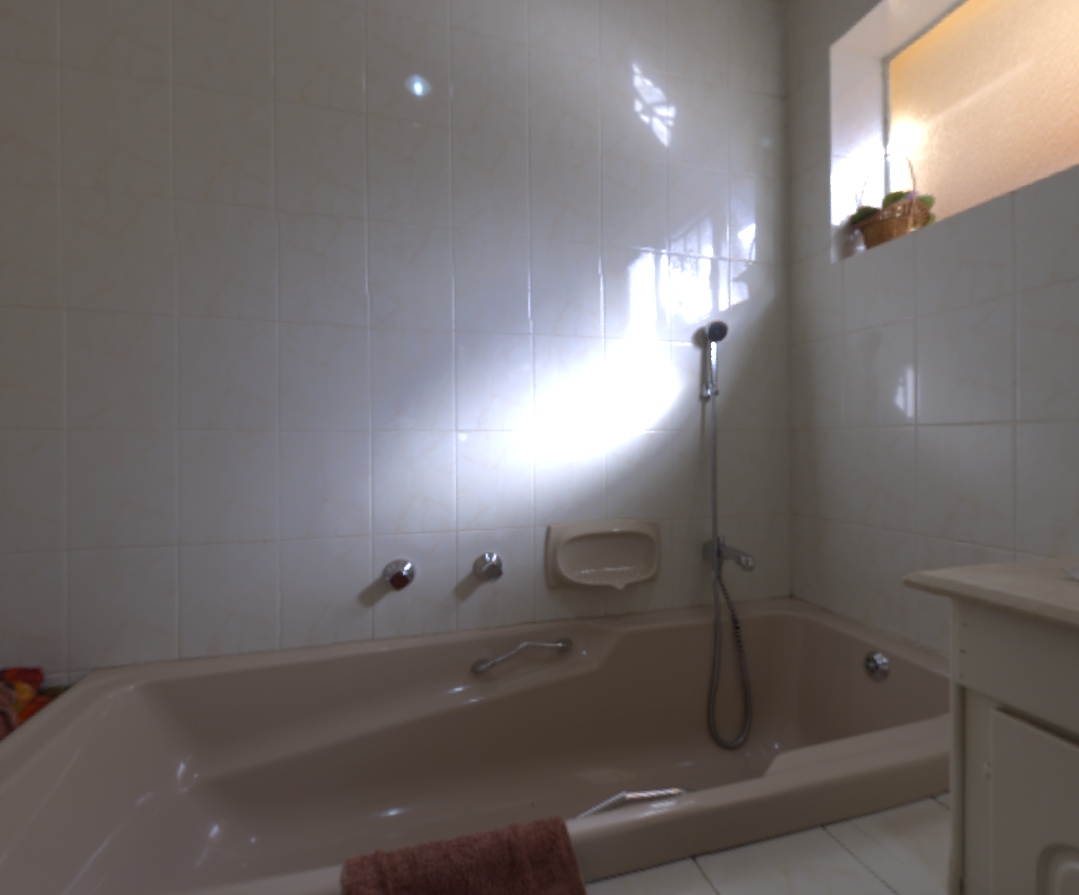} 
        \\
        \includegraphics[width=\resLen]{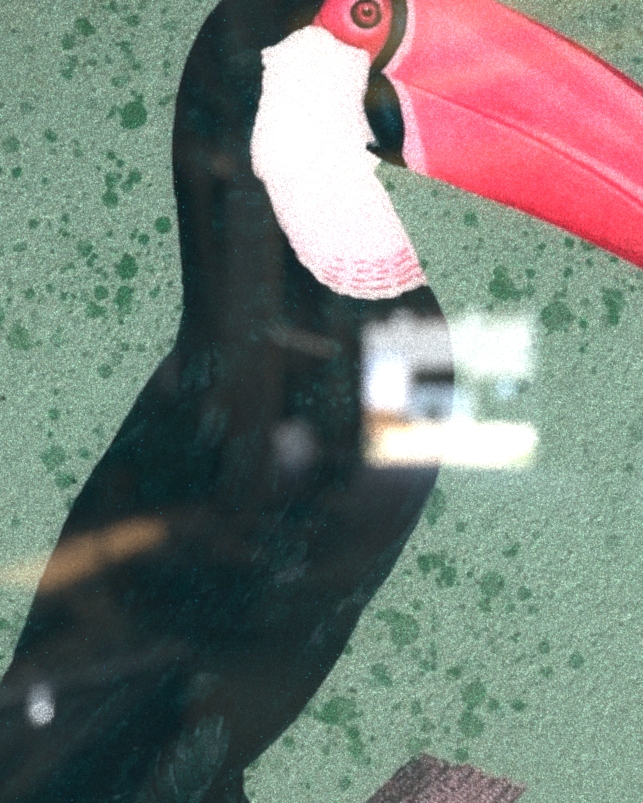} &
        \includegraphics[width=\resLen]{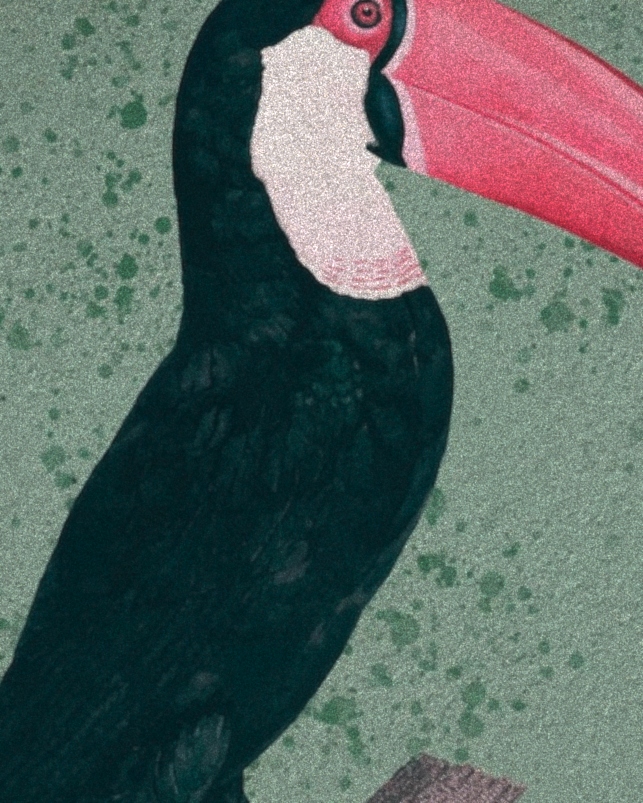} &
        \includegraphics[width=\resLen]{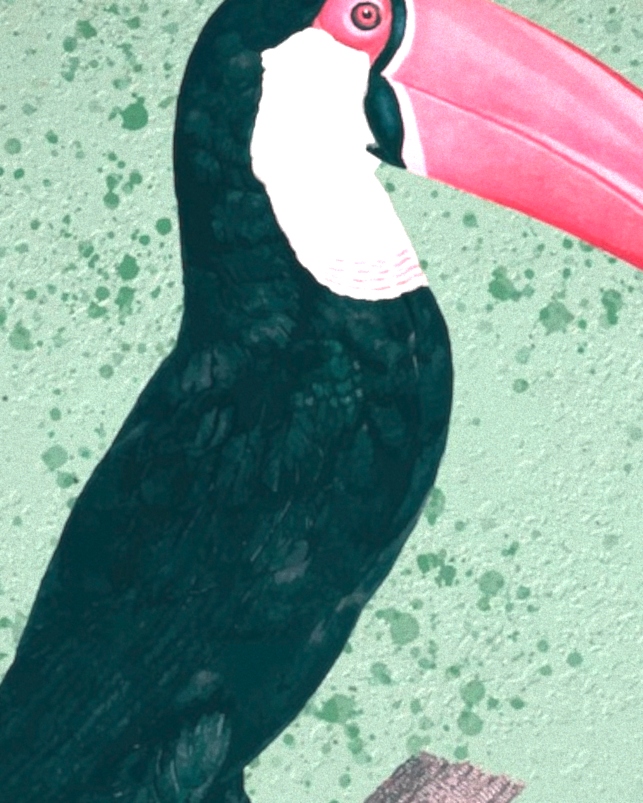} &
        \includegraphics[width=\resLen]{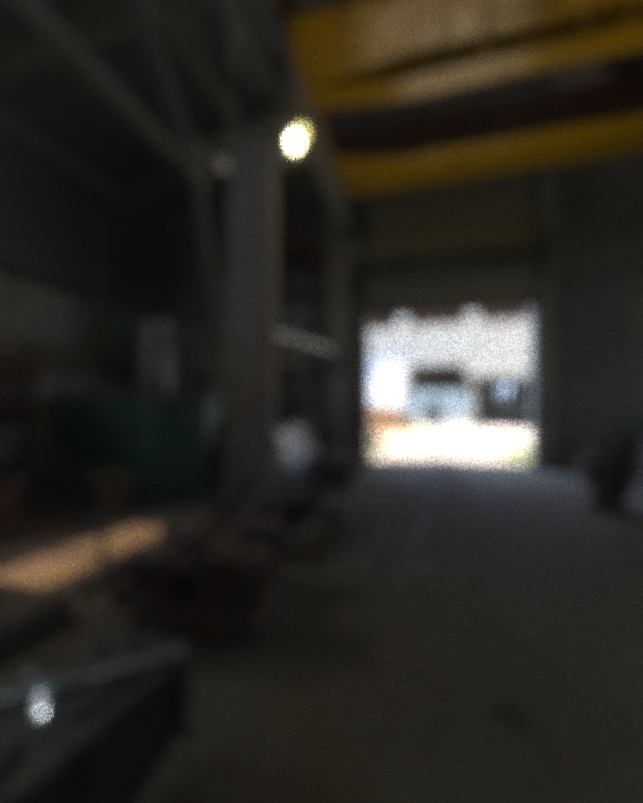} &
        \includegraphics[width=\resLen]{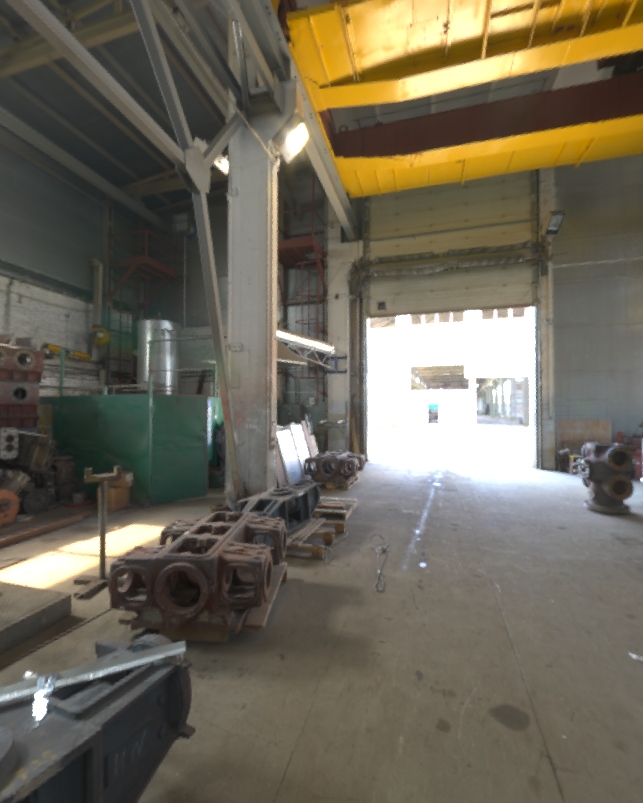} 
        \\
        \includegraphics[width=\resLen]{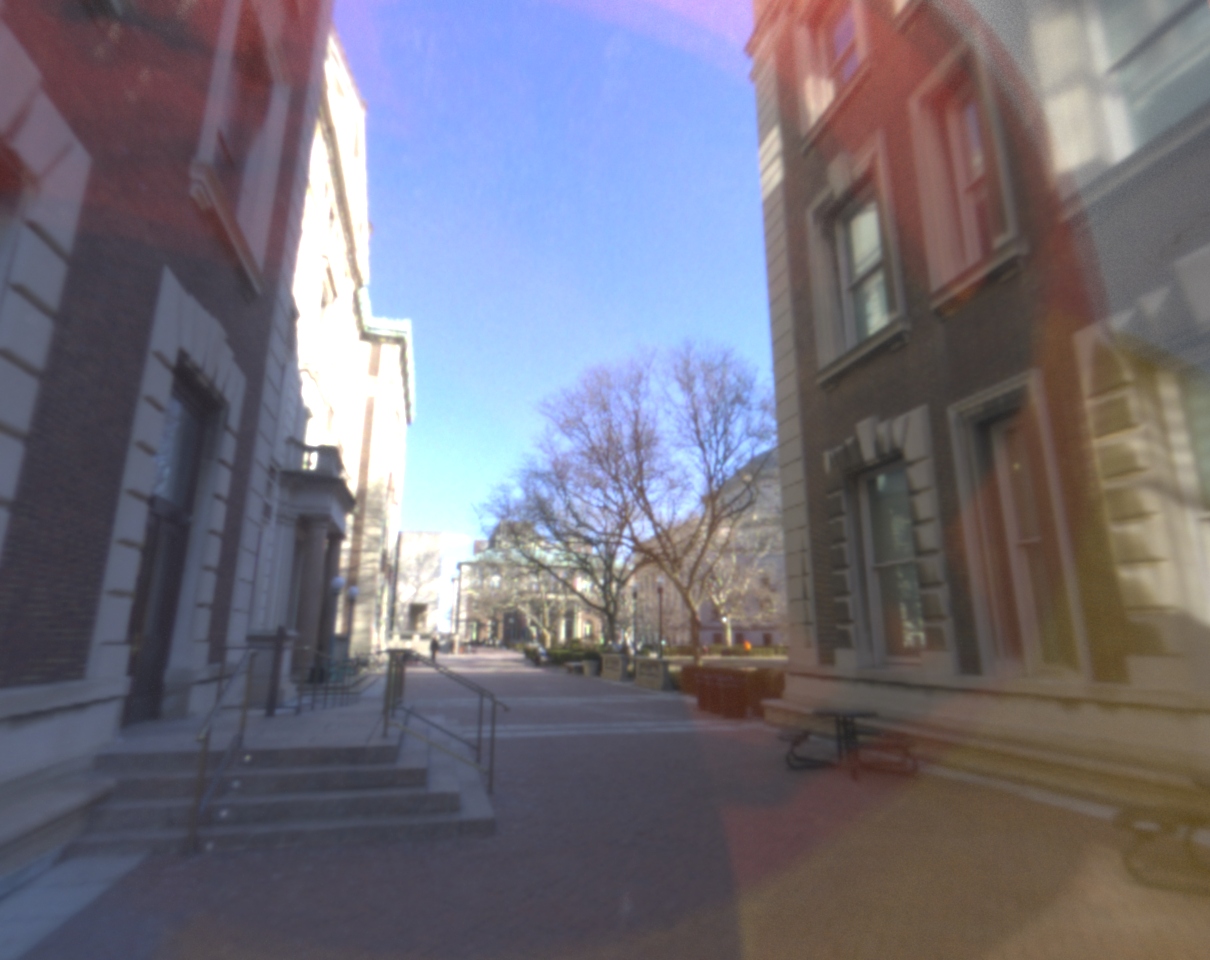} &
        \includegraphics[width=\resLen]{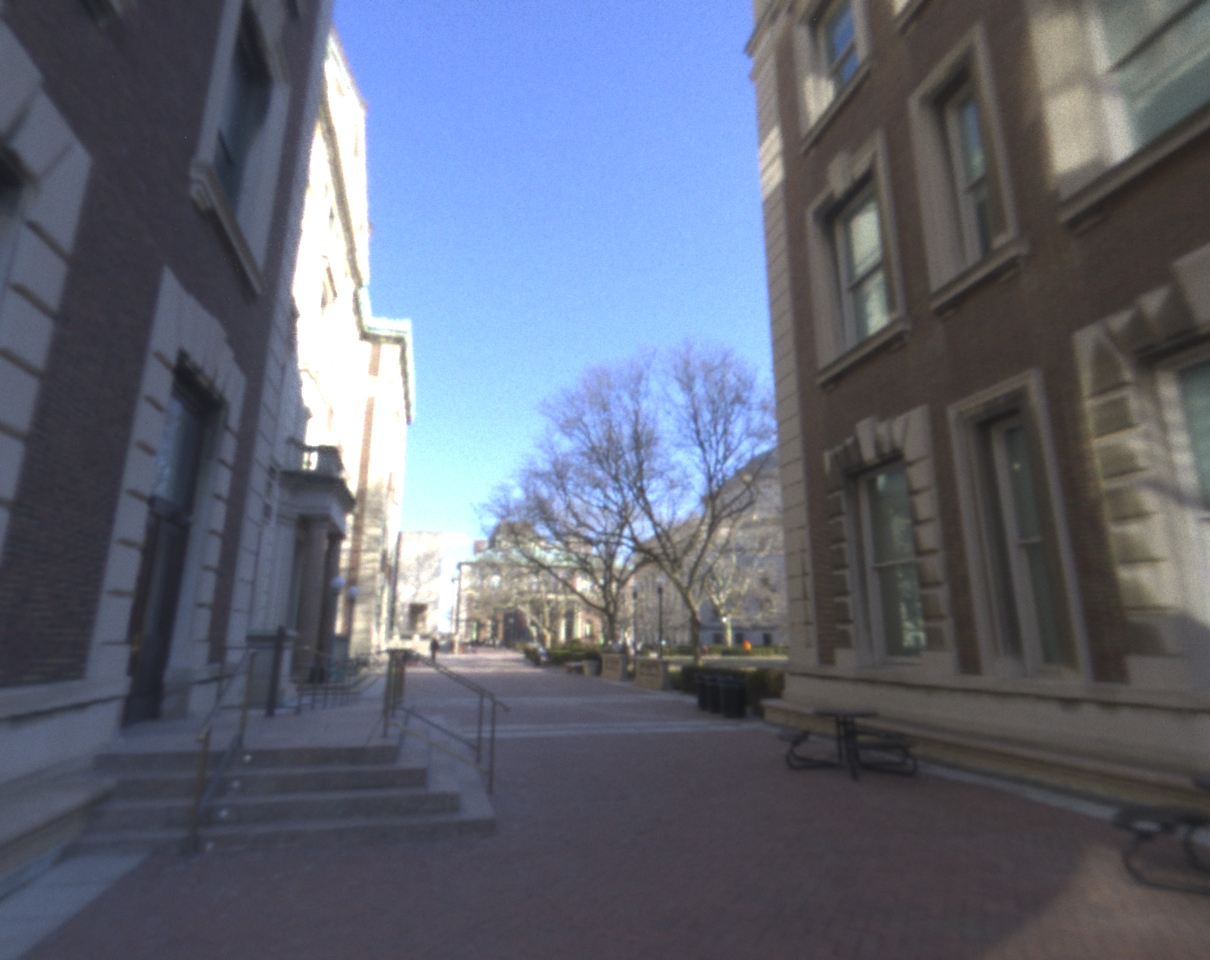} &
        \includegraphics[width=\resLen]{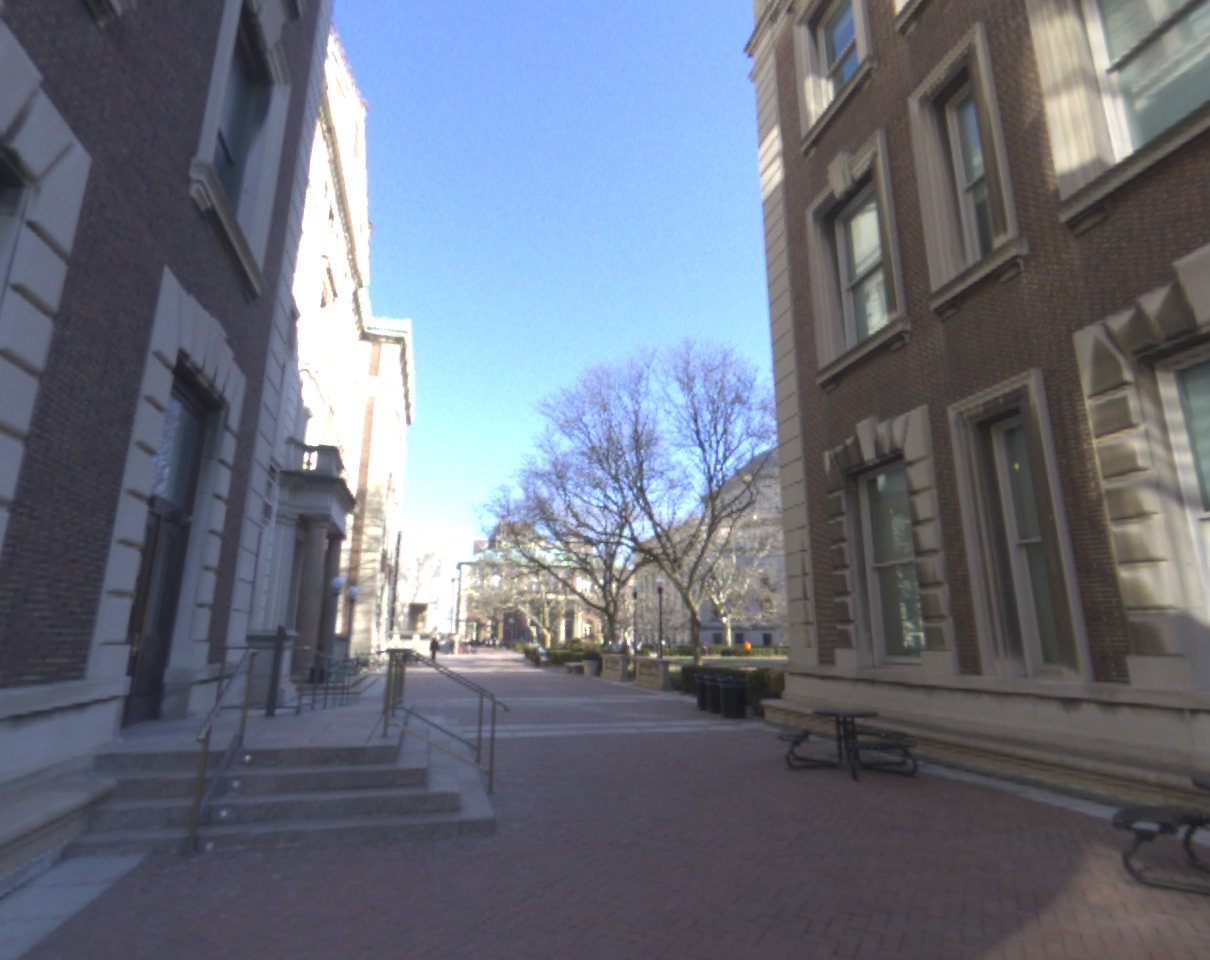} &
        \includegraphics[width=\resLen]{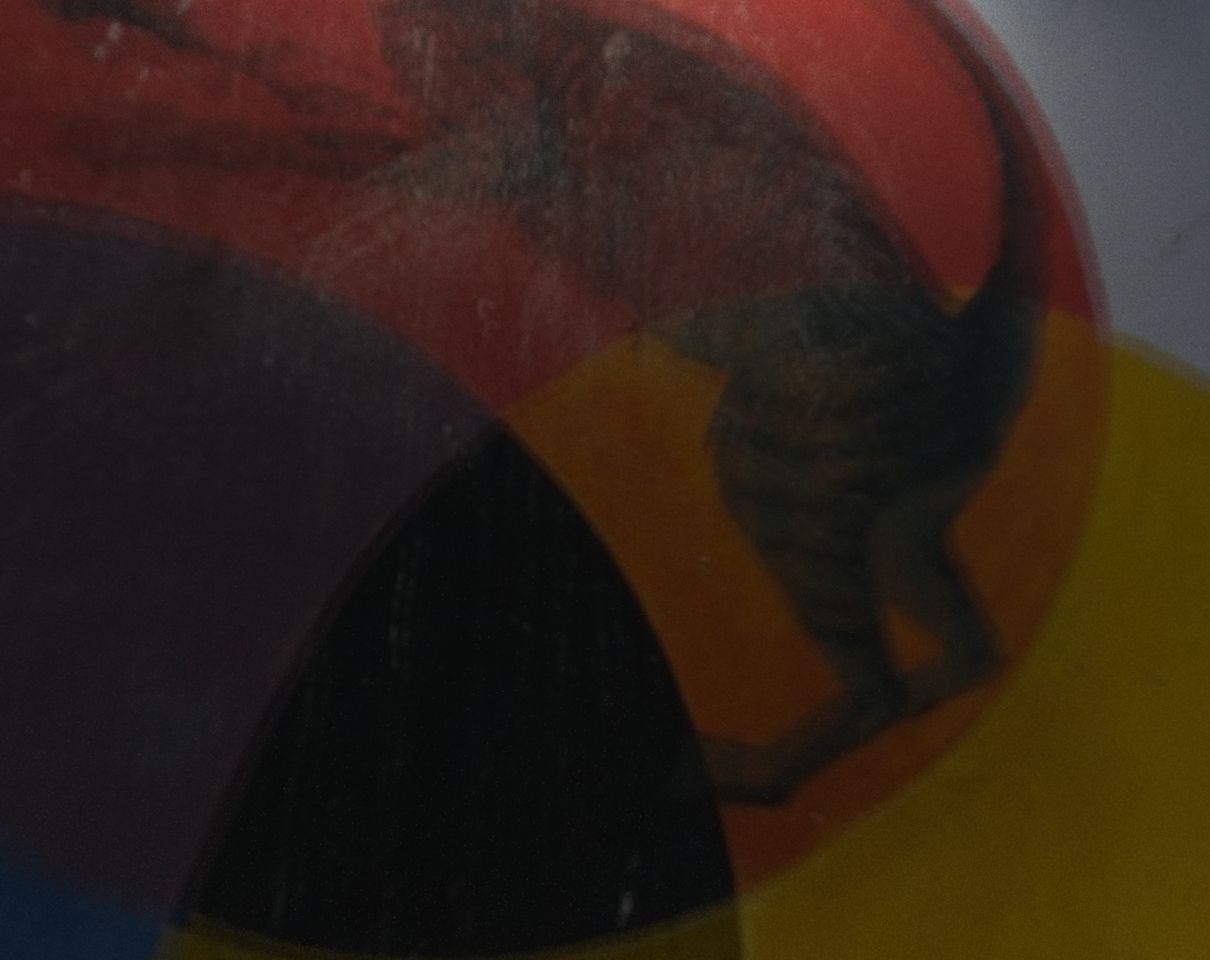} &
        \includegraphics[width=\resLen]{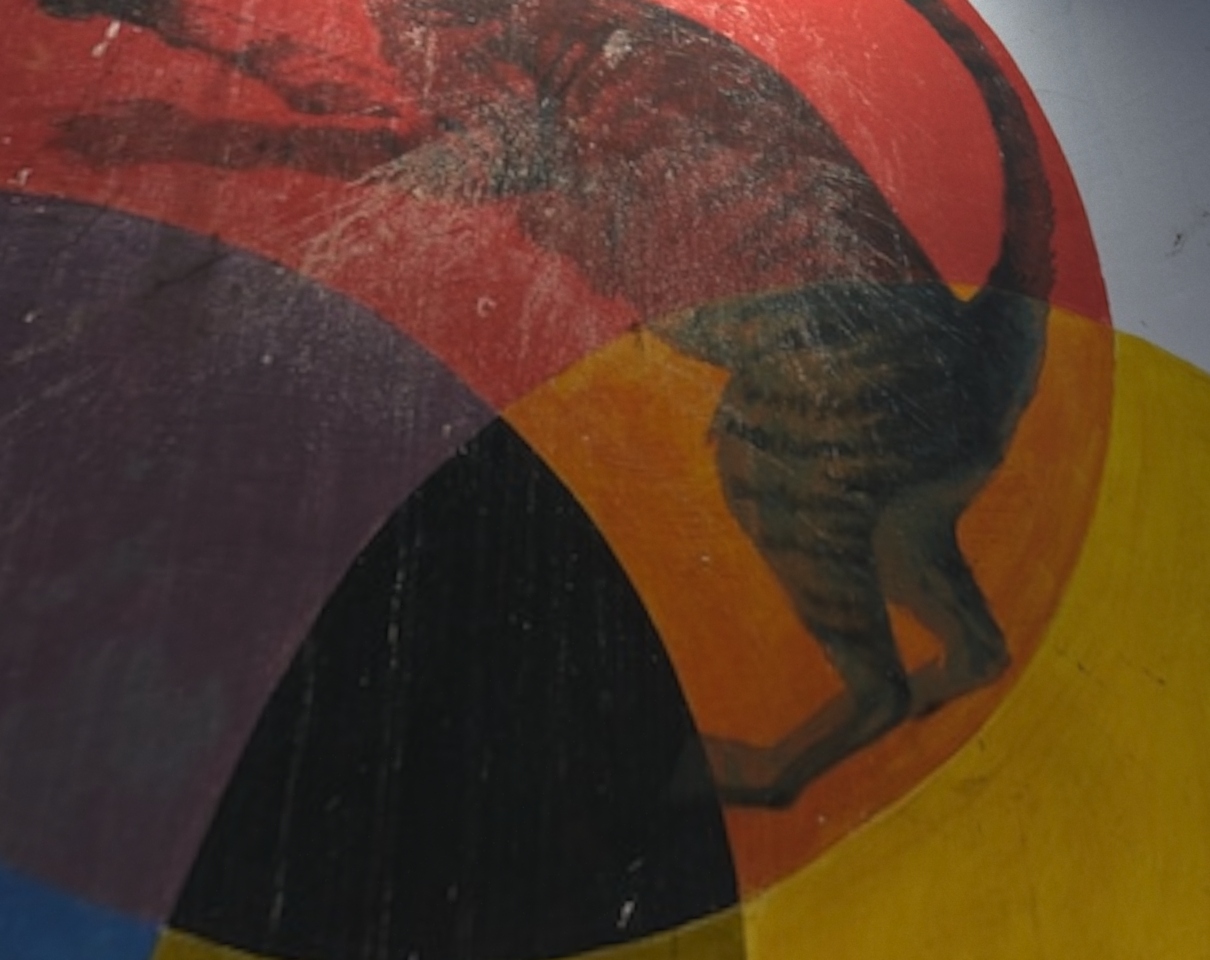} 
        \\
        \includegraphics[width=\resLen]{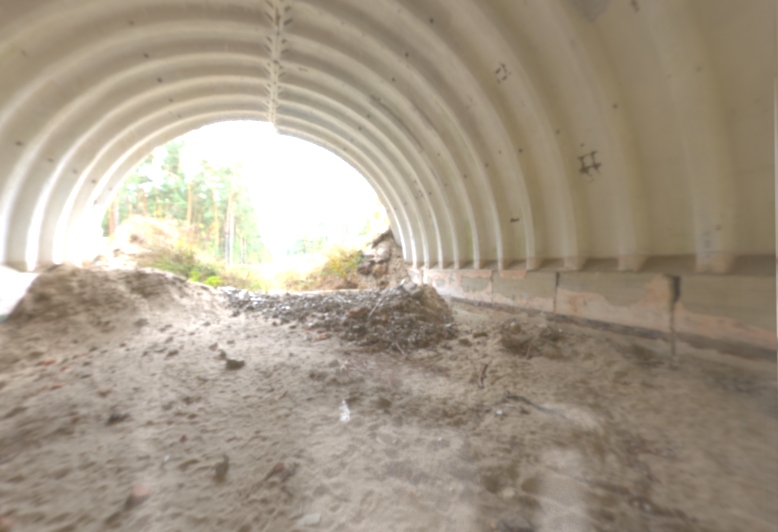} &
        \includegraphics[width=\resLen]{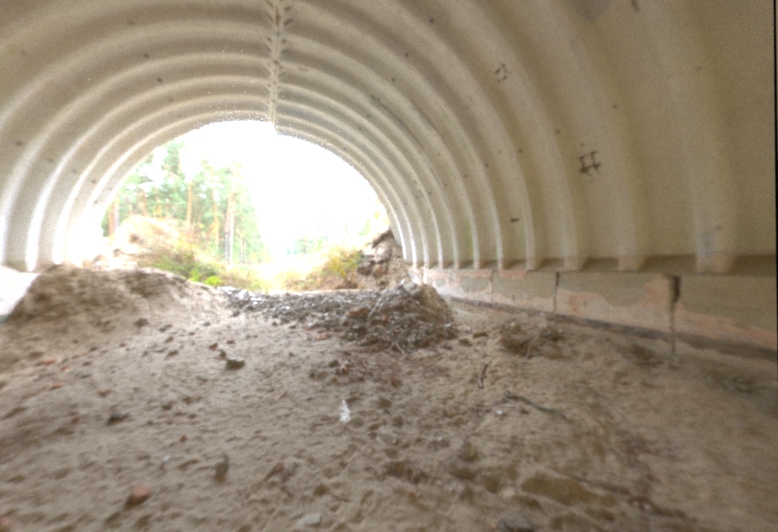} &
        \includegraphics[width=\resLen]{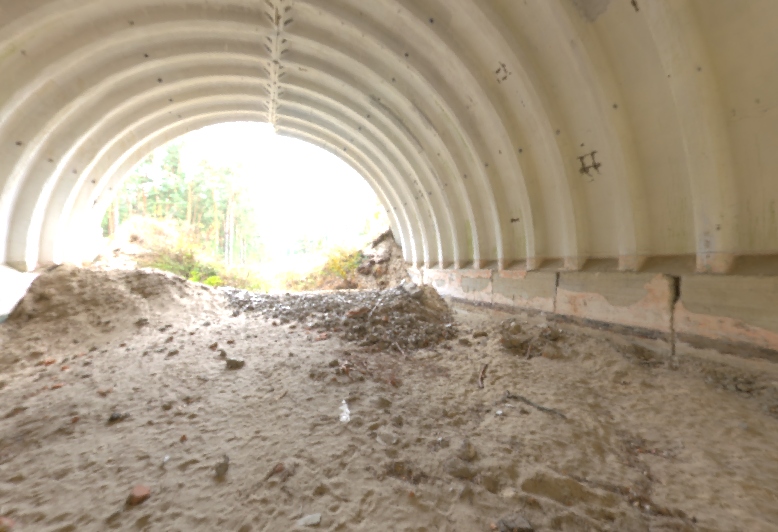} &
        \includegraphics[width=\resLen]{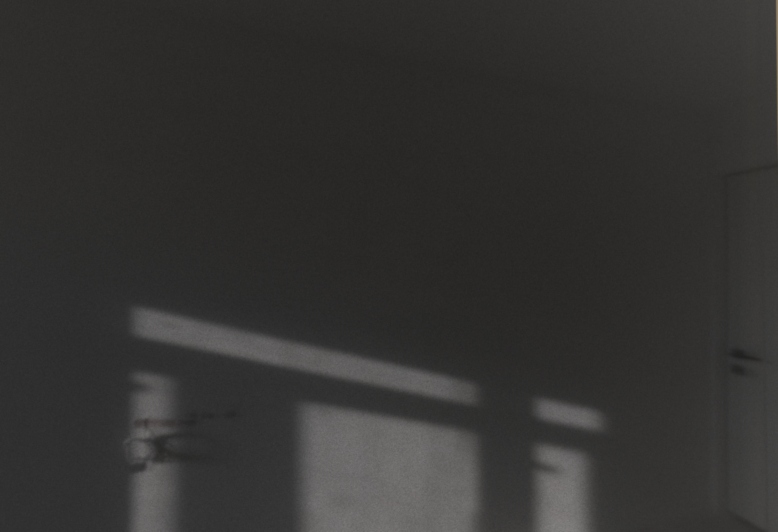} &
        \includegraphics[width=\resLen]{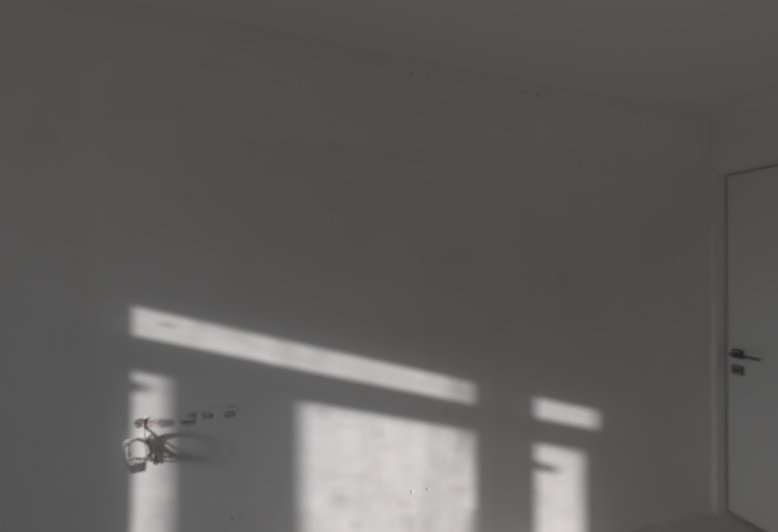} 
        \\
        \includegraphics[width=\resLen]{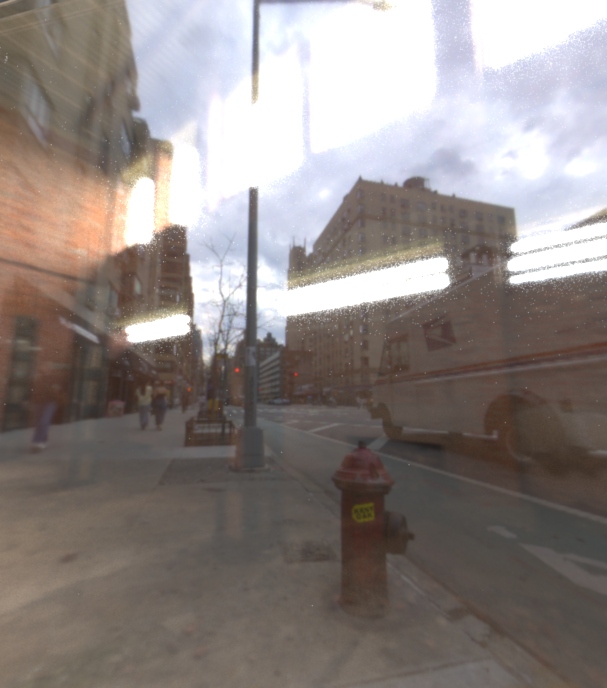} &
        \includegraphics[width=\resLen]{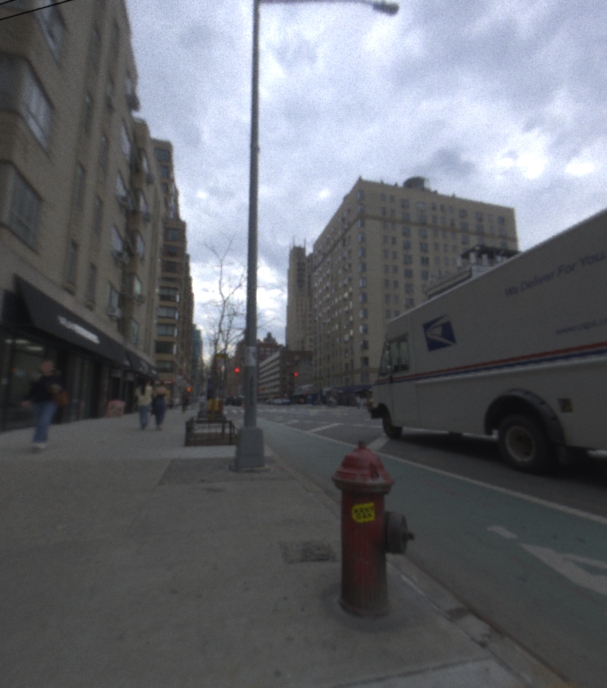} &
        \includegraphics[width=\resLen]{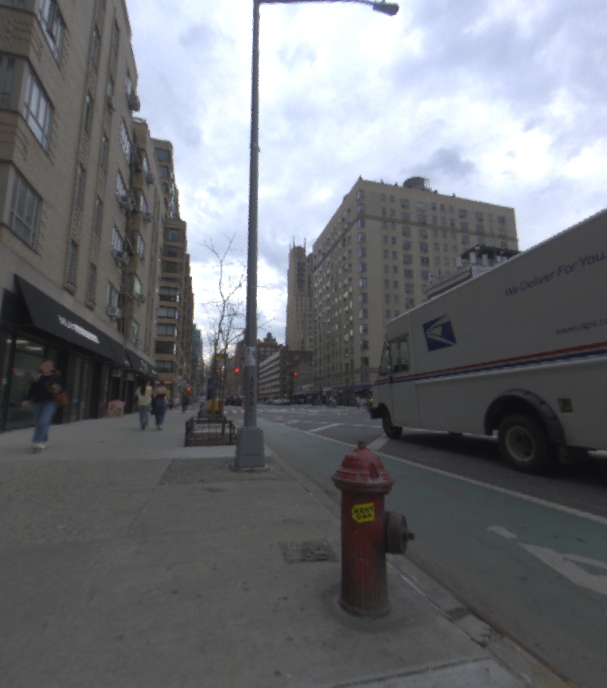} &
        \includegraphics[width=\resLen]{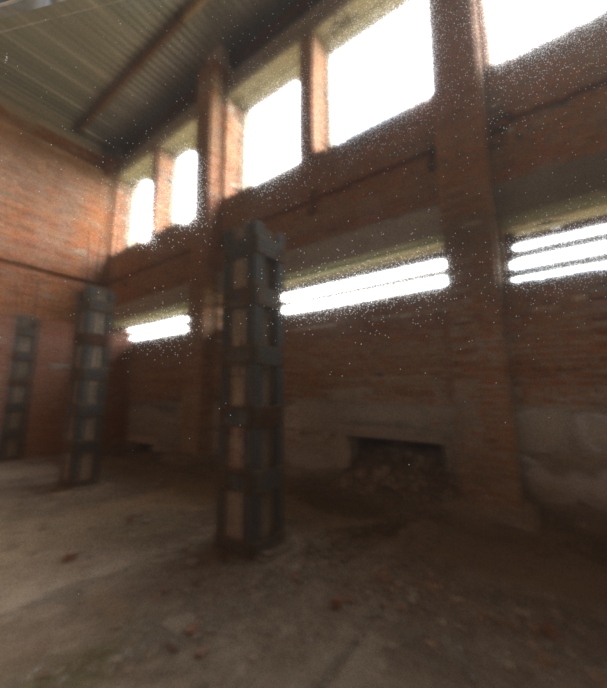} &
        \includegraphics[width=\resLen]{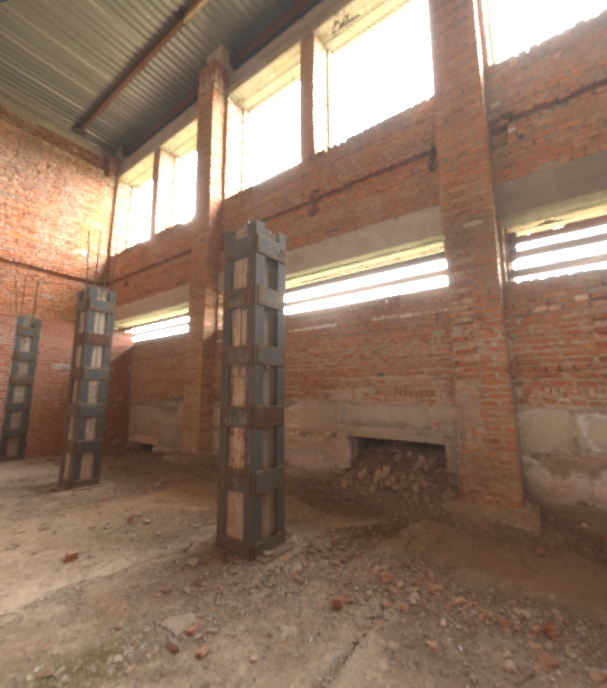} 
        \\
        \includegraphics[width=\resLen]{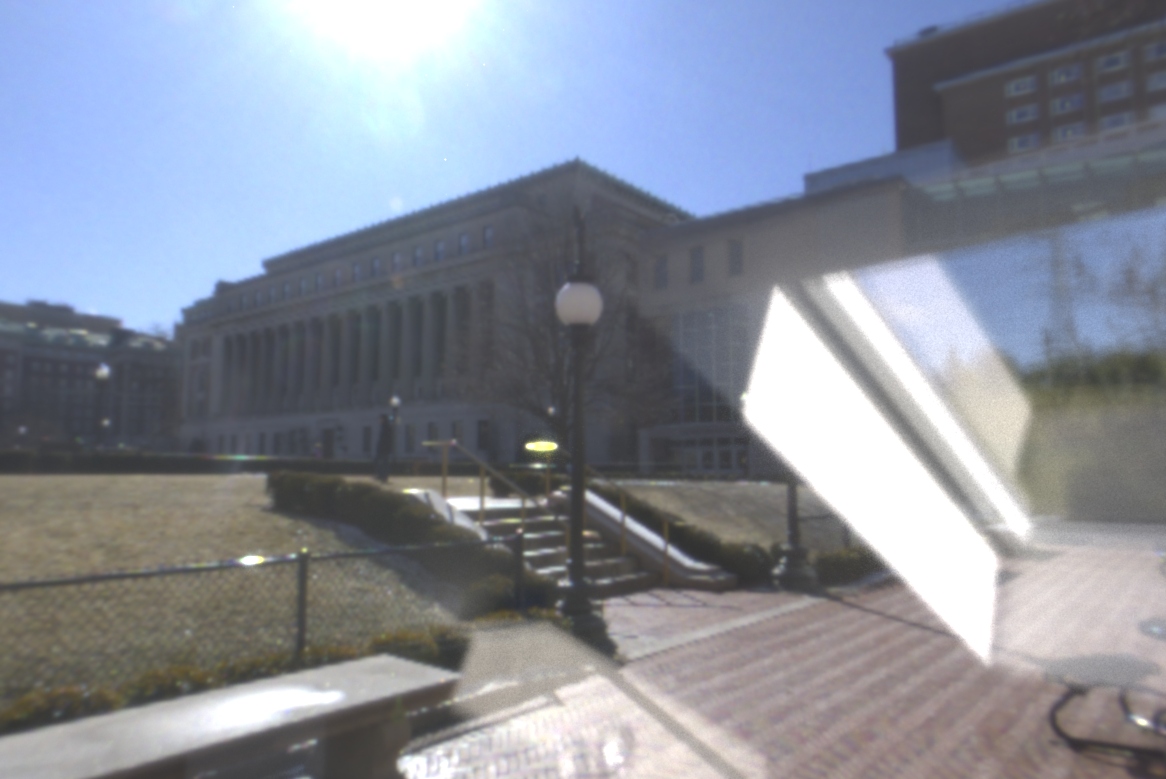} &
        \includegraphics[width=\resLen]{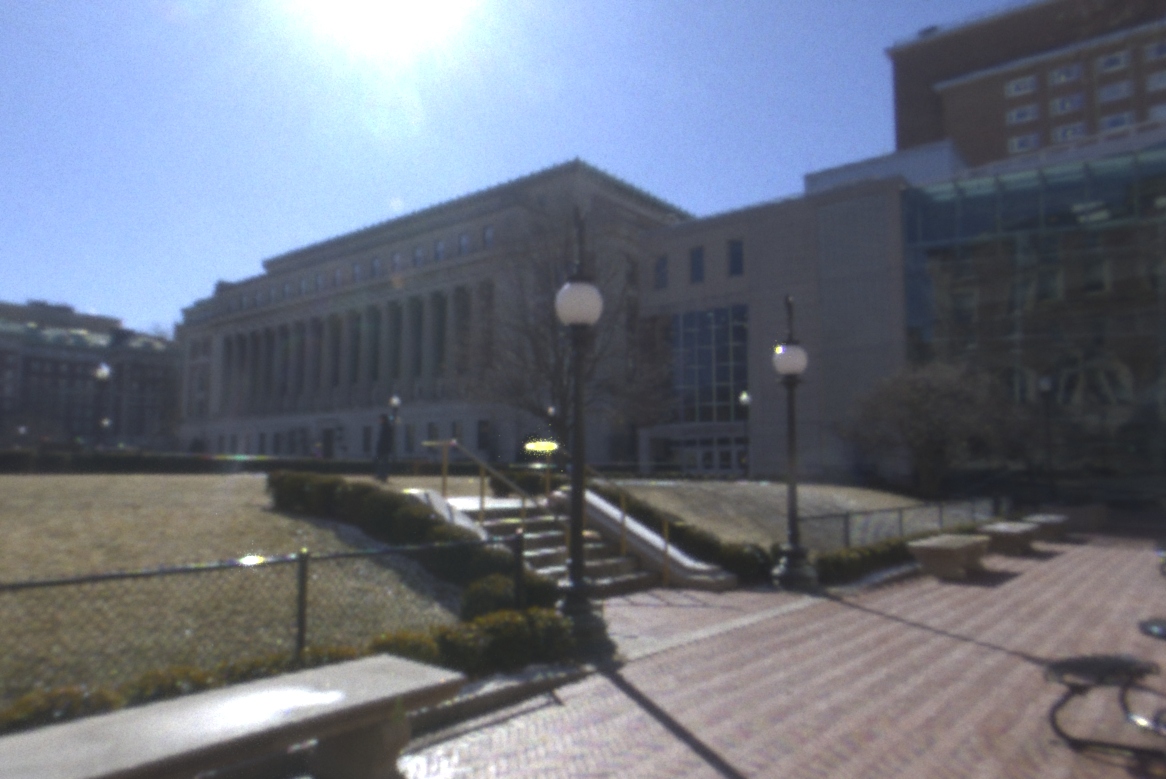} &
        \includegraphics[width=\resLen]{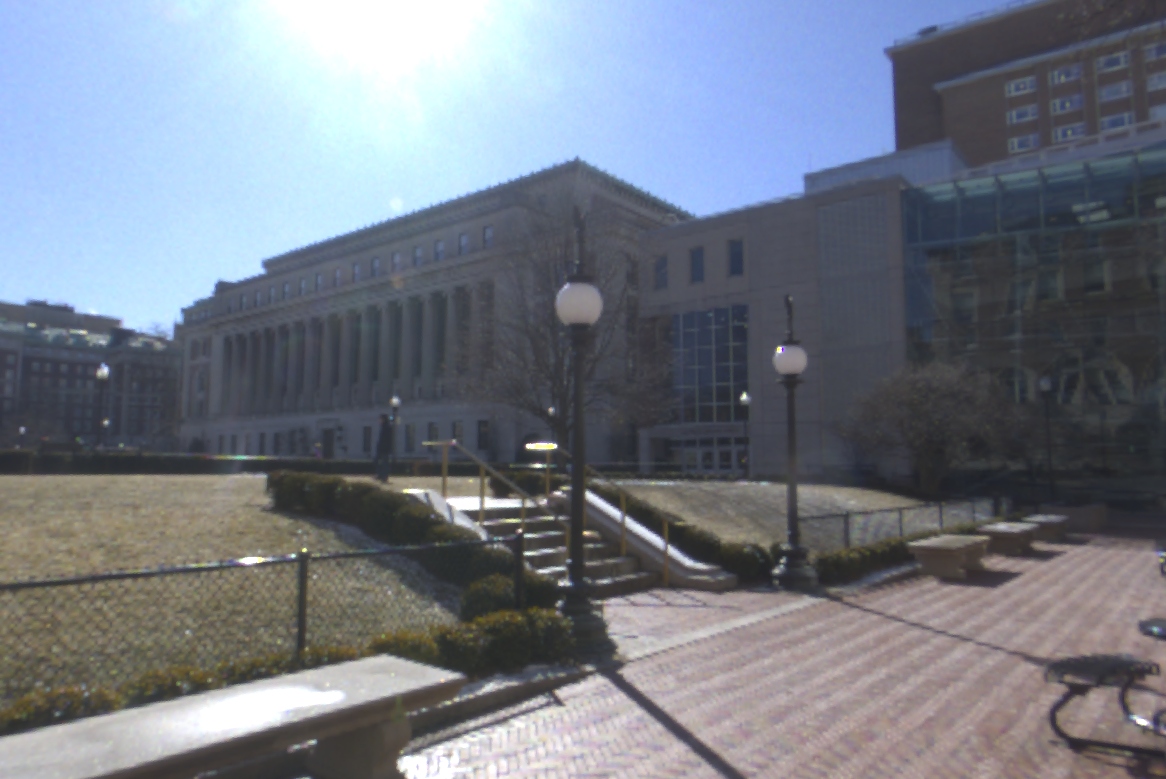} &
        \includegraphics[width=\resLen]{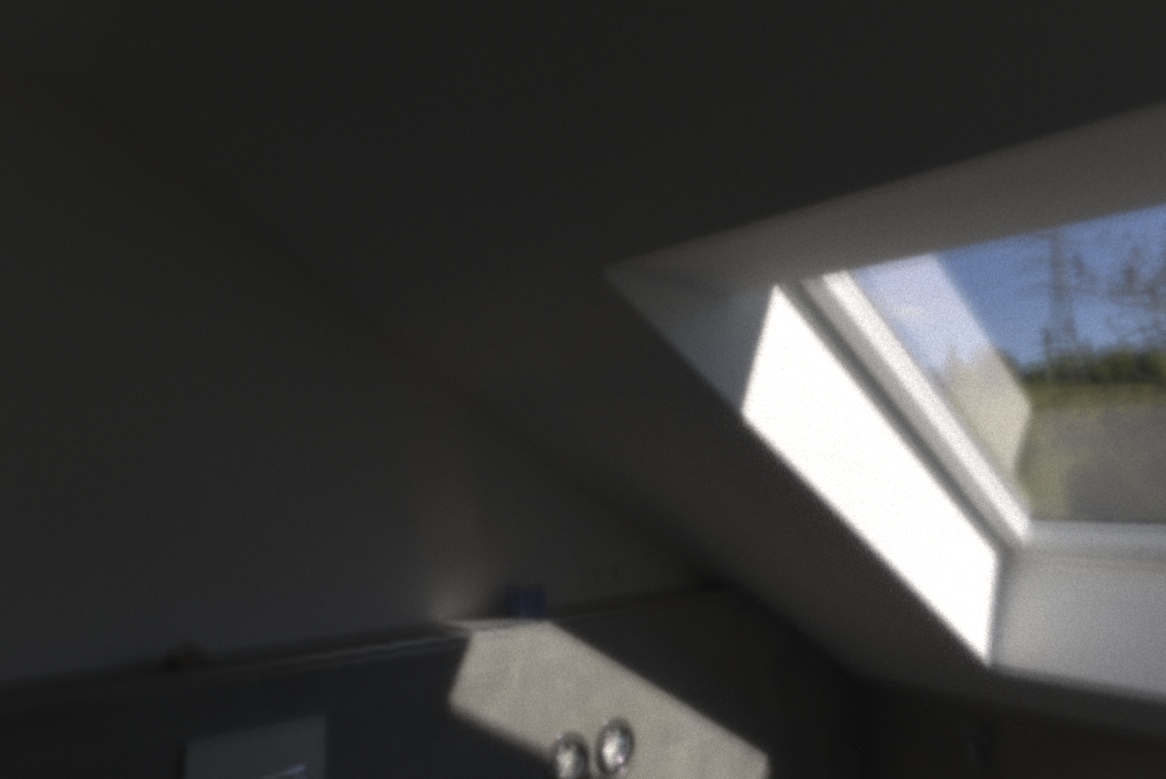} &
        \includegraphics[width=\resLen]{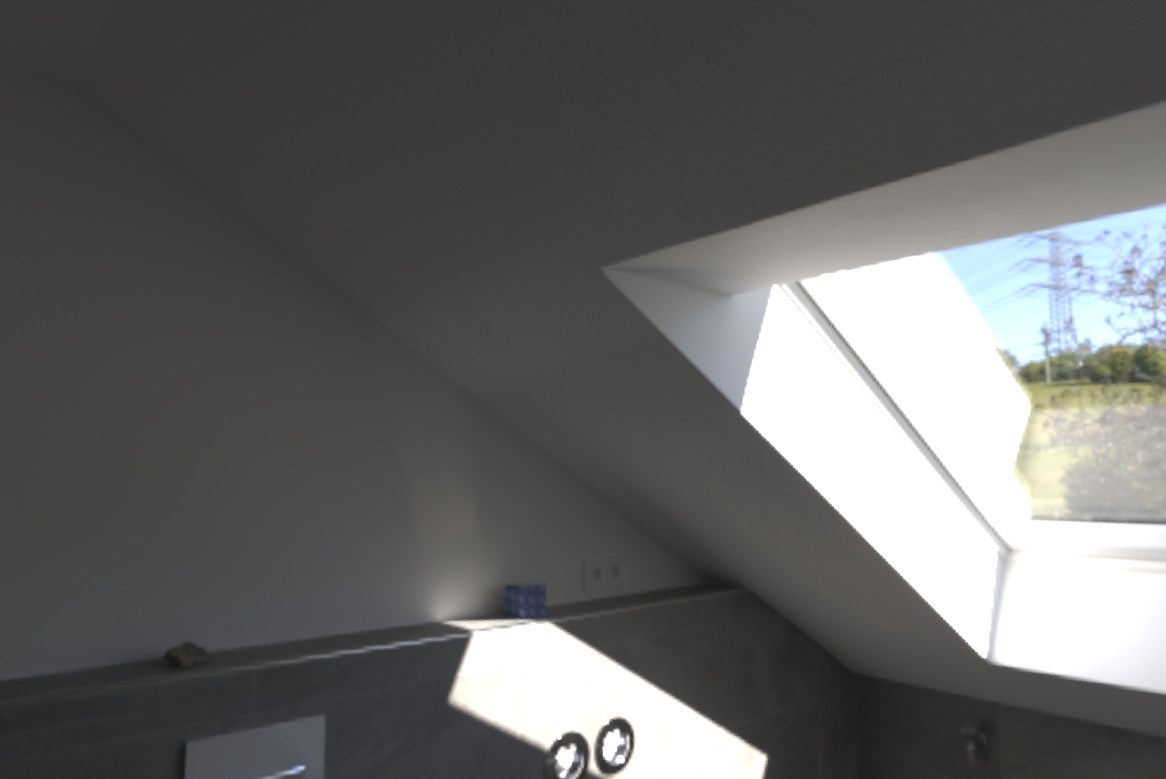} 
        \\        
        \I & \T & \B & \R & \MR
    \end{tabular}
    \vspace{-8pt}
    \caption{
        \textbf{Our synthetic data examples} 
    }
    \label{fig:data_preview}
\end{figure}

\begin{figure*}
    \centering
    \setlength{\resLen}{0.136\textwidth}
    \addtolength{\tabcolsep}{-5pt}
    \renewcommand{\arraystretch}{0.5}
    \begin{tabular}{ccccccc}
        \raisebox{20pt}{\multirow{2}{*}{
        \includegraphics[width=\resLen]{fig/img/more/input/000002.jpg}}} &
        \includegraphics[width=\resLen]{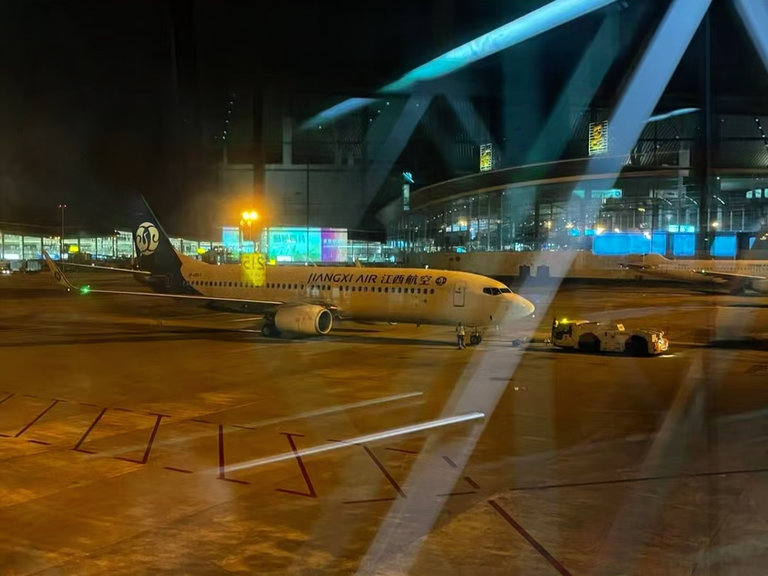} &
        \includegraphics[width=\resLen]{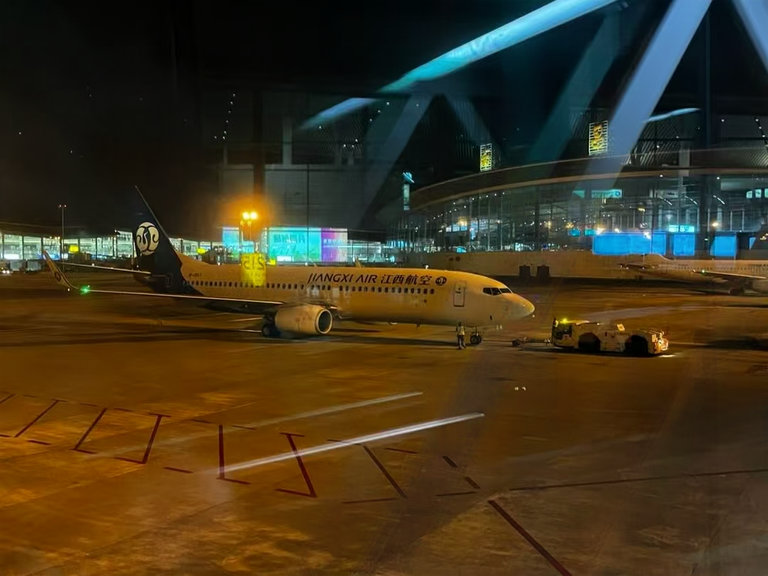} &
        \includegraphics[width=\resLen]{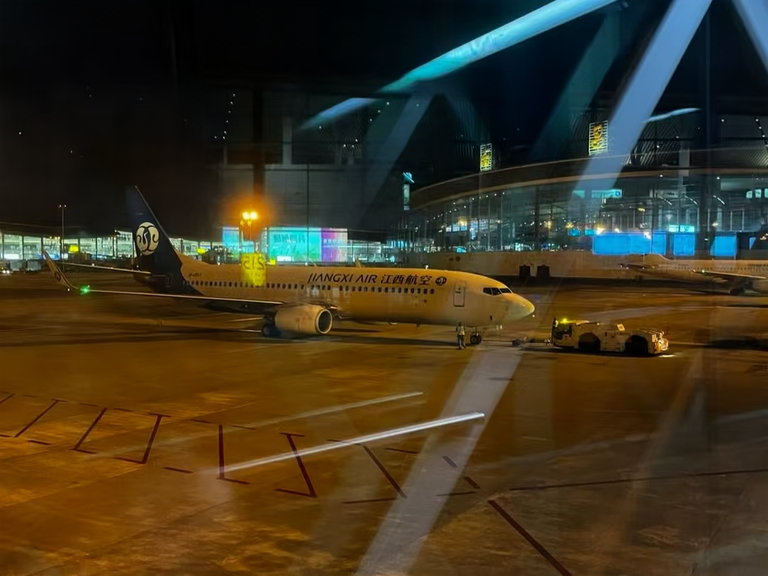} &
        \includegraphics[width=\resLen]{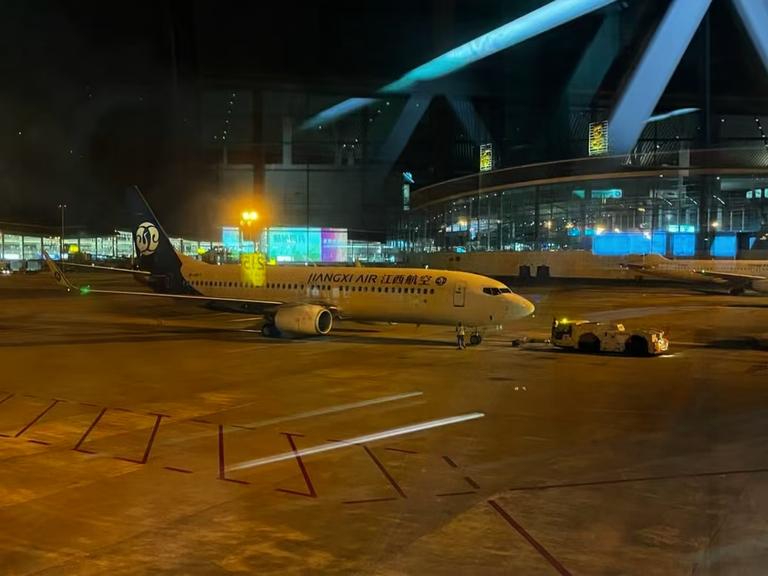} &
        \includegraphics[width=\resLen]{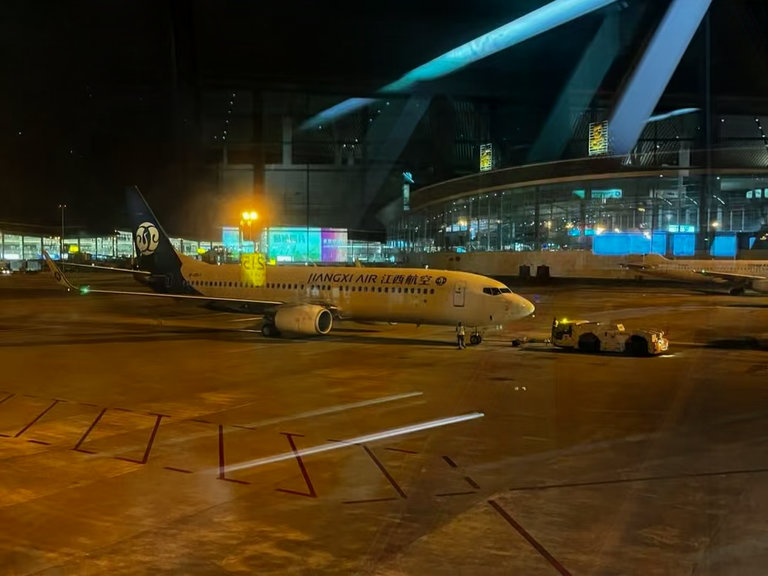} &
        \includegraphics[width=\resLen]{fig/img/more/ours/T/000002.jpg}
        \\ 
        \raisebox{18pt}{(a)} &
        \includegraphics[width=\resLen]{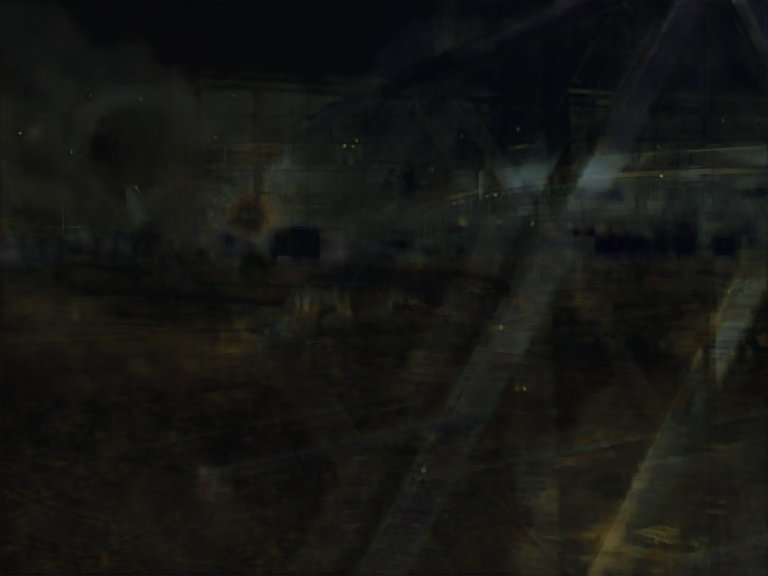} &
        \includegraphics[width=\resLen]{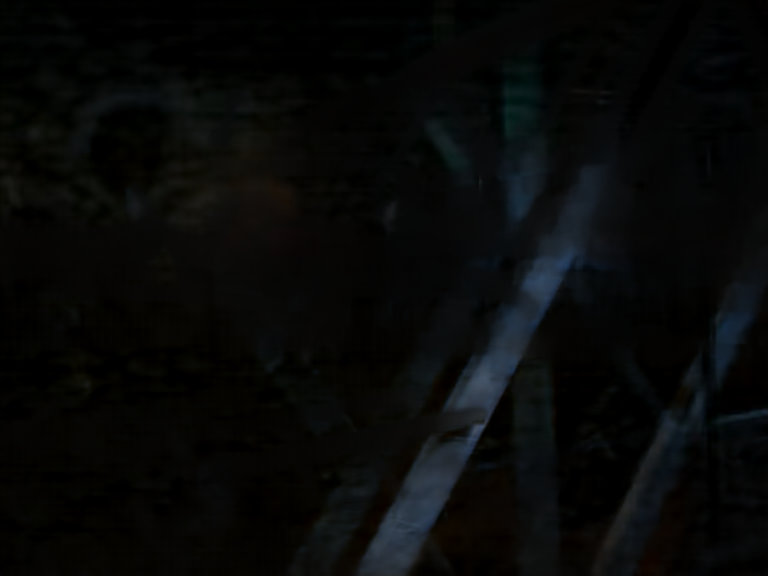} &
         &
        \includegraphics[width=\resLen]{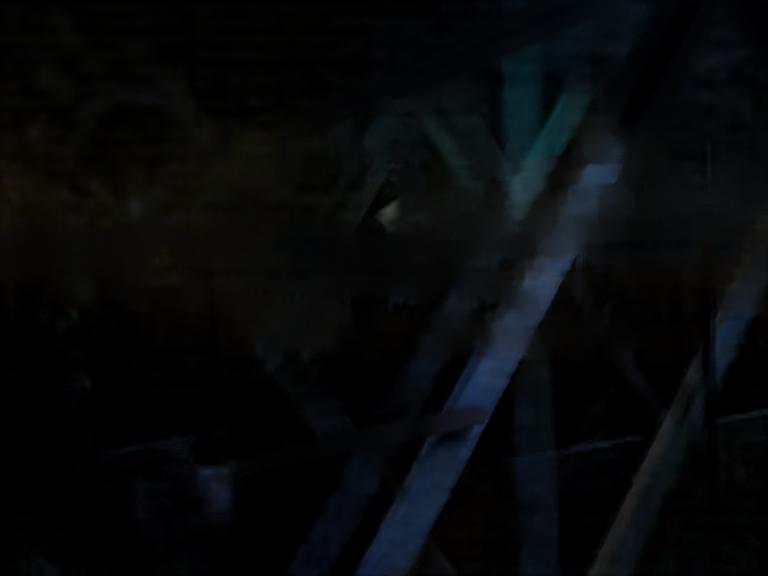} &
        \includegraphics[width=\resLen]{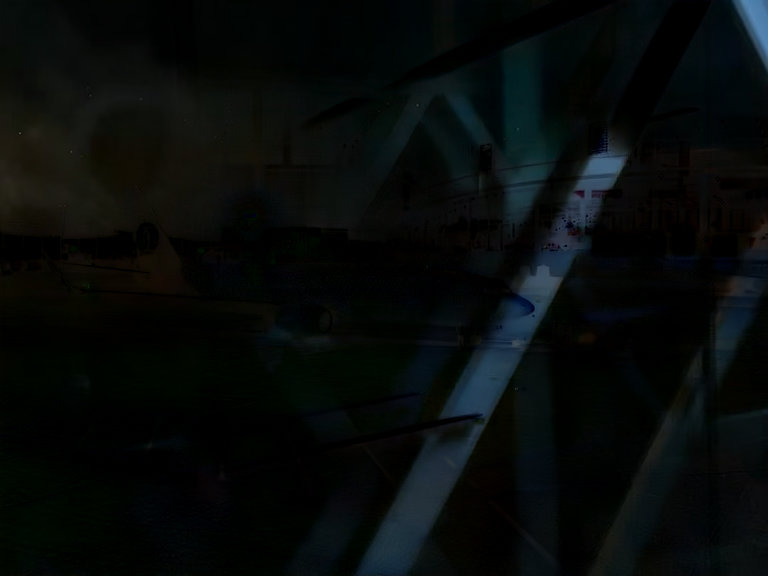} &
        \includegraphics[width=\resLen]{fig/img/more/ours/R/000002.jpg} 
        \\
        \raisebox{18pt}{\multirow{2}{*}{
        \includegraphics[width=\resLen]{fig/img/more/input/000003.jpg}}} &
        \includegraphics[width=\resLen]{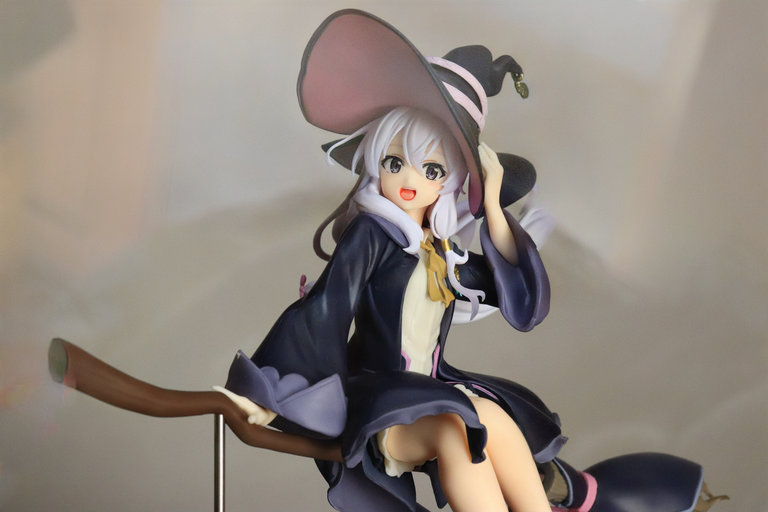} &
        \includegraphics[width=\resLen]{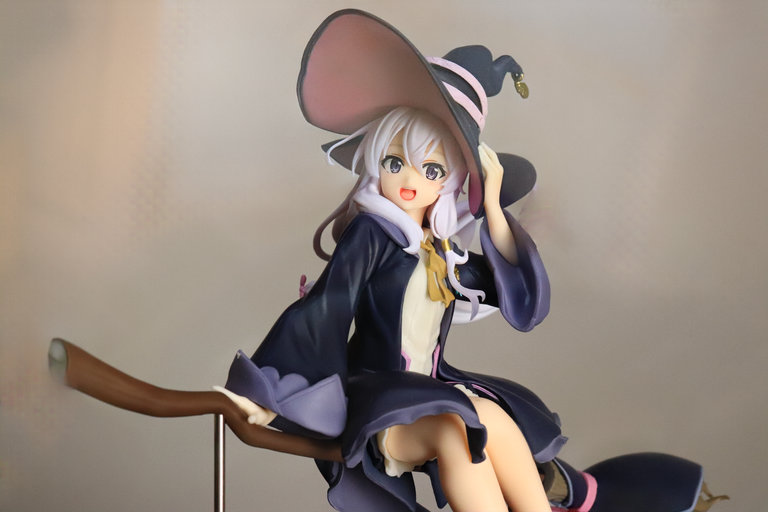} &
        \includegraphics[width=\resLen]{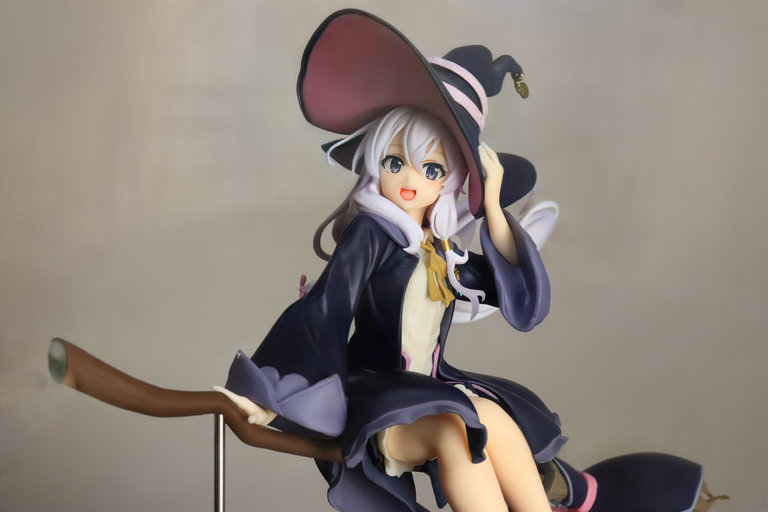} &
        \includegraphics[width=\resLen]{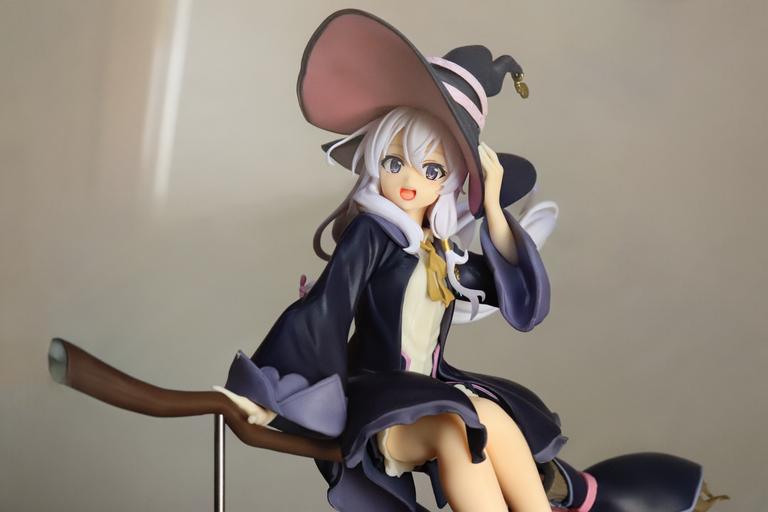} &
        \includegraphics[width=\resLen]{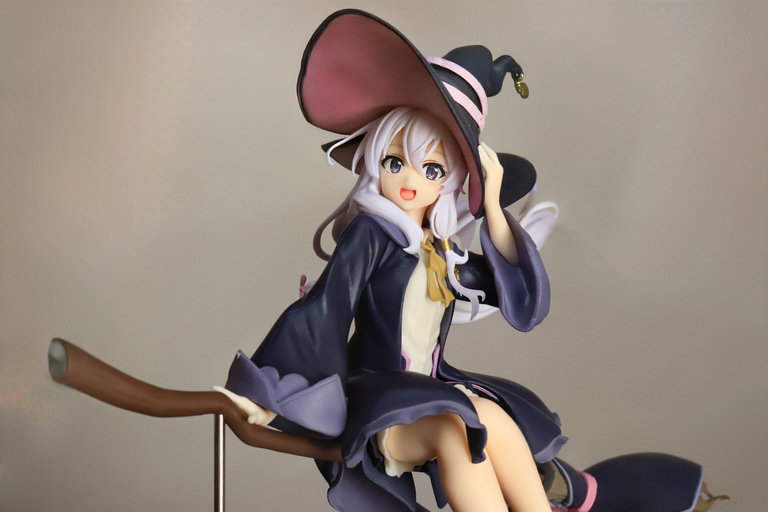} &
        \includegraphics[width=\resLen]{fig/img/more/ours/T/000003.jpg} 
        \\ 
        \raisebox{16pt}{(b)} &
        \includegraphics[width=\resLen]{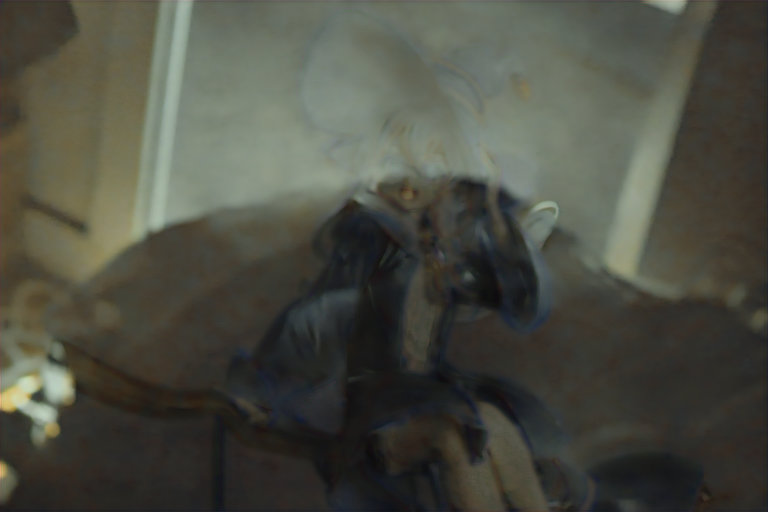} &
        \includegraphics[width=\resLen]{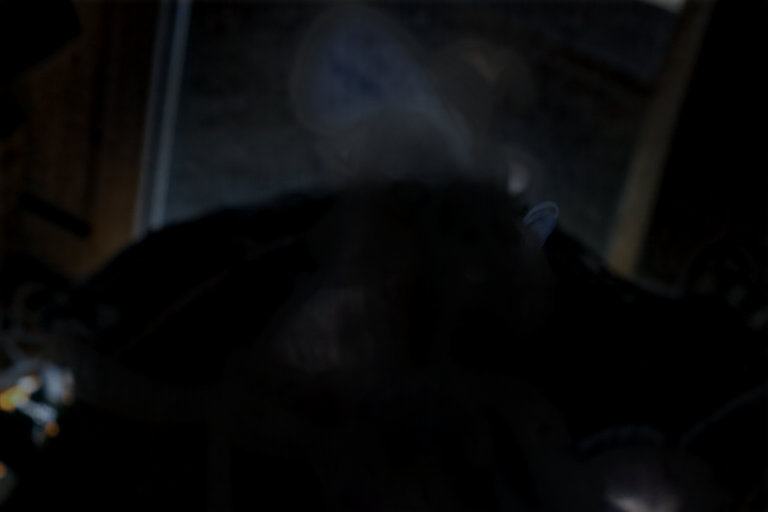} &
         &
        \includegraphics[width=\resLen]{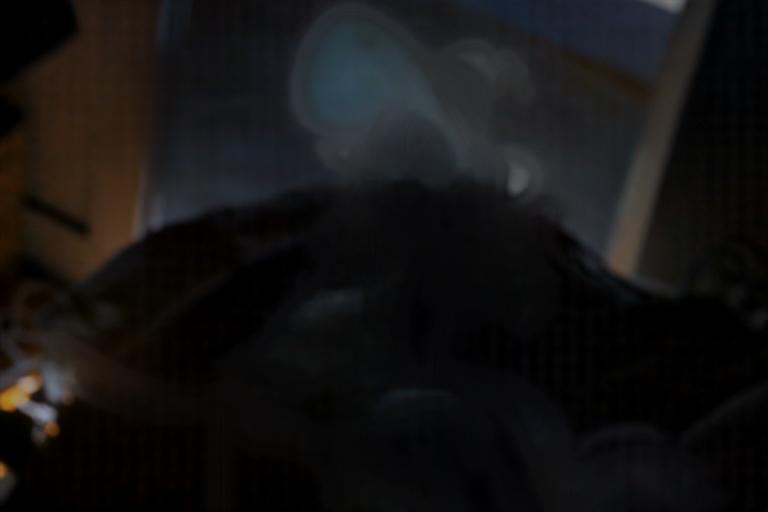} &
        \includegraphics[width=\resLen]{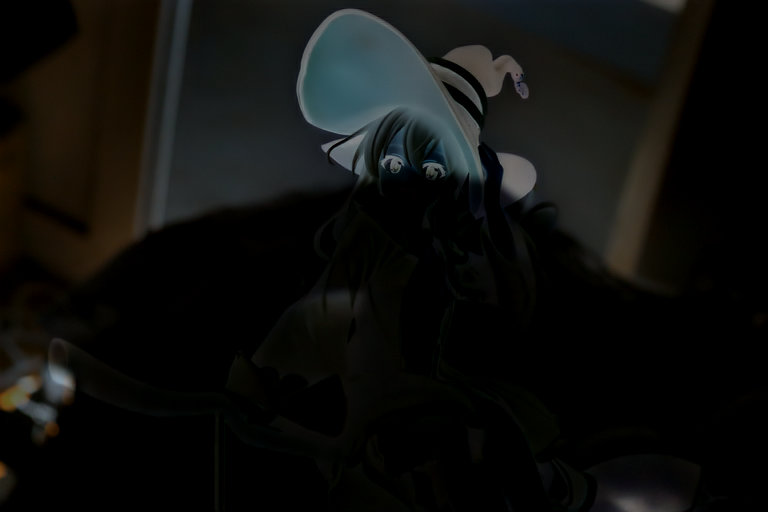} &
        \includegraphics[width=\resLen]{fig/img/more/ours/R/000003.jpg} 
        \\        
        \raisebox{20pt}{\multirow{2}{*}{
        \includegraphics[width=\resLen, height=0.75\resLen, keepaspectratio=False, trim=0 0 0 10,clip]{fig/img/more/input/000010.jpg}}} &
        \includegraphics[width=\resLen, height=0.75\resLen, keepaspectratio=False, trim=0 0 0 35,clip]{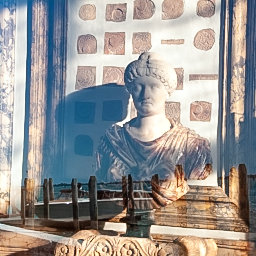} &
        \includegraphics[width=\resLen, height=0.75\resLen, keepaspectratio=False, trim=0 0 0 35,clip]{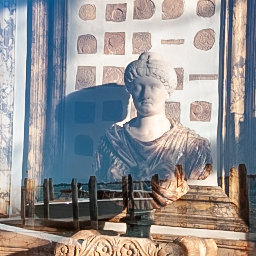} &
        \includegraphics[width=\resLen, height=0.75\resLen, keepaspectratio=False, trim=0 0 0 35,clip]{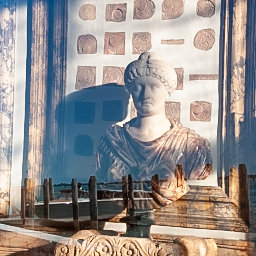} &
        \includegraphics[width=\resLen, height=0.75\resLen, keepaspectratio=False, trim=0 0 0 35,clip]{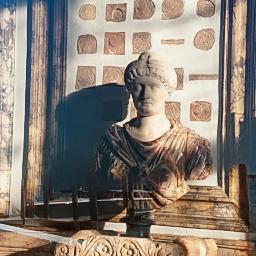} &
        \includegraphics[width=\resLen, height=0.75\resLen, keepaspectratio=False, trim=0 0 0 35,clip]{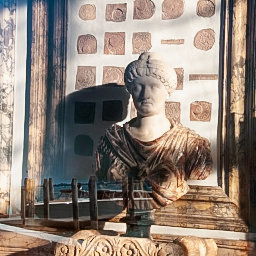} &
        \includegraphics[width=\resLen, height=0.75\resLen, keepaspectratio=False, trim=0 0 0 35,clip]{fig/img/more/ours/T/000010.jpg}
        \\ 
        \raisebox{18pt}{(c)} & 
        \includegraphics[width=\resLen, height=0.75\resLen, keepaspectratio=False, trim=0 0 0 35,clip]{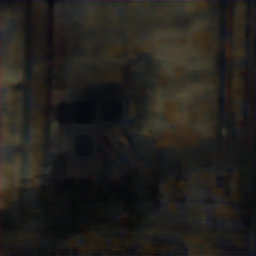} &
        \includegraphics[width=\resLen, height=0.75\resLen, keepaspectratio=False, trim=0 0 0 35,clip]{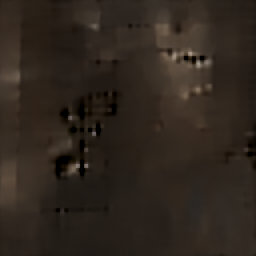} &
         &
        \includegraphics[width=\resLen, height=0.75\resLen, keepaspectratio=False, trim=0 0 0 35,clip]{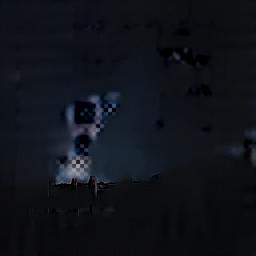} &
        \includegraphics[width=\resLen, height=0.75\resLen, keepaspectratio=False, trim=0 0 0 35,clip]{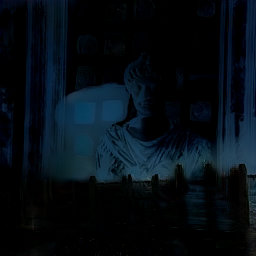} &
        \includegraphics[width=\resLen, height=0.75\resLen, keepaspectratio=False, trim=0 0 0 35,clip]{fig/img/more/ours/R/000010.jpg}
        \\
        \raisebox{20pt}{\multirow{2}{*}{
        \includegraphics[width=\resLen, height=0.8\resLen, keepaspectratio=False, trim=0 0 0 10,clip]{fig/img/more/input/000011.jpg}}} &
        \includegraphics[width=\resLen, height=0.8\resLen, keepaspectratio=False, trim=0 0 0 35,clip]{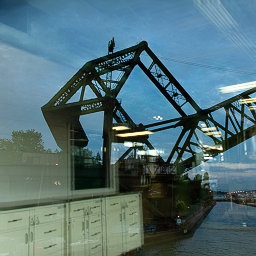} &
        \includegraphics[width=\resLen, height=0.8\resLen, keepaspectratio=False, trim=0 0 0 35,clip]{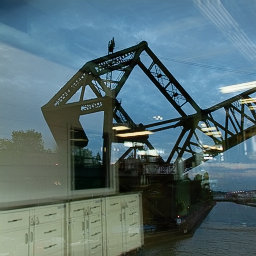} &
        \includegraphics[width=\resLen, height=0.8\resLen, keepaspectratio=False, trim=0 0 0 35,clip]{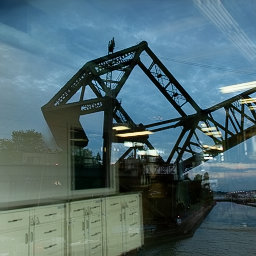} &
        \includegraphics[width=\resLen, height=0.8\resLen, keepaspectratio=False, trim=0 0 0 35,clip]{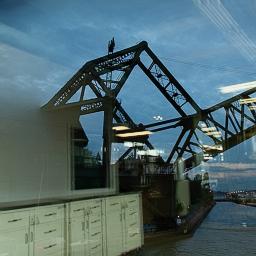} &
        \includegraphics[width=\resLen, height=0.8\resLen, keepaspectratio=False, trim=0 0 0 35,clip]{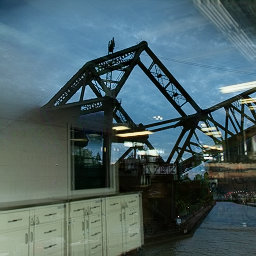} &
        \includegraphics[width=\resLen, height=0.8\resLen, keepaspectratio=False, trim=0 0 0 35,clip]{fig/img/more/ours/T/000011.jpg}
        \\ 
        \raisebox{18pt}{(d)} & 
        \includegraphics[width=\resLen, height=0.8\resLen, keepaspectratio=False, trim=0 0 0 35,clip]{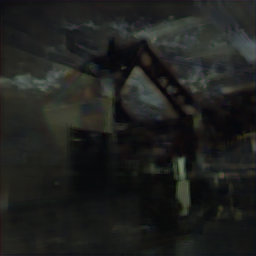} &
        \includegraphics[width=\resLen, height=0.8\resLen, keepaspectratio=False, trim=0 0 0 35,clip]{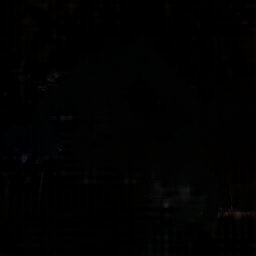} &
         &
        \includegraphics[width=\resLen, height=0.8\resLen, keepaspectratio=False, trim=0 0 0 35,clip]{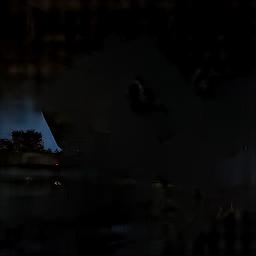} &
        \includegraphics[width=\resLen, height=0.8\resLen, keepaspectratio=False, trim=0 0 0 35,clip]{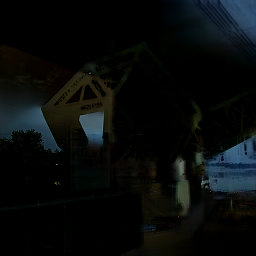} &
        \includegraphics[width=\resLen, height=0.8\resLen, keepaspectratio=False, trim=0 0 0 35,clip]{fig/img/more/ours/R/000011.jpg}
        \\
        \raisebox{20pt}{\multirow{2}{*}{
        \includegraphics[width=\resLen]{fig/img/more/input/000023.jpg}}} &
        \includegraphics[width=\resLen]{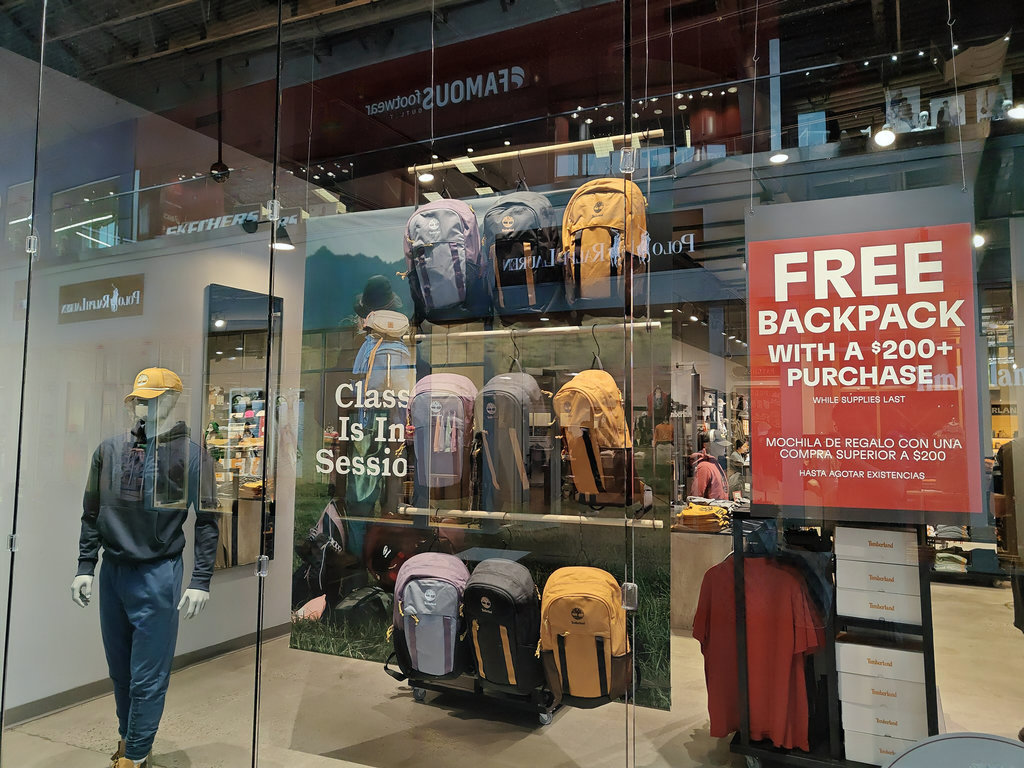} &
        \includegraphics[width=\resLen]{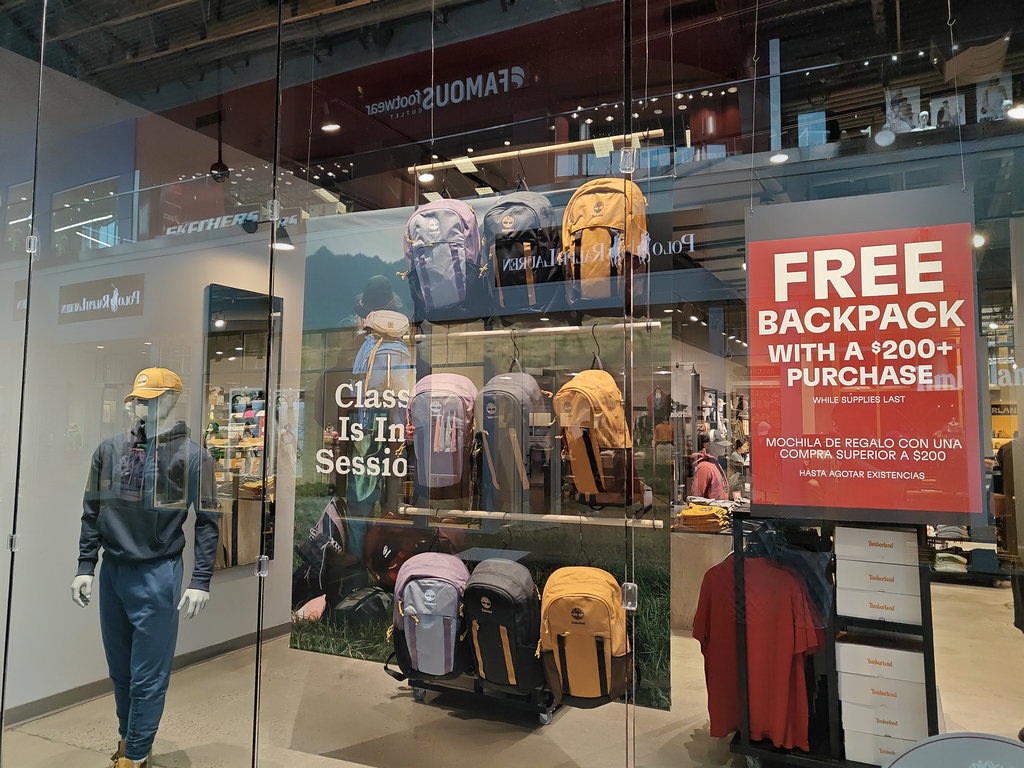} &
        \includegraphics[width=\resLen]{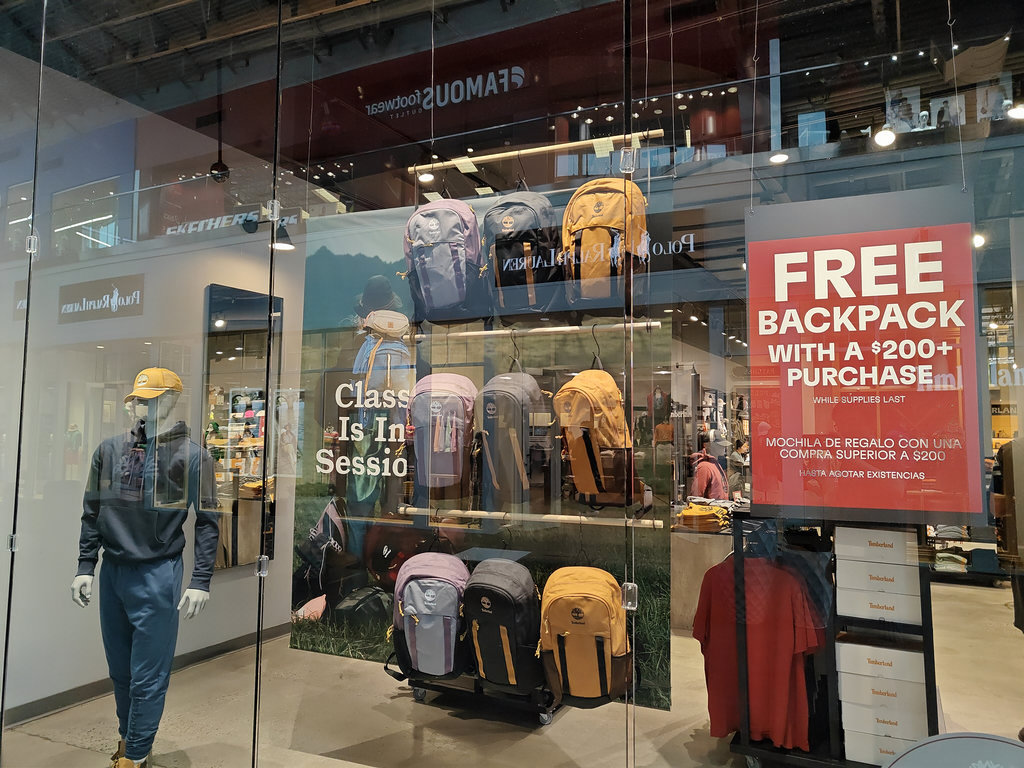} &
        \includegraphics[width=\resLen]{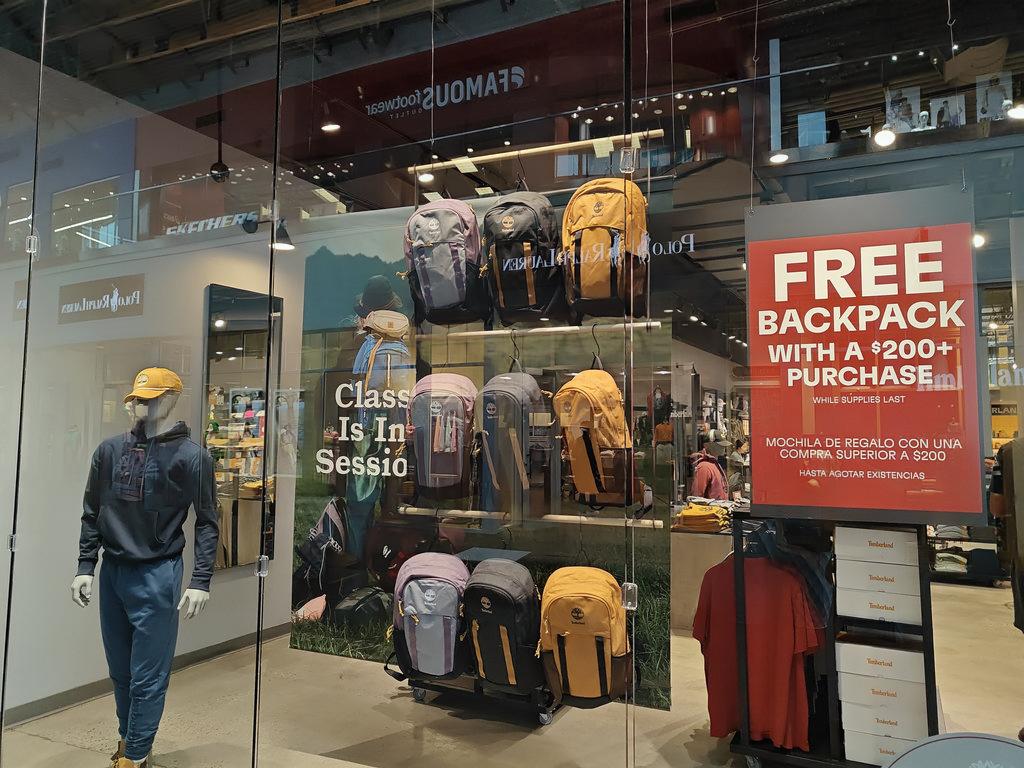} &
        \includegraphics[width=\resLen]{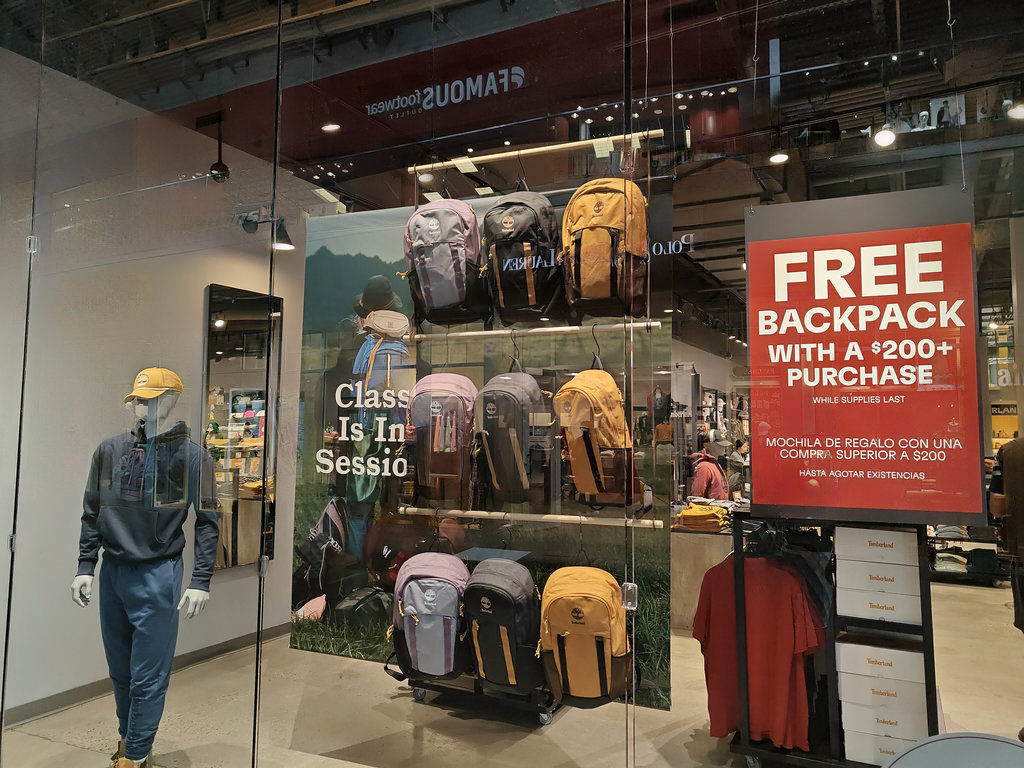} &
        \includegraphics[width=\resLen]{fig/img/more/ours/T/000023.jpg}
        \\
        \raisebox{18pt}{(e)} &
        \includegraphics[width=\resLen]{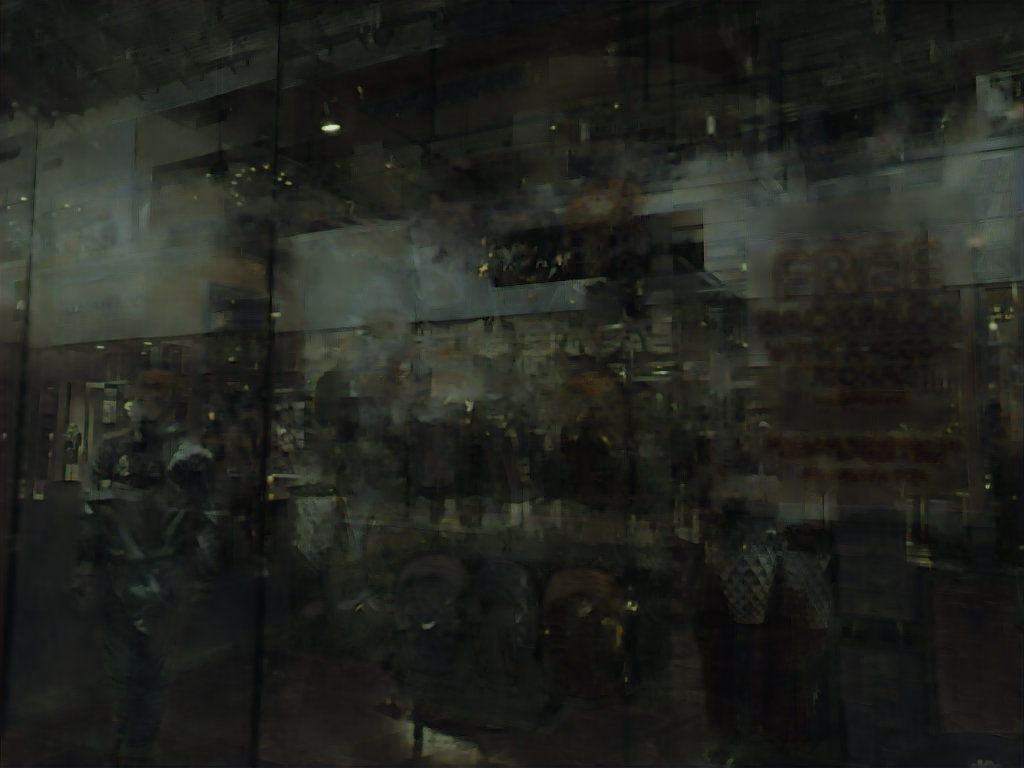} &
        \includegraphics[width=\resLen]{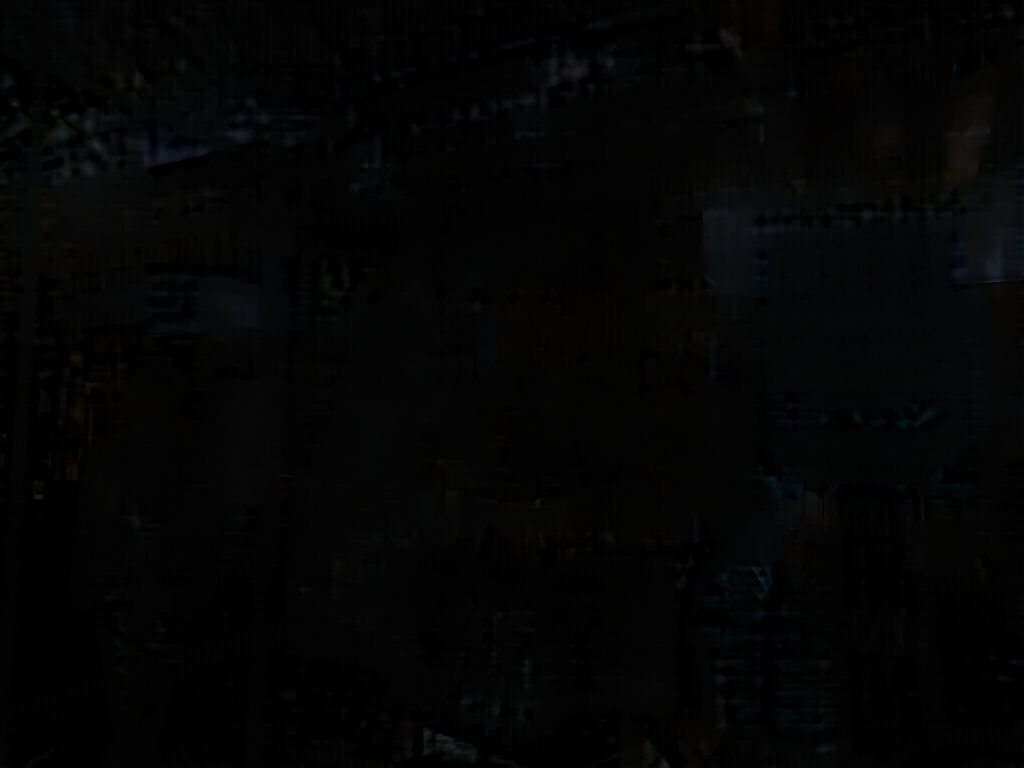} &
         &
        \includegraphics[width=\resLen]{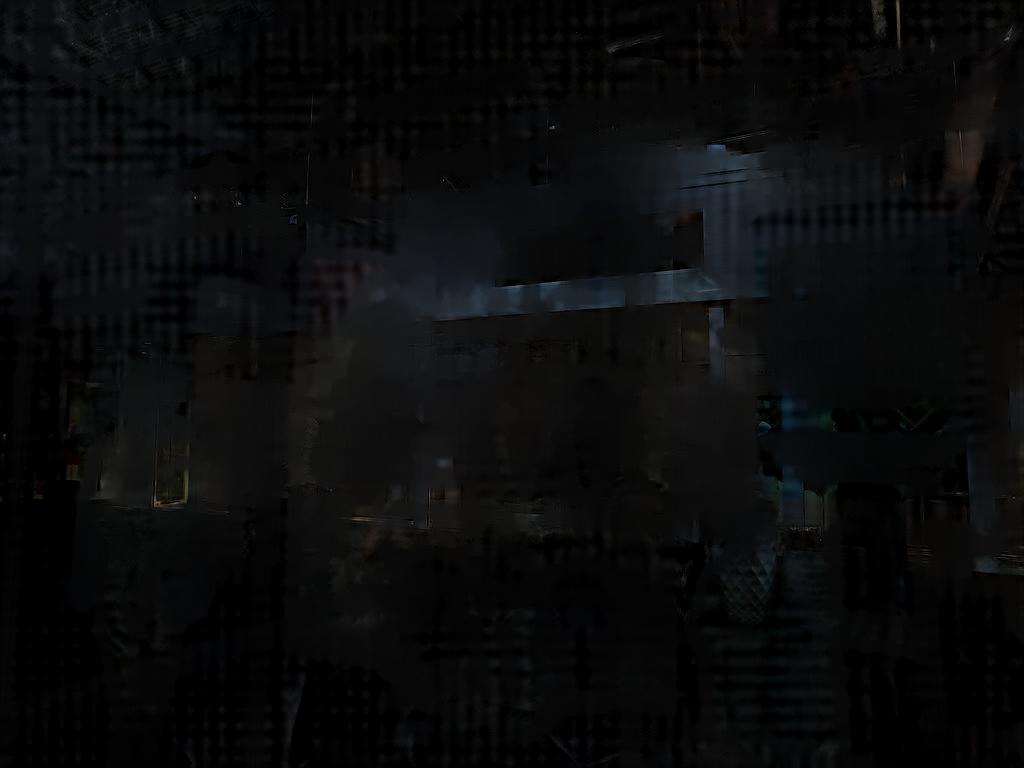} &
        \includegraphics[width=\resLen]{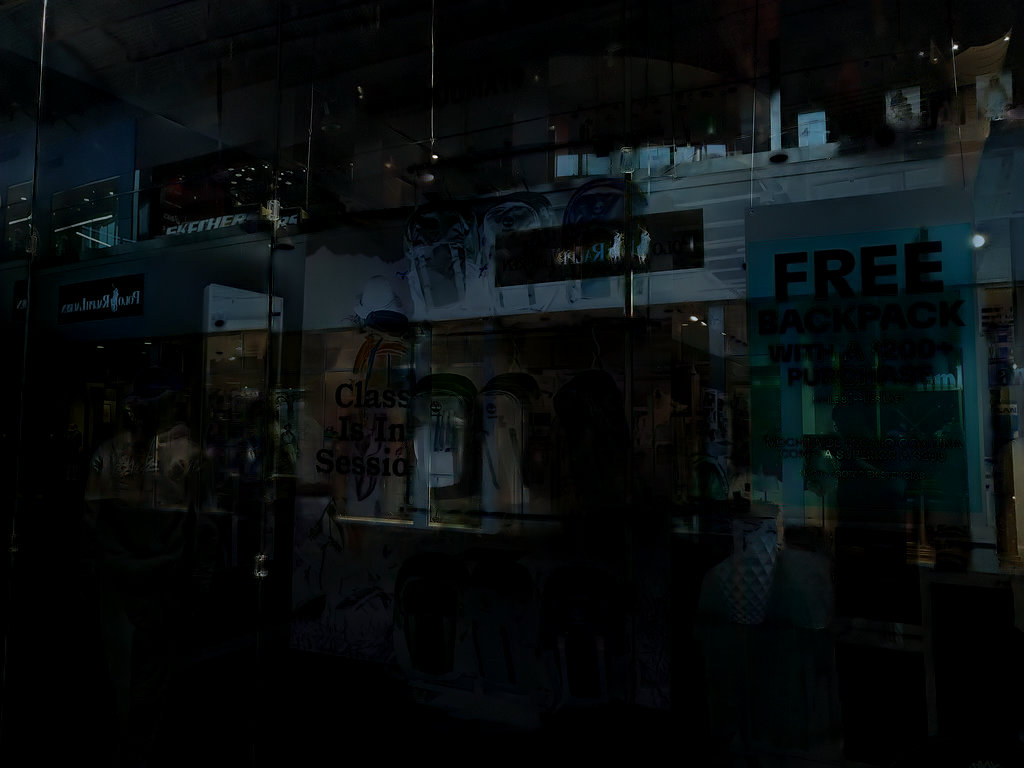} &
        \includegraphics[width=\resLen]{fig/img/more/ours/R/000023.jpg}
        \\
        \raisebox{20pt}{\multirow{2}{*}{
        \includegraphics[width=\resLen]{fig/img/more/input/000016.jpg}}} &
        \includegraphics[width=\resLen]{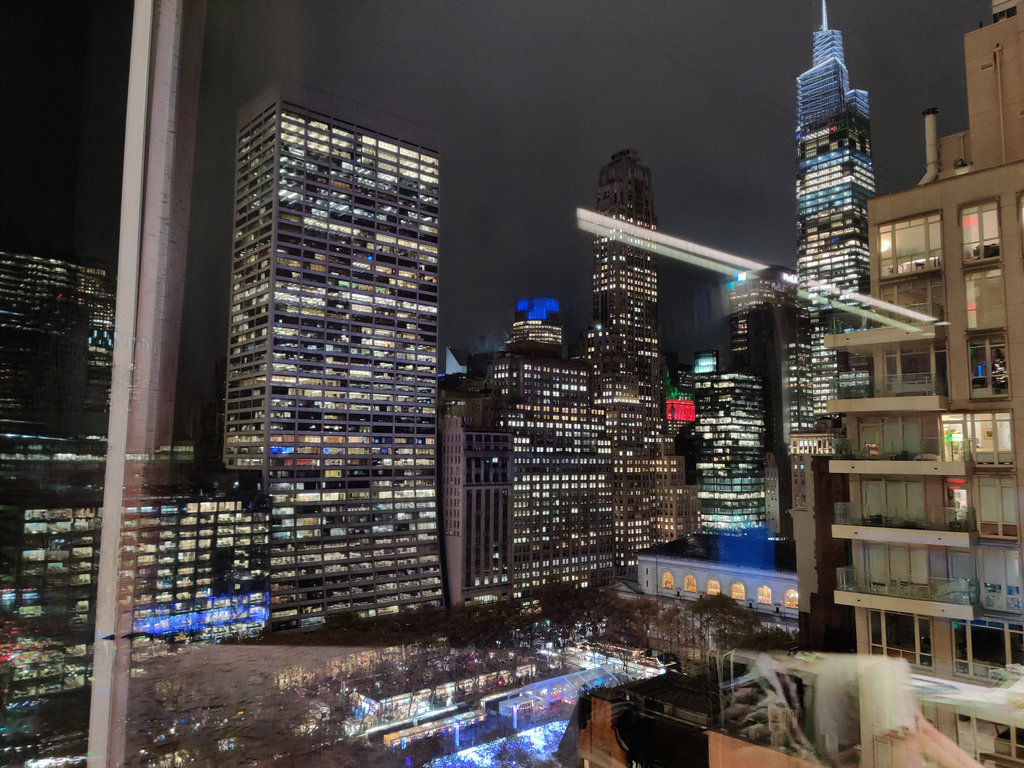} &
        \includegraphics[width=\resLen]{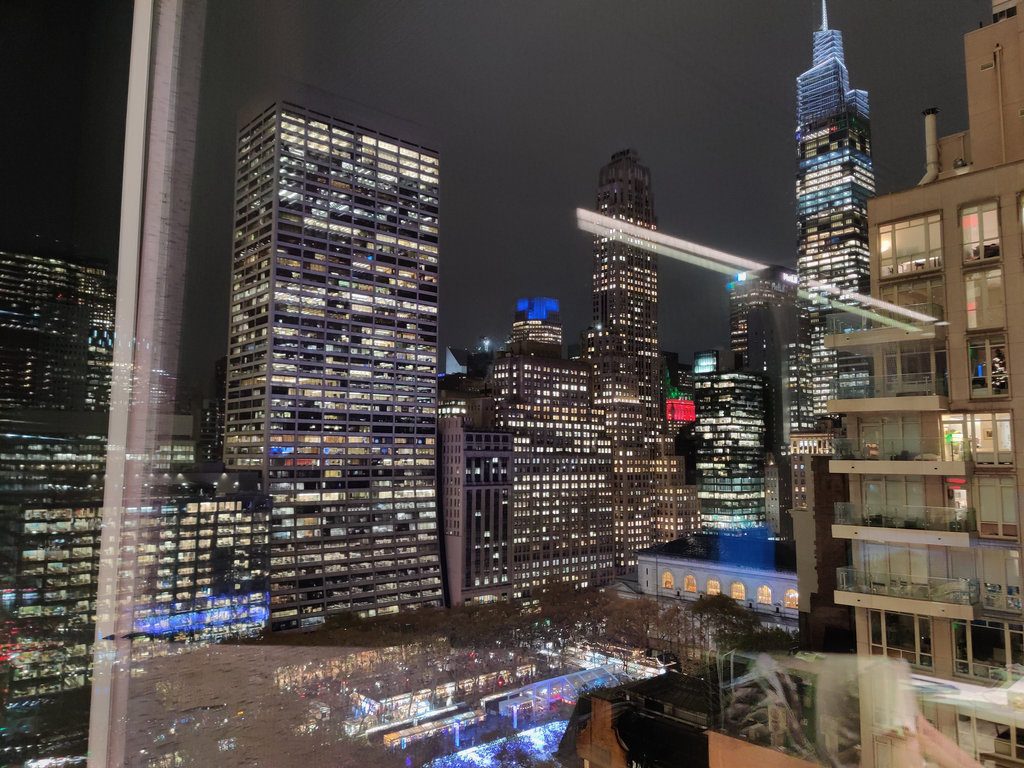} &
        \includegraphics[width=\resLen]{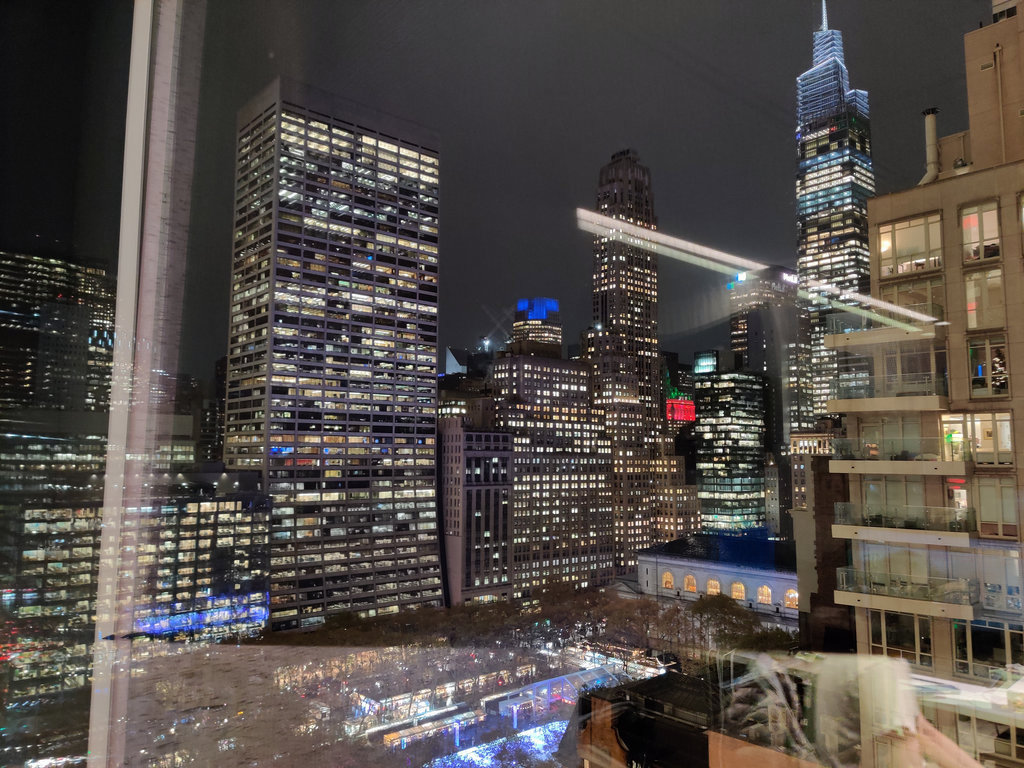} &
        \includegraphics[width=\resLen]{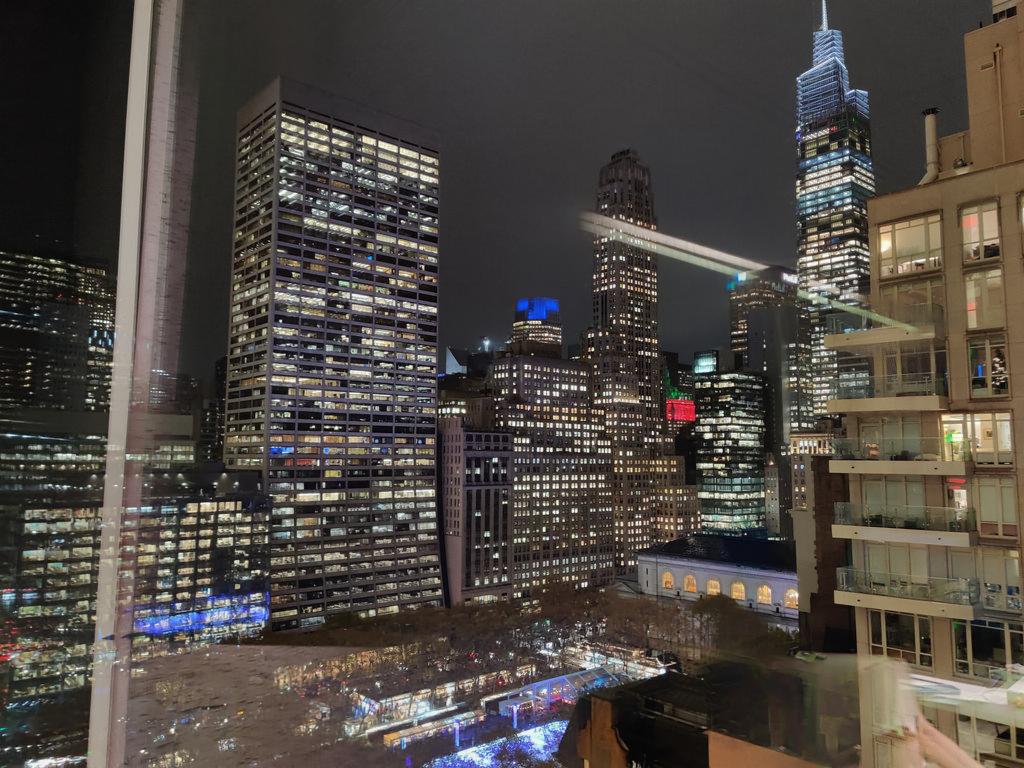} &
        \includegraphics[width=\resLen]{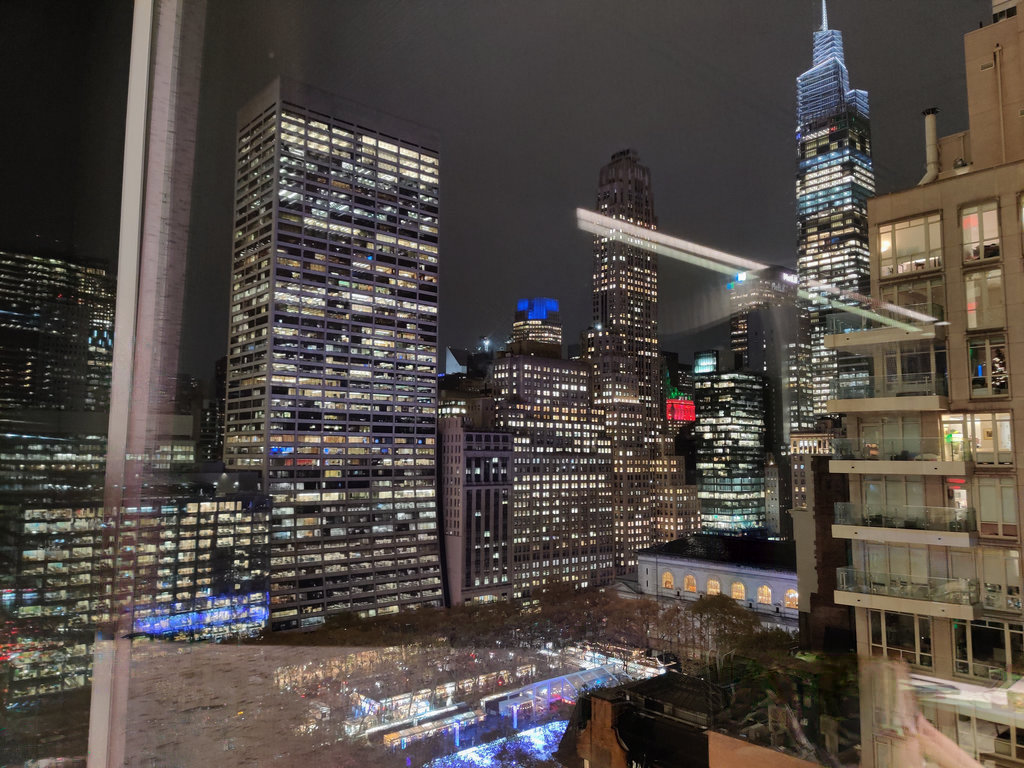} &
        \includegraphics[width=\resLen]{fig/img/more/ours/T/000016.jpg}
        \\
        \raisebox{18pt}{(f)} &
        \includegraphics[width=\resLen]{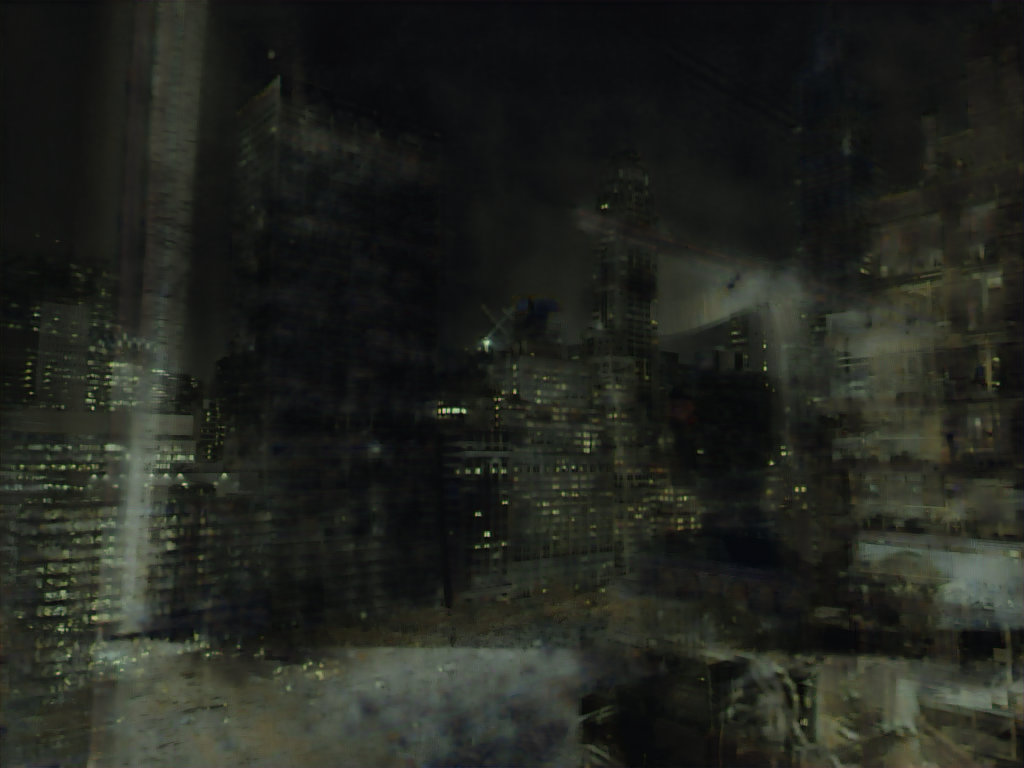} &
        \includegraphics[width=\resLen]{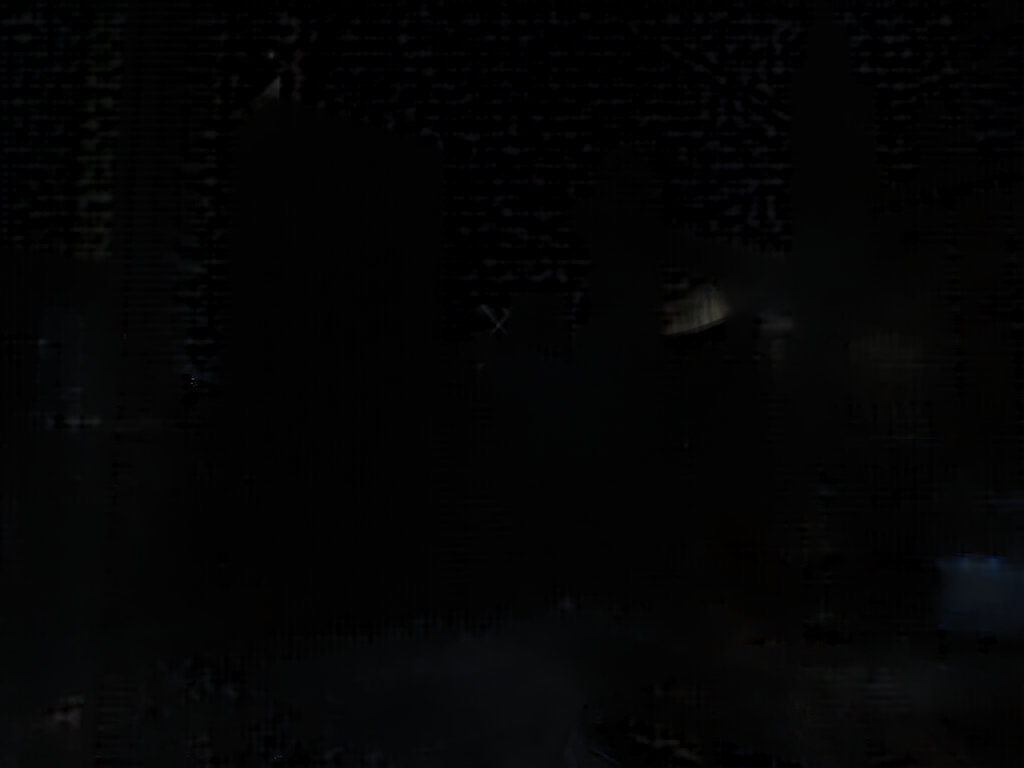} &
         &
        \includegraphics[width=\resLen]{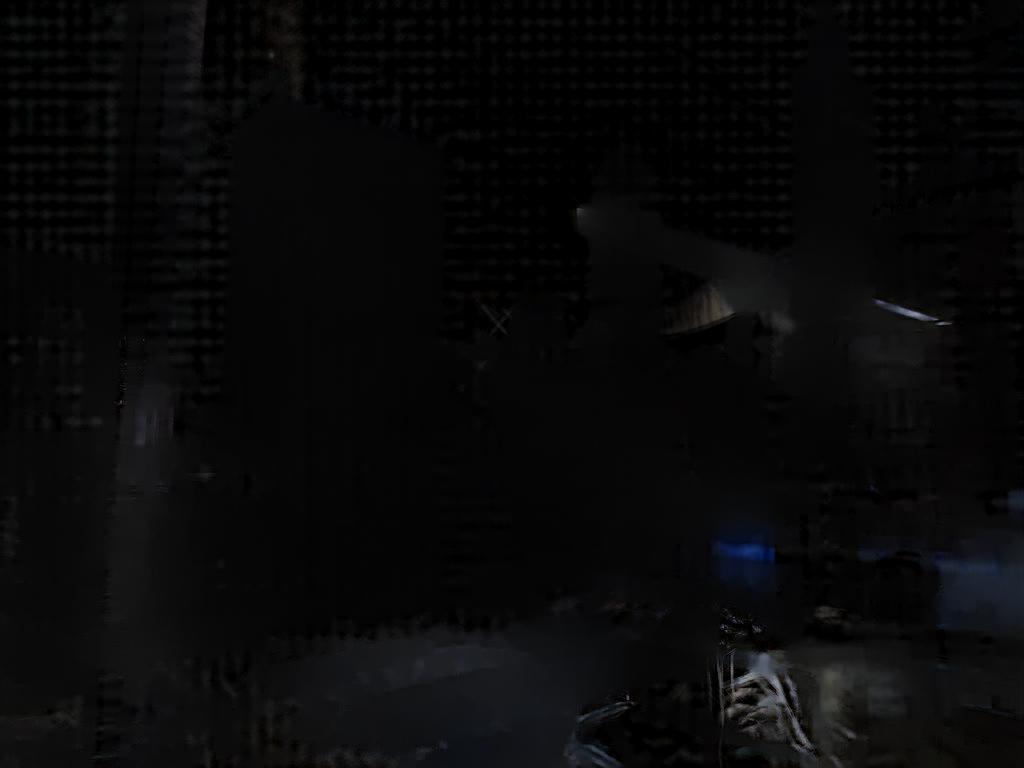} &
        \includegraphics[width=\resLen]{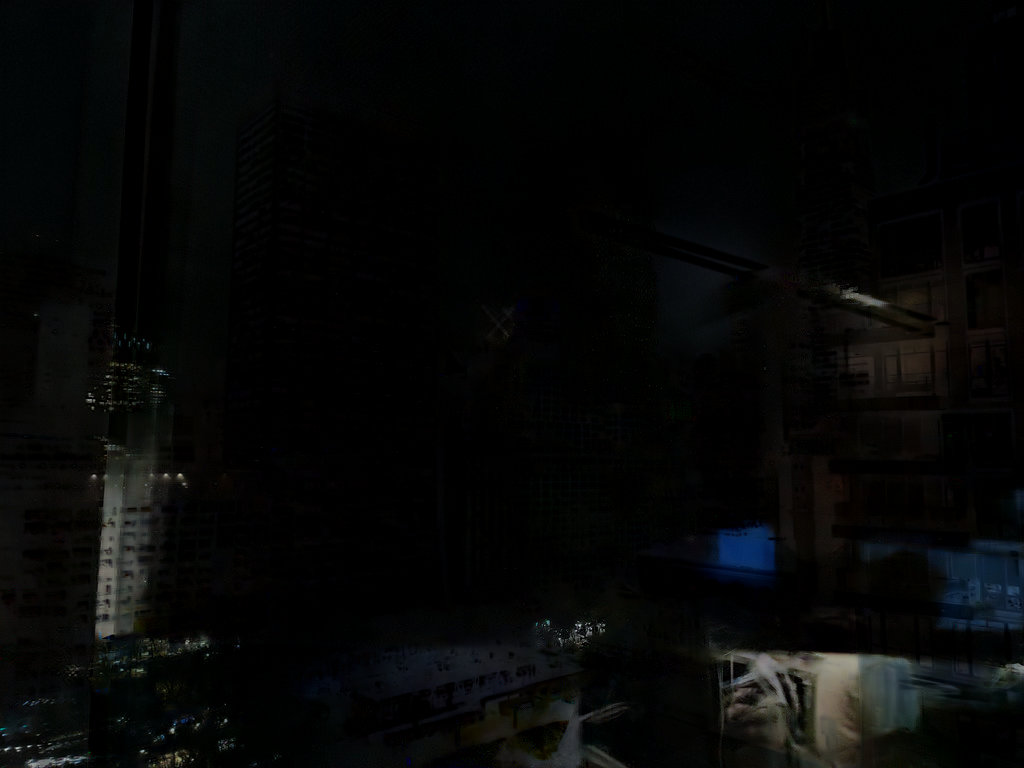} & 
        \includegraphics[width=\resLen]{fig/img/more/ours/R/000016.jpg}
        \\
        Input
        & Dong \cite{dong2021location} 
        & DSRNet \cite{hu2023single} 
        & Zhu \cite{zhu2024revisiting} 
        & DSIT \cite{hu2024single} 
        & RDNet \cite{zhao2025reversible}
        & Ours
    \end{tabular}
    \vspace{-8pt}
    \caption{
        \textbf{More results comparison.} We compared both \textbf{T}ransmission and \textbf{R}eflection of a given image with state-of-the-art methods. (a) (b) from \cite{hu2025dereflection}; (c) (d) from \cite{kee2025removing}; (e) (f) from our smartphone captures. 
    }
    \label{fig:main_other_compare}
\end{figure*}

\end{document}